\documentclass[10pt,twocolumn,letterpaper]{article}

\usepackage[pagenumbers]{cvpr}

\newcommand{\method}[0]{Stereo Anywhere\xspace}
\newcommand{\dataset}[0]{MonoTrap\xspace}

\usepackage{colortbl}
\usepackage{multirow}

\definecolor{firstcolor}{HTML}{BDE6CD}
\definecolor{secondcolor}{HTML}{E2EEBC}
\definecolor{thirdcolor}{HTML}{FFF8C5}

\newcommand{\fst}[1]{\cellcolor{firstcolor}\bfseries #1}
\newcommand{\snd}[1]{\cellcolor{secondcolor} #1}
\newcommand{\trd}[1]{\cellcolor{thirdcolor}#1}

\usepackage{overpic}
\usepackage{pifont}
\usepackage[accsupp]{axessibility}

\definecolor{cvprblue}{rgb}{0.21,0.49,0.74}
\usepackage[pagebackref,breaklinks,colorlinks,allcolors=cvprblue]{hyperref}

\newcommand{\notsosmall}{\fontsize{10pt}{12pt}\selectfont}

\definecolor{somegray}{rgb}{0.5, 0.5, 0.5}
\newcommand{\darkgrayed}[1]{\textcolor{somegray}{#1}}
\makeatletter
\newcommand*\titleheader[1]{\gdef\@titleheader{#1}}
\AtBeginDocument{
  \let\st@red@title\@title
  \def\@title{
    \vskip-3em
    \bgroup\normalfont\large\centering\@titleheader\par\egroup
    \vskip1.5em\st@red@title}
}
\makeatother
\titleheader{\darkgrayed{This paper has been accepted for publication at the \\
IEEE/CVF Conference on Computer Vision and Pattern Recognition (CVPR), Nashville, 2025.
\copyright IEEE}}

\title{Stereo Anywhere: Robust Zero-Shot Deep Stereo Matching \\ Even Where Either Stereo or Mono Fail}

\author{Luca Bartolomei$^{*,\dagger}$ \hspace{0.7cm} Fabio Tosi$^\dagger$ \hspace{0.7cm} Matteo Poggi$^{*,\dagger}$  \hspace{0.7cm} Stefano Mattoccia$^{*,\dagger}$ \\
\notsosmall $^*$Advanced Research Center on Electronic System (ARCES) \\ 
\notsosmall $^\dagger$Department of Computer Science and Engineering (DISI) \vspace{-0.1cm}\\
\notsosmall University of Bologna, Italy \\
\small\url{https://stereoanywhere.github.io/}
}

\begin{document}

\twocolumn[{
\renewcommand\twocolumn[1][]{#1}
\maketitle
\vspace{-2.5em}
\centering
    \begin{tabular}{c@{\hskip 1pt}c@{\hskip 4pt}c@{\hskip 4pt}c@{\hskip 4pt}c@{\hskip 4pt}}
        
        & \small RGB
        & \small Depth Anything v2 \cite{depth_anything_v2} 
        & \small RAFT-Stereo \cite{lipson2021raft} 
        & \small Stereo Anywhere (Ours) \\

        \rotatebox[origin=c]{90}{\raisebox{0.08\textwidth}{\parbox[c][0.10\textwidth][c]{0.10\textwidth}{\centering\small Middlebury}}}\hspace{-3.5em} &
        \includegraphics[width=0.18\textwidth]{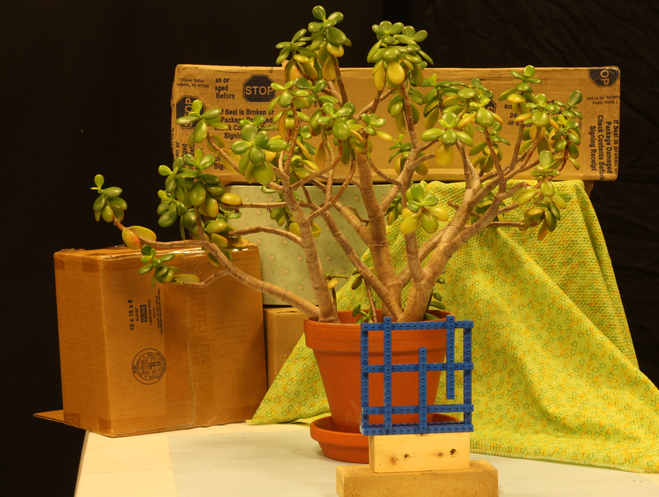} &
        \begin{overpic}[width=0.18\textwidth]{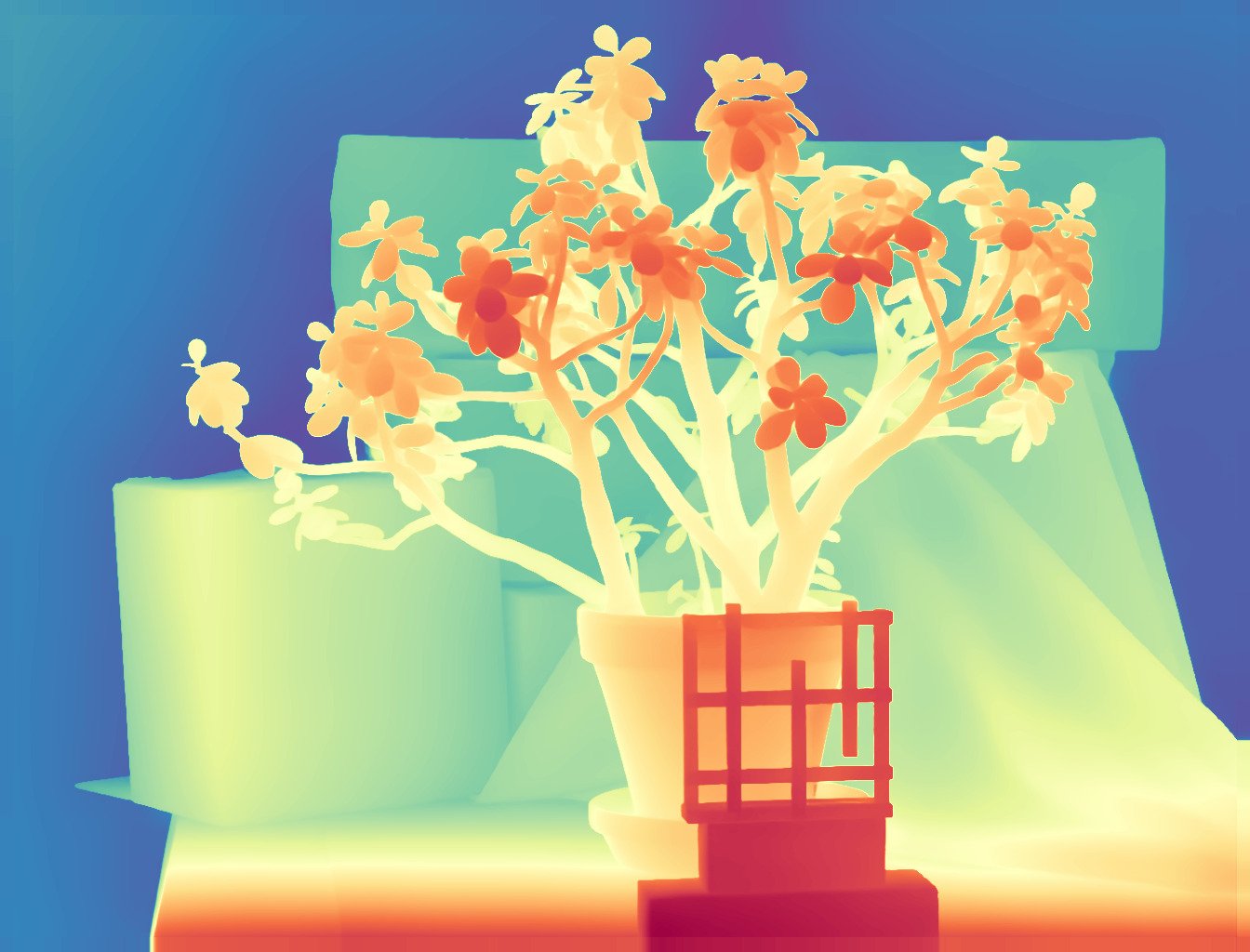}
        \put(85,-5){\Huge\textbf{\color{green}\ding{51}}}
        \end{overpic}&
        \begin{overpic}
        [width=0.18\textwidth]{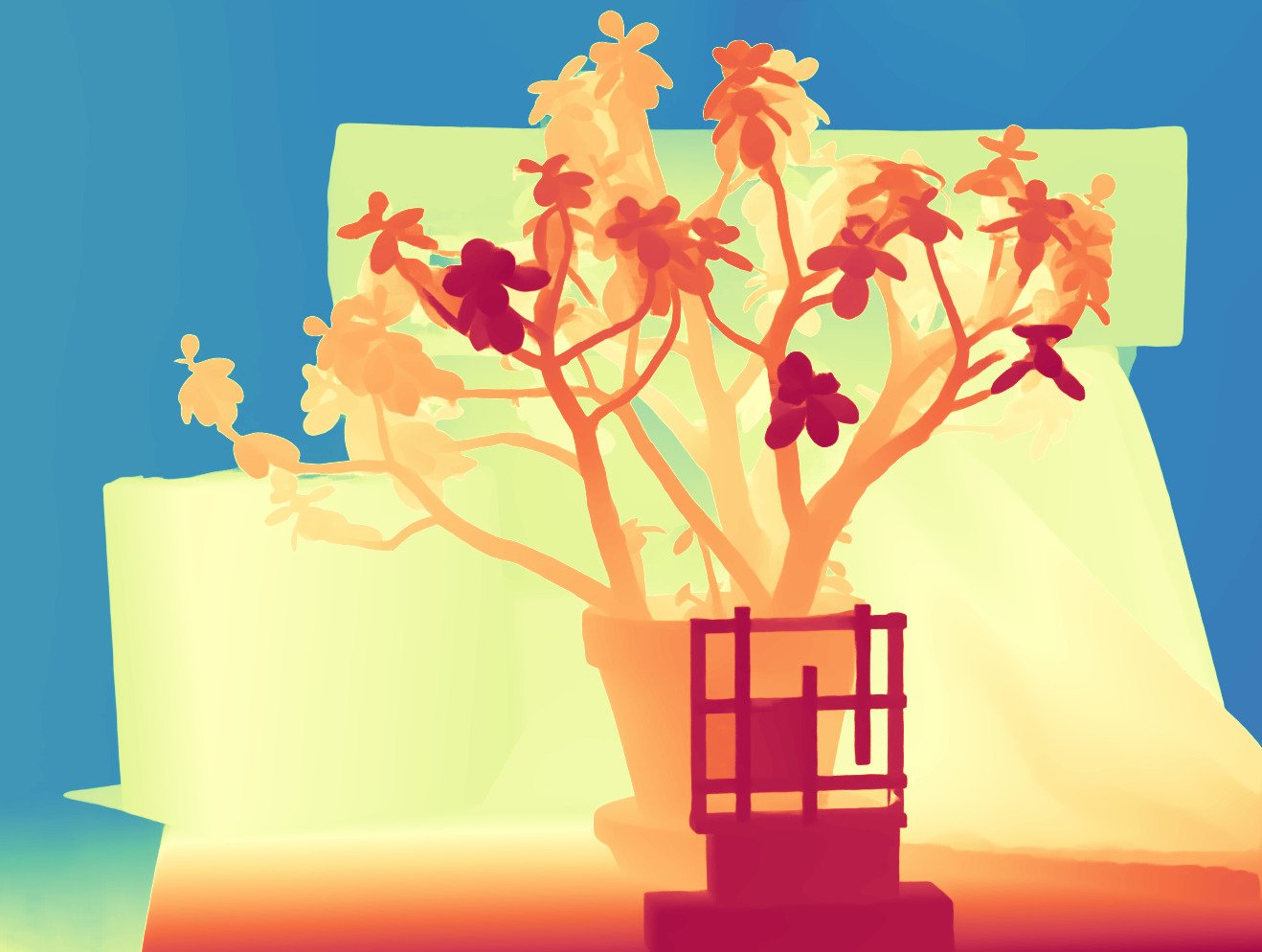}
        \put(85,-5){\Huge\textbf{\color{green}\ding{51}}}
        \end{overpic}&
        \begin{overpic}[width=0.18\textwidth]{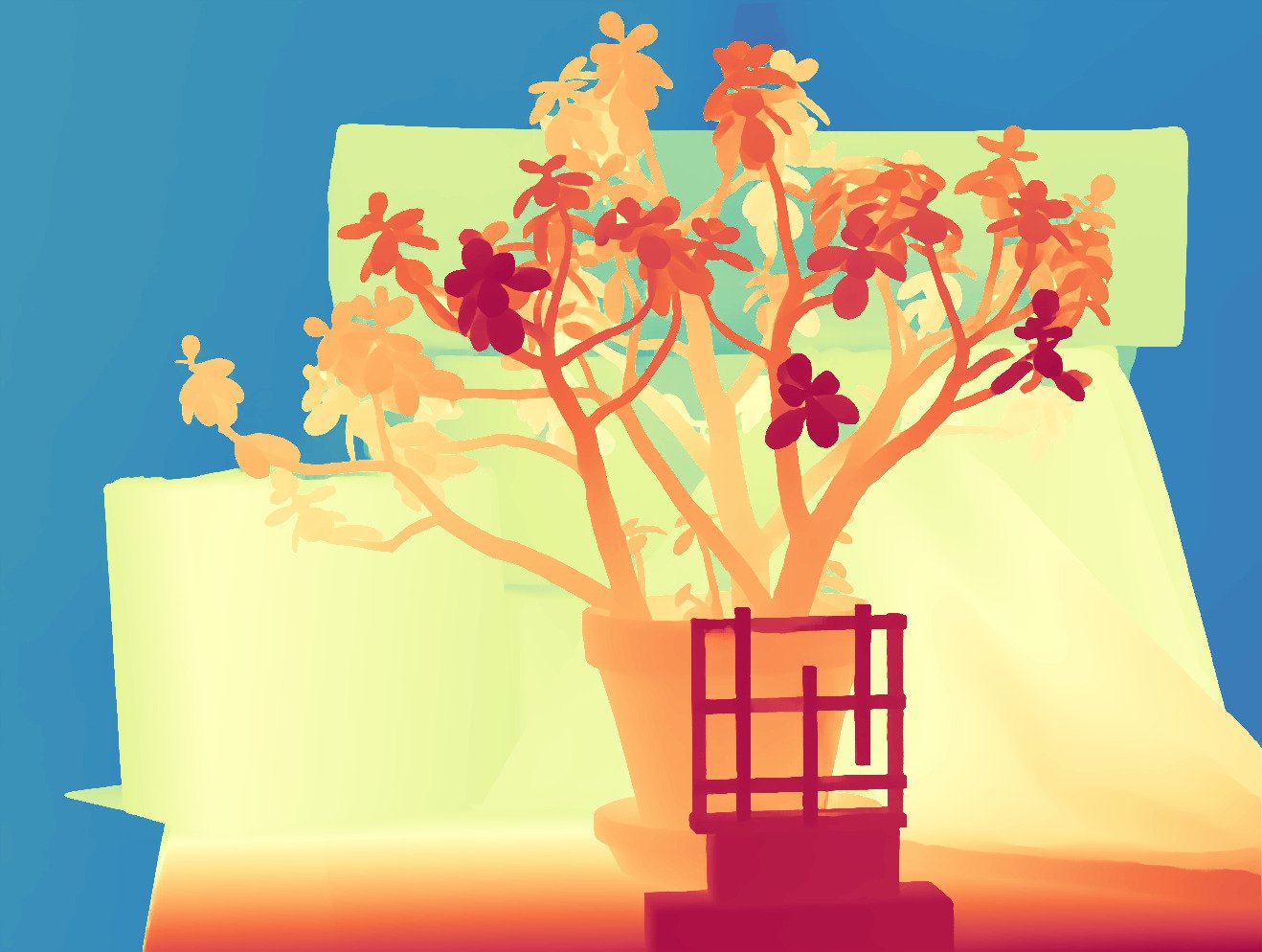}
        \put(85,-5){\Huge\textbf{\color{green}\ding{51}}}
        \end{overpic} \vspace{0.12cm}\\

        \rotatebox[origin=c]{90}{\raisebox{0.08\textwidth}{\parbox[c][0.10\textwidth][c]{0.10\textwidth}{\centering\small Booster }}}\hspace{-3.5em}  &
        \includegraphics[clip,trim=0cm 4cm 0cm 0cm,width=0.18\textwidth]{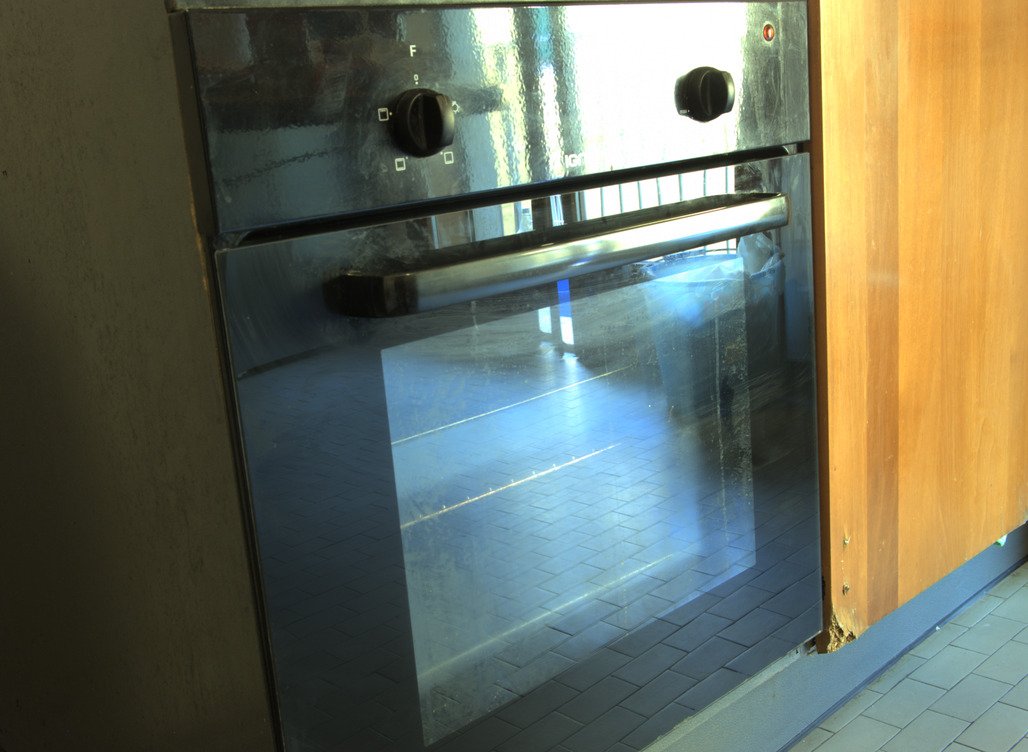} &
        \begin{overpic}[clip,trim=0cm 4cm 0cm 0cm,width=0.18\textwidth]{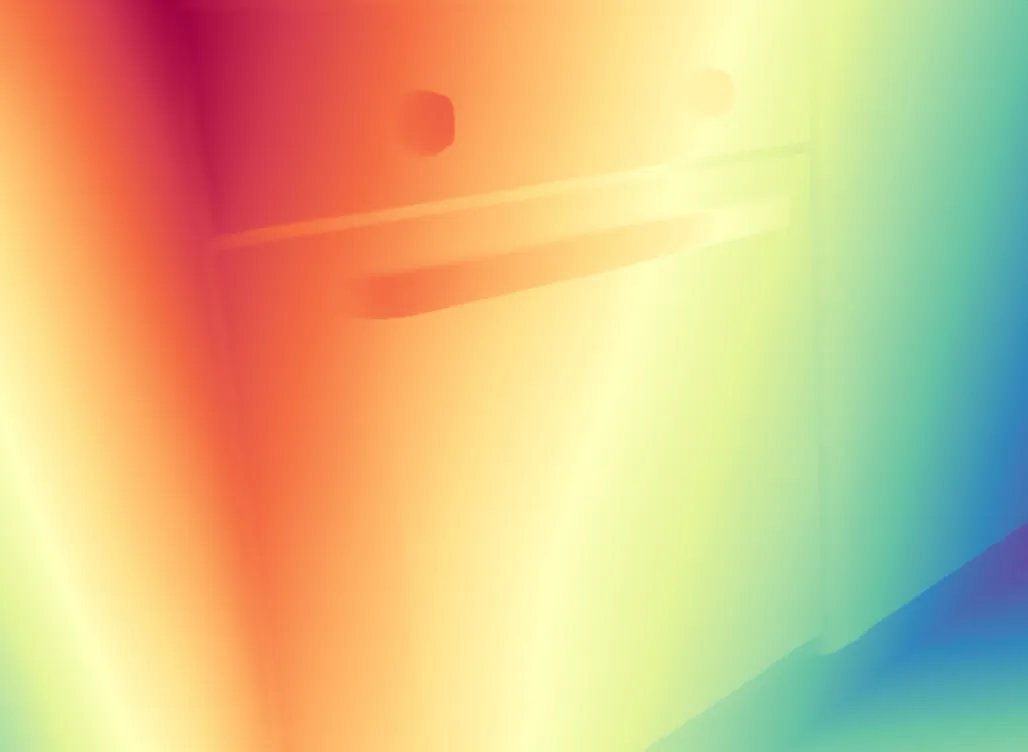}
        \put(85,-5){\Huge\textbf{\color{green}\ding{51}}}
        \end{overpic}&
        \begin{overpic}
        [clip,trim=0cm 4cm 0cm 0cm,width=0.18\textwidth]{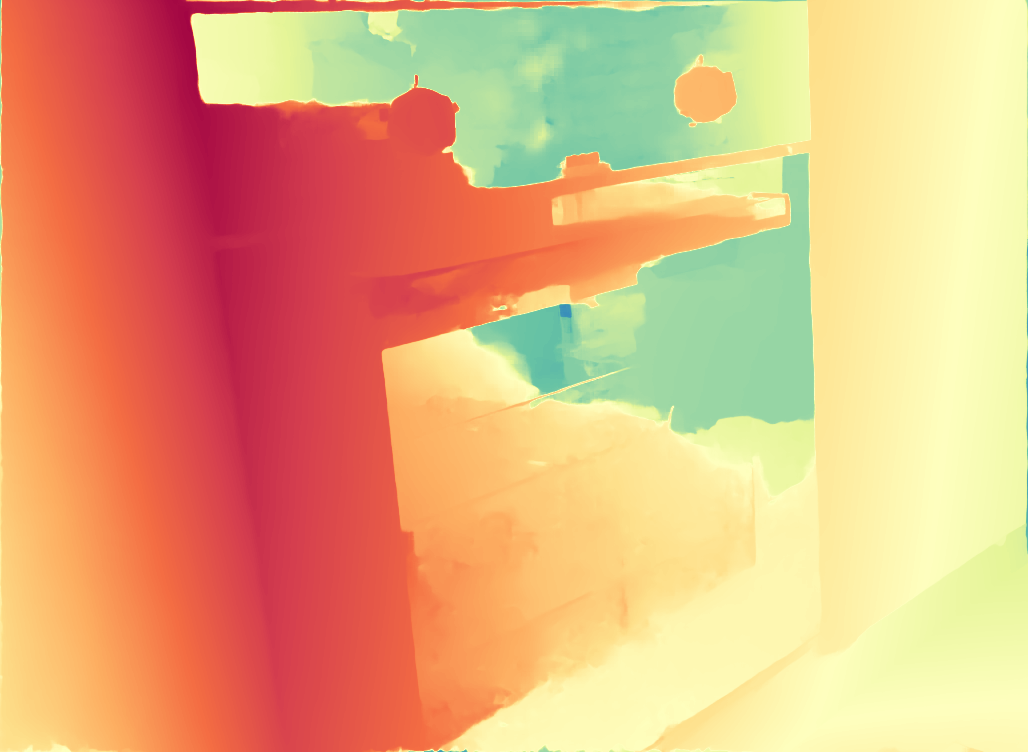}
        \put(85,-5){\Huge\textbf{\color{red}\ding{55}}}
        \end{overpic}&
        \begin{overpic}[clip,trim=0cm 4cm 0cm 0cm,width=0.18\textwidth]{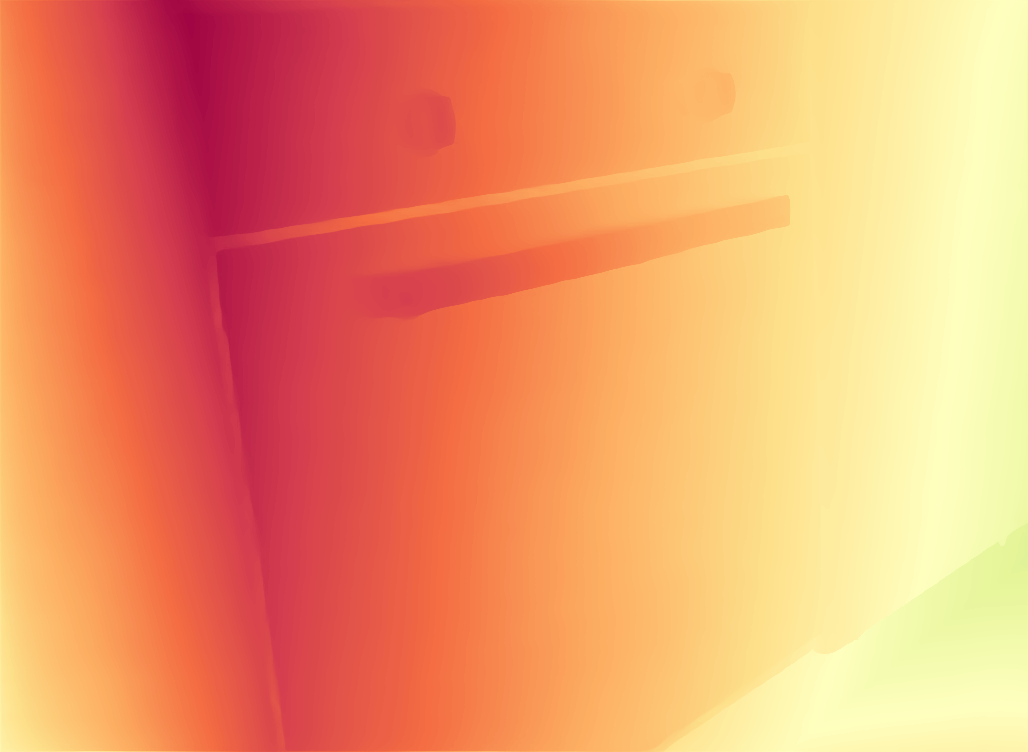}
        \put(85,-5){\Huge\textbf{\color{green}\ding{51}}}
        \end{overpic} \vspace{0.12cm}\\

        \rotatebox[origin=c]{90}{\raisebox{0.08\textwidth}{\parbox[c][0.10\textwidth][c]{0.10\textwidth}{\centering\small MonoTrap }}}\hspace{-3.5em}  &
        \includegraphics[width=0.18\textwidth]{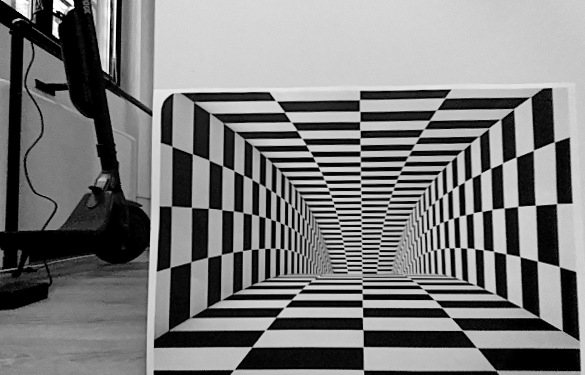} &
        \begin{overpic}[width=0.18\textwidth]{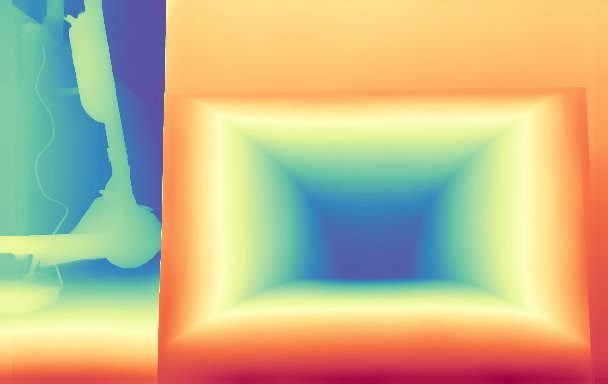}
        \put(85,-5){\Huge\textbf{\color{red}\ding{55}}}
        \end{overpic}&
        \begin{overpic}
        [width=0.18\textwidth]{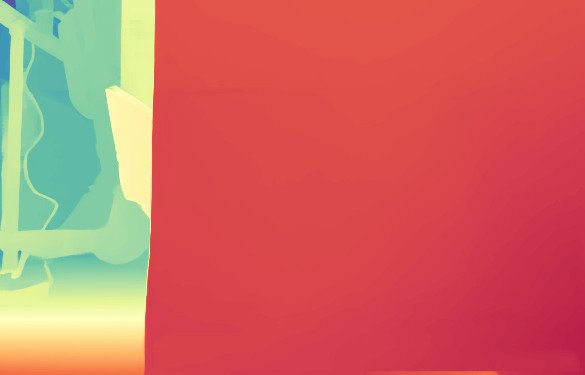}
        \put(85,-5){\Huge\textbf{\color{green}\ding{51}}}
        \end{overpic}&
        \begin{overpic}[width=0.18\textwidth]{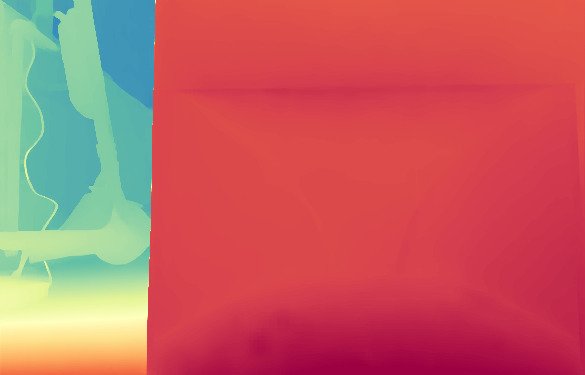}
        \put(85,-5){\Huge\textbf{\color{green}\ding{51}}}
        \end{overpic} \vspace{0.10cm}\\      
    \end{tabular}
   \vspace{-0.4cm}
   \captionof{figure}{\textbf{\method: Combining Monocular and Stereo Strenghts for Robust Depth Estimation.} Our model achieves accurate results on standard conditions (on Middlebury \cite{scharstein2014high}), while effectively handling non-Lambertian surfaces where stereo networks fail (on Booster \cite{zamaramirez2022booster}) and perspective illusions that deceive monocular depth foundation models (on \dataset, our novel dataset).
   }\vspace{0.1cm}
  \label{fig:teaser}  
}]

\begin{abstract}
We introduce Stereo Anywhere, a novel stereo-matching framework that combines geometric constraints with robust priors from monocular depth Vision Foundation Models (VFMs). By elegantly coupling these complementary worlds through a dual-branch architecture, we seamlessly integrate stereo matching with learned contextual cues. Following this design, our framework introduces novel cost volume fusion mechanisms that effectively handle critical challenges such as textureless regions, occlusions, and non-Lambertian surfaces. Through our novel optical illusion dataset, MonoTrap, and extensive evaluation across multiple benchmarks, we demonstrate that our synthetic-only trained model achieves state-of-the-art results in zero-shot generalization, significantly outperforming existing solutions while showing remarkable robustness to challenging cases such as mirrors and transparencies.
\end{abstract}    
\section{Introduction}
\label{sec:intro}

Stereo is a fundamental task that computes depth from a synchronized, rectified image pair by finding pixel correspondences to measure their horizontal offset (\textit{disparity}). Due to its effectiveness and minimal hardware requirements, stereo has become prevalent in numerous applications, from autonomous navigation to augmented reality.

Although in principle single-image depth estimation \cite{arampatzakis2023monocular} requires an even simpler acquisition setup, its ill-posed nature leads to scale ambiguity and perspective illusion issues that stereo methods inherently overcome through well-established geometric multi-view constraints.

However, despite significant advances through deep learning \cite{laga2020survey,poggi2021synergies}, stereo models still face two main challenges: (i) limited generalization across different scenarios, and (ii) critical conditions that hinder matching or proper depth triangulation.
Regarding (i), despite the initial success of synthetic datasets in enabling deep learning for stereo, their limited variety and simplified nature poorly reflect real-world complexity, and the scarcity of real training data further hinders the ability to handle heterogeneous scenarios. As for (ii), large textureless regions common in indoor environments make pixel matching highly ambiguous, while occlusions and non-Lambertian surfaces \cite{zamaramirez2022booster,zamaramirez2024booster,wen2024layeredflow} violate the fundamental assumptions linking pixel correspondences to 3D geometry.

We argue that both challenges are rooted in the underlying limitations of stereo training data. Indeed, while data has scaled up to millions - or even billions - for several computer vision tasks, stereo datasets are still constrained in quantity and variety. 
This is particularly evident for non-Lambertian surfaces, which are severely underrepresented in existing datasets as their material properties prevent reliable depth measurements from active sensors (e.g. LiDAR).

In contrast, single-image depth estimation has recently witnessed a significant scale-up in data availability, reaching the order of \textit{millions} of samples and enabling the emergence of Vision Foundation Models (VFMs) \cite{depth_anything_v1,depth_anything_v2,ke2023repurposing,fu2024geowizard}. Such data abundance has influenced these models in different ways, either through direct training on large-scale depth datasets  \cite{depth_anything_v1,depth_anything_v2} or indirectly by leveraging networks pre-trained on \textit{billions} of images for diverse tasks \cite{ke2023repurposing,fu2024geowizard}. 
Since these models rely on contextual cues for depth estimation, they show better capability in handling textureless regions and non-Lambertian materials \cite{roberts2021,Ramirez_2023_CVPR,Ramirez2024,zamaramirez2024tricky} while being inherently immune to occlusions.
Modern graphics engines have further accelerated this progress, enabling rapid generation of high-quality synthetic data with dense depth annotations. However, although synthetic datasets featuring non-Lambertian surfaces like HyperSim \cite{roberts2021} have proven effective for monocular depth estimation \cite{Ramirez_2023_CVPR,Ramirez2024,zamaramirez2024tricky}, this data abundance has not translated to stereo. Despite efforts in generating stereo pairs via novel view synthesis \cite{Tosi_2023_CVPR,gjerde2024nerf,ling2024self}, available data remains insufficient for robust stereo matching.

In this paper, rather than focusing on costly real-world data collection or generating additional synthetic datasets, we propose to bridge this gap by leveraging existing VFMs for single-view depth estimation.
To this end, we develop a novel dual-branch deep architecture that combines stereo matching principles with monocular depth cues.
Specifically, while one branch of the proposed network constructs a cost volume from learned stereo image features, the other branch processes depth predictions from the VFM on both left and right images to build a second cost volume that incorporates depth priors to guide the disparity estimation process. These complementary signals are then iteratively combined \cite{lipson2021raft}, along with novel augmentation strategies applied to both cost volumes, to predict the final disparity map. Through this design, our network achieves robust performance on challenging cases like textureless regions, occlusions, and non-Lambertian surfaces, while requiring minimal synthetic stereo data. Importantly, while leveraging monocular cues, our approach preserves stereo matching geometric guarantees, effectively handling scenarios where monocular depth estimation typically fails, such as in the presence of perspective illusions. We validate this through our novel dataset of optical illusions, comprising 26 scenes with ground-truth depth maps. 

We dub our framework \textit{\method}, highlighting its ability to overcome the individual limitations of stereo and monocular approaches, as depicted in Fig. \ref{fig:teaser}. To summarize, our main contributions are:

\begin{itemize}
    \item A novel deep stereo architecture leveraging monocular depth VFMs to achieve strong generalization capabilities and robustness to challenging conditions.
    \item Novel data augmentation strategies designed to enhance the robustness of our model to textureless regions and non-Lambertian surfaces. 
    \item A challenging dataset with optical illusion, which is particularly challenging for monocular depth with VFMs. 
    \item Extensive experiments showing \method's superior generalization and robustness to conditions critical for either stereo or monocular approaches.
\end{itemize}

\begin{figure*}[t]
    \centering
    \includegraphics[width=0.98\linewidth]{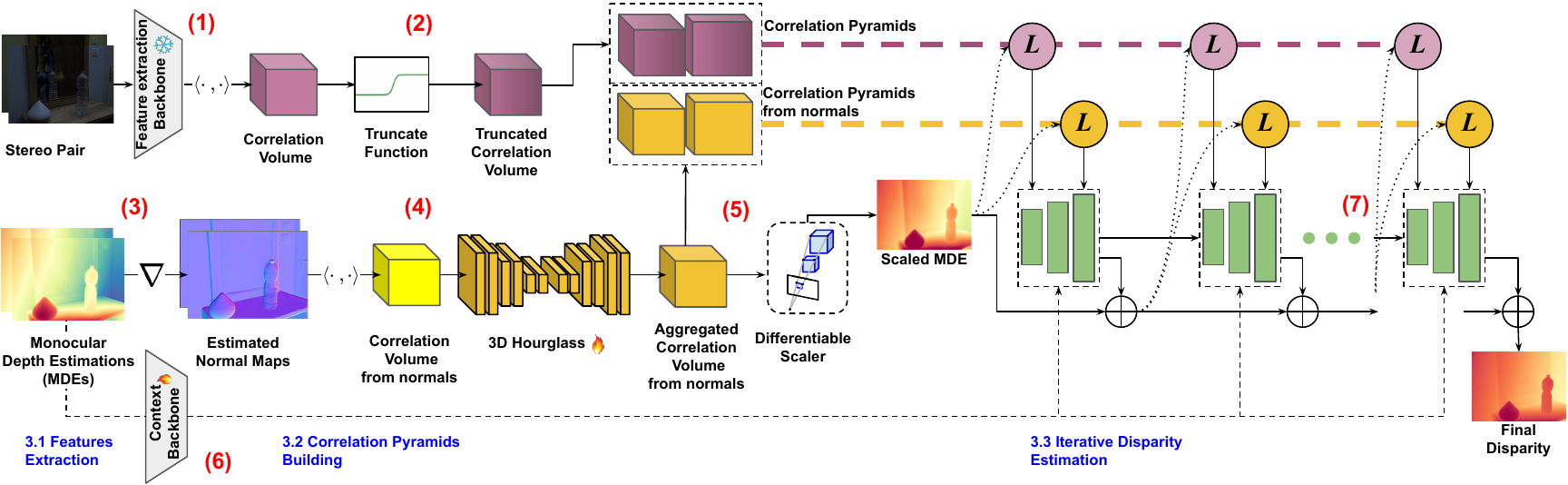}\vspace{-0.2cm}
    \caption{\textbf{\method Architecture.} Given a stereo pair, \textcolor{red}{\bf(1)} a pre-trained backbone is used to extract features and then build a correlation volume. Such a volume is then truncated \textcolor{red}{\bf(2)} to reject matching costs computed for disparity hypotheses being \textit{behind} non-Lambertian surfaces -- glasses and mirrors. On a parallel branch, the two images are processed by a monocular VFM to obtain two depth maps \textcolor{red}{\bf(3)}: these are used to build a second correlation volume from retrieved normals \textcolor{red}{\bf(4)}. This volume is then aggregated through a 3D CNN to predict a new disparity map, used to align the original monocular depth to metric scale through a differentiable scaling module \textcolor{red}{\bf(5)} for it. In parallel, the monocular depth map from left images is processed by another backbone \textcolor{red}{\bf(6)} to extract context features.
    Finally, the two volumes and the context features from monocular depth guide the iterative disparity prediction \textcolor{red}{\bf(7)}.}
    \label{fig:arch}
\end{figure*}

\section{Related Works}
\label{sec:related}

We briefly review the literature relevant to our work.

    \textbf{Deep Stereo Matching.} In the last decade, stereo matching has transitioned from classical hand-crafted algorithms \cite{scharstein2002taxonomy} to deep learning solutions, leading to unprecedented accuracy in depth estimation. Early deep learning efforts focused on replacing individual components of the conventional pipeline \cite{zbontar2015computing,vzbontar2016stereo_MC-CNN,seki2017sgm, spyropoulos2014learning, tosi2024neural}. Since DispNetC \cite{mayer2016large}, end-to-end architectures have evolved into 2D \cite{yin2019hierarchical,liang2018learning_iResNet,song2019edgestereo, yin2019hierarchical} and 3D \cite{kendall2017end_GC-NET,yang2019hierarchical,zhang2019ga,bangunharcana2021correlate,Zeng_2023_ICCV,chang2018pyramid,guo2019group,shen2021cfnet,chen2023learning,shen2022pcw} approaches, processing cost volumes through correlation layers or 3D convolutions respectively. More recent advances, thoroughly reviewed in \cite{poggi2021synergies,laga2020survey,tosi2024survey}, include recurrent architectures for stereo matching~\cite{lipson2021raft,wang2024selective, chen2024mocha, zhao2023high, xu2023iterative, li2022practical, Jing_2023_ICCV, gong2024learning} inspired by RAFT~\cite{teed2020raft}, Transformer-based solutions \cite{Li_2021_ICCV_STTR, guo2022context_CEST, xu2023unifying, Su_2022_CVPR_Chitransformer, croco_v2,lou2023elfnet,zhang2024learning} for capturing long-range dependencies, and fully data-driven MRF models \cite{guan2024neural}. Among them, some methods specifically address temporal consistency in stereo videos \cite{Zhang2023TemporalStereo, Karaev_2023_CVPR, jing2024match, zeng2024temporally}. Domain generalization remains a major challenge, with various approaches proposed including domain-invariant feature learning \cite{zhang2019domaininvariant, Liu_2022_CVPR, Rao_2023_CVPR, Chuah_2022_CVPR, Song_2021_CVPR}, hand-crafted matching costs \cite{cai2020matchingspace, cheng2022revisiting}, integration of additional geometric cues \cite{aleotti2021neural, Pilzer_2023_WACV, tosi2024neural}, and exploitation of sparse depth measurements from active sensors \cite{poggi2019guided, bartolomei2023active, li2024stereo}. In parallel, self-supervised approaches \cite{godard2017unsupervised, Liu_2020_CVPR_Flow2Stereo} have emerged as effective alternatives to supervised learning, even using pseudo-labels from traditional algorithms \cite{tonioni2017unsupervised, aleotti2020reversing} or deploying neural radiance fields \cite{Tosi_2023_CVPR}. Despite the numerous attempts to improve specific aspects through the aforementioned techniques, recent architectures achieve remarkable generalization by combining their architectural advances with the increasing availability of diverse training data, while online adaptation techniques enable further improvements during deployment through self-supervised learning \cite{tonioni2019real, kim2022pointfix, poggi2021continual, Poggi_2024_CVPR}. However, although progress on challenges like over-smoothing \cite{Tosi2021CVPR_SMD,Xu_CVPR_2024_ADL} and visually imbalanced stereo \cite{liu2020visually,Chen_2022_CVPR,aleotti2021neural,tosi2024neural}, handling non-Lambertian surfaces remains particularly challenging due to limited annotated data and complex appearance, with rare works like Depth4ToM \cite{costanzino2023learning} specifically addressing this through semantic guidance. Among all the aforementioned approaches, there have been limited attempts to integrate stereo with monocular cues \cite{Chen_2021_ICCV, aleotti2020reversing, watson2020learning}, mostly in self-supervised settings or through loose coupling between modalities.
    
    \textbf{Monocular Depth Estimation.} Parallel to developments in stereo matching, single-image depth estimation has evolved from hand-crafted features~\cite{Saxena2008} to deep learning methods~\cite{chen2016single, eigen2014depth, laina2016deeper, Ramamonjisoa_2020_CVPR, wang2020cliffnet}, with self-supervised approaches~\cite{godard2017unsupervised, zhou2017unsupervised, mahjourian2018unsupervised, godard2019monodepth2, poggi2018learning, watson2019depthhints, zhao2022monovit}  reframing the task as an image reconstruction problem. This led to multi-task approaches incorporating flow~\cite{zou2018df, yin2018geonet, ranjan2019competitive, tosi2020distilled} and semantics~\cite{zama2019geometry, guizilini2020semantically}, alongside advances in uncertainty estimation~\cite{poggi2020uncertainty, hornauer2022gradient} and dynamic object handling~\cite{klingner2020self, sun2023dynamodepth, moon2023ground}.
    Affine-invariant models~\cite{Ranftl2022, Ranftl2021, Yin2020, Eftekhar2021, wang2024moge} marked a breakthrough in cross-domain generalization, pioneered by MiDaS~\cite{Ranftl2022} and followed by works like DPT~\cite{Ranftl2021} and, more recently, the Depth Anything series \cite{depth_anything_v1}. These approaches used different data sources, from internet photos~\cite{li2018megadepth,Yin2020,Spencer2023c,Spencer2024} to car sensors~\cite{geiger2012we,menze2015object} and RGB-D devices~\cite{Silberman2012, Cho2021}, representing the first generation of VFMs for monocular depth estimation. Recent works have focused on metric depth estimation through camera parameter integration~\cite{Yin2023, hu2024metric3dv2,Guizilini2023}, diffusion models~\cite{Ji2023, Duan2023, Saxena2023, Saxena2023b, ke2023repurposing, fu2024geowizard, he2024lotus}, and temporal consistency~\cite{shao2024learning, hu2024depthcrafter}.
    Moreover, material-aware methods~\cite{costanzino2023learning}, diffusion models~\cite{tosi2024diffusion}, and large-scale synthetic datasets have enabled robust monocular depth estimation for non-Lambertian surfaces~\cite{depth_anything_v2}. Stereo methods, however, still struggle with these surfaces due to limited real-world and synthetic annotated data, affecting generalization. We address this by integrating robust monocular VFMs into a stereo architecture.

    \textbf{Concurrent Works.} Finally, we mention some solutions for stereo \cite{wen2025stereo,cheng2025monster,jiang2025defom} and for multi-view stereo \cite{izquierdo2025mvsanywhere}, developed in parallel with ours and sharing similar rationale.

\section{Method Overview}
\label{sec:method}

Given a rectified stereo pair $\mathbf{I}_L, \mathbf{I}_R \in \mathbb{R}^{3 \times H \times W}$, we first obtain monocular depth estimates (MDEs) $\mathbf{M}_L, \mathbf{M}_R \in \mathbb{R}^{1 \times H \times W}$ using a generic VFM $\phi_M$ for monocular depth estimation. We aim to estimate a disparity map $\mathbf{D}=\phi_S(\mathbf{I}_L, \mathbf{I}_R, \mathbf{M}_L, \mathbf{M}_R)$, incorporating VFM priors to provide accurate results even under challenging conditions, such as texture-less areas, occlusions, and non-Lambertian surfaces. 
At the same time, our stereo network $\phi_S$ is designed to avoid depth estimation errors that could arise from relying solely on contextual cues, which can be ambiguous, like in the presence of visual illusions.

Following recent advances in iterative models \cite{lipson2021raft}, \method comprises three main stages, as shown in \cref{fig:arch}: I) Feature Extraction, II) Correlation Pyramids Building, and III) Iterative Disparity Estimation.

\subsection{Feature Extraction}

Two distinct types of features are extracted \cite{lipson2021raft}: image features and context features -- (1) and (6) in \cref{fig:arch}.
The image features are obtained through a feature encoder processing the stereo pair, yielding feature maps $\mathbf{F}_L, \mathbf{F}_R \in \mathbb{R}^{D \times \frac{H}{4} \times \frac{W}{4}}$, which are used to build a stereo correlation volume at $\frac{1}{4}$ of the original input resolution.
These encoders are initialized with pre-trained weights \cite{lipson2021raft} and the image encoder is kept frozen during training.
For context features, we employ a context encoder with identical architecture to the feature encoder, but processing the monocular depth estimate aligned with the reference image $\mathbf{M}_L$ -- (3) in \cref{fig:arch} -- instead of $\mathbf{I}_L$ to capture strong geometry priors. Accordingly, during training the context encoder is optimized to extract meaningful features from these depth maps.

\subsection{Correlation Pyramids Building}
\label{subsec:corr_pyramids}

As a standard practice in stereo matching, the \textit{cost volume} is the data structure encoding the similarity between pixels across two images. Accordingly, our model utilizes cost volumes—specifically Correlation Pyramids \cite{lipson2021raft}—but in a novel manner.
Indeed, \method constructs two correlation pyramids: a \textit{stereo correlation volume} derived from $\mathbf{I}_L, \mathbf{I}_R$ to encode image similarities, and a \textit{monocular correlation volume} from $\mathbf{M}_L, \mathbf{M}_R$ to encode geometric similarities—(2) and (4) in \cref{fig:arch}.
Unlike the former, the latter remains unaffected by non-Lambertian surfaces, assuming a robust $\phi_M$.

\textbf{Stereo Correlation Volume.} Given $\mathbf{F}_L, \mathbf{F}_R$, we construct a 3D correlation volume $\mathbf{V}_S$ using dot product between feature maps:
\begin{equation}
    (\mathbf{V}_S)_{ijk} = \sum_{h} (\mathbf{F}_L)_{hij} \cdot (\mathbf{F}_R)_{hik}, \ \mathbf{V}_S \in \mathbb{R}^{\frac{H}{4} \times \frac{W}{4} \times \frac{W}{4}}
    \label{eq:dot_corr}
\end{equation}

\textbf{Monocular Correlation Volume.} Given $\mathbf{M}_L, \mathbf{M}_R$, 
we downsample them to 1/4, compute their normals $\nabla_L, \nabla_R$,
and construct a 3D correlation volume $\mathbf{V}_M$ using dot product between normal maps:
\begin{equation}
    (\mathbf{V}_M)_{ijk} = \sum_{h} (\nabla_L)_{hij} \cdot (\nabla_R)_{hik}, \ \mathbf{V}_M \in \mathbb{R}^{\frac{H}{4} \times \frac{W}{4} \times \frac{W}{4}}
    \label{eq:dot_corr_mono}
\end{equation}

Given the absence of texture in $\nabla_L$ and $\nabla_R$, the resulting monocular volume $\mathbf{V}_M$ will be less informative. 
To alleviate this problem we segment $\mathbf{V}_M$ using the relative depth priors from $\mathbf{M}_L$ and $\mathbf{M}_R$: to do so, we generate left and right segmentation masks $\mathcal{M}_L \in \{0,1\}^{\frac{H}{4} \times \frac{W}{4} \times 1}$, $\mathcal{M}_R \in \{0,1\}^{\frac{H}{4} \times 1 \times \frac{W}{4}}$.
We refer the reader to the \textbf{supplementary material} for a detailed description.
Given the segmentation masks, we can generate masked volumes as:
\begin{equation}
    ({\mathbf{V}_M}^n)_{ijk} = ({\mathcal{M}_L}^n)_{ij} \cdot ({\mathcal{M}_R}^n)_{ik} \cdot (\mathbf{V}_M)_{ijk}
    \label{eq:vol_masking}
\end{equation}
Next, we insert a 3D Convolutional Regularization module $\phi_A$ to aggregate ${\mathbf{V}_M}^n$, resulting in ${\mathbf{V}'}_M=\phi_A({\mathbf{V}_M}^1,\dots,{\mathbf{V}_M}^{N},\mathbf{M}_L,\mathbf{M}_R)$, with $N=8$. The architecture of $\phi_A$ follows the one in \cite{xu2023iterative}, with a simple permutation to match the structure of the correlation volumes.
We propose an adapted version of CoEx \cite{bangunharcana2021correlate} correlation volume excitation that exploits both views. 
The resulting feature volumes ${\mathbf{V}'}_M \in \mathbb{R}^{F \times \frac{H}{4} \times \frac{W}{4} \times \frac{W}{4}}$ are fed to two different shallow 3D conv layers $\phi_D$ and $\phi_C$ to obtain two aggregated volumes $\mathbf{V}^D_M = \phi_D({\mathbf{V}'}_M)$ and $\mathbf{V}^C_M = \phi_C({\mathbf{V}'}_M)$ with $\mathbf{V}^D_M,\mathbf{V}^C_M \in \mathbb{R}^{\frac{H}{4} \times \frac{W}{4} \times \frac{W}{4}}$.

\textbf{Differentiable Monocular Scaling.} Volume $\mathbf{V}^D_M$ will be used not only as a monocular guide for the iterative refinement unit but also to estimate the coarse disparity maps $\hat{\mathbf{D}}_L$ $\hat{\mathbf{D}}_R$, while $\mathbf{V}^C_M$ is used to estimate confidence maps $\hat{\mathbf{C}}_L$ $\hat{\mathbf{C}}_R$. These maps are then used to scale both $\mathbf{M}_L$ and $\mathbf{M}_R$ -- (5) in \cref{fig:arch}.
To estimate left disparity from a correlation volume, we first perform a \textit{softargmax} on the last $W$ dimension of $\mathbf{V}^D_M$ to extract the correlated pixel x-coordinate. Then, given the relationship between left disparity and correlation $d_L=j_L-j_R$, we obtain a coarse disparity map $\hat{\mathbf{D}}_L$: 
\begin{equation} 
    (\hat{\mathbf{D}}_L)_{ij} = j - \left(\text{softargmax}_L(\mathbf{V}^D_M)\right)_{ij}
    \label{eq:softmax_left}
\end{equation}
Similarly, we estimate $\hat{\mathbf{D}}_R$ from $\mathbf{V}^D_M$. We refer the reader to the supplementary for details.
We also estimate a pair of confidence maps $\hat{\mathbf{C}}_L, \hat{\mathbf{C}}_R \in [0,1]^{H \times W}$ to classify outliers and perform robust scaling.
Inspired by information entropy, we measure the \textit{chaos} within correlation curves: clear monomodal-like cost curves—those with low entropy—are reliable, while \textit{chaotic} curves with high entropy indicate uncertainty.
To estimate the left confidence map, we perform a \textit{softmax} operation on the last $W$ dimension of $\mathbf{V}^C_M$, then $\hat{\mathbf{C}}_L$ is obtained as follows:
\begin{equation}
    (\hat{\mathbf{C}}_L)_{ij} = 1  + \frac{\sum_{d}^{\frac{W}{4}} \frac{e^{(\mathbf{V}^C_M)_{ijd}}}{\sum_{f}^{\frac{W}{4}} e^{(\mathbf{V}^C_M)_{ijf}}} \cdot \log_2 \left( \frac{e^{(\mathbf{V}^C_M)_{ijd}}}{\sum_{f}^{\frac{W}{4}} e^{(\mathbf{V}^C_M)_{ijf}}} \right)}{\log_2(\frac{W}{4})}
    \label{eq:confidence_left}
\end{equation}
In the same way, we estimate $\hat{\mathbf{C}}_R$.
To further reduce outliers, we mask out occluded pixels from $\hat{\mathbf{C}}_L$ and $\hat{\mathbf{C}}_R$ using a \textit{SoftLRC} operator -- see the \textbf{supplementary material} for details.
Finally, we estimate the scale $\hat{s}$ and shift $\hat{t}$ using a differentiable weighted least-square approach:
\begin{equation}
    \min_{\hat{s}, \hat{t}} \sum_{}^{L,R} \left\lVert \sqrt{\hat{\mathbf{C}}}\odot\left[\left(\hat{s}\mathbf{M} + \hat{t}\right)  - \hat{\mathbf{D}} \right] \right\rVert_F 
    \label{eq:scale_shift}
\end{equation}
where $\lVert\cdot\rVert_F$ denotes the Frobenius norm.
Using the scaling coefficients, we obtain two disparity maps $\hat{\mathbf{M}}_L$, $\hat{\mathbf{M}}_R$:
\begin{equation}
    \hat{\mathbf{M}}_L = \hat{s}\mathbf{M}_L + \hat{t},\ \hat{\mathbf{M}}_R = \hat{s}\mathbf{M}_R + \hat{t}
    \label{eq:scaling_op}
\end{equation}
It is crucial to optimize both left and right scaling jointly to obtain consistency between $\hat{\mathbf{M}}_L$ and $\hat{\mathbf{M}}_R$.

\begin{figure*}[t]
    \centering
    \renewcommand{\tabcolsep}{1pt}
    \hspace*{-0.5cm}\begin{tabular}{ccccccc}
    \small Image
    & \small Ground-Truth
    & \small Depth Anything v2 \cite{depth_anything_v2} & 
    & \small Image
    & \small Ground-Truth
    & \small Depth Anything v2 \cite{depth_anything_v2} \\
    \includegraphics[width=0.16\textwidth]{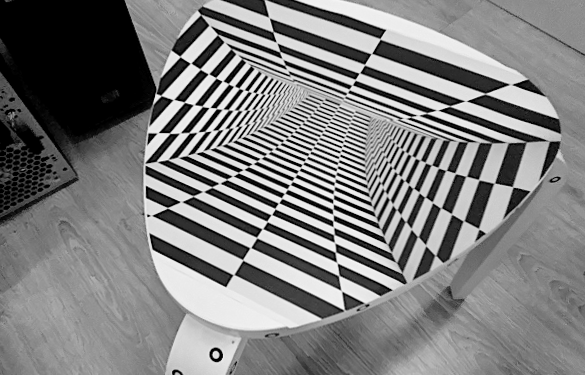}
    &\includegraphics[width=0.16\textwidth]{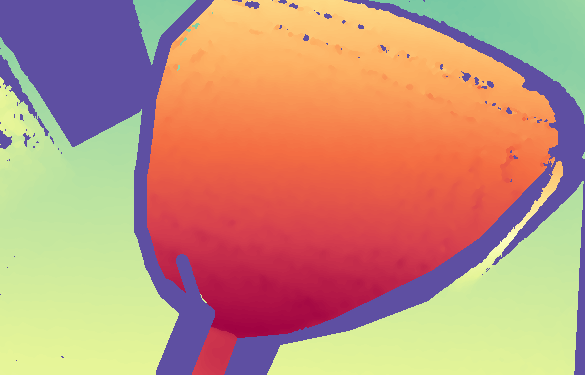}
    &\includegraphics[width=0.16\textwidth]{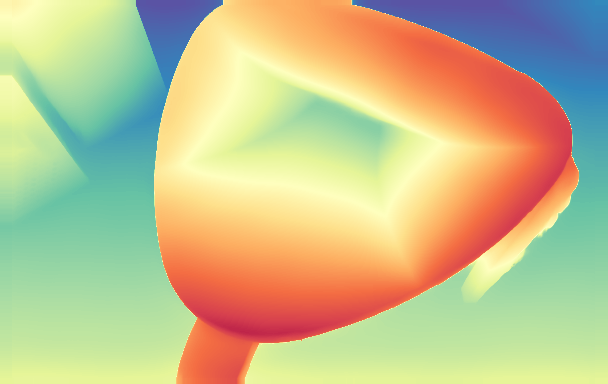}
    & \hspace{0.1cm} &

    \includegraphics[width=0.16\textwidth]{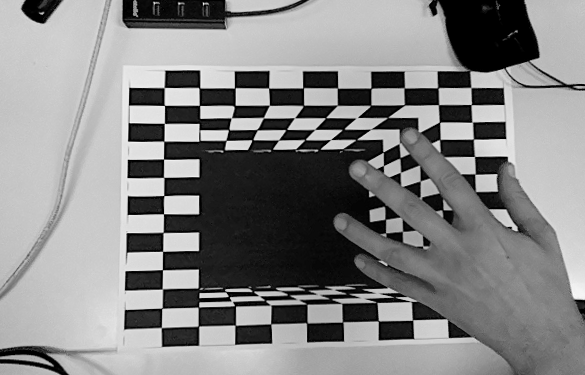}
    &\includegraphics[width=0.16\textwidth]{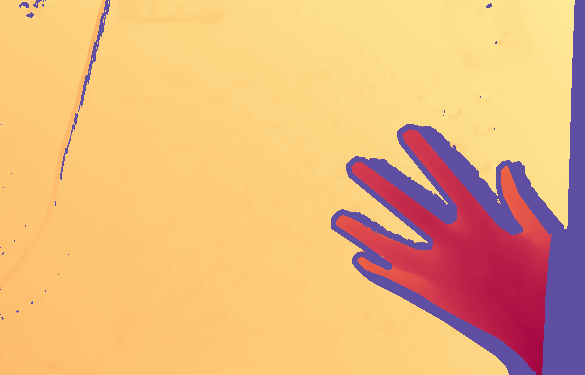}
    &\includegraphics[width=0.16\textwidth]{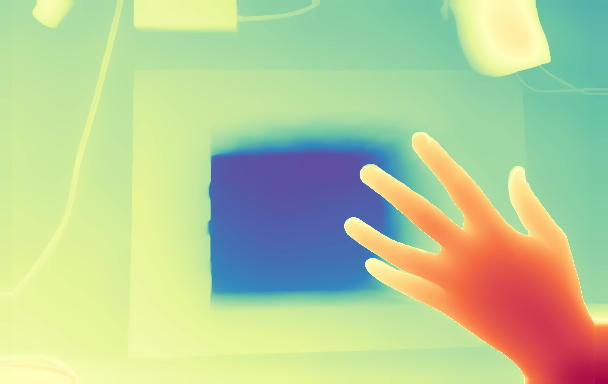}
    \\
    \end{tabular}\vspace{-0.2cm}
    \caption{\textbf{Samples from \dataset Dataset.} We report two scenes featured in our dataset, showing the left image, the ground-truth depth, and the predictions by Depth Anything v2 \cite{depth_anything_v2}, highlighting how it fails in the presence of visual illusions.}\vspace{-0.3cm}
    \label{fig:monotrap}
\end{figure*}

\textbf{Volume Augmentations.} Unfortunately, \method cannot properly learn when to choose stereo or mono information from \cite{mayer2016large} alone.
Hence, we propose three volume augmentations and a monocular augmentation to overcome this issue: 1) \textit{Volume Rolling}: we randomly apply a rolling operation to the last $W$ dimension of ${\mathbf{V}^D}_M$ or ${\mathbf{V}}_S$; 2) \textit{Volume Noising}: we apply random noise sampled from the interval $[0,1)$ using a uniform distribution; 3) \textit{Volume Zeroing}: we apply a Gaussian-like curve with the peak where disparity equals zero. Furthermore, 
we randomly substitute the monocular depth with ground truth normalized between $[0,1]$ as an additional augmentation.
We apply only one volume augmentation to ${\mathbf{V}^D}_M$ or ${\mathbf{V}}_S$ and only for a section of the volume, randomly selecting an $\mathcal{M}_L^n$ mask.

\textbf{Volume Truncation.} To further help \method to handle mirror surfaces, we introduce a hand-crafted volume truncation operation on ${\mathbf{V}}_S$. Firstly, we extract left confidence $\mathbf{C}_M=\text{softLRC}_L(\hat{\mathbf{M}}_L, \hat{\mathbf{M}}_R)$ to classify reliable monocular predictions. Then, we create a truncate mask $\mathbf{T} \in [0,1]^{\frac{H}{4} \times \frac{W}{4}}$ using the following logic condition: $(\mathbf{T})_{ij}=\left[\left((\hat{\mathbf{M}}_L)_{ij} >(\hat{\mathbf{D}}_L)_{ij}\right) \land (\mathbf{C}_M)_{ij} \right] \lor \left[ (\mathbf{C}_M)_{ij} \land \neg(\hat{\mathbf{C}}_L)_{ij} \right]$.
We implement this logic using fuzzy operators (more details in the \textbf{supplementary material}).
The rationale is that stereo predicts farther depths on mirror surfaces: the mirror is perceived as a window into a new environment, specular to the real one.
Finally, for values of $\mathbf{T}>T_\text{m}=0.98$, we truncate ${\mathbf{V}_S}$ using a sigmoid curve centered at the correlation value predicted by $\hat{\mathbf{M}}_L$ -- \ie, the real disparity of mirror surfaces -- preserving only the stereo correlation curve not ``piercing" mirrors.

\subsection{Iterative Disparity Estimation}
We aim to estimate a series of refined disparity maps $\{\mathbf{D}^1=\hat{\mathbf{M}}_L, \mathbf{D}^2,\dots\,\mathbf{D}^l,\dots\}$ exploiting the guidance from both stereo and mono branches. 
Starting from the Multi-GRU update operator by \cite{lipson2021raft}, we introduce a second lookup operator that extracts correlation features $\mathbf{G}_M$ from the additional volume $\mathbf{V}^D_M$ -- (7) in \cref{fig:arch}.
The two sets of correlation features from $\mathbf{G}_S$ and $\mathbf{G}_M$ are processed by the same two-layer encoder and concatenated with features derived from the current disparity estimation $\mathbf{D}^l$. This concatenation is further processed by a 2D conv layer, and then by the ConvGRU operator.
We inherit the convex upsampling module \cite{lipson2021raft} to upsample final disparity to full resolution.

\subsection{Training Supervision}
We supervise the iterative module using the well-known L1 loss with exponentially increasing weights \cite{lipson2021raft}, then $\hat{\mathbf{D}}_L$, $\hat{\mathbf{D}}_R$, $\hat{\mathbf{M}}_L$ and $\hat{\mathbf{M}}_R$ using the L1 loss, 
finally $\hat{\mathbf{C}}_L$ and $\hat{\mathbf{C}}_R$ using the Binary Cross Entropy loss.
We invite the reader to read the \textbf{supplementary material} for additional details.

\begin{table*}[t]
\centering
\renewcommand{\tabcolsep}{12pt}
\scalebox{0.72}{
\begin{tabular}{|ll|rrrrr|rrrr|}
\multicolumn{2}{c}{} & \multicolumn{5}{c}{Booster (Q)} & \multicolumn{4}{c}{Middlebury 2014 (H)} \\
\hline
 & \multirow{2}{*}{Experiment} & \multicolumn{4}{c}{bad} & Avg. & \multicolumn{3}{c}{bad $>2$} & Avg. \\
 & & $>2$ & $>4$ & $>6$ & $>8$ & (px) & All & Noc & Occ & (px) \\
\hline\hline

(A) & Baseline \cite{lipson2021raft}
& 17.84 & 13.06 & 10.76 & 9.24 & 3.59
& 11.15 & 8.06 & 29.06 & 1.55 \\
\hline\hline
(B) & (A) + Monocular Context w/o re-train
& 15.85 & 10.98 & 8.89 & 7.69 & 3.05
& 14.96 & 11.70 & 34.38 & 2.82 \\
(C) & (A) + Monocular Context w/ re-train
& \trd 14.94 & \trd 10.40 & \trd 8.61 & \trd 7.63 & \trd 3.03
& \trd 9.62 & \trd 6.98 & \trd 25.39 & \trd 1.13 \\
(D) & (C) + Normals Correlation Volume / Scaled Depth
& \snd 11.33 & \snd 6.88 & \snd 5.32 & \snd 4.59 & \snd 1.87
& \snd 7.67 & \snd 5.24 & \snd 21.51 & \snd 0.96 \\
(E) & (D) + Volume augmentation / truncation
&\fst 9.01 &\fst 5.40 &\fst 4.12 &\fst 3.34 &\fst 1.21
& \fst 6.96 &\fst 4.75 &\fst 20.34 &\fst 0.94 \\
\hline
\end{tabular}}\vspace{-0.3cm}
\caption{\textbf{Ablation Studies.} We measure the impact of different design strategies. Networks trained on SceneFlow \cite{mayer2016large}.
}\vspace{-0.3cm}
\label{tab:ablation}
\end{table*}

\section{The \dataset Dataset}

Monocular depth estimation is known for possibly failing in the presence of perspective illusions.
The reader may wonder how \method would behave in such cases: would it blindly trust the monocular VFM or rely on the stereo geometric principles to maintain robustness?

To answer these questions, we introduce MonoTrap, a novel stereo dataset specifically designed to challenge monocular depth estimation. Our dataset comprises 26 scenes featuring perspective illusions, captured with a calibrated stereo setup and annotated with ground-truth depth from an Intel Realsense L515 LiDAR.
The scenes contain carefully designed planar patterns that create visual illusions, such as apparent holes in walls or floors and simulated transparent surfaces that reveal content behind them. 
Figure \ref{fig:monotrap} shows examples from our dataset that illustrate how these visual illusions easily fool monocular methods.

\begin{table*}
\centering
\renewcommand{\tabcolsep}{6pt}
\scalebox{0.62}{
\begin{tabular}{|l||rrrr|rrrr|rrrr|rrrr|rrrr|}
\multicolumn{1}{c}{} & \multicolumn{4}{c}{Middlebury 2014 (H)} & \multicolumn{4}{c}{Middlebury 2021} & \multicolumn{4}{c}{ETH3D} & \multicolumn{4}{c}{KITTI 2012} & \multicolumn{4}{c}{KITTI 2015} \\ 
\hline
 \multirow{2}{*}{Model} & \multicolumn{3}{c}{bad $>2$} & Avg. & \multicolumn{3}{c}{bad $>2$} & Avg. & \multicolumn{3}{c}{bad $>1$} & Avg. & \multicolumn{3}{c}{bad $>3$} & Avg. & \multicolumn{3}{c}{bad $>3$} & Avg. \\
  & All & Noc & Occ & (px) & All & Noc & Occ & (px) & All & Noc & Occ & (px) & All & Noc & Occ & (px) & All & Noc & Occ & (px) \\
\hline\hline
{RAFT-Stereo} \cite{lipson2021raft} 
& 11.15 & 8.06 & 29.06 & \trd 1.55 
& 12.05 & 9.38 & 37.89 & 1.81 
&\snd 2.59 &\snd 2.24 & \snd 8.78 &\snd 0.25 
& 4.80 & 4.23 & \trd 29.21 & \snd 0.89 
& 5.44 & 5.21 & 14.09 & \snd 1.16 \\

{PSMNet} \cite{chang2018pyramid} 
& 18.79 & 13.80 & 53.22 & 4.63 
& 23.67 & 20.61 & 53.75 & 5.70 
& 19.75 & 18.62 & 42.05 & 0.94 
& 6.73 & 5.81 & 46.24 & 1.22 
& 6.78 & 6.40 & 24.85 & 1.38 \\
{GMStereo} \cite{xu2023unifying} 
& 15.63 & 10.98 & 46.04 & 1.87 
& 25.43 & 22.43 & 54.70 & 2.86 
& 6.22 & 5.58 & 19.97 & 0.42 
& 5.68 & 4.87 & 38.84 & 1.10 
& 5.72 & 5.44 & 17.33 & \trd 1.21 \\
{ELFNet} \cite{lou2023elfnet} 
& 24.48 & 16.94 & 77.06 & 8.61 
& 27.08 & 21.77 & 85.56 & 11.01 
& 25.61 & 24.50 & 46.06 & 5.65 
& 10.52 & 8.67 & 88.21 & 2.30 
& 9.61 & 8.22 & 85.64 & 2.16 \\
{PCVNet} \cite{Zeng_2023_ICCV} 
& 16.79 & 13.54 & 35.66 & 2.96 
& 12.92 & 10.19 & 40.23 & 2.18 
& \trd 4.24 & \trd 3.61 & 14.01 & 0.41 
& \snd 4.44 & \snd 3.92 & \snd 27.70 & \snd 0.89 
& \snd 5.08 & \snd 4.88 & \trd 13.72 & 1.24 \\
{DLNR} \cite{zhao2023high} 
& \snd 9.46 & \snd 6.20 & 28.75 & \snd 1.45 
&\snd 8.44 &\snd 5.88 & \snd 32.71 & \snd 1.24 
& 23.12 & 22.94 & 26.93 & 9.89 
& 9.45 & 8.83 & 36.75 & 1.59 
& 15.74 & 15.41 & 34.32 & 2.83 \\
{Selective-RAFT} \cite{wang2024selective} 
& 12.05 & 9.46 & \trd 27.42 & 2.35 
& 15.69 & 13.86 & 36.32 & 5.92 
& 4.36 & 3.81 & \trd 10.23 & \trd 0.34 
& 5.71 & 5.16 & 30.54 & 1.08 
& 6.50 & 6.22 & 18.44 & 1.27 \\
{Selective-IGEV} \cite{wang2024selective} 
& 9.98 & 7.09 & 27.62 & 1.60 
& \trd 8.89 & \trd 6.34 & \trd 32.88 & 1.60 
& 6.42 & 5.71 & 18.71 & 1.73 
& 6.22 & 5.54 & 34.78 & 1.09 
& 5.87 & 5.66 & 14.99 & 1.42 \\
{IGEV-Stereo} \cite{xu2023iterative} 
& \trd 9.91 & \trd 7.08 & \snd 26.26 & 1.84 
& 9.15 & 6.43 & 34.88 & \trd 1.53 
& 4.30 & 3.86 & 12.65 & 0.38 
& 5.65 & 4.43 & 33.38 & 1.03 
& 5.87 & 5.13 & 14.31 & 1.34 \\
{NMRF} \cite{guan2024neural} 
& 14.08 & 10.87 & 34.62 & 2.91 
& 23.36 & 21.69 & 42.51 & 8.57 
& 4.34 & 3.66 & 17.15 & 0.42 
& \trd 4.62 & \trd 4.05 & 30.65 & \trd 0.92 
& \trd 5.24 & \trd 5.07 & \snd 12.28 & \snd 1.16 \\
{\bf \method{} (ours)} 
&\fst 6.96 &\fst 4.75 &\fst 20.34 &\fst 0.94
& \fst 7.97 & \fst 5.71 &\fst 29.52 &\fst 1.08 
& \fst 1.66 & \fst 1.43 &\fst 5.29 & \fst 0.24 
&\fst 3.90 &\fst 3.52 &\fst 21.65 &\fst 0.83 
&\fst 3.93 &\fst 3.79 &\fst 11.01 &\fst 0.97 \\

\hline
\end{tabular}}\vspace{-0.3cm}
\caption{\textbf{Zero-shot Generalization.} Comparison with state-of-the-art deep stereo models. Networks trained on SceneFlow \cite{mayer2016large}. 
}\vspace{-0.3cm}
\label{tab:roundtable1}
\end{table*}

\begin{figure*}[t]
    \centering
    \renewcommand{\tabcolsep}{1pt}
    \scalebox{0.9}{
    \begin{tabular}{ccccccc}
        & \small RGB &
        \small RAFT-Stereo \cite{lipson2021raft} &
        \small DLNR \cite{zhao2023high} &
        \small NMRF \cite{guan2024neural} &
        \small Selective-IGEV \cite{wang2024selective} &
        \method \\
        \hspace{-3.5em}\rotatebox[origin=c]{90}{\raisebox{0.08\textwidth}{\parbox[c][0.10\textwidth][c]{0.10\textwidth}{\small KITTI 15}}}\hspace{-3.5em} &\includegraphics[clip,trim=3cm 0cm 3cm 0cm, width=0.16\textwidth]{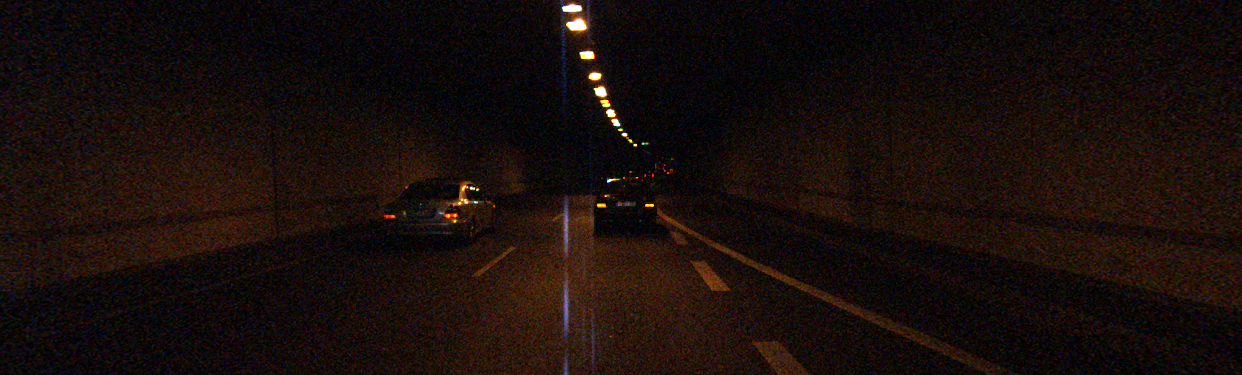} & 
        \includegraphics[clip,trim=12.5cm 0cm 12.5cm 0cm, width=0.16\textwidth]{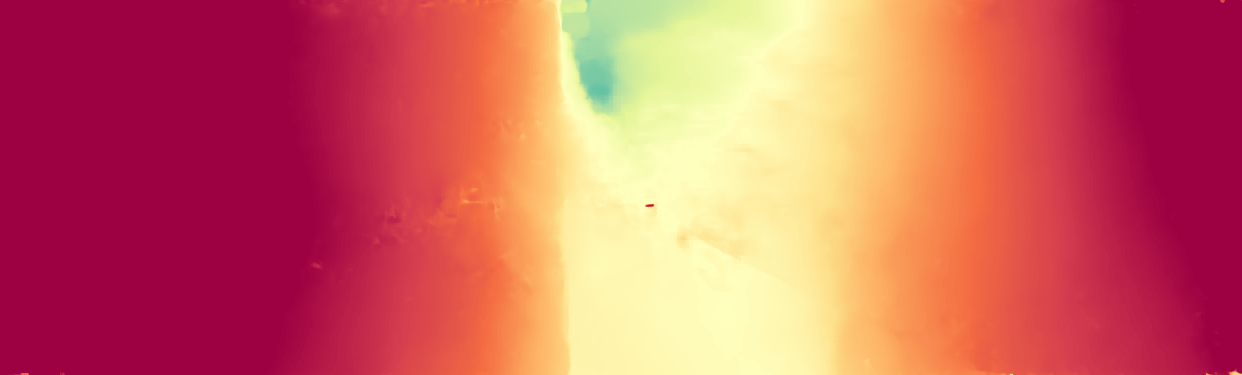} &
        \includegraphics[clip,trim=12.5cm 0cm 12.5cm 0cm, width=0.16\textwidth]{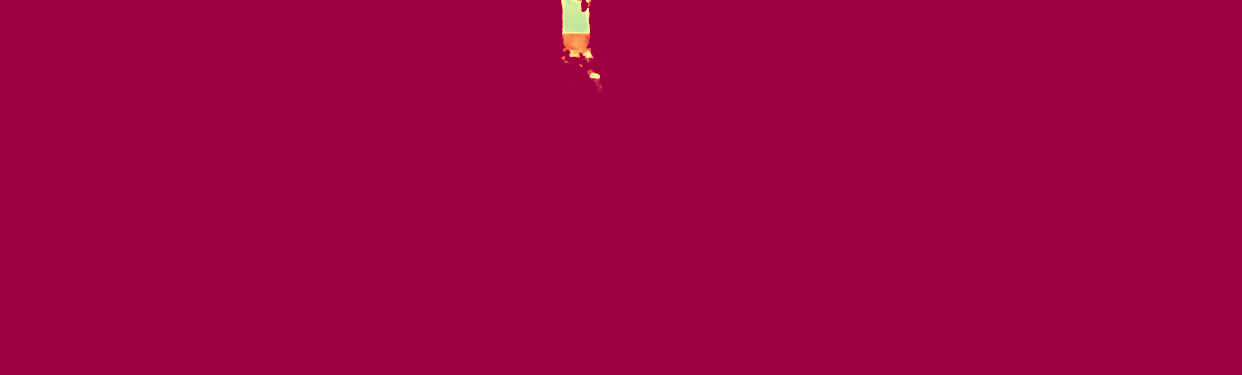} &
        \includegraphics[clip,trim=12.5cm 0cm 12.5cm 0cm, width=0.16\textwidth]{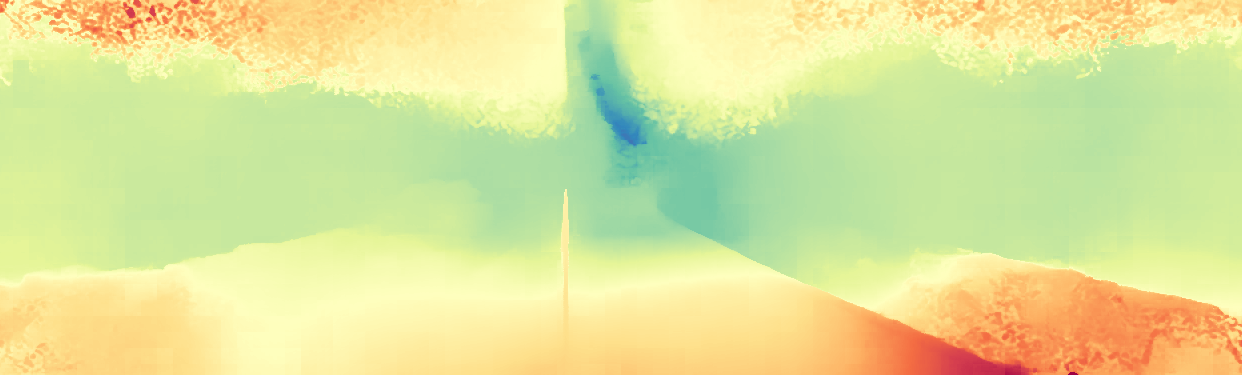} &
        \includegraphics[clip,trim=12.5cm 0cm 12.5cm 0cm, width=0.16\textwidth]{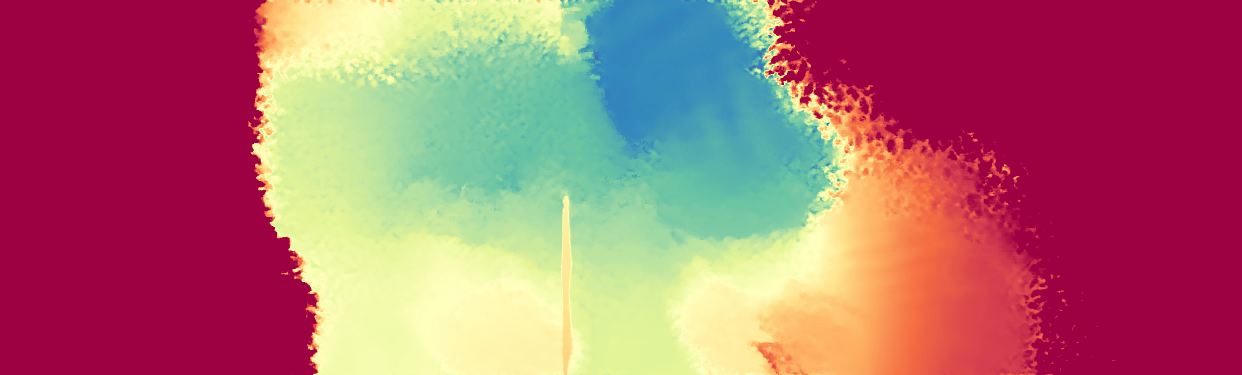} &
        \includegraphics[clip,trim=12.5cm 0cm 12.5cm 0cm, width=0.16\textwidth]{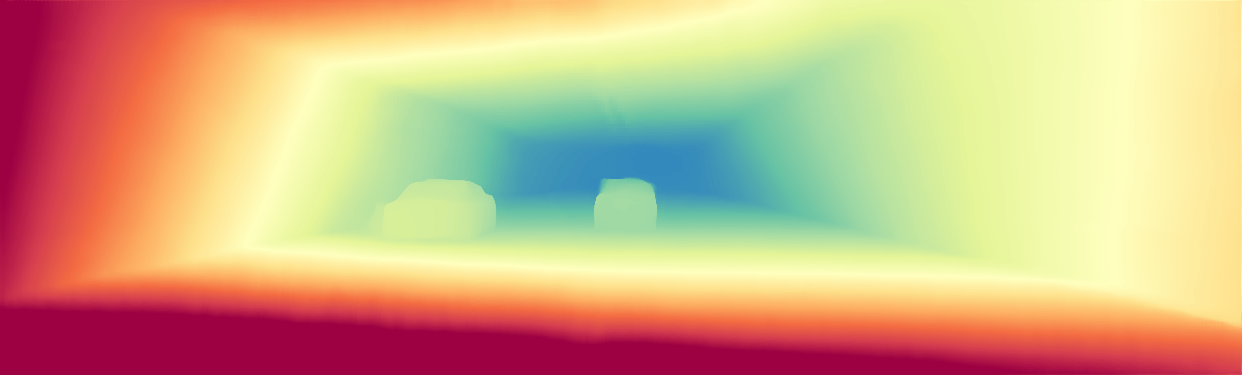} \\
        
        \hspace{-3.5em}\rotatebox[origin=c]{90}{\raisebox{0.08\textwidth}{\parbox[c][0.10\textwidth][c]{0.10\textwidth}{\centering\small Middlebury}}}\hspace{-3.5em} & \includegraphics[width=0.16\textwidth]{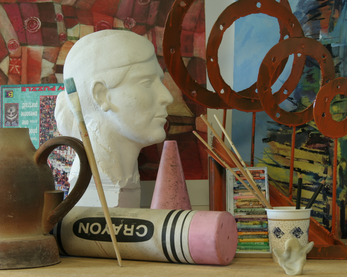} & 
        \includegraphics[width=0.16\textwidth]{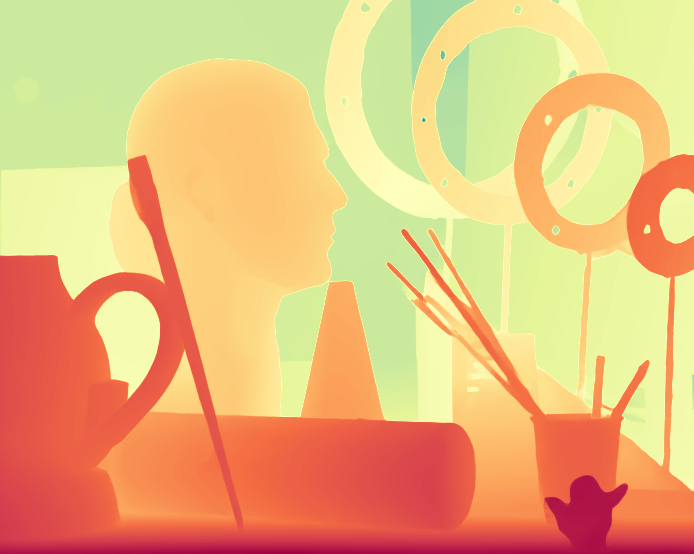} &
        \includegraphics[width=0.16\textwidth]{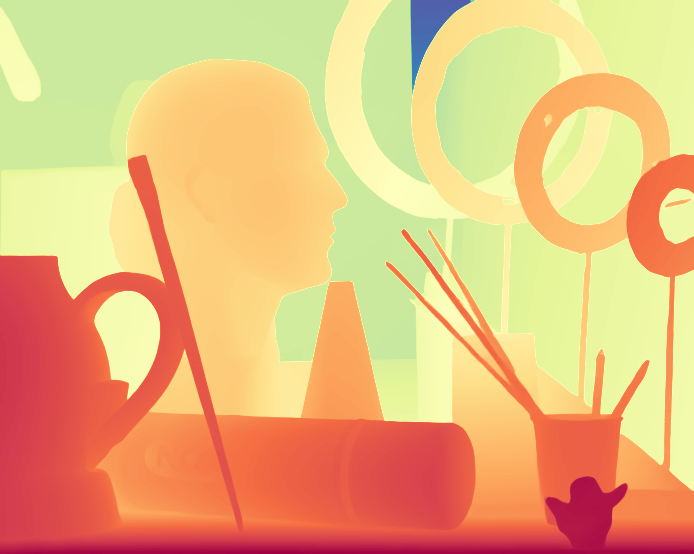} &
        \includegraphics[width=0.16\textwidth]{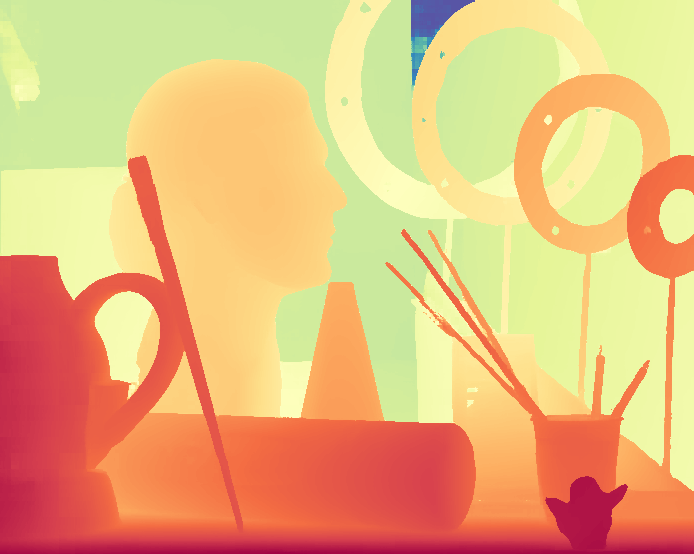} &
        \includegraphics[width=0.16\textwidth]{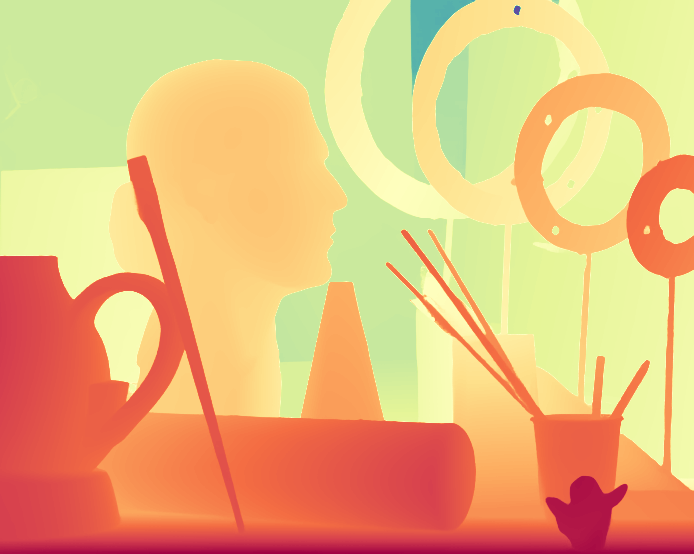} &
        \includegraphics[width=0.16\textwidth]{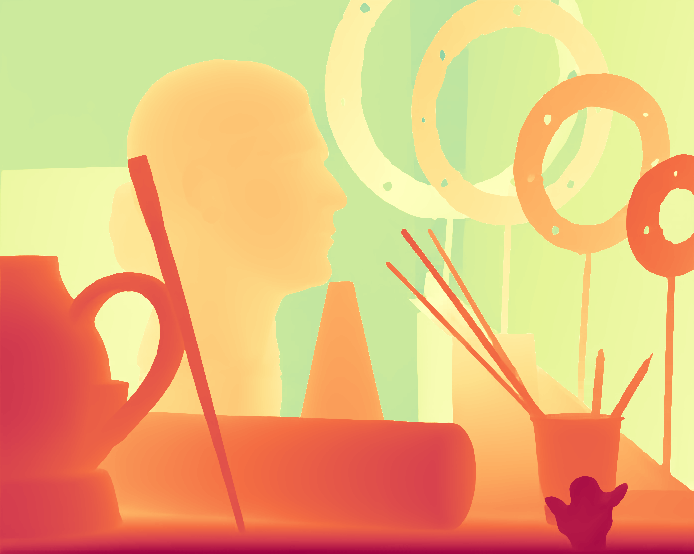} \vspace{-0.3cm}\\
               
        \hspace{-3.5em}\rotatebox[origin=c]{90}{\raisebox{0.08\textwidth}{\parbox[c][0.10\textwidth][c]{0.10\textwidth}{\small ETH3D}}}\hspace{-3.5em} &\includegraphics[width=0.16\textwidth]{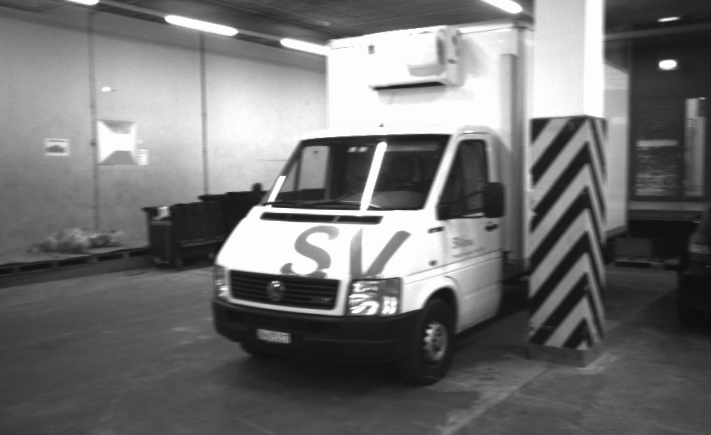} & 
        \includegraphics[width=0.16\textwidth]{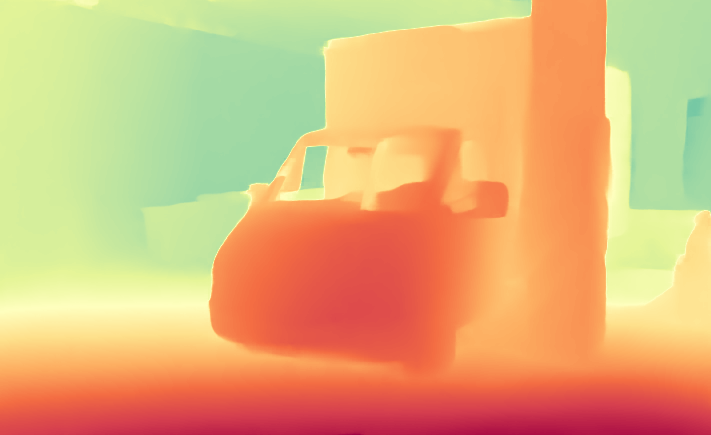} &
        \includegraphics[width=0.16\textwidth]{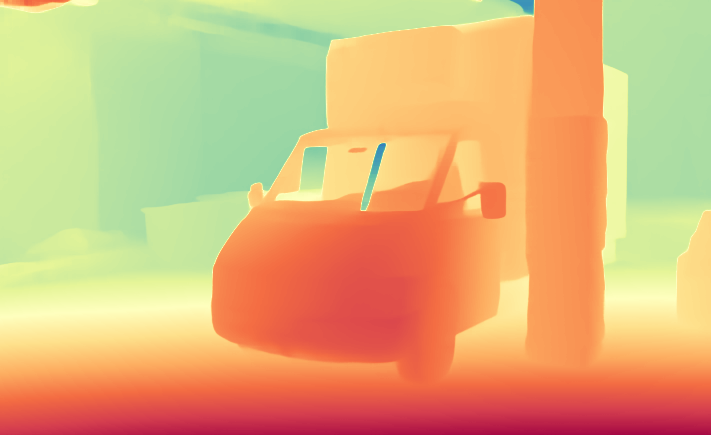} &
        \includegraphics[width=0.16\textwidth]{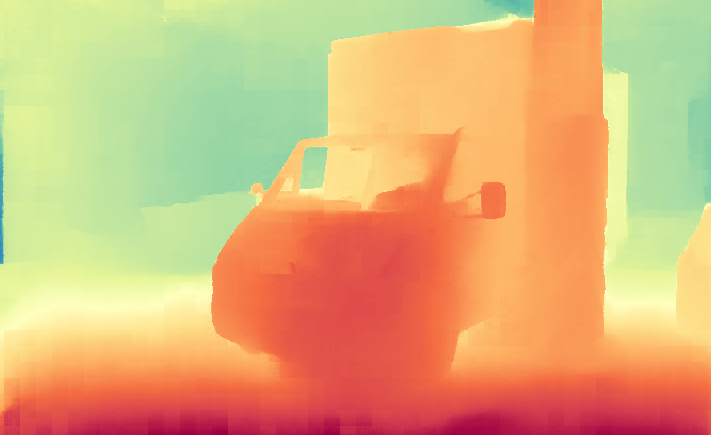} &
        \includegraphics[width=0.16\textwidth]{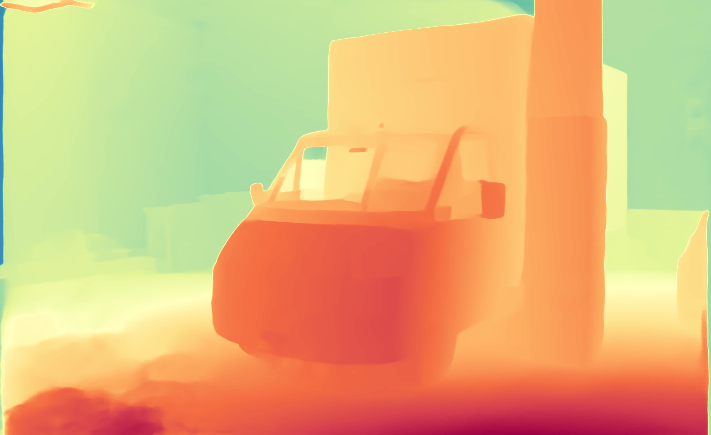} &
        \includegraphics[width=0.16\textwidth]{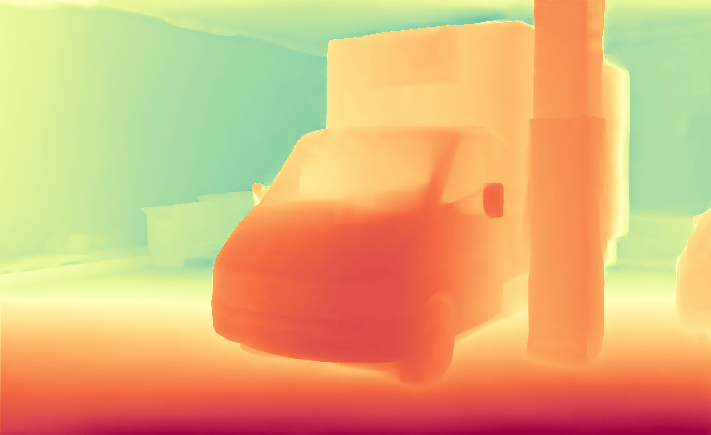} \\        
    \end{tabular}}\vspace{-0.3cm}
    \caption{\textbf{Qualitative Results -- Zero-Shot Generalization.} Predictions by state-of-the-art models and \method.}
    \label{fig:qual_zeroshot}\vspace{-0.3cm}
\end{figure*}

\section{Experiments}
\label{sec:experiments}

We describe our implementation details, datasets, and evaluation protocols, followed by experiments. We also refer the reader to the \textbf{supplementary material} for more results.

\subsection{Implementation and Experimental Settings}

We implement \method using PyTorch, starting from RAFT-Stereo codebase \cite{lipson2021raft}.
We use Depth Anything v2 \cite{depth_anything_v2} as the VFM fueling our model, using the \textit{Large} weights provided by the authors, trained on ground-truth labels from the HyperSim synthetic dataset \cite{roberts2021} only. 

Starting from the Sceneflow RAFT-Stereo checkpoint, we train \method on a single A100 GPU for 3 epochs, with learning rate 1e-4 and AdamW optimizer, on batches of 2 images. We extract random crops of size 320$\times$640 from images and apply standard color and spatial augmentations \cite{lipson2021raft}. 
The VFM is used only to source monocular depth maps, remaining frozen during training.
The number of iterations for GRUs is fixed to 12 during training and increased to 32 at inference time.

\subsection{Evaluation Datasets \& Protocol}

\textbf{Datasets}. We utilize SceneFlow \cite{mayer2016large} as our sole training dataset, comprising about 39k synthetic stereo pairs with dense ground-truth disparities. For evaluation, we employ several benchmarks: Middlebury 2014 \cite{scharstein2014high} and its 2021 extension \cite{middlebury2021} provide high-resolution indoor scenes with semi-dense labels (15 and 24 stereo pairs), KITTI 2012 \cite{geiger2012we} and 2015 \cite{menze2015object} feature outdoor driving scenarios ($\sim$200 pairs each at $1280 \times 384$ with sparse LiDAR ground truth), and ETH3D \cite{schops2017multi} contributes 27 low-resolution indoor/outdoor scenes. For non-Lambertian surfaces, we primarily use Booster \cite{zamaramirez2022booster}, containing 228 high-resolution (12 Mpx) indoor pairs with its 191-pair online benchmark, and LayeredFlow \cite{wen2024layeredflow}, featuring 400 pairs with transparent objects and sparse ground truth ($\sim$50 points per pair). Additionally, we include our newly proposed MonoTrap dataset focusing on optical illusions. For zero-shot evaluation, we test on KITTI 2015, Middlebury v3 at half (H) resolution, Middlebury 2021, and ETH3D, while non-Lambertian zero-shot testing relies on Booster at quarter (Q) resolution and LayeredFlow at eight (E) resolution.

\textbf{Evaluation Metrics}. We evaluate our method using two standard metrics: the average pixel error (Avg.), which computes the absolute difference between predicted and ground truth disparities averaged over all pixels, and the bad$>\tau$ error, which measures the percentage of pixels with a disparity error greater than $\tau$ pixels -- for the latter, we compute it considering all pixels or either non-occluded or occluded pixels, referred to as \textit{All}, \textit{Noc} or \textit{Occ} respectively. 

We evaluate on \dataset through standard monocular depth metrics \cite{godard2017unsupervised} - Absolute relative error (AbsRel), RMSE, and $\delta<1.05$ score.

\begin{table*}
\centering
\renewcommand{\tabcolsep}{18pt}
\scalebox{0.7}{
\begin{tabular}{|l||rrrrr|rrrr|}
\multicolumn{1}{c}{} & \multicolumn{5}{c}{Booster (Q)} & \multicolumn{4}{c}{LayeredFlow (E)} \\
\hline
 \multirow{2}{*}{Model} & \multicolumn{4}{c}{Error Rate (\%)} & Avg. & \multicolumn{3}{c}{Error Rate (\%)} & Avg. \\
  & $>2$ & $>4$ & $>6$ & $>8$ & (px) & $>1$ & $>3$ & $>5$ &(px) \\
\hline\hline
{RAFT-Stereo} \cite{lipson2021raft}
& \trd 17.84 & \snd 13.06 & \snd 10.76 & \snd 9.24 & \snd 3.59
& 89.21 & \trd 79.02 & 71.61 & 19.27 \\
{PSMNet} \cite{chang2018pyramid}
& 34.47 & 24.83 & 20.46 & 17.77 & 7.26
& 91.85 & 79.84 & \trd 70.04 & 21.18 \\
{GMStereo} \cite{xu2023unifying}
& 32.44 & 22.52 & 17.96 & 15.02 & 5.29
& 92.95 & 83.68 & 74.76 & 20.91 \\

{ELFNet} \cite{lou2023elfnet}
& 45.52 & 35.79 & 30.72 & 27.33 & 14.04
& 93.08 & 82.24 & 70.41 & 20.19 \\
{PCVNet} \cite{Zeng_2023_ICCV}
& 22.63 & 16.51 & 13.81 & 12.08 & 4.70
& \trd 88.27 & \snd 76.65 & \snd 66.79 & \snd 18.19 \\
{DLNR} \cite{zhao2023high}
& 18.56 & 14.55 & 12.61 & 11.22 & 3.97
& 89.90 & 79.46 & 72.72 & \trd 18.97 \\
{Selective-RAFT} \cite{wang2024selective}
& 20.01 & 15.08 & 12.52 & 10.88 & 4.12
& 92.69 & 86.32 & 78.82 & 20.18 \\
{Selective-IGEV} \cite{wang2024selective}
& 18.52 & 14.24 & 12.14 & 10.77 & 4.38
& 91.31 & 81.72 & 74.74 & 19.65 \\
{IGEV-Stereo} \cite{xu2023iterative}
& \snd 16.90 & \trd 13.23 & \trd 11.40 & \trd 10.20 & \trd 3.94
& \snd 87.28 & 80.07 & 72.91 & 19.07 \\
{NMRF} \cite{guan2024neural}
& 27.08 & 19.06 & 15.43 & 13.21 & 5.02
& 89.08 & 79.13 & 70.51 & 20.17 \\
{\bf \method{} (ours)}
&\fst 9.01 &\fst 5.40 &\fst 4.12 &\fst 3.34 &\fst 1.21
&\fst 81.83 &\fst 57.66 &\fst 45.12 &\fst 11.20 \\
\hline

\multicolumn{1}{c}{} & \multicolumn{5}{c}{Booster (Q) Online Benchmark} & \multicolumn{4}{c}{LayeredFlow (E)} \\
\hline
{DKT-RAFT} \cite{zhang2024robust} (*)
& \snd 10.32 & \snd 7.13 & \snd 5.65 & \snd 4.36 & \snd 1.70 
& \snd 66.05 & \snd 46.95 & \snd 37.77 & \snd 8.72 \\
{\bf \method{} (ours)} (*) &
 \fst 6.52 & \fst 2.82 & \fst 1.77 & \fst 1.27 & \fst 0.73 
& \fst 51.24 & \fst 25.63 & \fst 15.65 & \fst 4.84 \\
\hline

\end{tabular}}\vspace{-0.3cm}
\caption{\textbf{Zero-shot Non-Lambertian Generalization.} Comparison with state-of-the-art models. Networks trained on SceneFlow \cite{mayer2016large}. (*) means fine-tuned on Booster training set.
}\vspace{-0.3cm}
\label{tab:roundtable2}
\end{table*}

\begin{figure*}[t]
    \centering
    \renewcommand{\tabcolsep}{1pt}
    \scalebox{0.9}{
    \begin{tabular}{ccccccc}
        & \small RGB &
        \small RAFT-Stereo \cite{lipson2021raft} &
        \small DLNR \cite{zhao2023high} &
        \small NMRF \cite{guan2024neural} &
        \small Selective-IGEV \cite{wang2024selective} &
        \method \\
        
        \hspace{-3.5em}\rotatebox[origin=c]{90}{\raisebox{0.08\textwidth}{\parbox[c][0.10\textwidth][c]{0.10\textwidth}{\centering\small Booster}}}\hspace{-3.5em} &\includegraphics[width=0.16\textwidth]{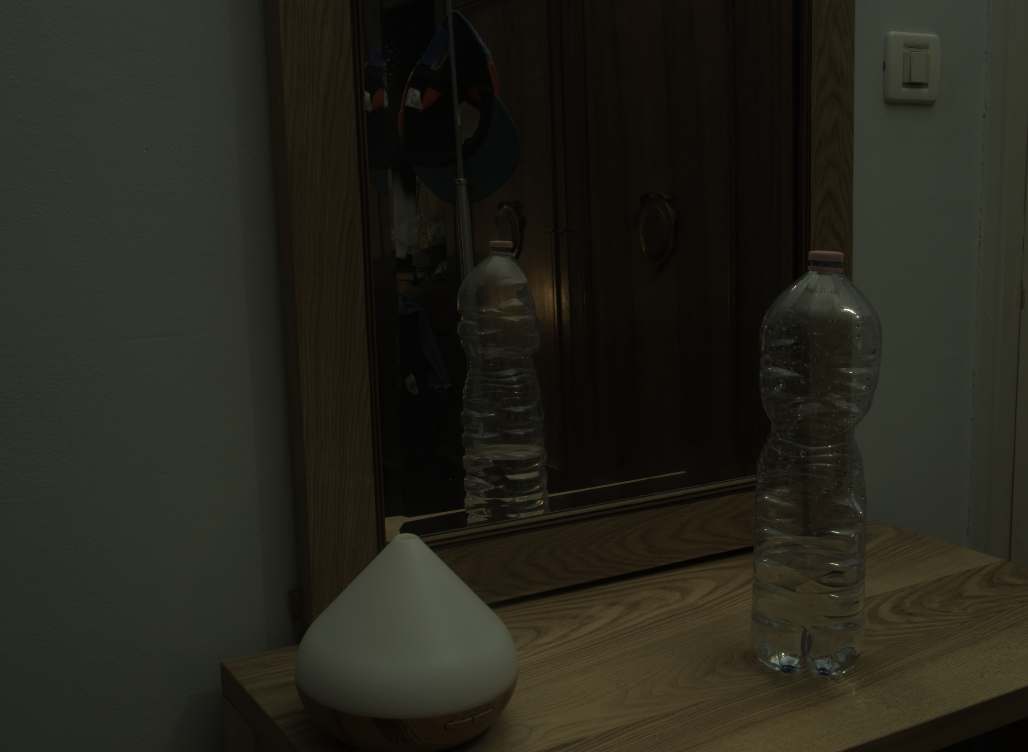} & 
        \includegraphics[width=0.16\textwidth]{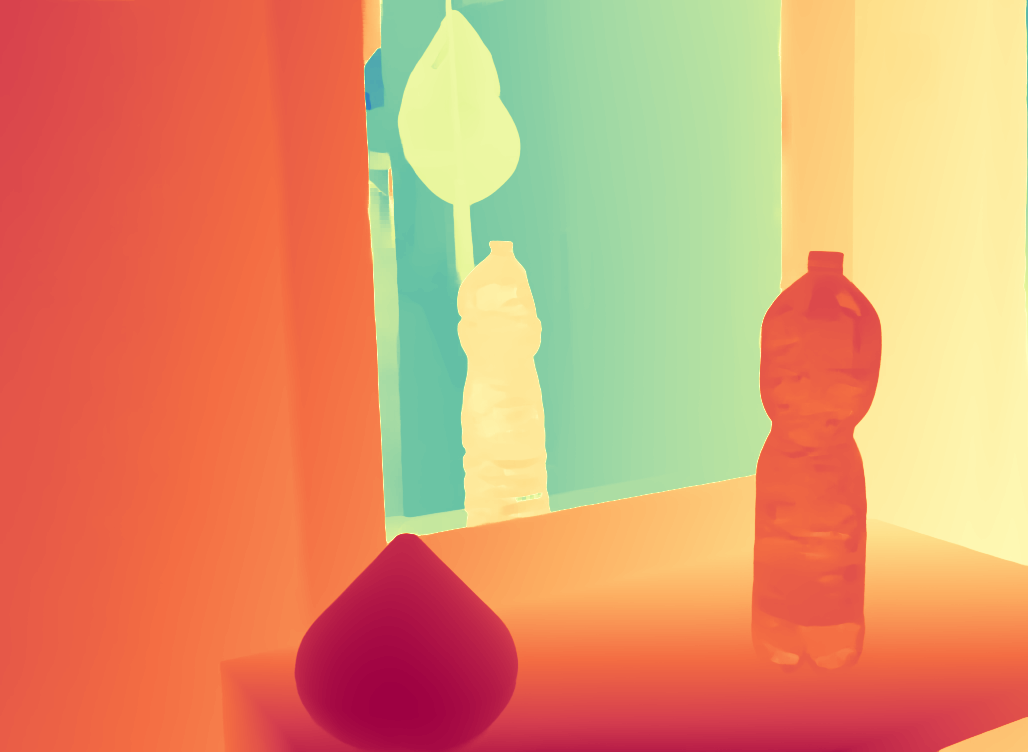} &
        \includegraphics[width=0.16\textwidth]{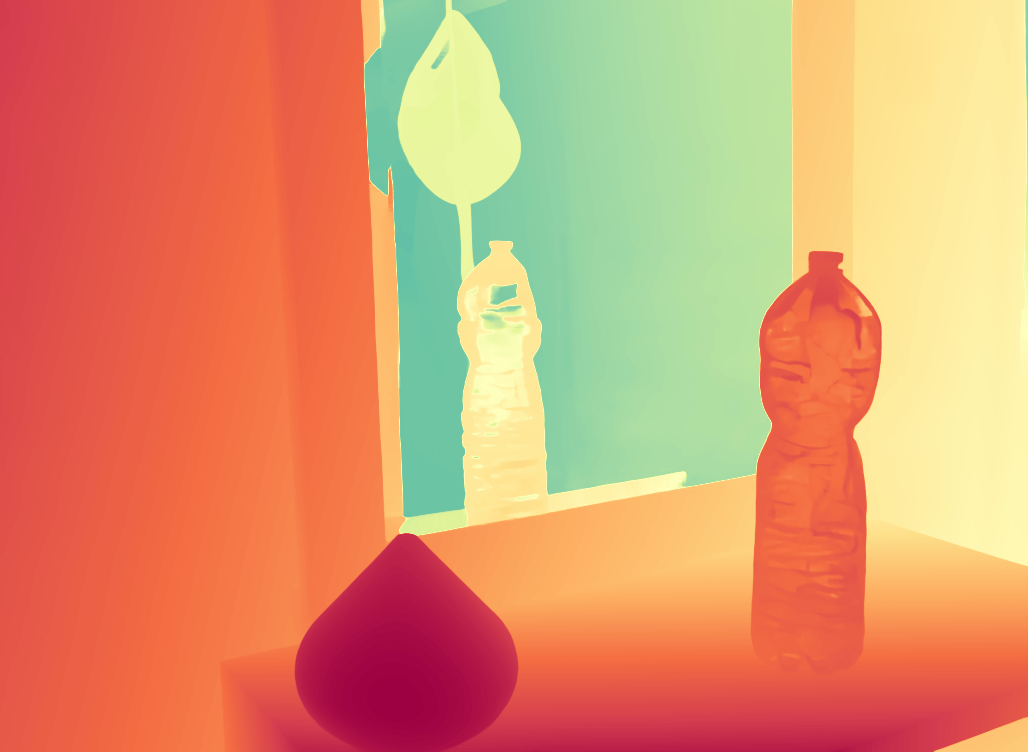} &
        \includegraphics[width=0.16\textwidth]{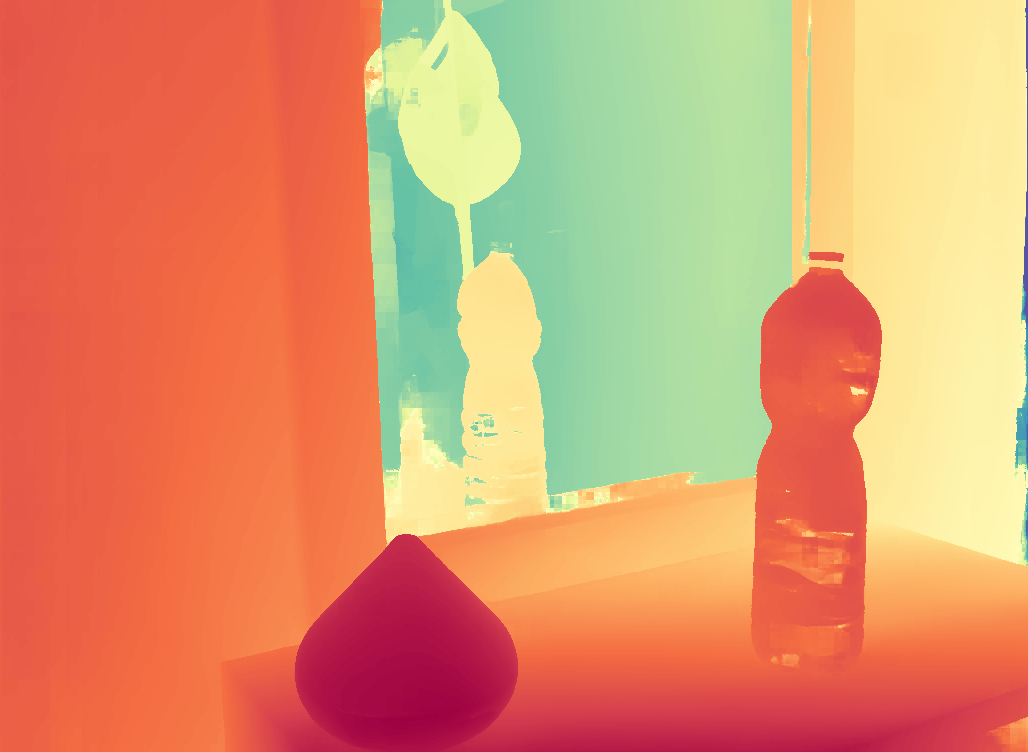} &
        \includegraphics[width=0.16\textwidth]{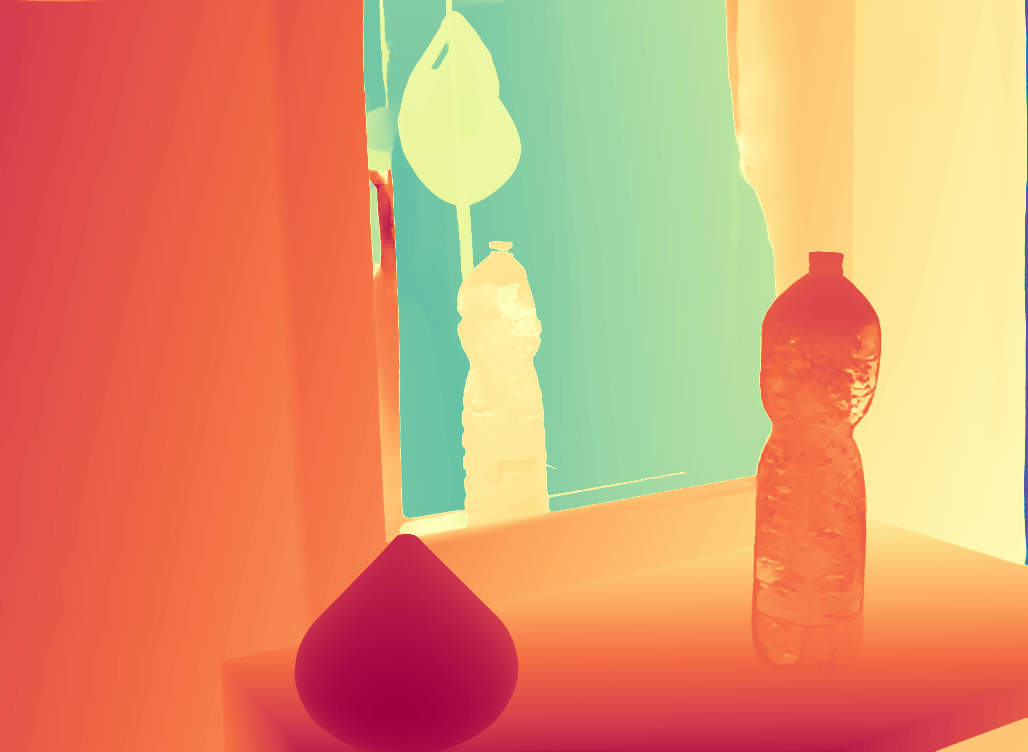} &
        \includegraphics[width=0.16\textwidth]{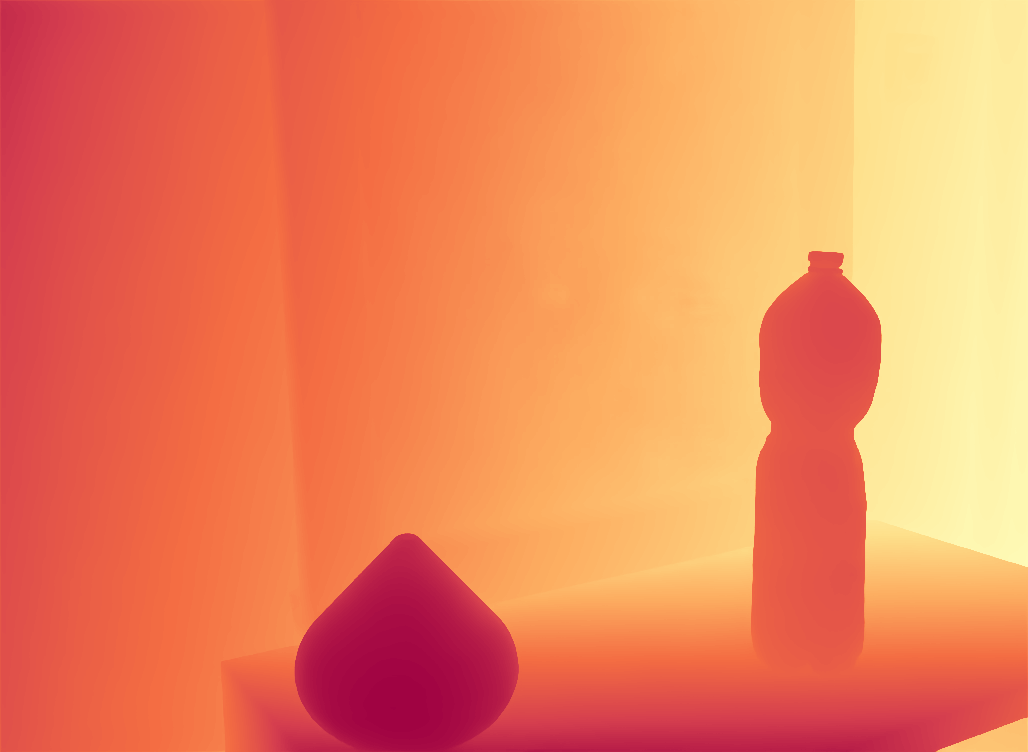} \vspace{-0.5cm}\\

        \hspace{-3.5em}\rotatebox[origin=c]{90}{\raisebox{0.08\textwidth}{\parbox[c][0.10\textwidth][c]{0.10\textwidth}{\hspace{-0.3cm}\small LayeredFlow}}}\hspace{-3.5em} &\includegraphics[width=0.16\textwidth]{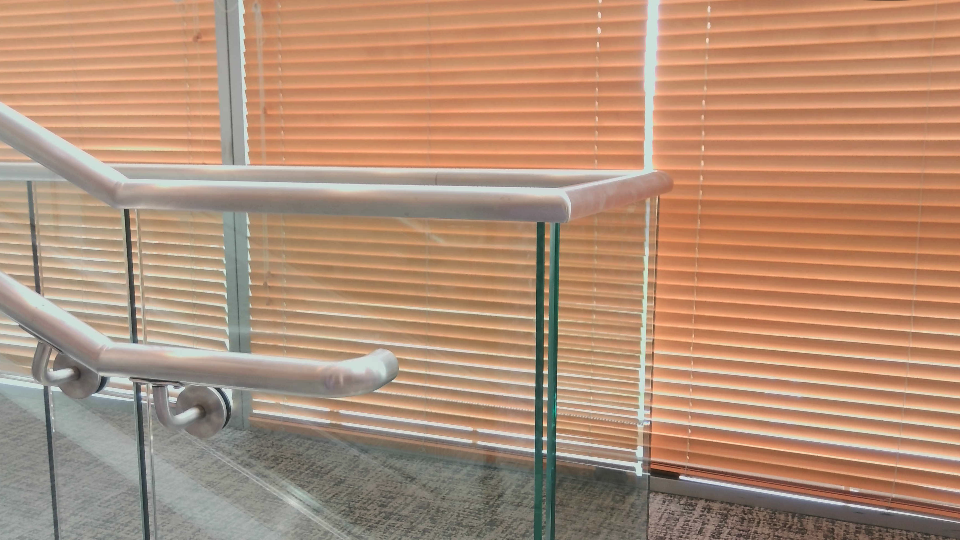} & 
        \includegraphics[width=0.16\textwidth]{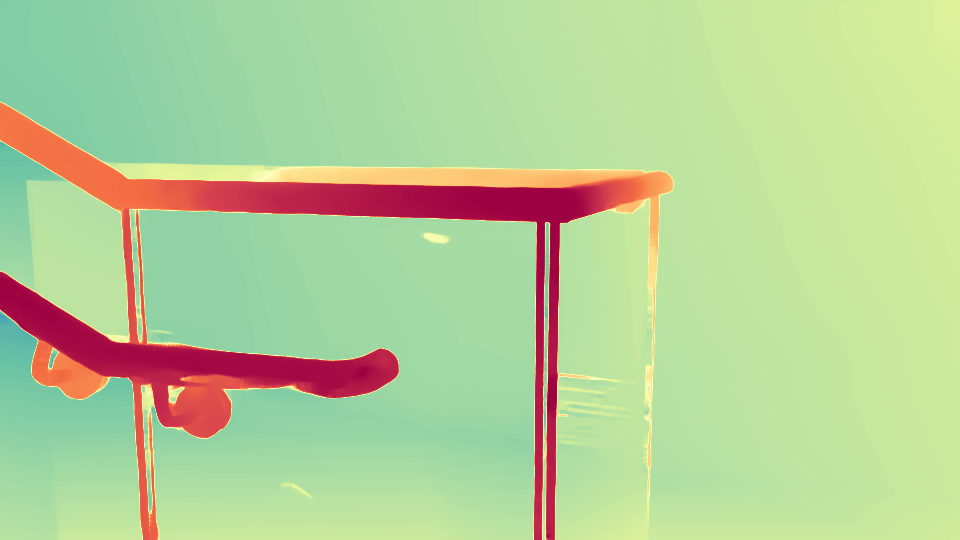} &
        \includegraphics[width=0.16\textwidth]{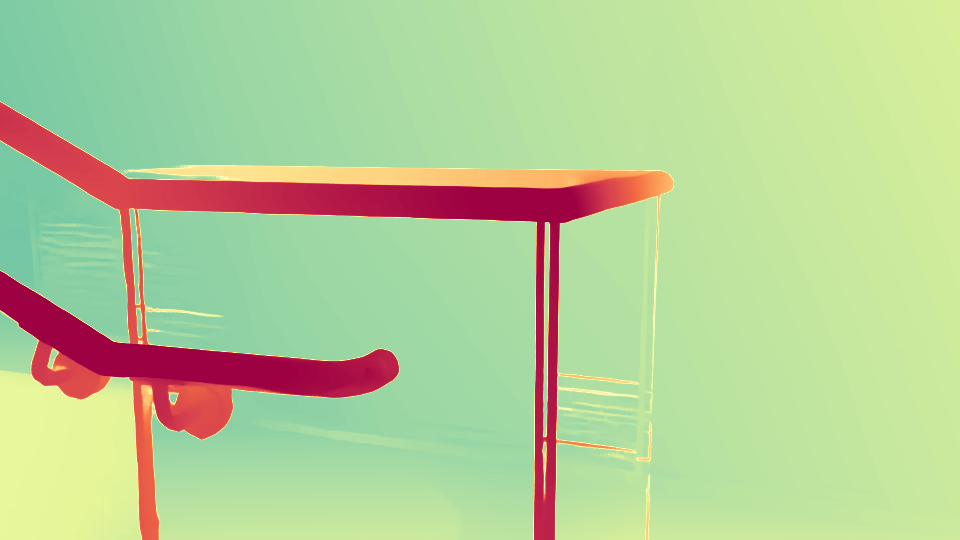} &
        \includegraphics[width=0.16\textwidth]{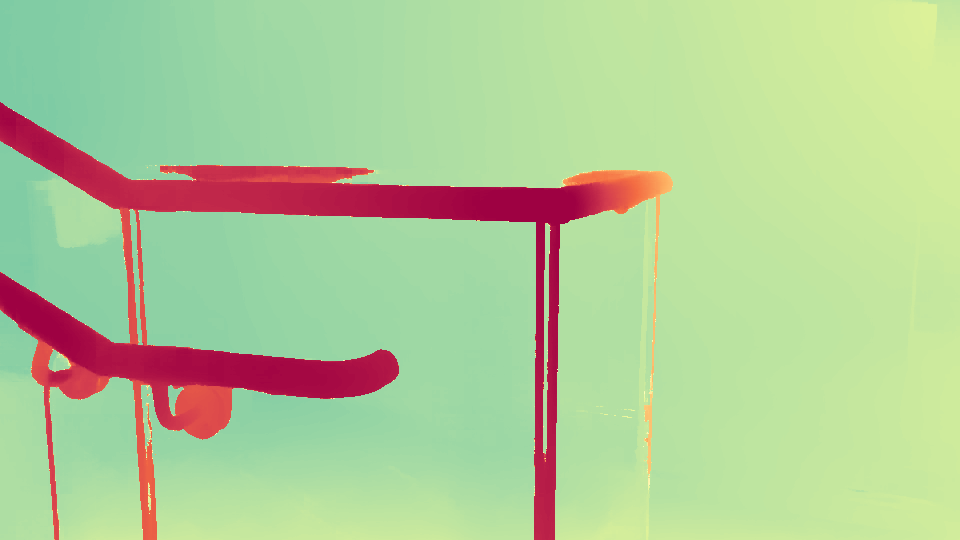} &
        \includegraphics[width=0.16\textwidth]{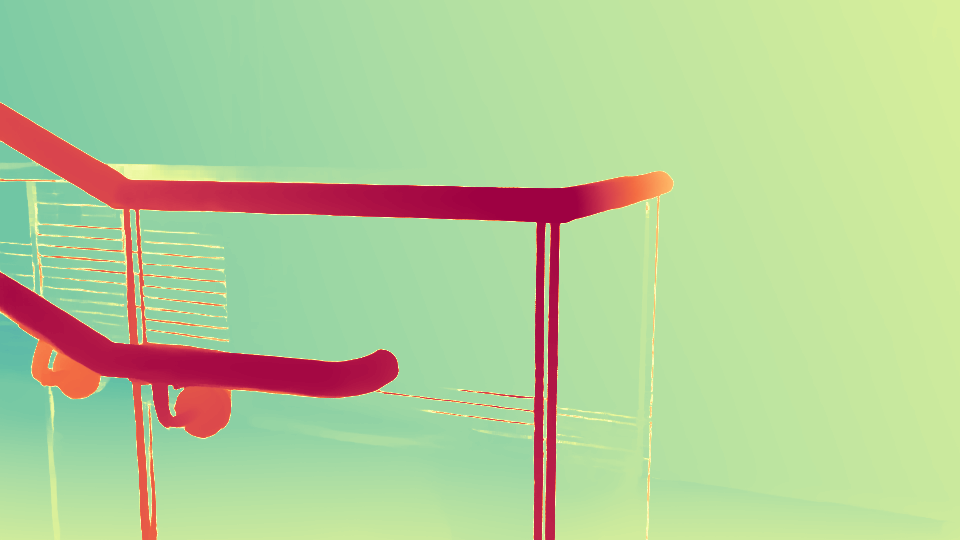} &
        \includegraphics[width=0.16\textwidth]{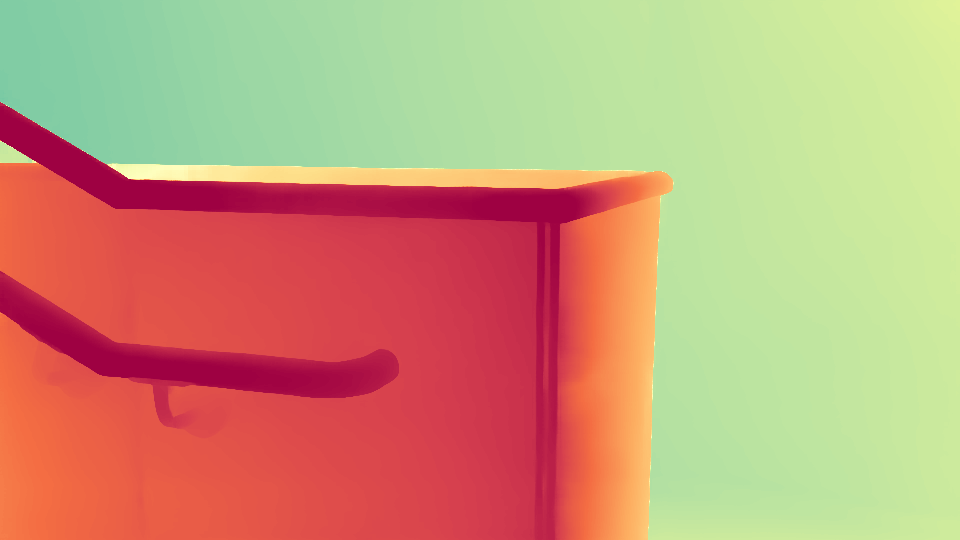} \\
        
    \end{tabular}}\vspace{-0.3cm}
    \caption{\textbf{Qualitative results -- Zero-Shot non-Lambertian Generalization.} Predictions by state-of-the-art models and \method.}
    \label{fig:qual_nonlambertian}\vspace{-0.3cm}
\end{figure*}

\subsection{Ablation Study}

We start our analysis by evaluating how individual components of our model contribute to the overall accuracy. All model variants are trained solely on the synthetic SceneFlow dataset and tested on Booster and Middlebury 2014, allowing us to examine their effectiveness on non-Lambertian surfaces and general scenes. 

Table \ref{tab:ablation} summarizes our findings. In (A), we report the performance of our baseline model, upon which we build \method -- i.e., RAFT-stereo \cite{lipson2021raft}. On the one hand, by adding monocular context from an off-the-shelf monocular depth network to the pre-trained context backbone (B), we observe improved performance on non-Lambertian surfaces, though at the expense of a general drop in accuracy on Middlebury. On the other hand, by re-training the context backbone to process depth maps obtained from the monocular network on SceneFlow (C), we can appreciate a consistent improvement in both datasets. 
Introducing the normals correlation volume with subsequent differentiable depth scaling (D) significantly enhances the accuracy on non-Lambertian surfaces, also showing improvements on indoor scenes.
Finally, cost volume augmentations and truncation (E) demonstrate positive effects on transparent surfaces and mirrors present in the Booster dataset by further reducing the bad-2 metric by approximately 1.5\% and Avg. by 0.7 pixels, with minimal influence on Middlebury.

According to these results, from now on, we will adopt (E) as the default setting for \method.

\subsection{Zero-Shot Generalization}

We now compare our \method model against state-of-the-art deep stereo networks, assessing zero-shot generalization capability when transferred from synthetic to real images.
Purposely, we follow a well-established benchmark in the literature \cite{lipson2021raft,Tosi_2023_CVPR}, evaluating on real datasets models pre-trained exclusively on SceneFlow \cite{mayer2016large}.

Table \ref{tab:roundtable1} compares \method with off-the-shelf stereo networks using authors' provided weights. Considering All, Noc, and Avg. metrics, we can notice how \method achieves consistently better results across most datasets, achieving almost 3\% lower bad-2 All on Middlebury 2014 versus the second-best method DLNR \cite{zhao2023high}, and breaking the 4\% barrier on KITTI's bad-3 All metric. 

The Occ metric further demonstrates how \method{} consistently outperforms other stereo models on any dataset, with substantial margins over the second-best -- i.e., approximately 6\% on Middlebury 2014 and KITTI 2012, and 3\% on ETH3D. This confirms that leveraging priors from VFMs for monocular depth estimation effectively improve the stereo matching estimation accuracy in challenging conditions where stereo matching is ill-posed, such as at occluded regions.

Figure \ref{fig:qual_zeroshot} shows predictions on KITTI 2015, Middlebury 2014, and ETH3D samples. In particular, the first row shows an extremely challenging case for SceneFlow-trained models, where \method achieves accurate disparity maps thanks to VFM priors.

\subsection{Zero-Shot Non-Lambertian Generalization}

We now assess the generalization capabilities of \method and existing stereo models when dealing with non-Lambertian materials, such as transparent surfaces or mirrors. To this end, we conduct a zero-shot generalization evaluation experiment on the Booster \cite{Ramirez_2022_CVPR} and LayeredFlow \cite{wen2024layeredflow} datasets, once again using models pre-trained on SceneFlow \cite{mayer2016large} -- with weights provided by the authors. 

Table \ref{tab:roundtable2} shows the outcome of this evaluation. This time, we can perceive even more clearly how \method is the absolute winner, demonstrating unprecedented robustness in the presence of non-Lambertian surfaces despite being trained only on synthetic stereo data, not even featuring such objects. These results further validate how leveraging strong priors from existing VFMs for monocular depth estimation can play a game-changing role in stereo matching as well, especially when lacking training data explicitly targeting critical conditions such as non-Lambertian surfaces.
At the bottom, we report results achieved by fine-tuning \method{} on the Booster training set and evaluating on the online benchmark. Our model ranks first when evaluated at quarter resolution.

Figure \ref{fig:qual_nonlambertian} shows examples from Booster and LayeredFlow, where \method is the only stereo model correctly perceiving the mirror and transparent railing.

\begin{table}[t]
\centering
\renewcommand{\tabcolsep}{10pt}
\scalebox{0.8}{
\begin{tabular}{|l||rrr|}
\multicolumn{1}{c}{} & \multicolumn{3}{c}{\dataset} \\
\hline
 \multirow{2}{*}{Model} & AbsRel & RMSE & $\sigma<1.05$ \\
  & (\%)$\downarrow$ & (m)$\downarrow$ & (\%)$\uparrow$ \\
\hline\hline

Depth Anything v2 \cite{depth_anything_v2} & 53.46 & 0.36 & 15.21 \\
Depth Anything v2 \cite{depth_anything_v2} $\dagger$ & 27.92 & 0.27 & 19.43 \\
DepthPro \cite{depthpro} & 47.77 & 0.32 & 21.90 \\
DepthPro \cite{depthpro} $\dagger$ & \trd 20.82 & \trd 0.22 & \trd 22.88 \\
\hline
RAFT-Stereo \cite{lipson2021raft} & \snd 5.01 & \snd 0.09 & \snd 77.05 \\
\textbf{\method} & \fst 3.50 & \fst 0.06 & \fst 80.27 \\

\hline

\end{tabular}}\vspace{-0.2cm}
\caption{\textbf{\dataset Benchmark.} Comparison with state-of-the-art monocular depth estimation models and RAFT-Stereo. Both RAFT-Stereo and \method are trained on SceneFlow \cite{mayer2016large}. $\dagger$ refers to robust scaling through RANSAC.
}\vspace{-0.3cm}
\label{tab:monotrap}
\end{table}

\subsection{\dataset Benchmark}

We conclude our evaluation by running experiments on our newly collected \dataset dataset to prove the robustness of \method in the presence of critical conditions harming the accuracy of monocular depth predictors.

Table \ref{tab:monotrap} collects the results achieved by state-of-the-art monocular depth estimation models, the baseline stereo model over which we built our framework (RAFT-Stereo) and \method. Regarding the former models, as they predict affine-invariant depth maps, following the literature \cite{Ranftl2022} we use least square errors to align them to the ground-truth. As these models are fooled by the visual illusions, this scaling procedure is likely to yield sub-optimal scale and shift parameters. Therefore, we alternatively align to ground-truth depth through a more robust RANSAC fitting -- denoted with $\dagger$ in the table.

On the one hand, by comparing monocular and stereo methods, we notice how the failures of the former negatively impact their evaluation metrics. Once again, we remark that a direct comparison across the two families of methods is not the main goal of this experiment.
On the other hand, we focus on the comparison between RAFT-Stereo and \method, with our model performing slightly better than its baseline. This fact proves that despite its strong reliance on the priors retrieved from VFMs for monocular depth estimation, \method can properly ignore such priors when unreliable. 

\begin{figure}[t]
    \centering
    \renewcommand{\tabcolsep}{1pt}
    \scalebox{0.9}{
    \begin{tabular}{ccc}
        \small RGB &
        \small D. Anything v2 \cite{depth_anything_v2} &
        \method \\
        \includegraphics[width=0.16\textwidth]{imgs/monotrap_samples/rgb/13.png} & 
        \includegraphics[width=0.16\textwidth]{imgs/monotrap_samples/mono/13.png} &
        \includegraphics[width=0.16\textwidth]{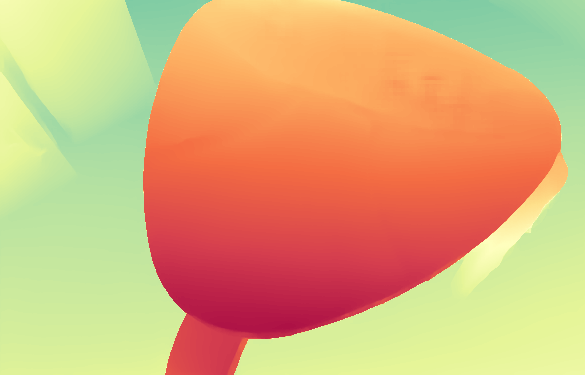} \\

        \includegraphics[width=0.16\textwidth]{imgs/monotrap_samples/rgb/2.png} & 
        \includegraphics[width=0.16\textwidth]{imgs/monotrap_samples/mono/2.png} &
        \includegraphics[width=0.16\textwidth]{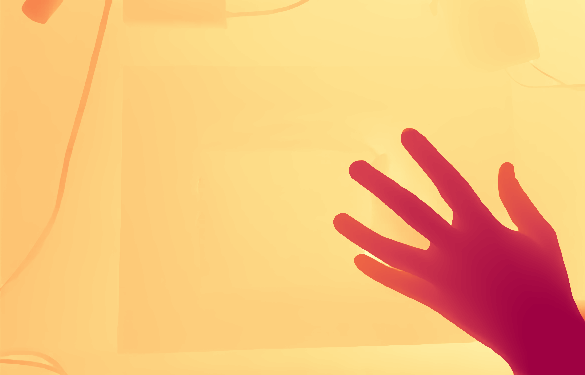} \\

        \includegraphics[width=0.16\textwidth]{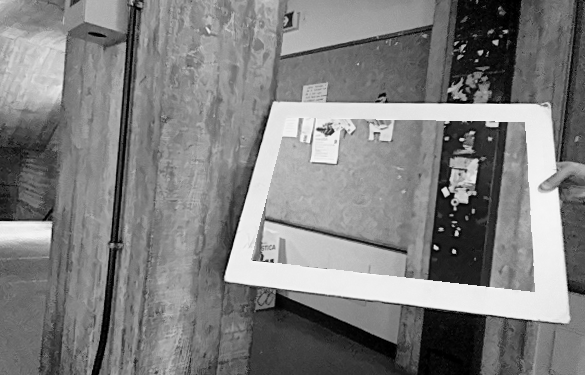} & 
        \includegraphics[width=0.16\textwidth]{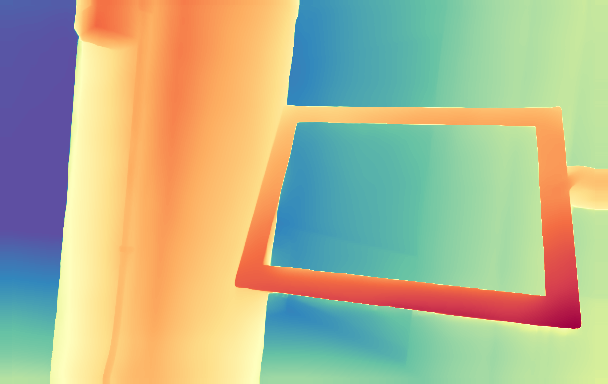} &
        \includegraphics[width=0.16\textwidth]{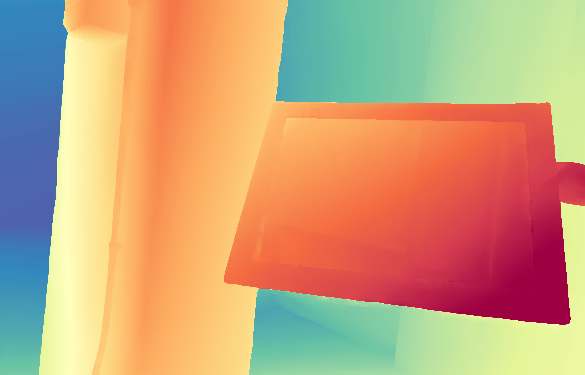} \\
        
    \end{tabular}}\vspace{-0.3cm}
    \caption{\textbf{Qualitative results -- \dataset.} \method is not fooled by erroneous predictions by its monocular engine \cite{depth_anything_v2}. }
    \label{fig:qual_monotrap}\vspace{-0.5cm}
\end{figure}

Figure \ref{fig:qual_monotrap} shows three samples where Depth Anything v2 fails while \method does not.

\section{Conclusion}
\label{sec:conclusion}

In this paper, we introduced \method, a novel stereo matching framework that leverages monocular depth VFMs to overcome traditional stereo matching limitations. Combining stereo geometric constraints with monocular priors, our approach demonstrates superior zero-shot generalization and robustness to challenging conditions like textureless regions, occlusions, and non-Lambertian surfaces. Furthermore, through our novel MonoTrap dataset, we showed that \method effectively combines the best of both worlds - maintaining stereo matching's geometric accuracy where monocular methods fail, while leveraging monocular priors to handle challenging stereo scenarios. Extensive comparisons against state-of-the-art networks in zero-shot settings validate these findings.

\small
{\textbf{Acknowledgement.} 
This study was carried out within the MOST – Sustainable Mobility National Research Center and received funding from the European Union Next-GenerationEU – PIANO NAZIONALE DI RIPRESA E RESILIENZA (PNRR) – MISSIONE 4 COMPONENTE 2, INVESTIMENTO 1.4 – D.D. 1033 17/06/2022, CN00000023. This manuscript reflects only the authors’ views and opinions, neither the European Union nor the European Commission can be considered responsible for them.
This study was funded by the European Union – Next Generation EU within the framework of the National Recovery and Resilience Plan NRRP – Mission 4 ``Education and Research" – Component 2 - Investment 1.1 ``National Research Program and Projects of Significant National Interest Fund (PRIN)" (Call D.D. MUR n. 104/2022) – PRIN2022 – Project reference: ``RiverWatch: a citizen-science approach to river pollution monitoring" (ID: 2022MMBA8X, CUP: J53D23002260006).

We also acknowledge the CINECA award under the ISCRA initiative, for the availability of high-performance computing resources and support.}

{
    \small
    \bibliographystyle{ieeenat_fullname}
    \bibliography{references}
}

\clearpage

\onecolumn

{
    \clearpage
    \centering
    \Large
    \textbf{\thetitle}\\
    \vspace{0.5em}Supplementary Material \\
    \vspace{1.0em}
}

This document reports additional material concerning CVPR paper ``Stereo Anywhere: Robust Zero-Shot Deep Stereo Matching Even Where Either Stereo or Mono Fail". Specifically:

\begin{itemize}
    \item First, we present an extended description of our proposed architecture in \cref{sec:method}, including detailed formulations of the monocular correlation volume (\cref{subsec:mono_corr}), differentiable monocular scaling (\cref{subsec:diff_mono}), cost volume augmentation (\cref{subsec:cost_aug}), volume truncation (\cref{subsec:vol_trunc}), and training supervision (\cref{subsec:training}). 

    \item We then report extensive ablation studies in \cref{sec:ablation} demonstrating how our stereo matching architecture effectively generalizes across different state-of-the-art monocular depth networks (\cref{subsec:ablation_VFM}), showing consistent improvements over baseline stereo methods regardless of the specific VFM employed. 
    Then, we show qualitatively the impact of the truncated cost volume augmentation on disparity estimation on non-Lambertian surfaces (\cref{subsec:vol_trunc_qual}). Furthermore, we include an analysis of runtime performance and memory consumption (\cref{subsec:runtime}) across different input resolutions and VFMs.  

    \item Finally, we present extensive qualitative results in \cref{subsec:qual} across multiple datasets, demonstrating the effectiveness of our method in dealing with challenging scenarios such as non-Lambertian surfaces, transparent objects and textureless regions.
\end{itemize}

\section{Method Overview: Additional Details}

In this section, we enrich the description of \method architecture.

\subsection{Monocular Correlation Volume}
\label{subsec:mono_corr}

Given the monocular depth estimations $\textbf{M}_L \in \mathbb{R}^{1 \times H \times W}$ and $\textbf{M}_R \in \mathbb{R}^{1 \times H \times W}$, we aim to estimate the normal maps $\nabla_L \in \mathbb{R}^{3 \times \frac{H}{4} \times \frac{W}{4}}$ and $\nabla_R \in \mathbb{R}^{3 \times \frac{H}{4} \times \frac{W}{4}}$ to construct the 3D correlation volume $\mathbf{V}_M \in \mathbb{R}^{\frac{H}{4} \times \frac{W}{4} \times \frac{W}{4}}$.
We decide to use $\nabla_L$ and $\nabla_R$ instead of extracting additional features from $\textbf{M}_L$ and $\textbf{M}_R$ because $\textbf{M}_L$ and $\textbf{M}_R$ already provide high-level information.
Furthermore, normal maps can handle depth inconsistencies between $\textbf{M}_L$ and $\textbf{M}_R$ that can occur for example when a foreground object is visible only in a single view.
We downsample $\textbf{M}_L$ and $\textbf{M}_R$ to 1/4  -- bilinear interpolation, then we estimate $\nabla_L$ and $\nabla_R$  -- spatial gradient:

\small\begin{equation}
    \nabla = \frac{\nabla^*}{\lVert \nabla^* \rVert}, \quad \nabla^* = \begin{bmatrix}
        -\frac{\partial \left(\lambda\mathbf{M}_\frac{1}{4}\right)}{\partial x} & -\frac{\partial \left(\lambda\mathbf{M}_\frac{1}{4}\right)}{\partial y} & 1 \\
    \end{bmatrix}, \quad \lambda = \frac{1}{10}\cdot\frac{W}{4}
    \label{eq:normal_estimation}
\end{equation}\normalsize
where $\lambda$ is a gain factor that is proportional to $W$, which permits to achieve scale-invariant normal maps.

Given the absence of texture in normal maps, $\textbf{V}_M$ will be not ambiguous only in edges.
To alleviate this problem, we segment $\textbf{V}_M$  -- the relative depth priors from $\textbf{M}_L$ and $\textbf{M}_R$: doing so we aim to reduce the ambiguity by forcing the matching only in similar depth regions (\eg, foreground objects cannot match with background object since the correlation score is masked to zero).
Considering \cref{eq:vol_masking}, we calculate masks ${\mathcal{M}_L}^n$ and ${\mathcal{M}_R}^n$ as follows:

\small\begin{equation}
    ({\mathcal{M}_L}^n)_{ij} = \begin{cases}
        1\ \text{if}\ \frac{n}{N} \leq (\mathbf{M}_L)_{ij} < \frac{n+1}{N}\\
        0\ \text{otherwise}
    \end{cases} \quad ({\mathcal{M}_R}^n)_{ik} = \begin{cases}
        1\ \text{if}\ \frac{n}{N} \leq (\mathbf{M}_R)_{ik} < \frac{n+1}{N}\\
        0\ \text{otherwise}
    \end{cases}
    \label{eq:mask}
\end{equation}\normalsize

To further deal with the ambiguity, we improve the 3D Convolutional Regularization model $\phi_A$  -- an adapted version of CoEx \cite{bangunharcana2021correlate} correlation volume excitation that exploits both views $\textbf{M}_L$, $\textbf{M}_R$:

\small\begin{equation}
    ({\mathbf{V'}^s}_M)_{fijk} = \sigma\left(({{\mathbf{f}_L}^s})_{fij}\right) \odot \sigma\left(({{\mathbf{f}_R}^s})_{fik}\right) \odot ({\mathbf{V}^s}_M)_{fijk}
    \label{eq:our_coex}
\end{equation}\normalsize
where ${\mathbf{V'}^s}_M$ is the excited volume, $\sigma(\cdot)$ is the sigmoid function, $\odot$ is the element-wise product, ${\mathbf{V}^s}_M \in \mathbb{R}^{F \times \frac{H}{s} \times \frac{W}{s} \times \frac{W}{s}}$ is an intermediate correlation feature volume at scale $s$ with $F$ features inside module $\phi_A$,  ${\mathbf{f}_L}^s \in \mathbb{R}^{F \times \frac{H}{s} \times \frac{W}{s} \times 1}$ and ${\mathbf{f}_R}^s \in \mathbb{R}^{F \times \frac{H}{s} \times 1 \times \frac{W}{s}}$ are shallow 2D conv-features extracted from $\mathbf{M}_L$ and $\mathbf{M}_R$ downsampled at proper scale.

\subsection{Differentiable Monocular Scaling}
\label{subsec:diff_mono}

As detailed in \cref{subsec:corr_pyramids}, volume $\mathbf{V}^D_M$ is used also to estimate the coarse disparity maps $\hat{\mathbf{D}}_L$ $\hat{\mathbf{D}}_R$, while volume $\mathbf{V}^C_M$ is utilized to estimate confidence maps $\hat{\mathbf{C}}_L$ $\hat{\mathbf{C}}_R$.
$\hat{\mathbf{D}}_L$ $\hat{\mathbf{C}}_L$ and $\hat{\mathbf{D}}_R$ $\hat{\mathbf{C}}_R$ are used to scale respectively $\mathbf{M}_L$ and $\mathbf{M}_R$.
As described in \cref{eq:softmax_left}, we can estimate left disparity from a correlation volume  -- a softargmax operation on the last W dimension of $\mathbf{V}^D_M$ and  -- the relationship between left disparity and correlation.
Here we report an extended version of \cref{eq:softmax_left} with the explicit formula for softargmax operator:

\small\begin{equation} 
    (\hat{\mathbf{D}}_L)_{ij} = j - \left(\text{softargmax}_L (\mathbf{V}^D_M )\right)_{ij} = j - \sum_{d}^{\frac{W}{4}} d \cdot \frac{e^{(\mathbf{V}^D_M)_{ijd}}}{\sum_{f}^{\frac{W}{4}} e^{(\mathbf{V}^D_M)_{ijf}}}
    \label{eq:softmax_left2}
\end{equation}\normalsize

At the same time, given the relationship between right disparity and correlation $d_R=k_L-k_R$ we can estimate the right disparity performing a softargmax on the first $W$ dimension of $\mathbf{V}^D_M$:
\small\begin{equation}
    (\hat{\mathbf{D}}_R)_{ik} = \left(\text{softargmax}_R(\mathbf{V}^D_M)\right)_{ik} - k = \sum_{d}^{\frac{W}{4}} d \cdot \frac{e^{(\mathbf{V}^D_M)_{idk}}}{\sum_{f}^{\frac{W}{4}} e^{(\mathbf{V}^D_M)_{ifk}}} - k
    \label{eq:softmax_right}
\end{equation}\normalsize

Disparity maps $\hat{\mathbf{D}}_L$ $\hat{\mathbf{D}}_R$ are used in combination with confidence maps $\hat{\mathbf{C}}_L$ $\hat{\mathbf{C}}_R$ to obtain a robust scaling.
We present an expanded version of the information entropy based confidence estimation (\cref{eq:confidence_left}), with the explicit formula for softmax operator:

\small\begin{equation}
    (\hat{\mathbf{C}}_L)_{ij} = 1  + \frac{\sum_{d}^{\frac{W}{4}} \left(\text{softmax}_L(\mathbf{V}_M^C)\right)_{ijd} \cdot \log_2 \left( \left(\text{softmax}_L(\mathbf{V}_M^C)\right)_{ijd} \right)}{\log_2(\frac{W}{4})} = 1  + \frac{\sum_{d}^{\frac{W}{4}} \frac{e^{(\mathbf{V}^C_M)_{ijd}}}{\sum_{f}^{\frac{W}{4}} e^{(\mathbf{V}^C_M)_{ijf}}} \cdot \log_2 \left( \frac{e^{(\mathbf{V}^C_M)_{ijd}}}{\sum_{f}^{\frac{W}{4}} e^{(\mathbf{V}^C_M)_{ijf}}} \right)}{\log_2(\frac{W}{4})}
    \label{eq:confidence_left2}
\end{equation}\normalsize

In the same way, we estimate right confidence map $\hat{\mathbf{C}}_R$ performing a softmax operation on the first $W$ dimension of $\mathbf{V}^C_M$:

\small\begin{equation}
    (\hat{\mathbf{C}}_R)_{ik} = 1  + \frac{\sum_{d}^{\frac{W}{4}} \left(\text{softmax}_R(\mathbf{V}_M^C)\right)_{idk} \cdot \log_2 \left( \left(\text{softmax}_R(\mathbf{V}_M^C)\right)_{idk} \right)}{\log_2(\frac{W}{4})} = 1 + \frac{\sum_{d}^{\frac{W}{4}} \frac{e^{(\mathbf{V}^C_M)_{idk}}}{\sum_{f}^{\frac{W}{4}} e^{(\mathbf{V}^C_M)_{ifk}}} \cdot \log_2 \left( \frac{e^{(\mathbf{V}^C_M)_{idk}}}{\sum_{f}^{\frac{W}{4}} e^{(\mathbf{V}^C_M)_{ifk}}} \right)}{\log_2(\frac{W}{4})}
    \label{eq:confidence_right}
\end{equation}\normalsize

To improve the robustness of the scaling, we introduce a softLRC operator to classify occlusions as low-confidence pixels and consequentially mask out them from $\hat{\mathbf{C}}_L$ and $\hat{\mathbf{C}}_R$.
We define the softLRC operator as follows:

\small\begin{equation}
    \text{softLRC}_L(\mathbf{D},\mathbf{D}_R) = \frac{\log\left(1+\exp\left(T_\text{LRC}-\left| \mathbf{D}_L - \mathcal{W}_L(\mathbf{D}_L,\mathbf{D}_R) \right|\right)\right)}{\log(1+\exp(T_\text{LRC}))}
    \label{eq:softlrc}
\end{equation}\normalsize
where $T_\text{LRC}=1$ is the LRC threshold and $\mathcal{W}_L(\mathbf{D}_L,\mathbf{D}_R)$ is the warping operator that uses the left disparity $\mathbf{D}_L$ to warp the right disparity $\mathbf{D}_R$ into the left view. 

Finally, we can estimate the scale $\hat{s}$ and shift $\hat{t}$  -- a differentiable weighted least-square approach. We report here the expanded form of \cref{eq:scale_shift}:
\small\begin{equation}
    \min_{\hat{s}, \hat{t}} \left\lVert \sqrt{\hat{\mathbf{C}}_L}\odot\left[\left(\hat{s}\mathbf{M}_L + \hat{t}\right)  - \hat{\mathbf{D}}_L \right] \right\rVert_F + \left\lVert \sqrt{\hat{\mathbf{C}}_R}\odot\left[\left(\hat{s}\mathbf{M}_R + \hat{t}\right)  - \hat{\mathbf{D}}_R \right] \right\rVert_F 
    \label{eq:scale_shift2}
\end{equation}\normalsize
where $\lVert\cdot\rVert_F$ denotes the Frobenius norm.

\subsection{Cost Volume Augmentations}
\label{subsec:cost_aug}

Volume augmentations are necessary when the training set -- \eg, Sceneflow \cite{mayer2016large} -- does not model particularly complex scenarios where a VFM could be useful, for example, when experiencing non-Lambertian surfaces.
Without any augmentation of this kind, the stereo network would simply overlook the additional information from the monocular branch.
As detailed in the main paper, we propose three volume augmentations and a monocular augmentation to overcome this issue.
In this supplementary section, we explain the rationale behind the introduction of each augmentation:

\begin{itemize}
    \item \textit{Volume Rolling}: non-Lambertian surfaces such as mirrors and glasses violate the geometry constraints, leading to a high matching peak in a wrong disparity bin. This augmentation emulates this behavior by shifting some among the matching peaks to a random position: consequentially, \method learns to retrieve the correct peak from the other branch. \\
    \item \textit{Volume Noising} and \textit{Volume Zeroing}: we introduce noise and false peaks into the correlation volume to simulate scenarios with texture-less regions, repeating patterns, and occlusions. \\
    \item \textit{Perfect Monocular Estimation}: instead of acting inside the correlation volumes, we can substitute the prediction of the VFM with a perfect monocular map  -- the ground truth normalized between $[0,1]$. This perfect prediction is noise-free and therefore the monocular branch of \method will likely gain importance during the training process.
\end{itemize}

\subsection{Volume Truncation}
\label{subsec:vol_trunc}

The proposed volume truncation strategy further helps \method to handle mirror surfaces.
Here we introduce additional details about fuzzy operators -- useful to make a boolean expression differentiable -- and the sigmoid curve used to truncate the volume $\mathbf{V}_S$  -- the truncate mask $(\mathbf{T})_{ij} = \left[\left((\hat{\mathbf{M}}_L)_{ij} >(\hat{\mathbf{D}}_L)_{ij}\right) \land (\mathbf{C}_M)_{ij} \right] \lor \left[ (\mathbf{C}_M)_{ij} \land \neg(\hat{\mathbf{C}}_L)_{ij} \right]$.

We can replace operators AND ($\land$), OR ($\lor$), NOT ($\neg$) and GREATER ($>$) inside $\mathbf{T}$ with the fuzzy counterparts $\text{AND}_\text{F}(A,B) = A \cdot B$, $\text{OR}_\text{F}(A,B) = A+B-A \cdot B$, $\text{NOT}_\text{F}(A) = 1- A$ and $\text{GREATER}_\text{F}(A,B) = \sigma(A-B)$, obtaining the fuzzy truncate mask $\mathbf{T}_\text{F}$:

\small\begin{equation}
    \begin{split}
        (\mathbf{T}_\text{F})_{ij} &= (\mathbf{T}^A_\text{F})_{ij} + (\mathbf{T}^B_\text{F})_{ij} - (\mathbf{T}^A_\text{F})_{ij} \cdot (\mathbf{T}^B_\text{F})_{ij}\\        
        (\mathbf{T}^A_\text{F})_{ij} &= (\mathbf{C}_M)_{ij} \cdot \sigma\left( (\hat{M}_L)_{ij} - (\hat{D}_L)_{ij} \right)\\
        (\mathbf{T}^B_\text{F})_{ij} &= (\mathbf{C}_M)_{ij} \cdot \left(1-(\hat{\mathbf{C}}_L)_{ij}\right)
    \end{split}
    \label{eq:truncate_mask_fuzzy}
\end{equation}\normalsize

where $\mathbf{T}^A_\text{F}$ and $\mathbf{T}^B_\text{F}$ are respectively the left section and the right section of the $\text{OR}_\text{F}$ of mask $\textbf{T}_\text{F}$.
Next, we can apply threshold $T_m$ to achieve the final fuzzy mask $\mathbf{T}'_\text{F}$ as follows:

\small\begin{equation}
    (\mathbf{T}'_\text{F})_{ij}=\sigma\left((\mathbf{T}_\text{F})_{ij}-T_m\right)
    \label{eq:truncate_mask_fuzzy_thresholded}
\end{equation}\normalsize.

Finally, we can use the fuzzy truncate mask $\mathbf{T}'_\text{F}$ and the scaled monocular map $\hat{\mathbf{M}}_L$ to generate the sigmoid-based truncation volume $\mathbf{V}_T$:

\small\begin{equation}
    (\mathbf{V}_T)_{ijk} = \left(1-(\mathbf{T}'_\text{F})_{ij}\right) + (\mathbf{T}'_\text{F})_{ij} \cdot \left[ \sigma\left(j - (\hat{\mathbf{M}}_L)_{ij} - k\right) \cdot (1-G) + G \right]
    \label{eq:truncate_vol}
\end{equation}\normalsize
where $G=0.9$ attenuates the impact of the truncation. 
The correlation volume $\mathbf{V}_S$ is truncated through an element-wise product with $\mathbf{V}_T$.

\subsection{Training Supervision}
\label{subsec:training}

We supervise the iterative module  -- the L1 loss with exponentially increasing weights \cite{lipson2021raft}:

\small\begin{equation}
    \mathcal{L}_\text{A} = \sum_{l=1}^L{\gamma^{L-l}\lVert \mathbf{D}^l-\mathbf{D}_\text{Lgt} \rVert_1}
    \label{eq:loss_raft}
\end{equation}\normalsize
where L is the total number of iterations made by the update operator and $\mathbf{D}_\text{Lgt}$ is the ground truth of the left disparity map.
Furthermore, we supervise the outputs $\hat{\mathbf{D}}_L$, $\hat{\mathbf{D}}_R$, $\hat{\mathbf{M}}_L$, $\hat{\mathbf{M}}_R$, $\hat{\mathbf{C}}_L$, $\hat{\mathbf{C}}_R$ of the monocular branch -- respectively L1 loss and normal loss for $\hat{\mathbf{D}}_L$ $\hat{\mathbf{D}}_R$, L1 loss for $\hat{\mathbf{M}}_L$ $\hat{\mathbf{M}}_R$ and Binary Cross Entropy (BCE) loss for $\hat{\mathbf{C}}_L$ $\hat{\mathbf{C}}_R$:

\small\begin{equation}
    \mathcal{L}_\text{B} = \lVert \hat{\mathbf{D}}_L - \mathbf{D}_\text{Lgt} \rVert_1 + \psi \left\lVert \mathbf{1} - \nabla_L\cdot\hat{\nabla}_L \right\rVert_1 \quad \left( \nabla_L\cdot\hat{\nabla}_L \right)_{ij} = \sum_h (\nabla_L)_{hij} \cdot (\hat{\nabla}_L)_{hij}
    \label{eq:loss_coarse_disp_left}
\end{equation}\normalsize
\small\begin{equation}
    \mathcal{L}_\text{C} = \lVert \hat{\mathbf{D}}_R - \mathbf{D}_\text{Rgt} \rVert_1 + \psi \left\lVert \mathbf{1} - \nabla_R\cdot\hat{\nabla}_R \right\rVert_1 \quad \left( \nabla_R\cdot\hat{\nabla}_R \right)_{ik} = \sum_h (\nabla_L)_{hik} \cdot (\hat{\nabla}_L)_{hik}
    \label{eq:loss_coarse_disp_right}
\end{equation}\normalsize
\small\begin{equation}
    \mathcal{L}_\text{D} = \lVert \hat{\mathbf{M}}_L - \mathbf{D}_\text{Lgt} \rVert_1  \quad \mathcal{L}_\text{E} = \lVert \hat{\mathbf{M}}_R - \mathbf{D}_\text{Rgt} \rVert_1 
    \label{eq:loss_scaled_disp}
\end{equation}\normalsize
\small\begin{equation}
    \mathcal{L}_\text{F} = \text{BCE}(\hat{\mathbf{C}}_L,\mathbf{C}_\text{Lgt}), \quad (\mathbf{C}_\text{Lgt})_{ij} = \frac{\log\left(1+\exp\left(T_\text{LRC}-\left| (\hat{\mathbf{D}}_L)_{ij} - (\mathbf{D}_\text{Lgt})_{ij}  \right|\right)\right)}{\log(1+\exp(T_\text{LRC}))}
    \label{eq:loss_coarse_conf_left}
\end{equation}\normalsize
\small\begin{equation}
    \mathcal{L}_\text{G} = \text{BCE}(\hat{\mathbf{C}}_R,\mathbf{C}_\text{Rgt}), \quad (\mathbf{C}_\text{Rgt})_{ik} = \frac{\log\left(1+\exp\left(T_\text{LRC}-\left| (\hat{\mathbf{D}}_R)_{ik} - (\mathbf{D}_\text{Rgt})_{ik}  \right|\right)\right)}{\log(1+\exp(T_\text{LRC}))}
    \label{eq:loss_coarse_conf_right}
\end{equation}\normalsize
where $\psi=10$ is the normal loss weight, $\mathbf{D}_\text{Rgt}$ is the ground truth of the right disparity map, $\hat{\nabla}_L$ $\hat{\nabla}_R$ are the normal maps estimated respectively from $\hat{\mathbf{D}}_L$ $\hat{\mathbf{D}}_R$, $\nabla_L\cdot\hat{\nabla}_L$ and $\nabla_R\cdot\hat{\nabla}_R$ are the dot product between normal maps, and $\mathbf{C}_\text{Lgt}$ $\mathbf{C}_\text{Rgt}$ are the confidence ground truth.
The final supervision loss $\mathcal{L}$ is the sum of all previous partial losses:
\small\begin{equation}
    \mathcal{L} = \mathcal{L}_A + \mathcal{L}_B + \mathcal{L}_C + \mathcal{L}_D + \mathcal{L}_E + \mathcal{L}_F + \mathcal{L}_G
    \label{eq:total_loss}
\end{equation}\normalsize

\section{Additional Ablation Studies}
\label{sec:ablation}
In this section, we report additional studies concerning the performance of \method.

\subsection{Generalization to Different VFMs}
\label{subsec:ablation_VFM}

In the main paper, we assumed Depth Anything v2 \cite{depth_anything_v2} as the VFM fueling \method, since it is the latest state-of-the-art model being published at the time of this submission. 
However, any VFM for monocular depth estimation would be suitable for this purpose, either current or future ones. To confirm this argument, we conducted some experiments by replacing Depth Anything v2 with other VFMs that appeared on arXiv in the last months, yet that are not been officially published. Among them, we select DepthPro \cite{depthpro}, MoGe \cite{wang2024moge} and Lotus \cite{he2024lotus}.

Table \ref{tab:morevfm2} shows the results achieved by \method variants -- different VFMs on Booster and LayeredFlow. We can appreciate how the different flavors of \method always outperform the baseline stereo model \cite{lipson2021raft}. In general, Depth Anything v2 remains the best choice to deal with non-Lambertian surfaces, with DepthPro allowing for small improvements on some metrics over the LayeredFlow dataset.

\begin{table*}[ht]
\centering
\renewcommand{\tabcolsep}{14pt}
\scalebox{0.78}{
\begin{tabular}{|l||rrrrr|rrrr|}
\multicolumn{1}{c}{} & \multicolumn{5}{c}{Booster (Q)} & \multicolumn{4}{c}{LayeredFlow (E)} \\
\hline
 \multirow{2}{*}{Model} & \multicolumn{4}{c}{Error Rate (\%)} & Avg. & \multicolumn{3}{c}{Error Rate (\%)} & Avg. \\
  & $>2$ & $>4$ & $>6$ & $>8$ & (px) & $>1$ & $>3$ & $>5$ &(px) \\
\hline\hline
{Baseline} \cite{lipson2021raft}
&  17.84 &  13.06 &  10.76 &  9.24 &  3.59
&  89.21 &  79.02 & 71.61 &  19.27 \\
{\method{}  -- DAv2 \cite{depth_anything_v2}}
& 9.01 & 5.40 & 4.12 & 3.34 & 1.21
& 81.83 & 57.66 & 45.12 & 11.20 \\
{\method{}  -- DepthPro \cite{depthpro}}
& 10.53 & 7.02 & 5.79 & 5.13 & 2.40
& 78.76 & 61.11 & 51.04 & 14.43 \\
{\method{}  -- MoGe \cite{wang2024moge}}
& 9.47 & 5.77 & 4.49 & 3.84 & 1.44
& 84.27 & 68.67 & 58.89 & 16.22 \\
{\method{}  -- Lotus \cite{he2024lotus}}
& 12.44 & 8.71 & 7.58 & 6.98 & 3.21
& 86.04 & 62.75 & 49.47 & 13.98 \\
\hline

\end{tabular}}\vspace{-0.2cm}
\caption{\textbf{Non-Lambertian Generalization of \method w.r.t VFMs.} We measure the impact of different monocular depth estimation networks. Networks trained on SceneFlow \cite{mayer2016large}.
}
\label{tab:morevfm2}
\end{table*}

Table \ref{tab:morevfm1} shows the results achieved by the different VFMs on the zero-shot generalization benchmark. Also in this case, we can appreciate how any \method variant yields comparable accuracy, with some VFMs like Moge yielding some improvements over Depth Anything v2 on ETH3D, KITTI 2012 and 2015 at the expense of lowering the accuracy on Middlebury 2014 and 2021. Interestingly, we can observe an important drop in accuracy by using DepthPro on Middlebury 2021, due to several failures by the model itself on the scenes of this dataset.

\begin{table*}[ht]
\centering
\renewcommand{\tabcolsep}{6pt}
\scalebox{0.62}{
\begin{tabular}{|l||rrrr|rrrr|rrrr|rrrr|rrrr|}
\multicolumn{1}{c}{} & \multicolumn{4}{c}{Middlebury 2014 (H)} & \multicolumn{4}{c}{Middlebury 2021} & \multicolumn{4}{c}{ETH3D} & \multicolumn{4}{c}{KITTI 2012} & \multicolumn{4}{c}{KITTI 2015} \\
\hline
 \multirow{2}{*}{Model} & \multicolumn{3}{c}{bad $>2$} & Avg. & \multicolumn{3}{c}{bad $>2$} & Avg. & \multicolumn{3}{c}{bad $>1$} & Avg. & \multicolumn{3}{c}{bad $>3$} & Avg. & \multicolumn{3}{c}{bad $>3$} & Avg. \\
  & All & Noc & Occ & (px) & All & Noc & Occ & (px) & All & Noc & Occ & (px) & All & Noc & Occ & (px) & All & Noc & Occ & (px) \\
\hline\hline
{Baseline} \cite{lipson2021raft} 
& 11.15 & 8.06 & 29.06 &  1.55 
& 12.05 & 9.38 & 37.89 & 1.81 
& 2.59 & 2.24 &  8.78 & 0.25 
& 4.80 & 4.23 & 29.21 & 0.89 
& 5.44 & 5.21 & 14.09 & 1.16 \\
\hline\hline
{\method{}  -- DAv2 \cite{depth_anything_v2}} 
&6.96 & 4.75 & 20.34 & 0.94
&7.97 &  5.71 & 29.52 & 1.08 
&1.66 & 1.43 & 5.29 &  0.24 
&3.90 & 3.52 & 21.65 & 0.83 
&3.93 & 3.79 & 11.01 & 0.97 \\
{\method{}  -- DepthPro \cite{depthpro}} 
& 6.58 & 4.32 & 20.05 & 0.99 
& 15.13 & 12.52 & 41.16 & 8.97 
& 2.74 & 2.54 & 6.09 & 0.44 
& 3.13 & 2.25 & 18.25 & 0.75 
& 3.79 & 3.10 & 10.53 & 0.95 \\
{\method{}  -- MoGe \cite{wang2024moge}} 
& 7.79 & 5.23 & 22.86 & 1.21 
& 9.86 & 7.30 & 33.48 & 1.28 
& 1.28 & 1.09 & 3.78 & 0.21 
& 2.85 & 2.00 & 17.40 & 0.73 
& 3.22 & 2.57 & 8.97 & 0.89 \\
{\method{}  -- Lotus \cite{he2024lotus}} 
& 7.35 & 4.96 & 21.71 & 1.07 
& 9.62 & 7.01 & 34.92 & 1.29 
& 2.68 & 2.44 & 6.04 & 0.31 
& 4.54 & 3.58 & 22.71 & 0.92 
& 3.88 & 3.21 & 10.36 & 0.95 \\

\hline
\end{tabular}}\vspace{-0.2cm}
\caption{\textbf{Generalization of \method w.r.t VFMs.} We measure the impact of different monocular depth estimation networks. Networks trained on SceneFlow \cite{mayer2016large}.
}
\label{tab:morevfm1}
\end{table*}

\begin{figure*}[h]
    \centering 
    \renewcommand{\tabcolsep}{1pt}
    \begin{tabular}{cccccc}

    \multirow{2}{*}{\small RGB} & \multirow{2}{*}{\small RAFT-Stereo \cite{lipson2021raft}} & \small \method & \small \method & \small \method & \small \method \\

     &  & \small -- DAv2 \cite{depth_anything_v2} & \small -- DepthPro \cite{depthpro} & \small -- MoGe \cite{wang2024moge} & \small -- Lotus \cite{he2024lotus} \\

    \includegraphics[width=0.16\linewidth]{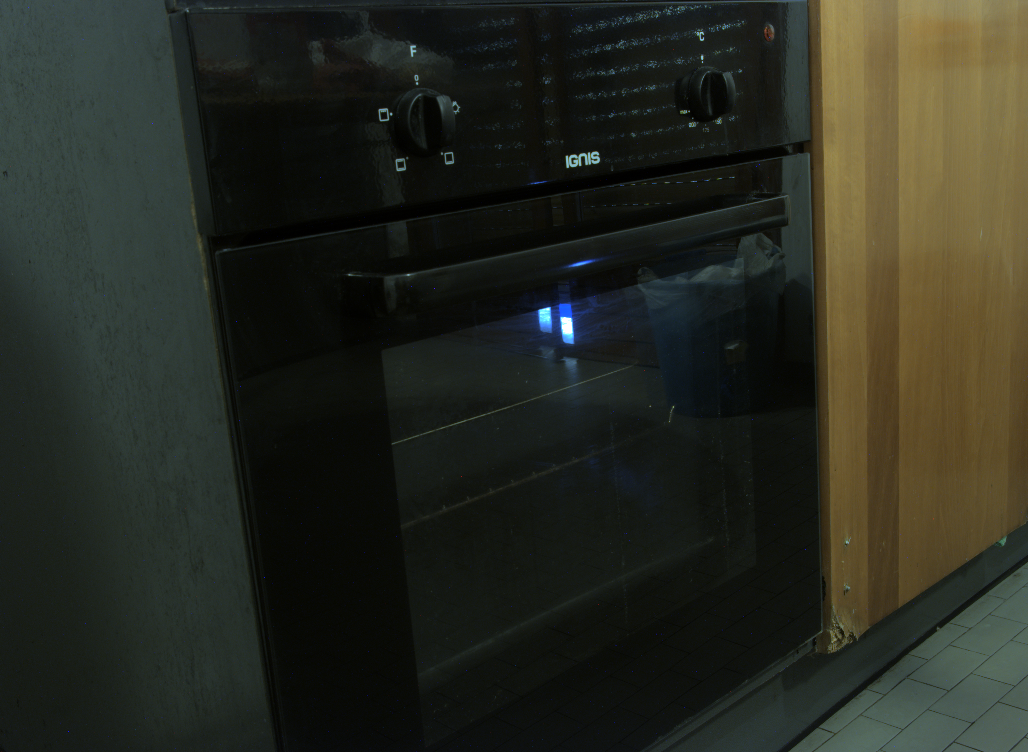} & 
    \includegraphics[width=0.16\linewidth]{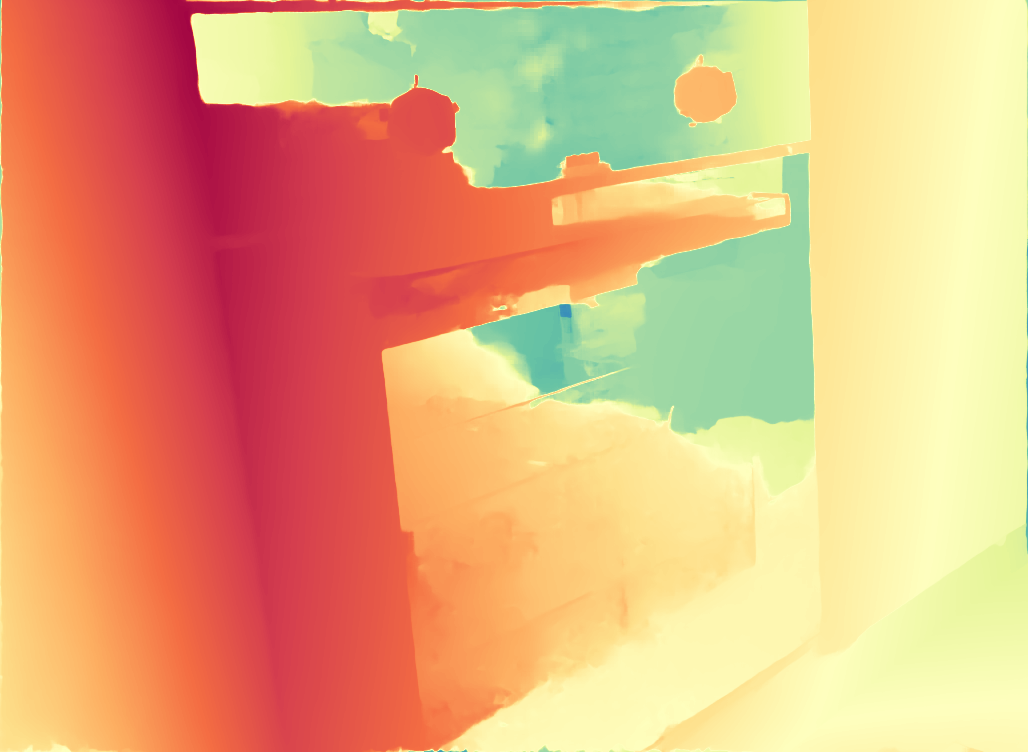} & 
    \includegraphics[width=0.16\linewidth]{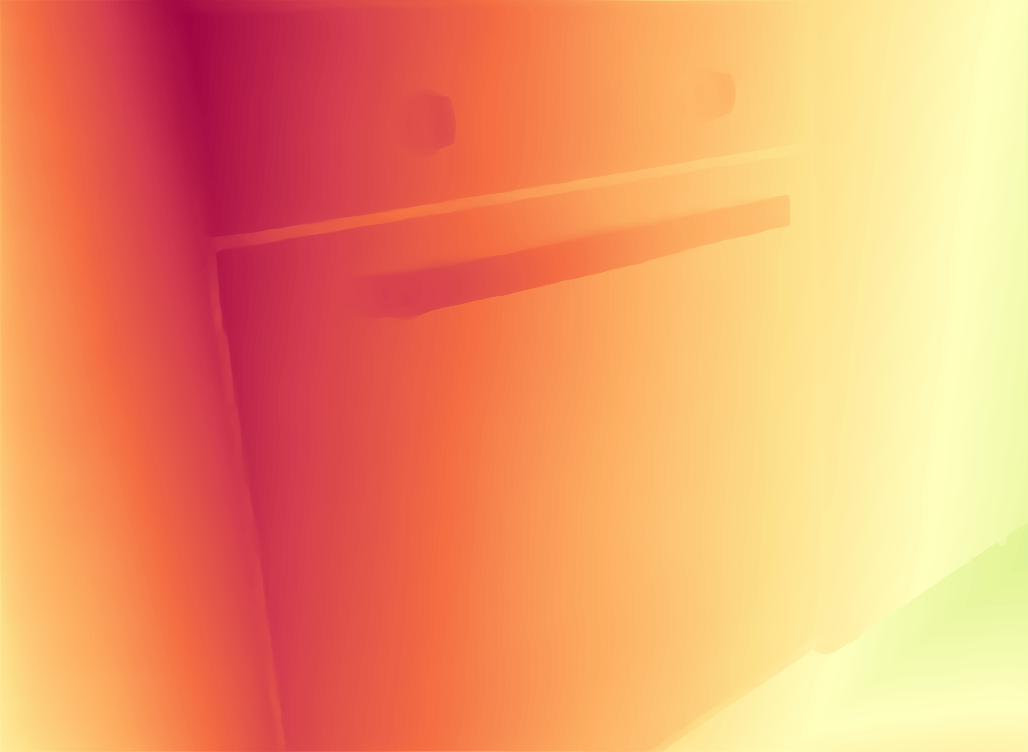} & 
    \includegraphics[width=0.16\linewidth]{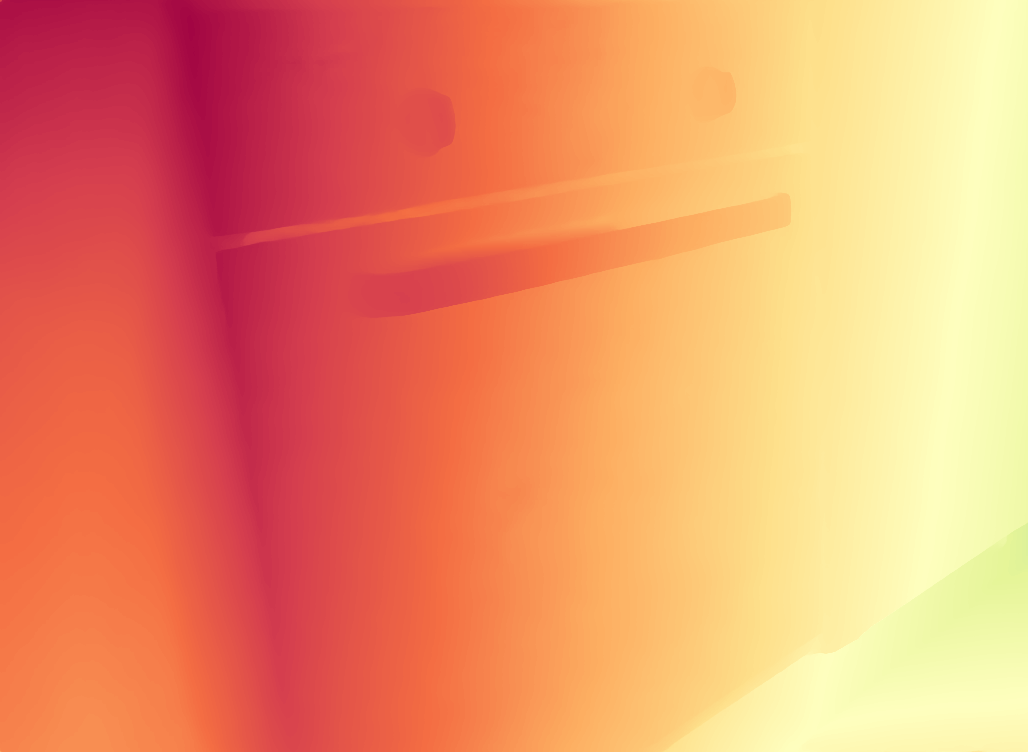} & 
    \includegraphics[width=0.16\linewidth]{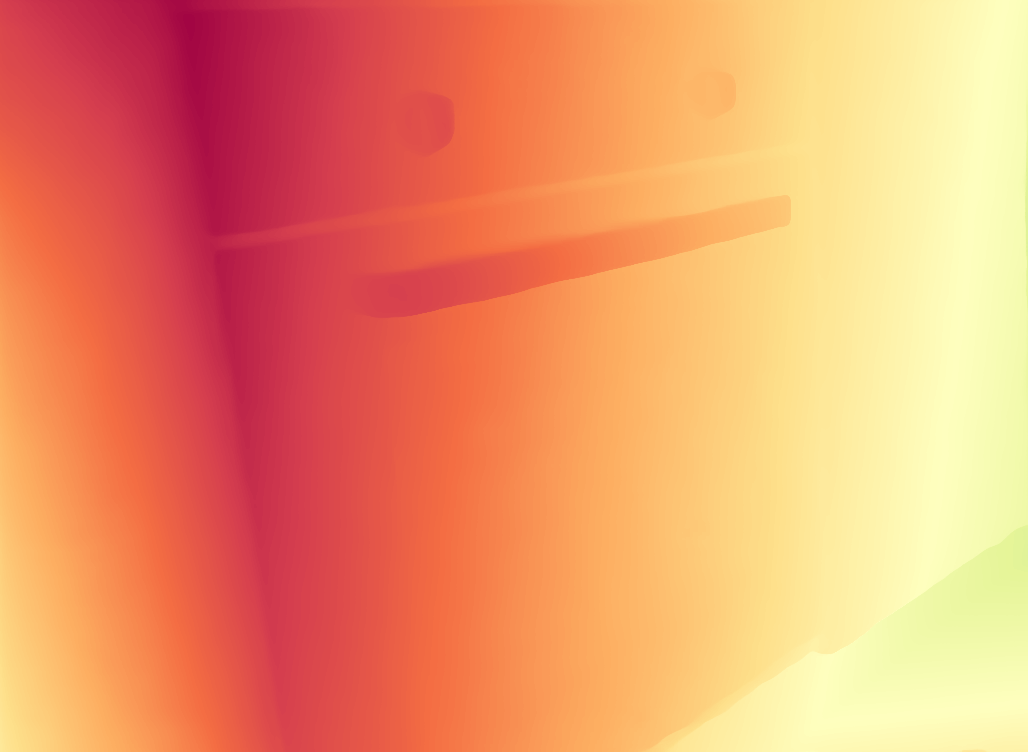} & 
    \includegraphics[width=0.16\linewidth]{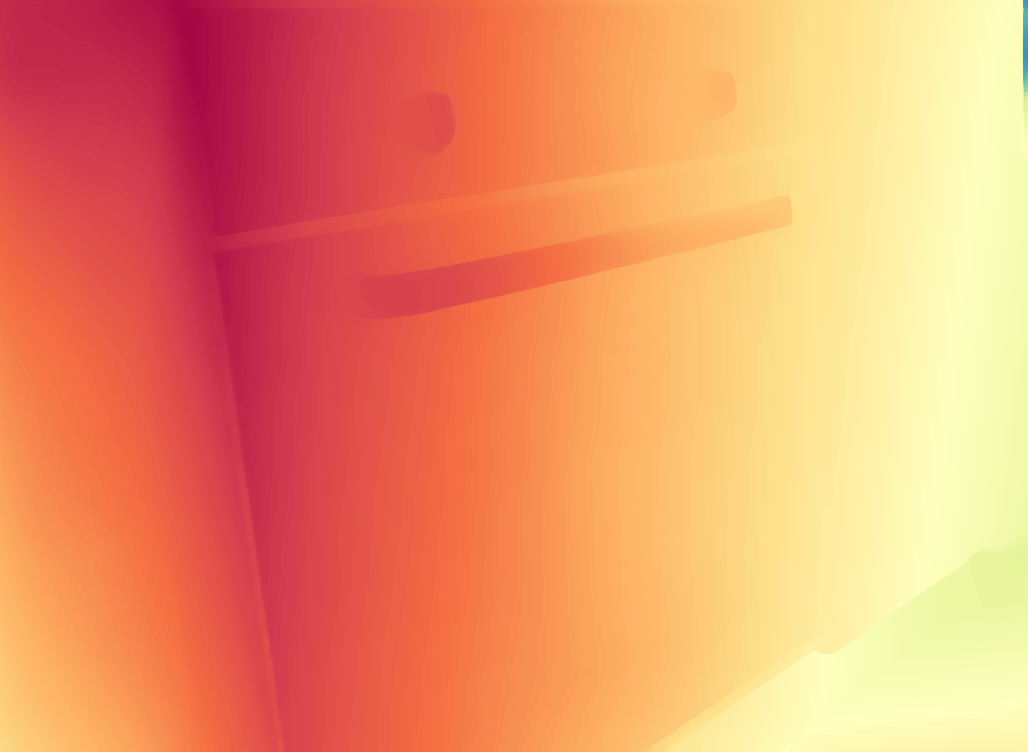} \\ 

    \includegraphics[width=0.16\linewidth]{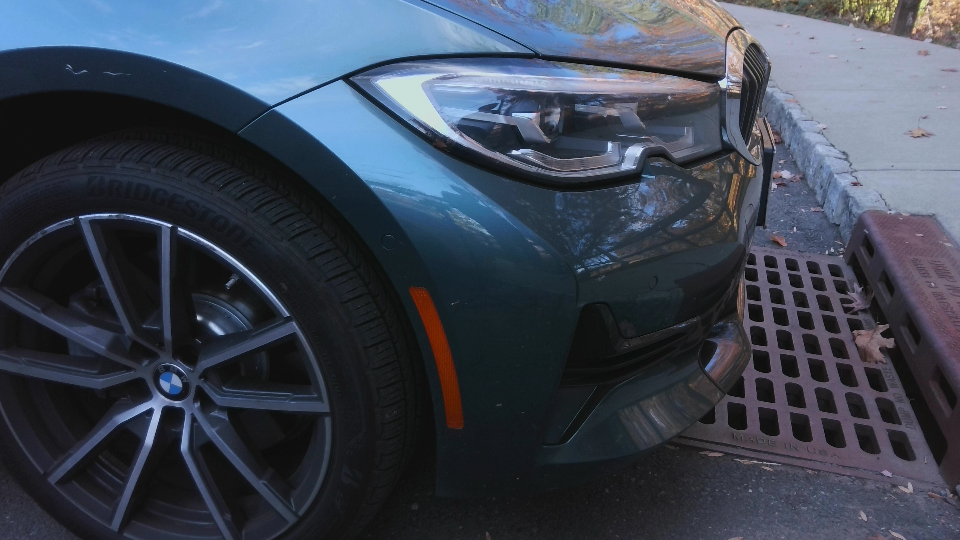} &
    \includegraphics[width=0.16\linewidth]{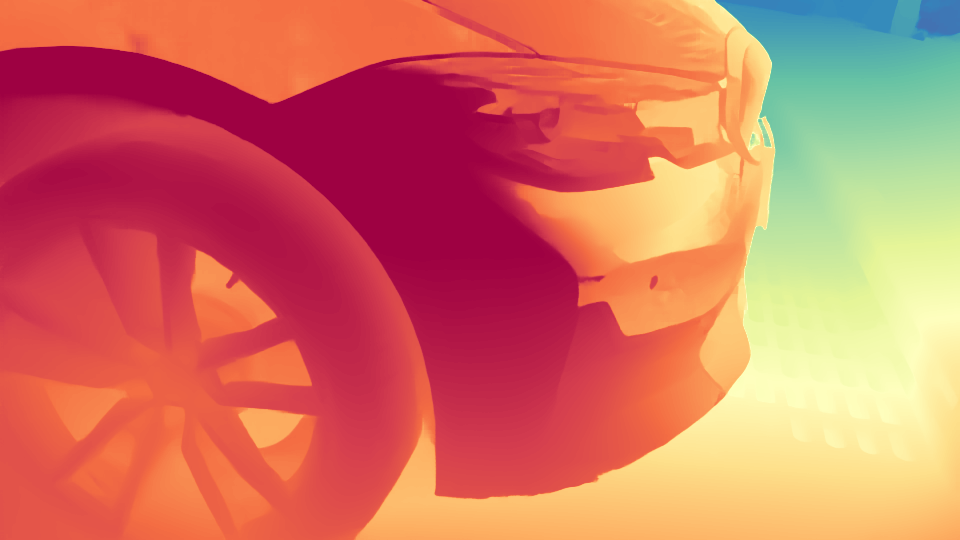} &
    \includegraphics[width=0.16\linewidth]{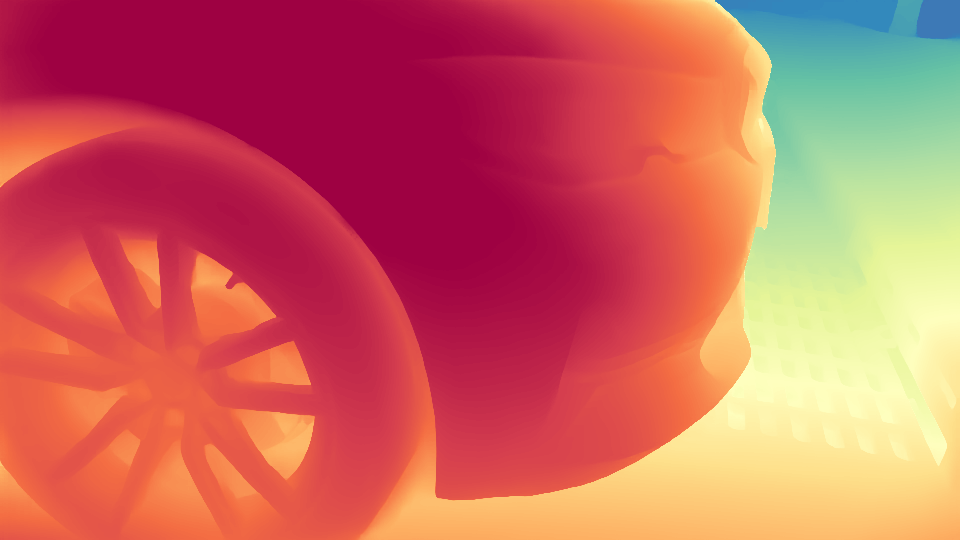} &
    \includegraphics[width=0.16\linewidth]{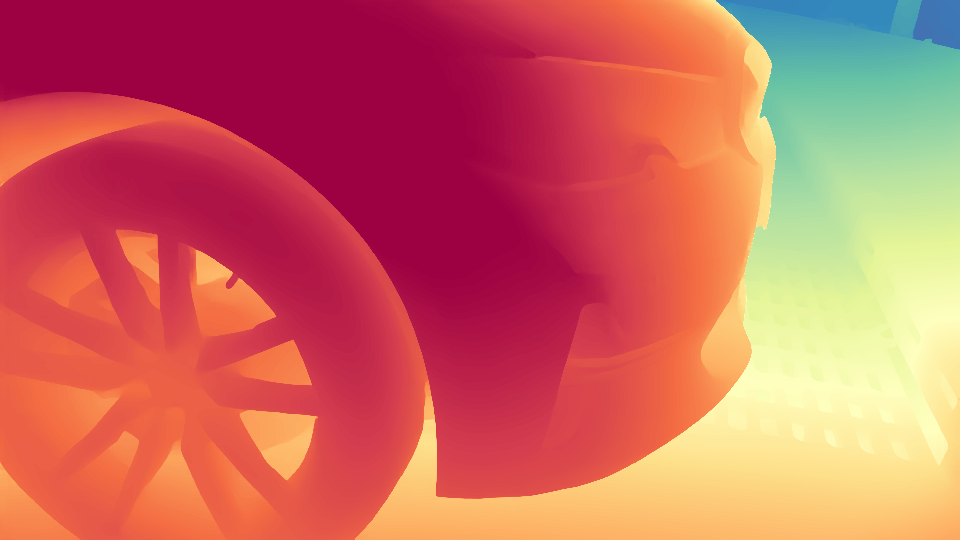} &
    \includegraphics[width=0.16\linewidth]{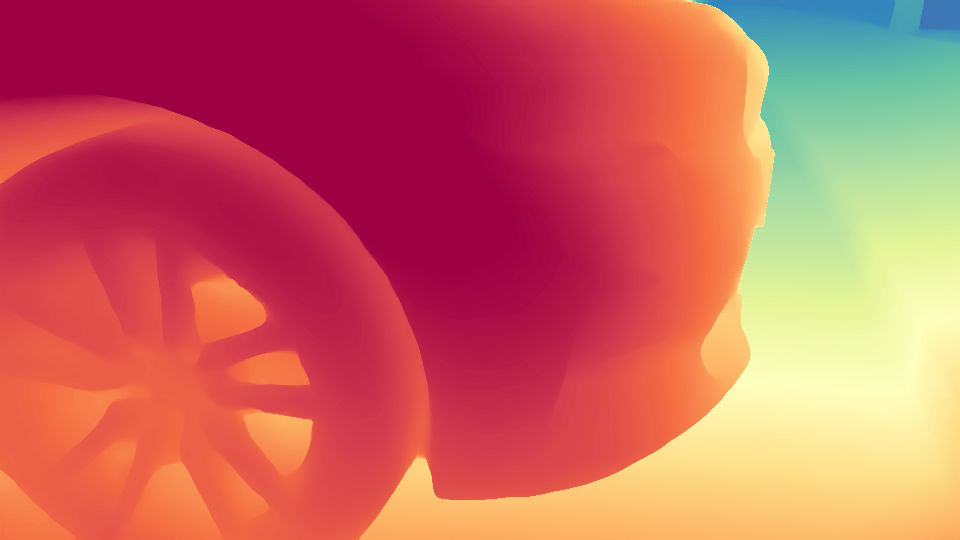} &
    \includegraphics[width=0.16\linewidth]{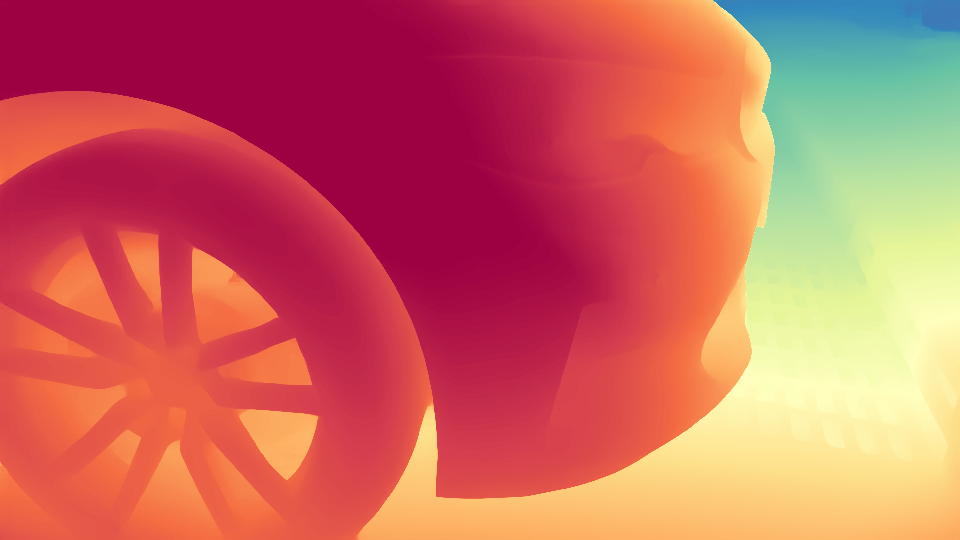} \\
    
    \end{tabular}\vspace{-0.3cm}
    \caption{\textbf{Qualitative Results -- Booster and LayeredFlow.} Predictions by RAFT-Stereo and \method{} -- different VFMs.}
    \label{fig:multiple_vfms}\vspace{-0.3cm}
\end{figure*}

Finally, Figure \ref{fig:multiple_vfms} shows qualitative results obtained by the different variants of \method, highlighting only minor differences among the different predictions.

\subsection{Impact of Cost Volume Truncation}
\label{subsec:vol_trunc_qual}

\begin{figure*}[h]
    \centering 
    \renewcommand{\tabcolsep}{1pt}
    \begin{tabular}{ccc}

    \multirow{2}{*}{\small RGB} & \small \method & \small \method \\
    & \small w/o volume truncation & \small w/ volume truncation \\
    \includegraphics[width=0.3\linewidth]{imgs/booster/rgb/19.png} &
    \includegraphics[width=0.3\linewidth]{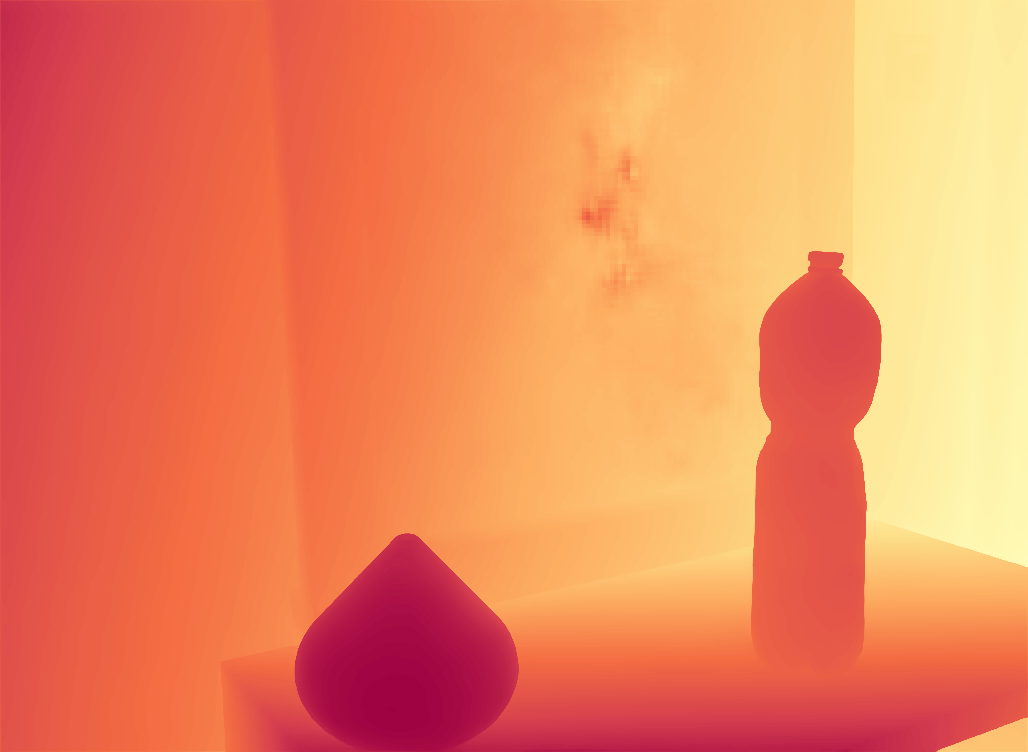} &
    \includegraphics[width=0.3\linewidth]{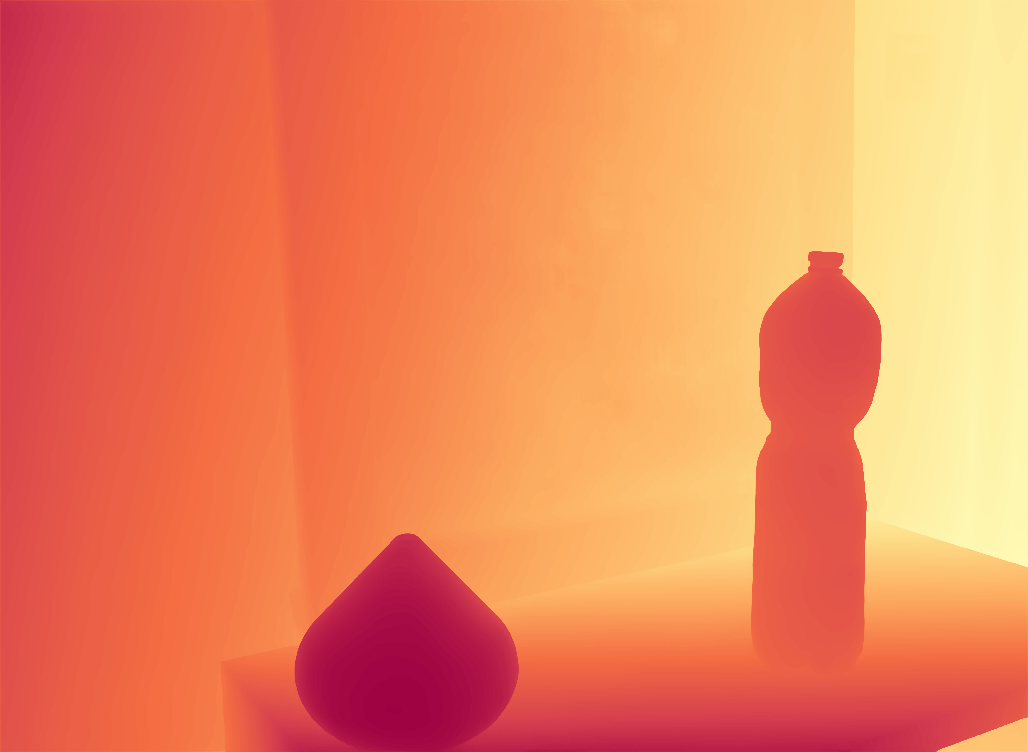} \\

    \end{tabular}\vspace{-0.3cm}
    \caption{\textbf{Qualitative Results -- Volume Truncation.} Predictions by \method{}.}
    \label{fig:truncation}\vspace{-0.3cm}
\end{figure*}

Cost volume truncation is a specific augmentation we apply to improve the results in the presence of mirrors. Figure \ref{fig:truncation} shows a qualitative example of predictions by \method (using Depth Anything v2) obtained by either not applying or by applying such augmentation. 
While \method alone cannot entirely restore the surface of the mirror starting from the priors provided by the VFM, applying cost volume truncation allows for predicting a much smoother and consistent surface.

\subsection{Runtime \& Memory Consumption Analysis}
\label{subsec:runtime}
Table \ref{tab:runtime} reports the processing time (in seconds) and memory consumption (in GB) required by \method during inference, comparing it with the baseline stereo backbone, RAFT-Stereo.
We measure the runtime on a single A100 GPU, repeating the experiment with three different input resolutions, specifically $256\times256$, $512\times512$, and $1024\times1024$, as well as by deploying the different VFMs studied before to fuel \method{} -- specifically, for each variant we report standalone runtime and memory usage by the VFM and the stereo backbone separately, as well as their sum. 

Concerning runtime, Depth Anything v2 is the fastest among the VFMs, taking about 30ms to process a single image at any resolution, with Moge requiring more than $10\times$ the time for a single inference when processing 1Mpx images. 
The stereo backbone requires about 50\% additional time compared to the baseline, RAFT-Stereo \cite{lipson2021raft}, because of the additional branch deployed to process the depth maps by the VFM.

For what concerns memory consumption, once again Depth Anything v2 is the most efficient among the VFMs, requiring as few as 2GB, with Moge sharing similar requirements. Our stereo backbone introduces additional memory consumption because of the second branch processing monocular cues: this overhead is negligible with $256\time256$ images, raising to about $2\times$ the memory required by RAFT-Stereo alone when dealing with 1Mpx images.

\begin{table}[ht]
\centering
\renewcommand{\tabcolsep}{6pt}
\scalebox{0.9}{
\begin{tabular}{|lll|rrr|rrr|}
\hline
  Image Size & Stereo Model Name & VFM Name & \multicolumn{3}{c|}{Processing Time (s)} & \multicolumn{3}{c|}{Memory Consumption (GB)} \\
  $(H \times W)$ &  &  & VFM & Stereo & Total & VFM & Stereo & Total \\
\hline\hline
\multirow{4}{*}{$256 \times 256$} & \multirow{4}{*}{\textbf{\method (ours)}} & DAv2 \cite{depth_anything_v2} & 0.03 & 0.15 & 0.18 & 0.57 & 0.18 & 0.76 \\
 & & DepthPro \cite{depthpro} & 0.21 & 0.15 & 0.36 & 1.92 & 0.18 & 2.09 \\
 & & MoGe \cite{wang2024moge} & 0.38 & 0.15 & 0.52 & 0.38 & 0.19 & 0.57 \\
 & & Lotus \cite{he2024lotus} & 0.13 & 0.15 & 0.29 & 0.22 & 0.18 & 0.41 \\
 \hline
 $256\times256$ & RAFT-Stereo \cite{lipson2021raft} & - & - & 0.10 & 0.10  & - & 0.17 & 0.17 \\
\hline\hline
 \multirow{4}{*}{$512\times512$} & \multirow{4}{*}{\textbf{\method (ours)}} & DAv2 \cite{depth_anything_v2} & 0.03 & 0.21 & 0.24 & 0.57 & 0.77 & 1.34 \\
 &  & DepthPro \cite{depthpro} & 0.20 & 0.21 & 0.41 & 1.84 & 0.77 & 2.60 \\
 &  & MoGe \cite{wang2024moge} & 0.38 & 0.21 & 0.59 & 0.38 & 0.78 & 1.17 \\
 &  & Lotus \cite{he2024lotus} & 0.16 & 0.22 & 0.38 & 0.85 & 0.77 & 1.62 \\
 \hline
 $512\times512$ & RAFT-Stereo \cite{lipson2021raft} & - & - & 0.14 & 0.14 & - & 0.66 & 0.66 \\
\hline\hline
 \multirow{4}{*}{$1024\times1024$} & \multirow{4}{*}{\textbf{\method (ours)}} & DAv2 \cite{depth_anything_v2} & 0.03 & 0.61 & 0.63 & 0.58 & 5.73 & 6.31 \\
 &  & DepthPro \cite{depthpro} & 0.21 & 0.61 & 0.82 & 1.85 & 5.73 & 7.59 \\
 &  & MoGe \cite{wang2024moge} & 0.38 & 0.60 & 0.98 & 0.42 & 5.77 & 6.19 \\
 &  & Lotus \cite{he2024lotus} & 0.49 & 0.61 & 1.10 & 3.40 & 5.73 & 9.13 \\
 \hline
 $1024\times1024$ & RAFT-Stereo \cite{lipson2021raft} & - & - & 0.36 & 0.36 & - & 2.63 & 2.63 \\
\hline
\end{tabular}}\vspace{-0.2cm}
\caption{\textbf{Runtime \& Memory Consumption Analysis.} 
}\vspace{-0.3cm}
\label{tab:runtime}
\end{table}

\clearpage

\section{Qualitative Results}
\label{subsec:qual}

We conclude with additional qualitative results by \method on the different datasets involved in our experiments.

\begin{figure*}[h]
    \centering 
    \renewcommand{\tabcolsep}{1pt}
    \begin{tabular}{cc}

        \small RGB &
        \small RAFT-Stereo \cite{lipson2021raft} \\
        \includegraphics[width=0.48\textwidth]{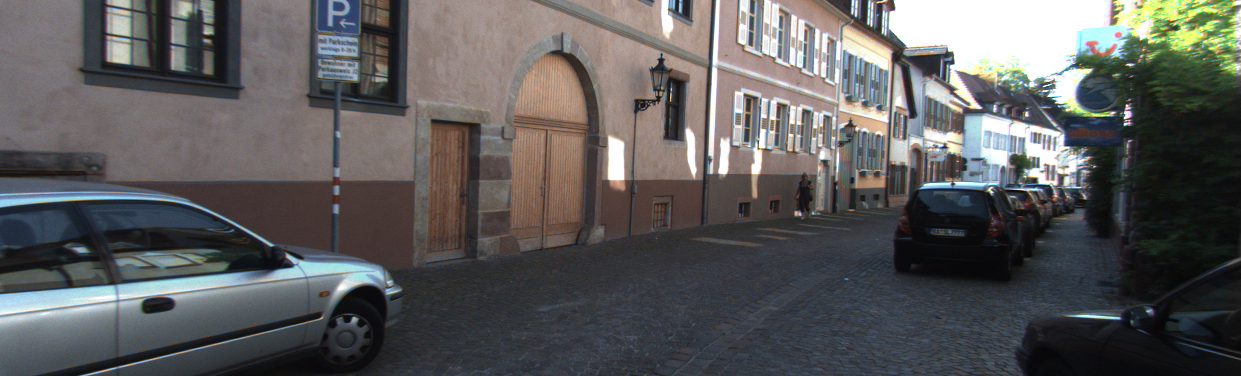} & 
        \includegraphics[width=0.48\textwidth]{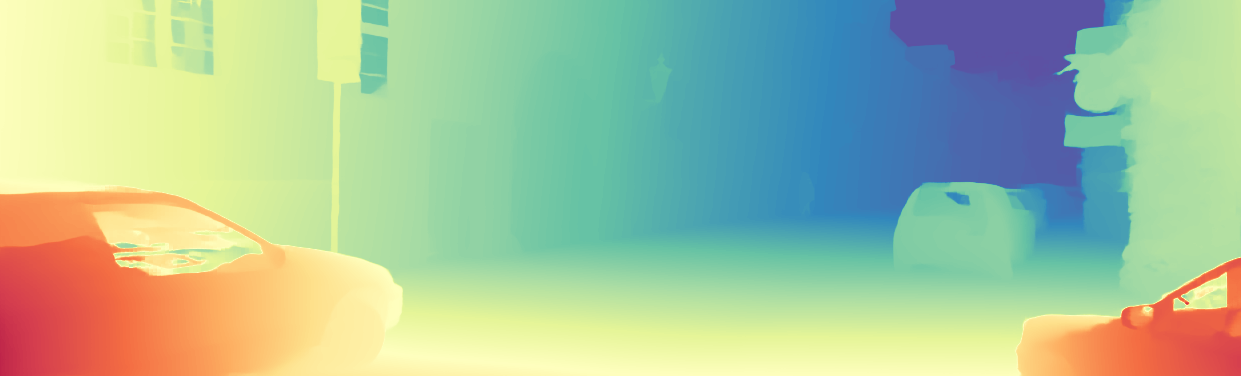} \\
        \small DLNR \cite{zhao2023high} &
        \small NMRF \cite{guan2024neural} \\
        \includegraphics[width=0.48\textwidth]{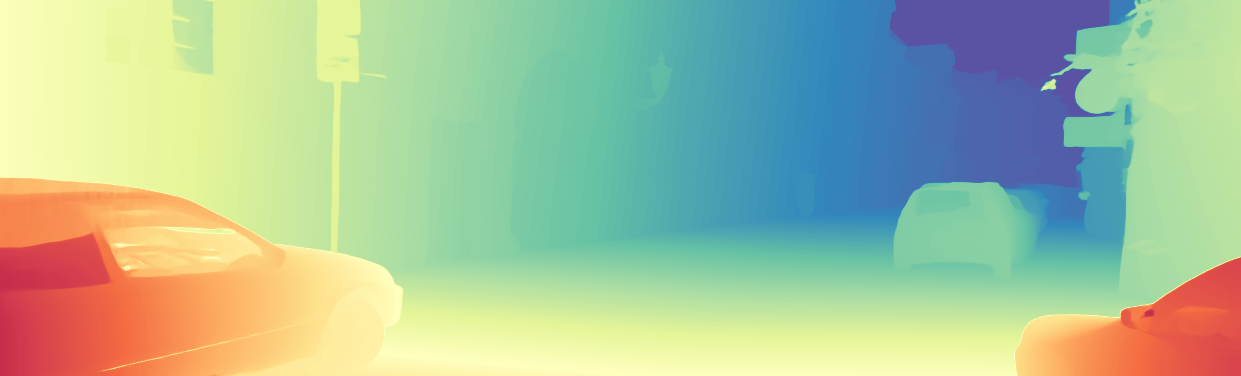} &
        \includegraphics[width=0.48\textwidth]{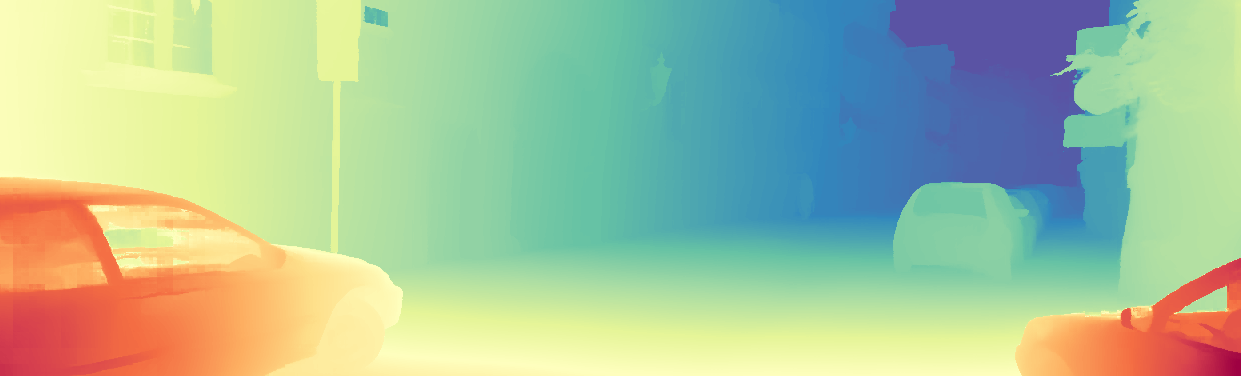} \\ 
        \small Selective-IGEV \cite{wang2024selective} &
        \textbf{\method (ours)} \\
        \includegraphics[width=0.48\textwidth]{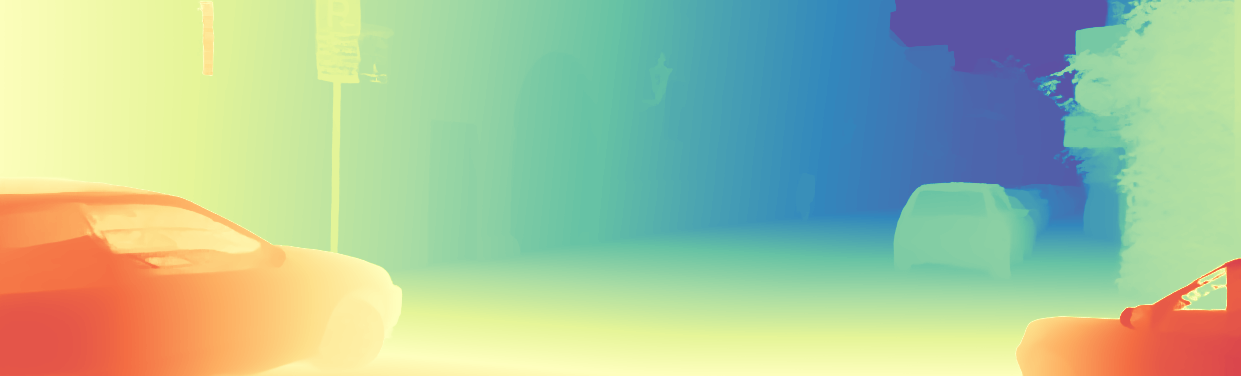} &
        \includegraphics[width=0.48\textwidth]{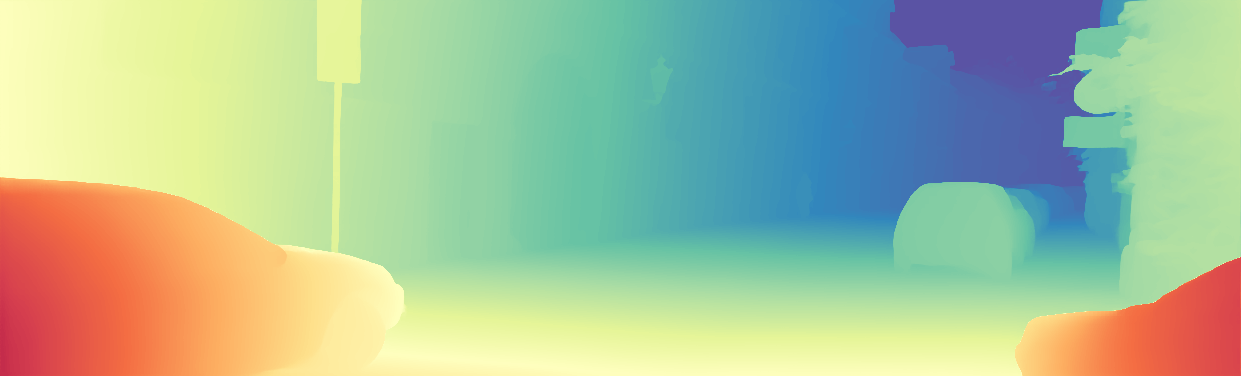} \\ \\

        \small RGB &
        \small RAFT-Stereo \cite{lipson2021raft} \\
        \includegraphics[width=0.48\textwidth]{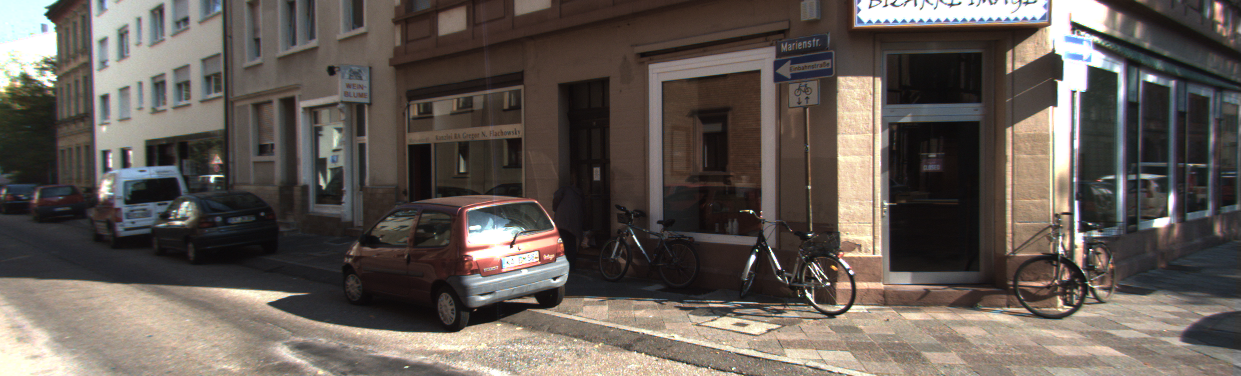} & 
        \includegraphics[width=0.48\textwidth]{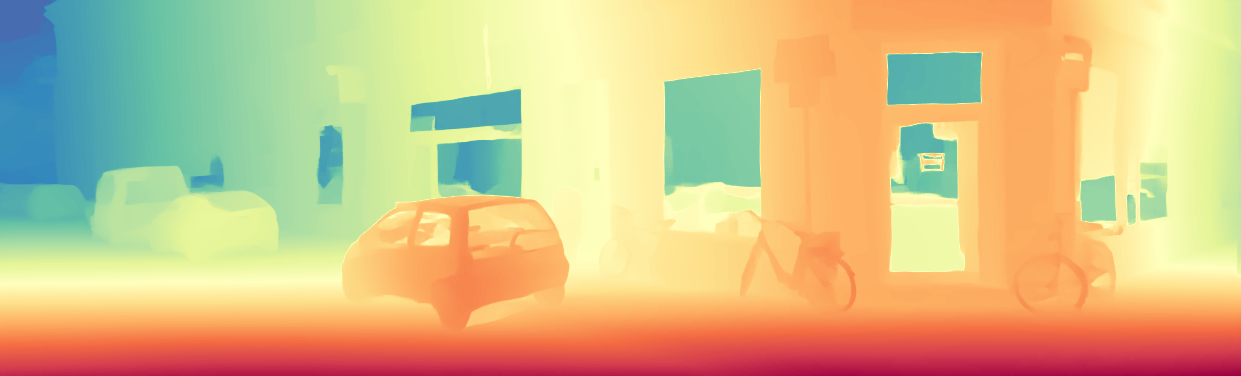} \\
        \small DLNR \cite{zhao2023high} &
        \small NMRF \cite{guan2024neural} \\
        \includegraphics[width=0.48\textwidth]{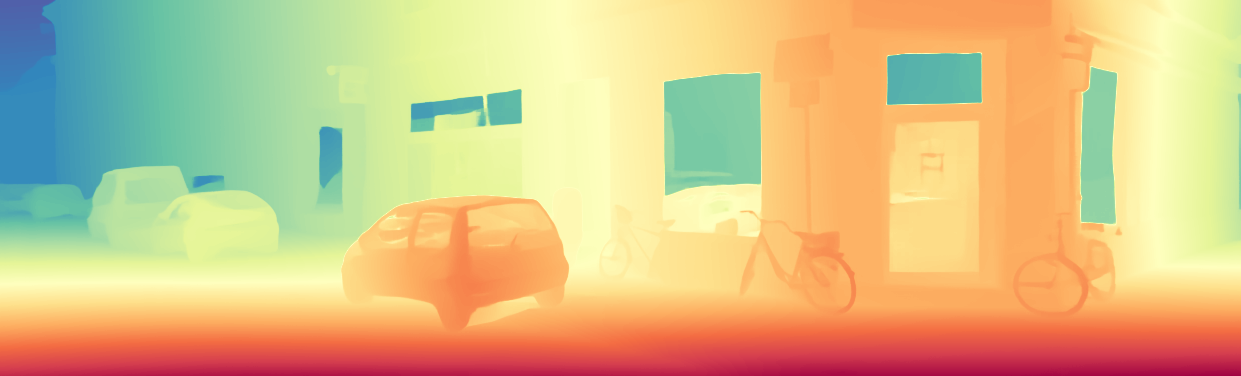} &
        \includegraphics[width=0.48\textwidth]{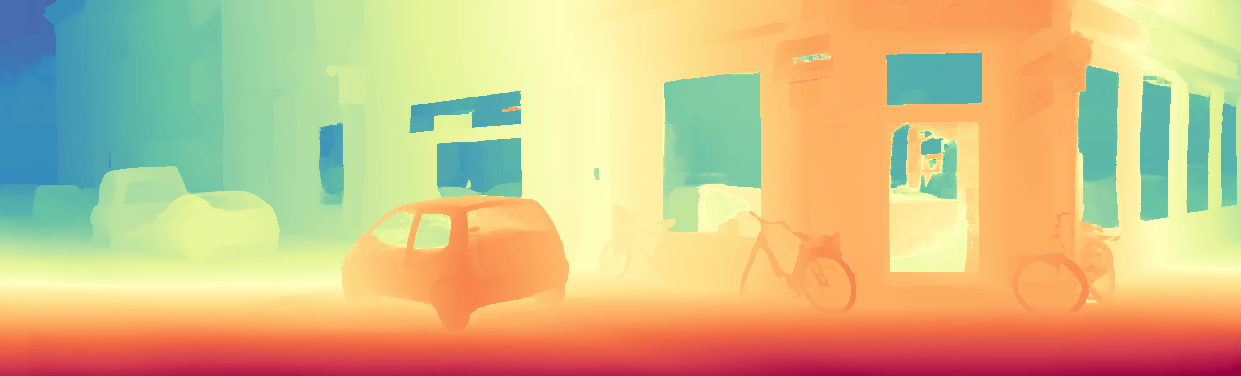} \\ 
        \small Selective-IGEV \cite{wang2024selective} &
        \textbf{\method (ours)} \\
        \includegraphics[width=0.48\textwidth]{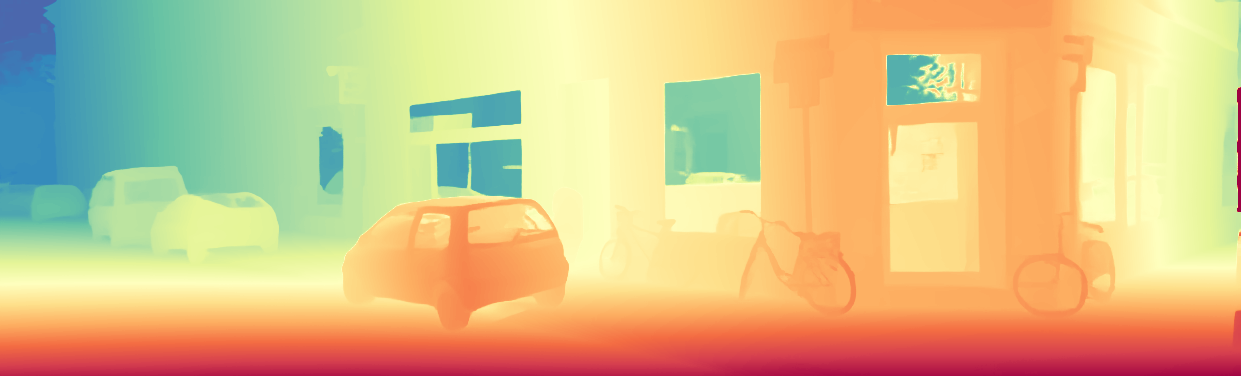} &
        \includegraphics[width=0.48\textwidth]{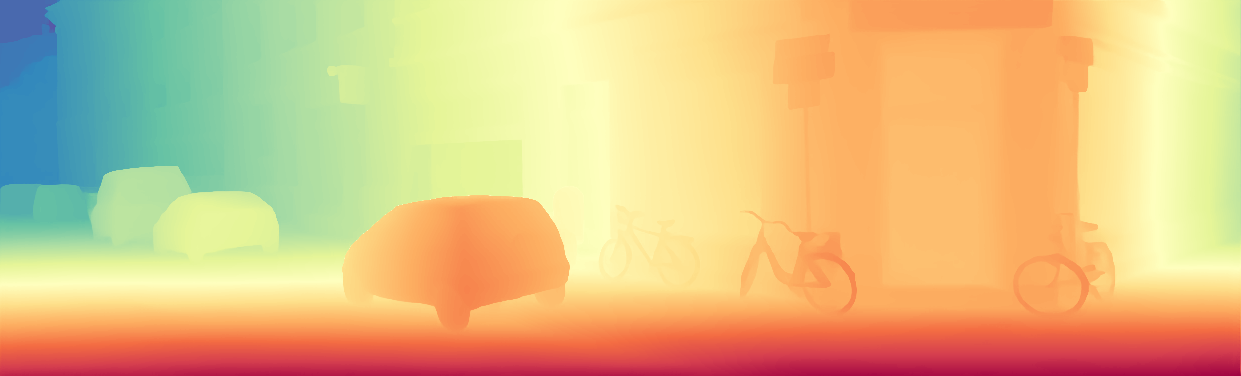} \\ 
    \end{tabular}\vspace{-0.3cm}
    \caption{\textbf{Qualitative Results -- KITTI 2012 (part 1).} Predictions by state-of-the-art models and \method.}
    \label{fig:qual_kitti12_1}\vspace{-0.3cm}
\end{figure*}

Figure \ref{fig:qual_kitti12_1} shows two examples from the KITTI 2012 dataset (respectively, stereo pairs \textit{000040} and \textit{000068}). We can notice how any existing stereo model is unable to properly perceive the presence of transparent surfaces, as in correspondence of the windows on buildings and cars. On the contrary \method{}, driven by the priors injected through the VFM, properly predicts the disparity corresponding to the transparent surfaces.   

\clearpage

\begin{figure*}[h]
    \centering 
    \renewcommand{\tabcolsep}{1pt}
    \begin{tabular}{cc}
        
        \small RGB &
        \small RAFT-Stereo \cite{lipson2021raft} \\
        \includegraphics[width=0.48\textwidth]{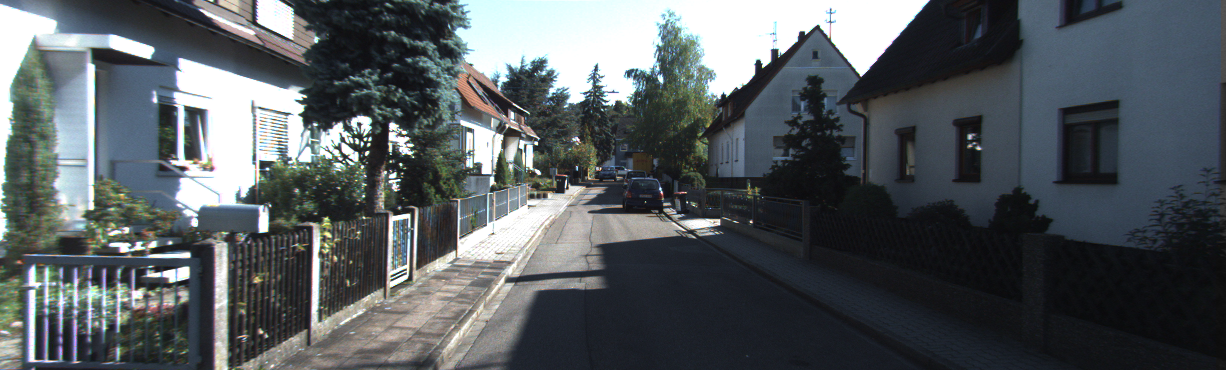} & 
        \includegraphics[width=0.48\textwidth]{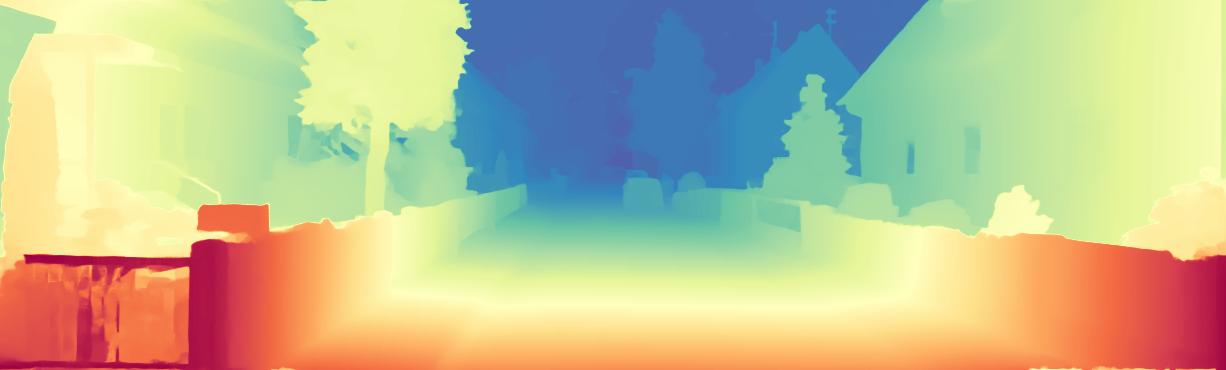} \\
        \small DLNR \cite{zhao2023high} &
        \small NMRF \cite{guan2024neural} \\
        \includegraphics[width=0.48\textwidth]{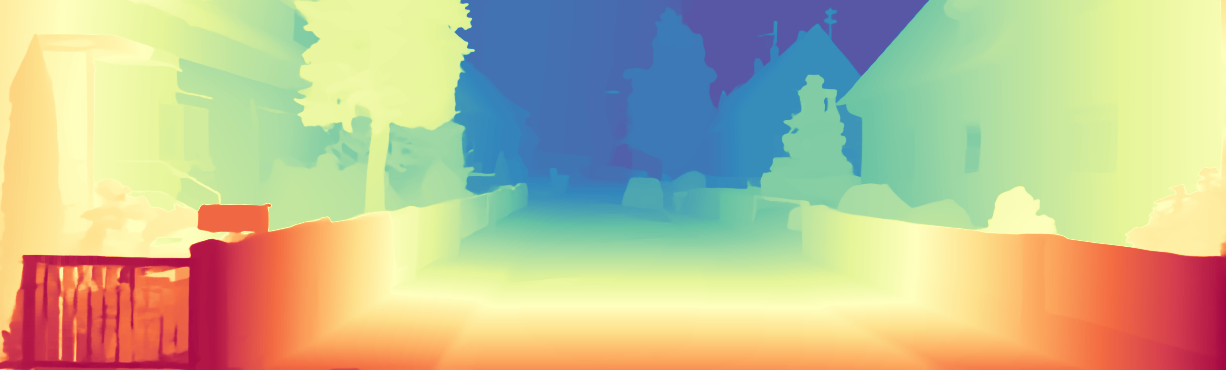} &
        \includegraphics[width=0.48\textwidth]{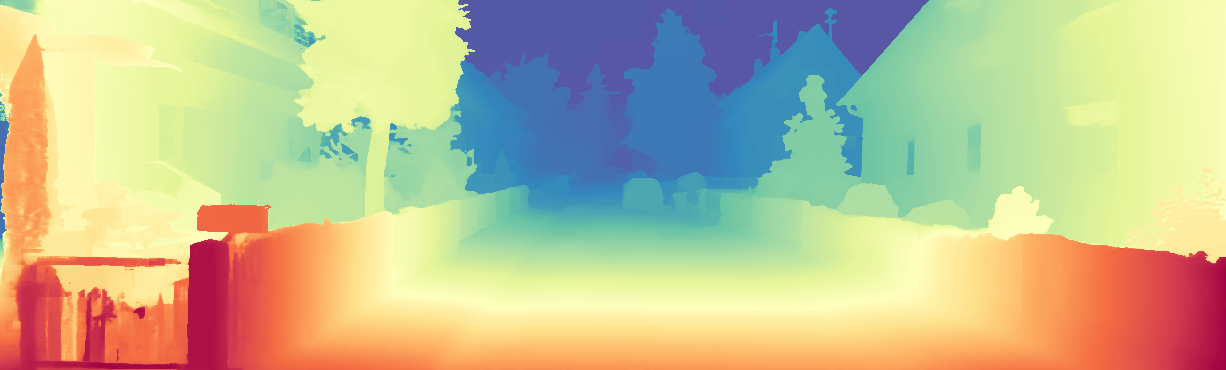} \\ 
        \small Selective-IGEV \cite{wang2024selective} &
        \textbf{\method (ours)} \\
        \includegraphics[width=0.48\textwidth]{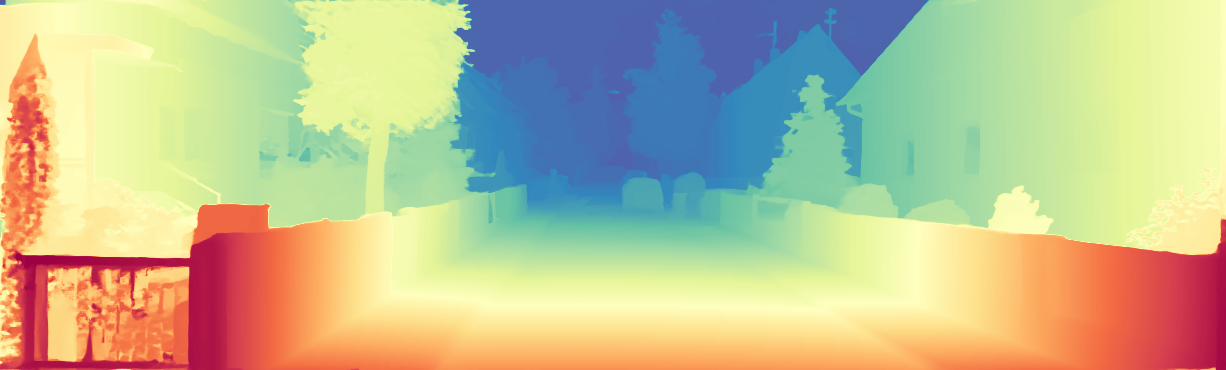} &
        \includegraphics[width=0.48\textwidth]{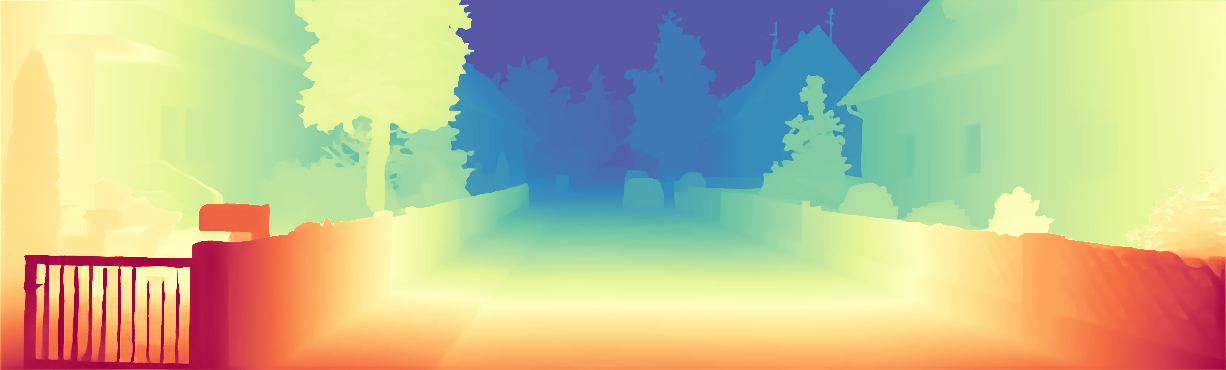} \\ \\

        \small RGB &
        \small RAFT-Stereo \cite{lipson2021raft} \\
        \includegraphics[width=0.48\textwidth]{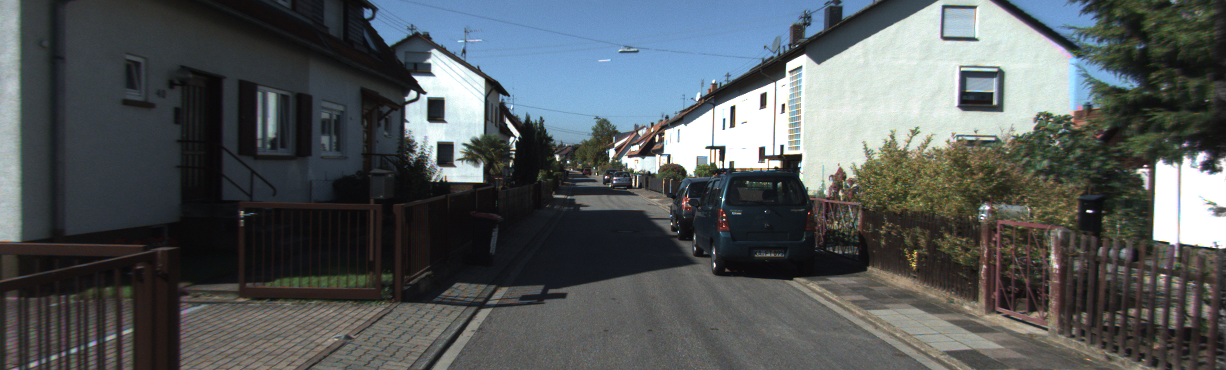} & 
        \includegraphics[width=0.48\textwidth]{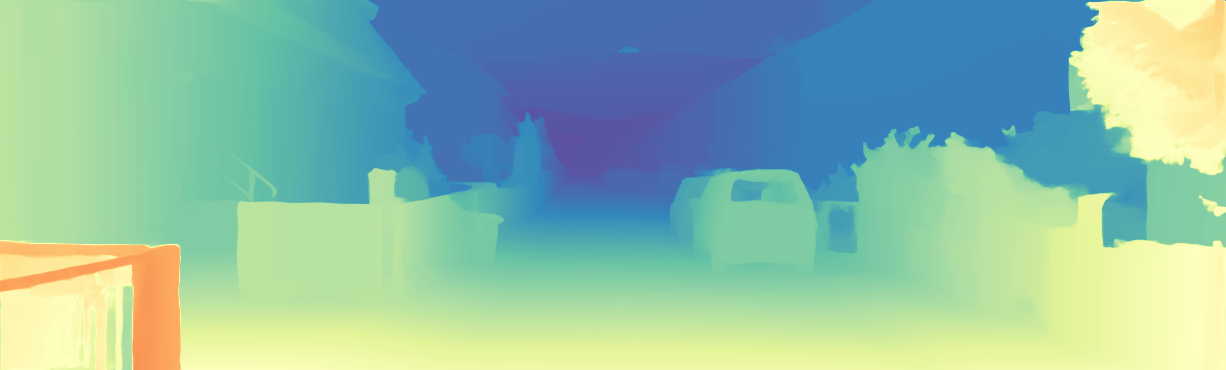} \\
        \small DLNR \cite{zhao2023high} &
        \small NMRF \cite{guan2024neural} \\
        \includegraphics[width=0.48\textwidth]{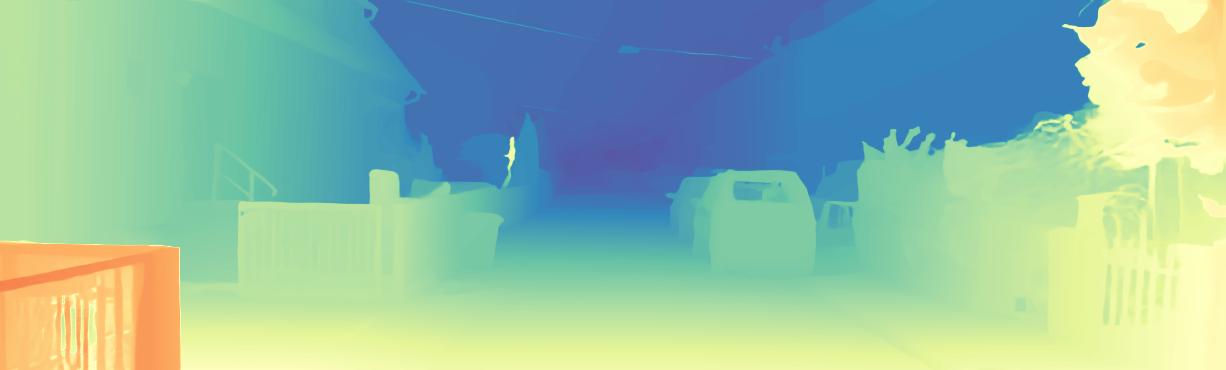} &
        \includegraphics[width=0.48\textwidth]{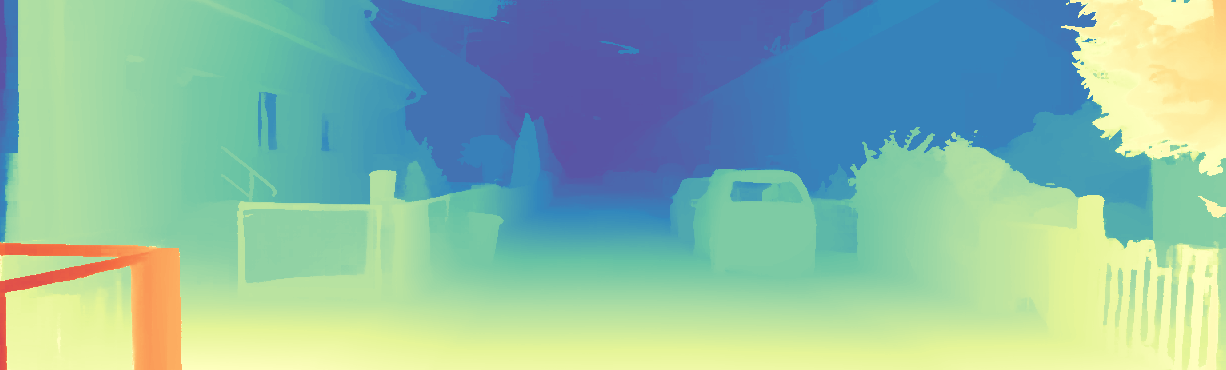} \\ 
        \small Selective-IGEV \cite{wang2024selective} &
        \textbf{\method (ours)} \\
        \includegraphics[width=0.48\textwidth]{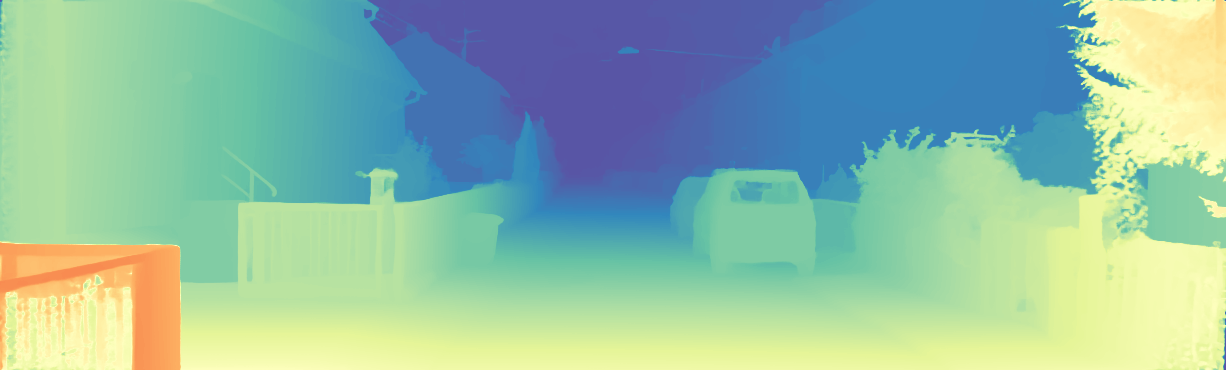} &
        \includegraphics[width=0.48\textwidth]{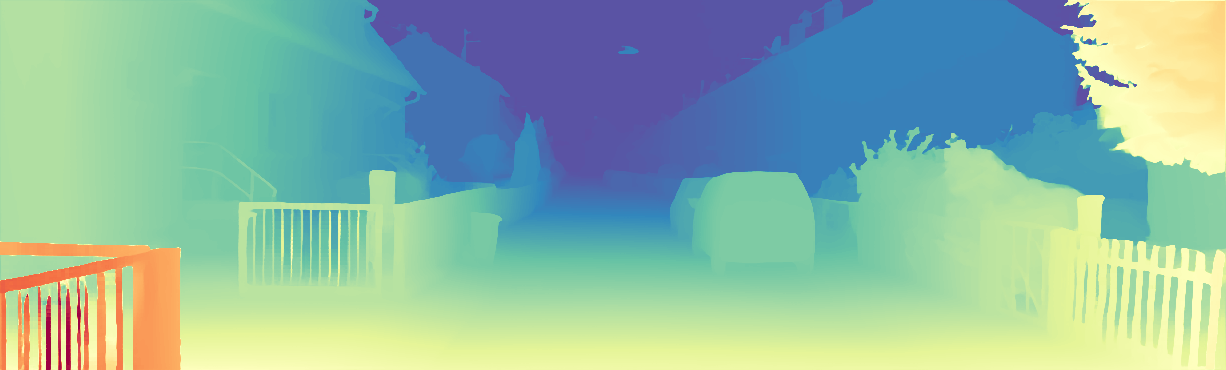} \\ 
    \end{tabular}\vspace{-0.3cm}
    \caption{\textbf{Qualitative Results -- KITTI 2012 (part 2).} Predictions by state-of-the-art models and \method.}
    \label{fig:qual_kitti12_2}\vspace{-0.3cm}
\end{figure*}

Figure \ref{fig:qual_kitti12_2} shows two further examples from KITTI 2012 (respectively, stereo pairs \textit{000073} and \textit{000127}). In this case, we can appreciate the much higher level of detail in the disparity maps predicted by \method, with extremely thin structures in fences and gates being preserved.

\clearpage

\begin{figure*}[h]
    \centering 
    \renewcommand{\tabcolsep}{1pt}
    \begin{tabular}{cc}

        \small RGB &
        \small RAFT-Stereo \cite{lipson2021raft} \\
        \includegraphics[width=0.48\textwidth]{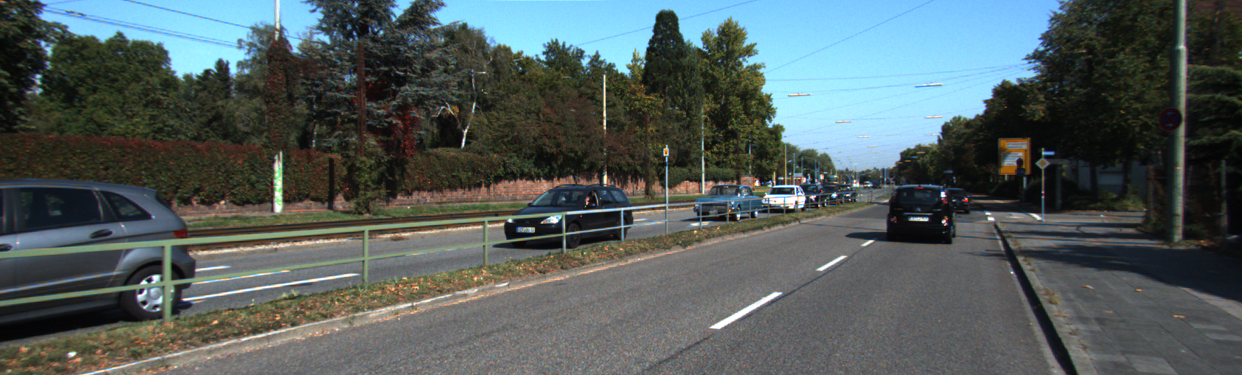} & 
        \includegraphics[width=0.48\textwidth]{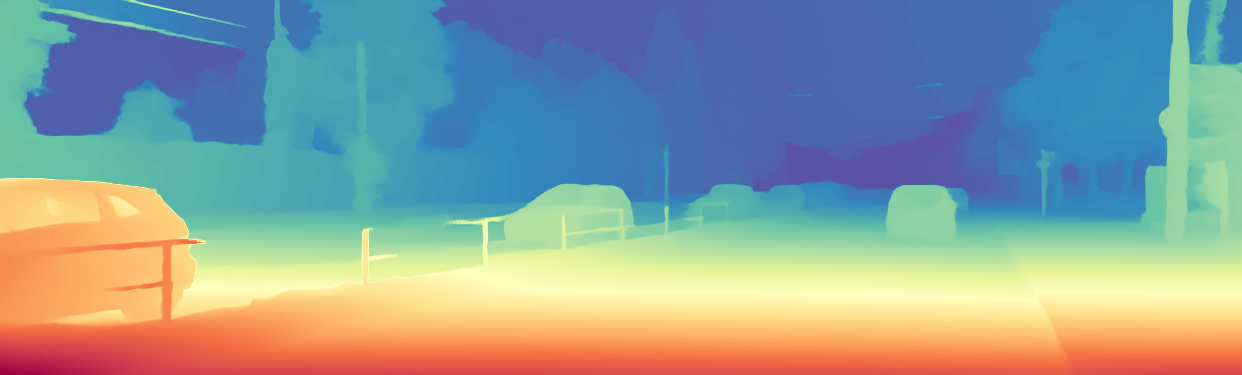} \\
        \small DLNR \cite{zhao2023high} &
        \small NMRF \cite{guan2024neural} \\
        \includegraphics[width=0.48\textwidth]{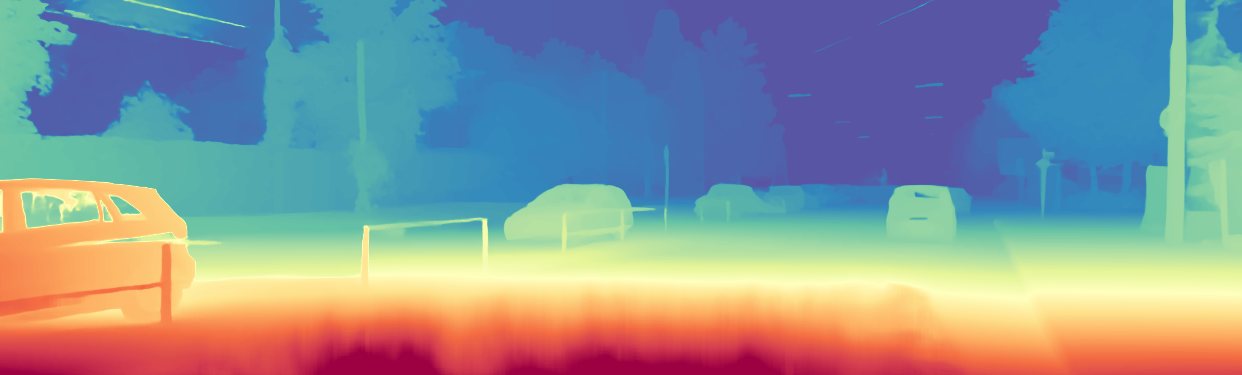} &
        \includegraphics[width=0.48\textwidth]{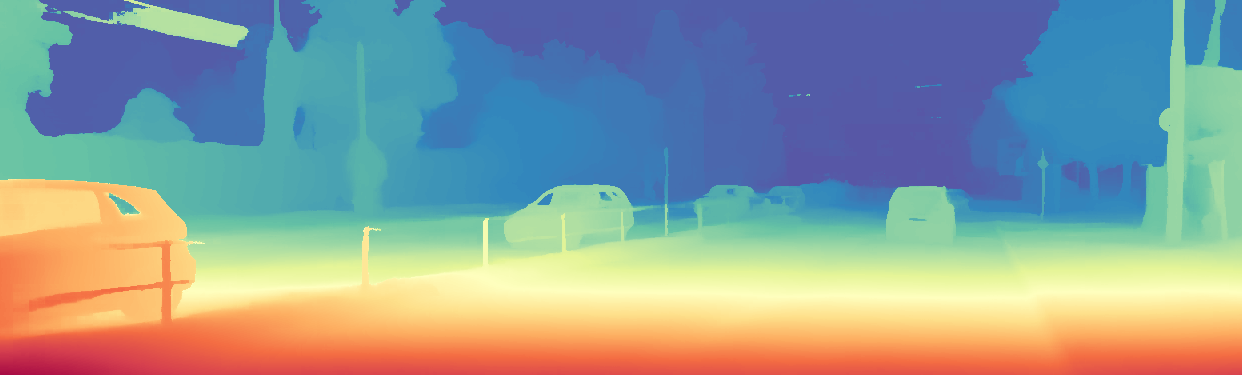} \\ 
        \small Selective-IGEV \cite{wang2024selective} &
        \textbf{\method (ours)} \\
        \includegraphics[width=0.48\textwidth]{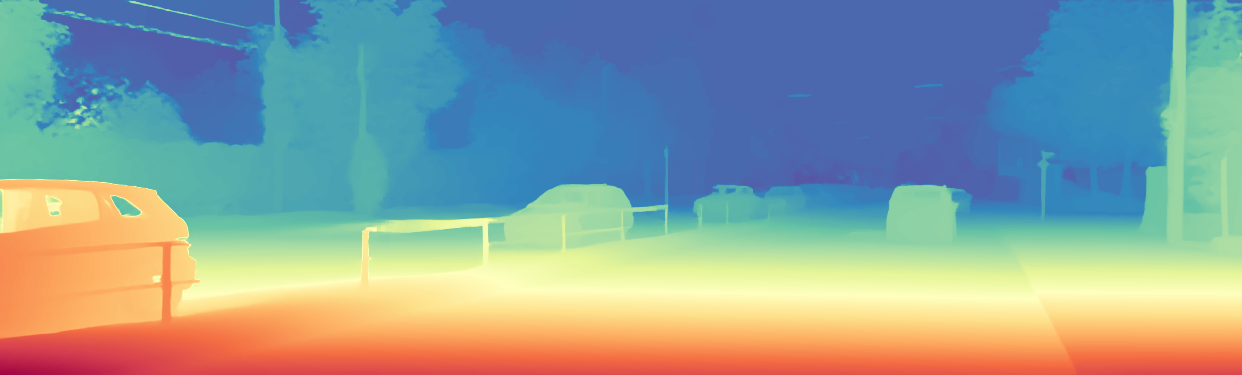} &
        \includegraphics[width=0.48\textwidth]{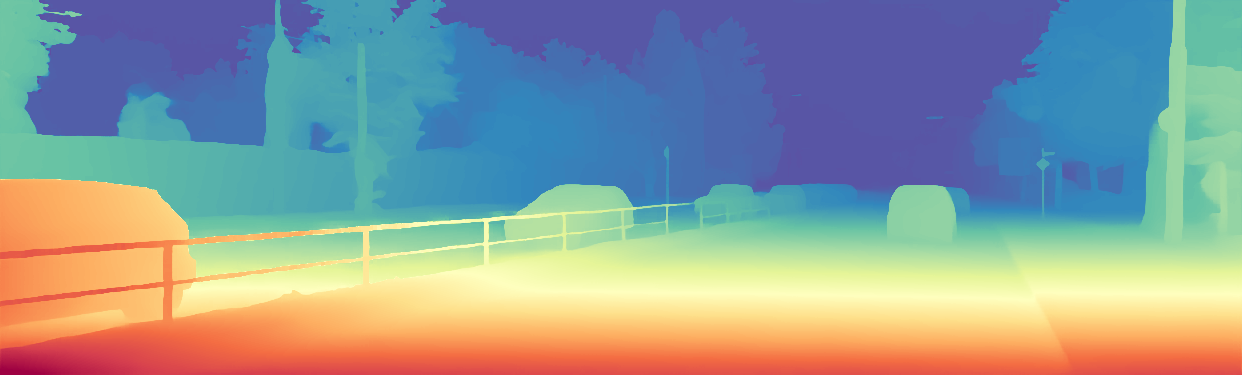} \\ \\

        \small RGB &
        \small RAFT-Stereo \cite{lipson2021raft} \\
        \includegraphics[width=0.48\textwidth]{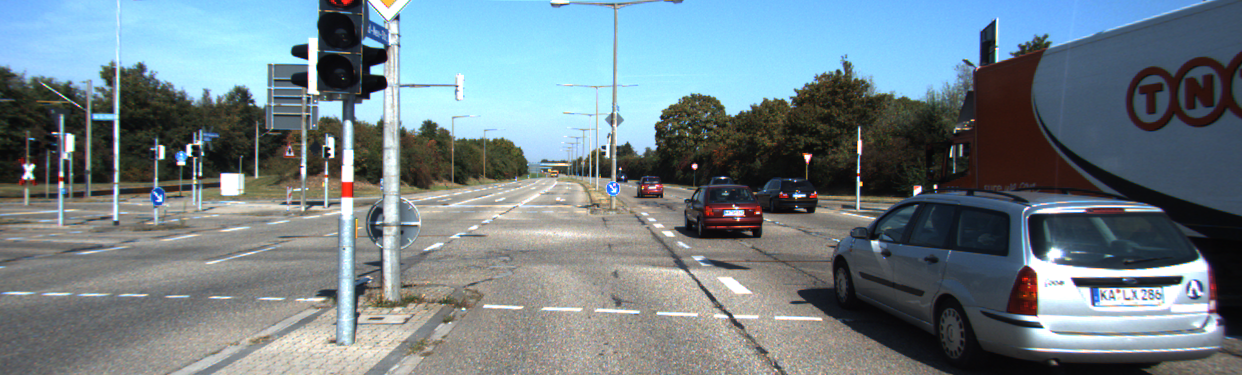} & 
        \includegraphics[width=0.48\textwidth]{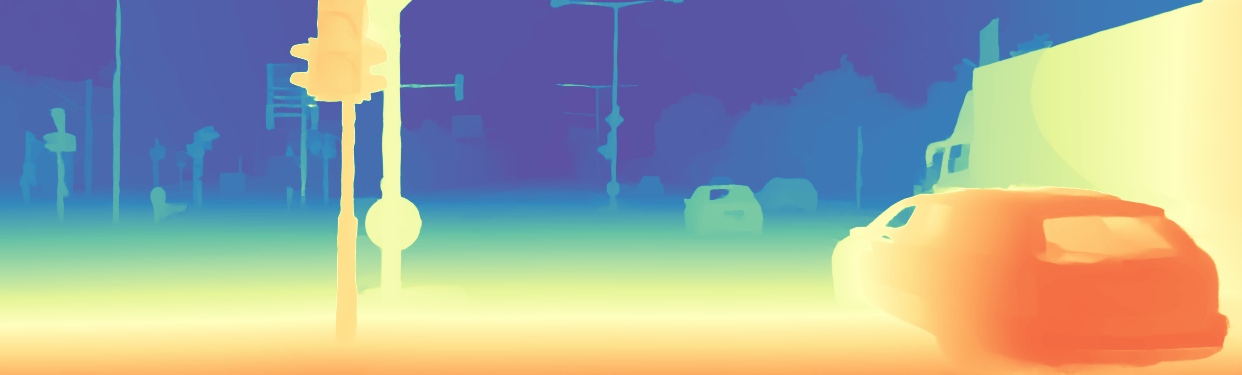} \\
        \small DLNR \cite{zhao2023high} &
        \small NMRF \cite{guan2024neural} \\
        \includegraphics[width=0.48\textwidth]{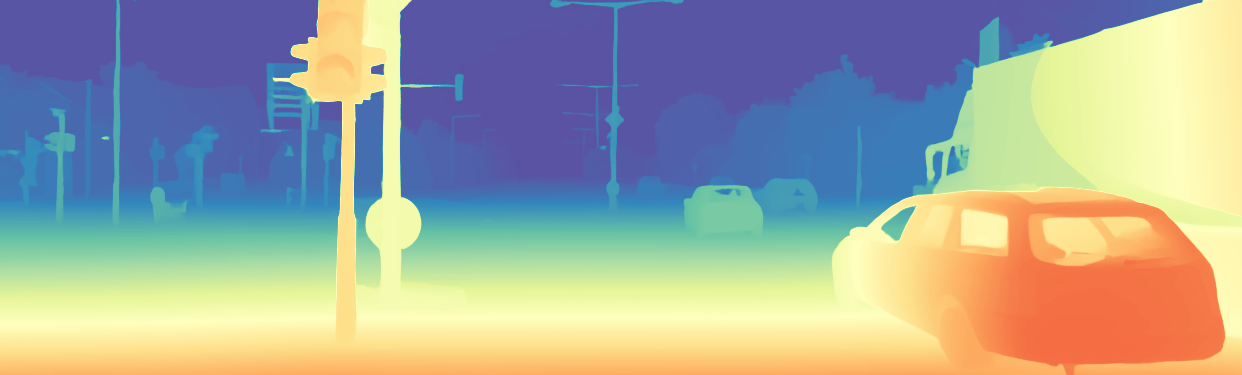} &
        \includegraphics[width=0.48\textwidth]{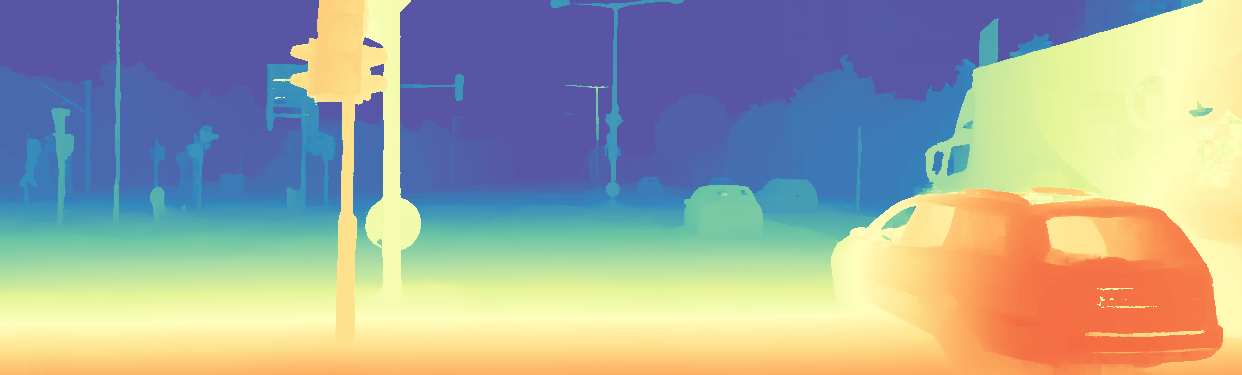} \\ 
        \small Selective-IGEV \cite{wang2024selective} &
        \textbf{\method (ours)} \\
        \includegraphics[width=0.48\textwidth]{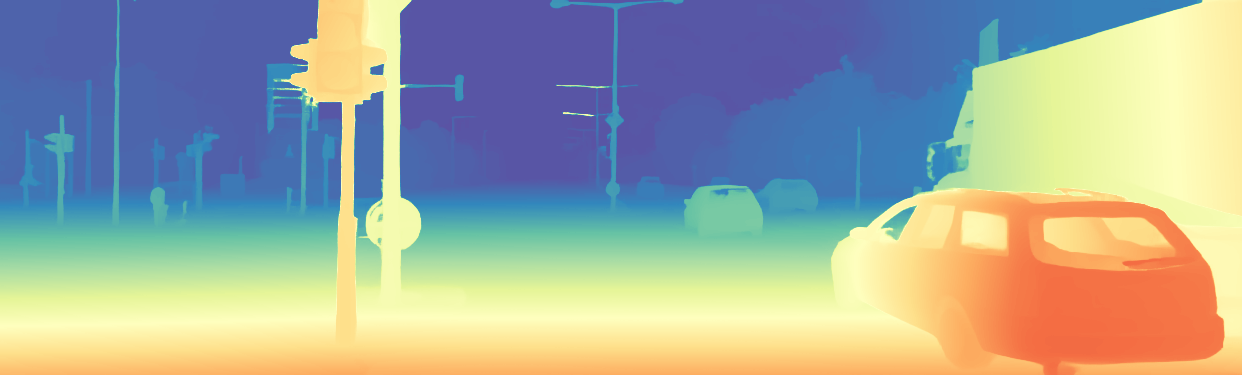} &
        \includegraphics[width=0.48\textwidth]{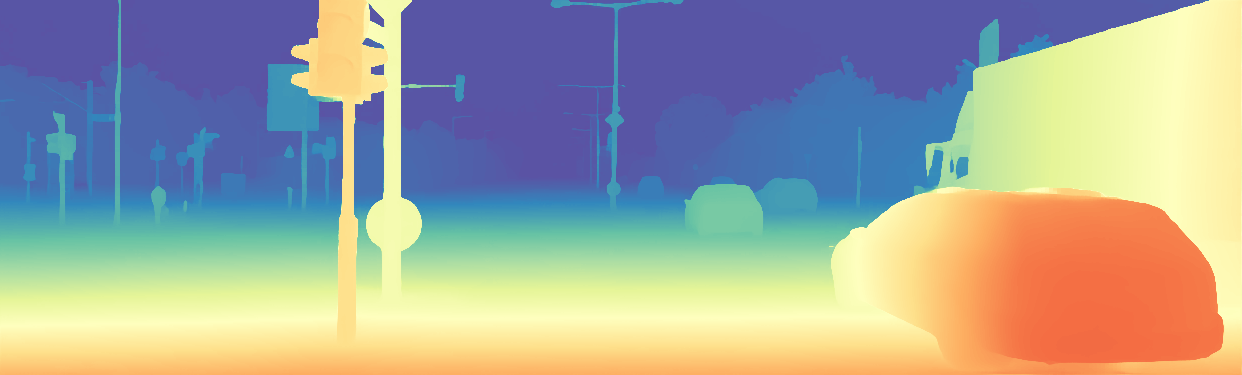} \\ 
    \end{tabular}\vspace{-0.3cm}
    \caption{\textbf{Qualitative Results -- KITTI 2015 (part 1).} Predictions by state-of-the-art models and \method.}
    \label{fig:qual_kitti15_1}\vspace{-0.3cm}
\end{figure*}

Figure \ref{fig:qual_kitti15_1} reports two stereo pairs from KITTI 2015 (respectively, \textit{000024} and \textit{000049}). These examples confirm the ability to recover both thin structures and transparent surfaces already appreciated in KITTI 2012.

\clearpage

\begin{figure*}[h]
    \centering 
    \renewcommand{\tabcolsep}{1pt}
    \begin{tabular}{cc}
        
        \small RGB &
        \small RAFT-Stereo \cite{lipson2021raft} \\
        \includegraphics[width=0.48\textwidth]{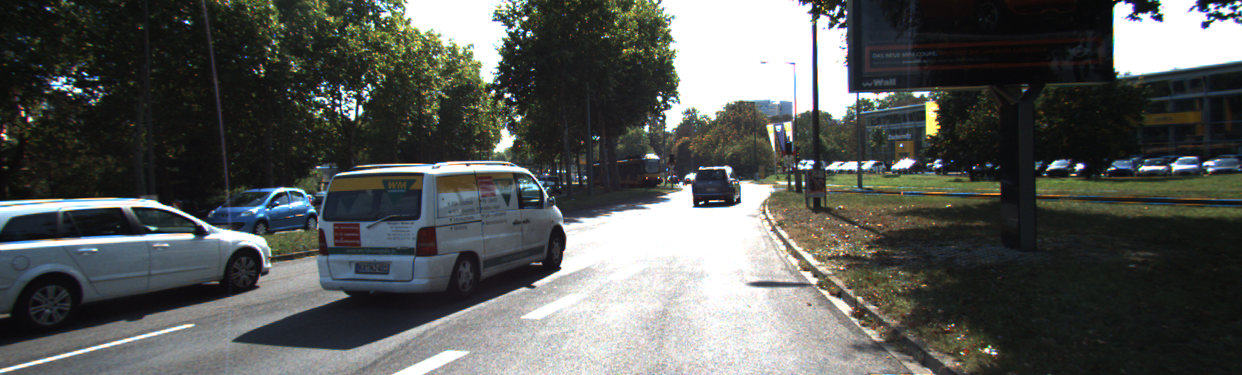} & 
        \includegraphics[width=0.48\textwidth]{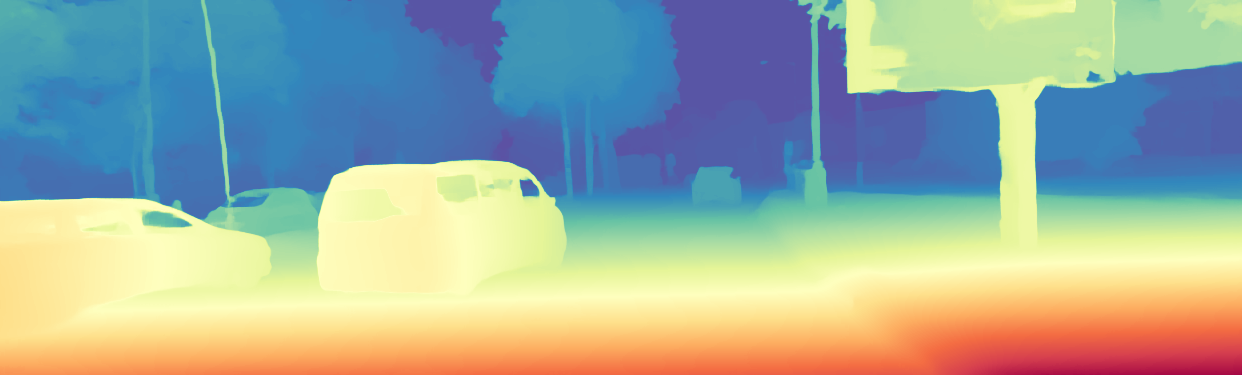} \\
        \small DLNR \cite{zhao2023high} &
        \small NMRF \cite{guan2024neural} \\
        \includegraphics[width=0.48\textwidth]{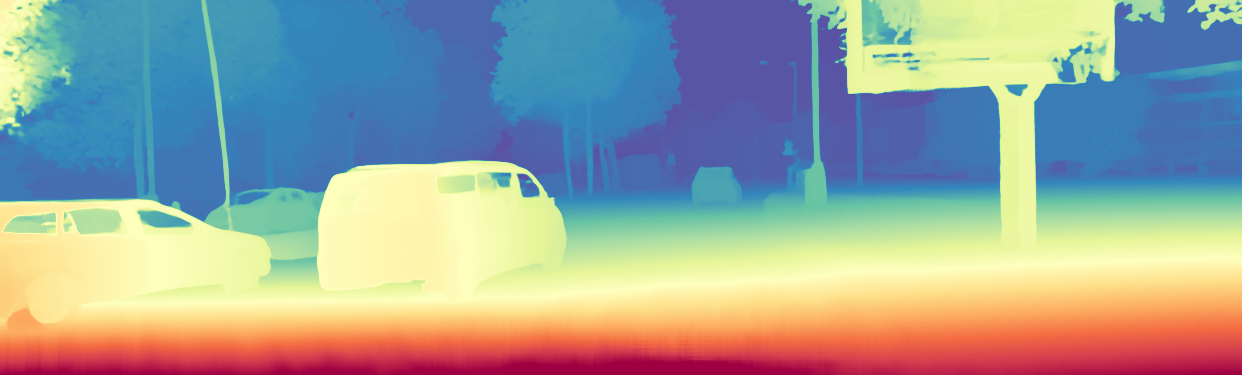} &
        \includegraphics[width=0.48\textwidth]{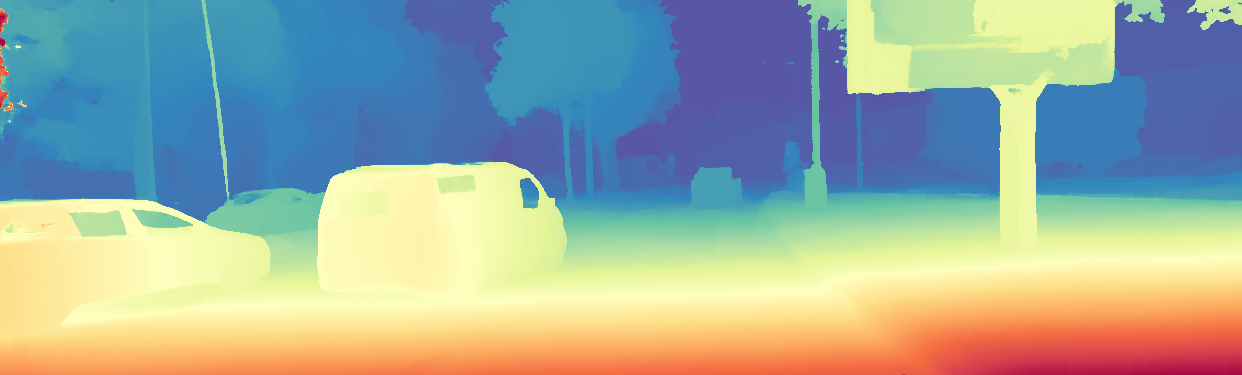} \\ 
        \small Selective-IGEV \cite{wang2024selective} &
        \textbf{\method (ours)} \\
        \includegraphics[width=0.48\textwidth]{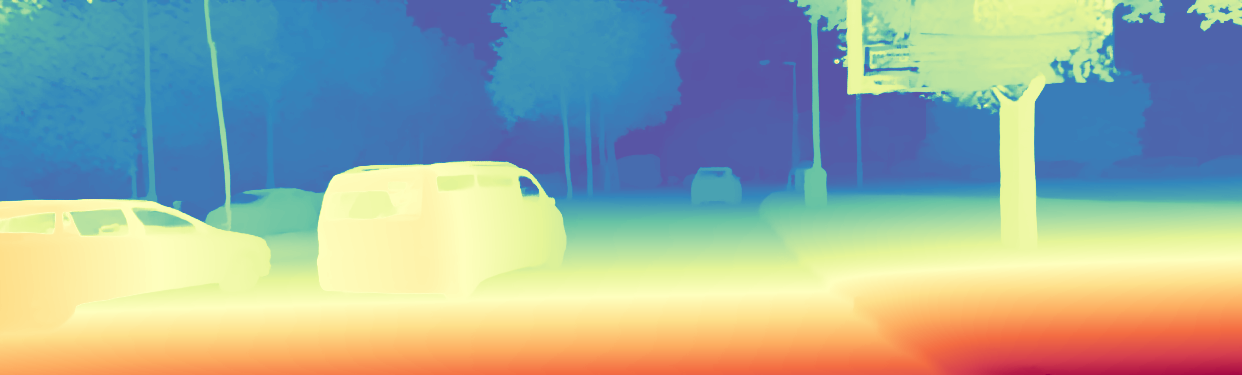} &
        \includegraphics[width=0.48\textwidth]{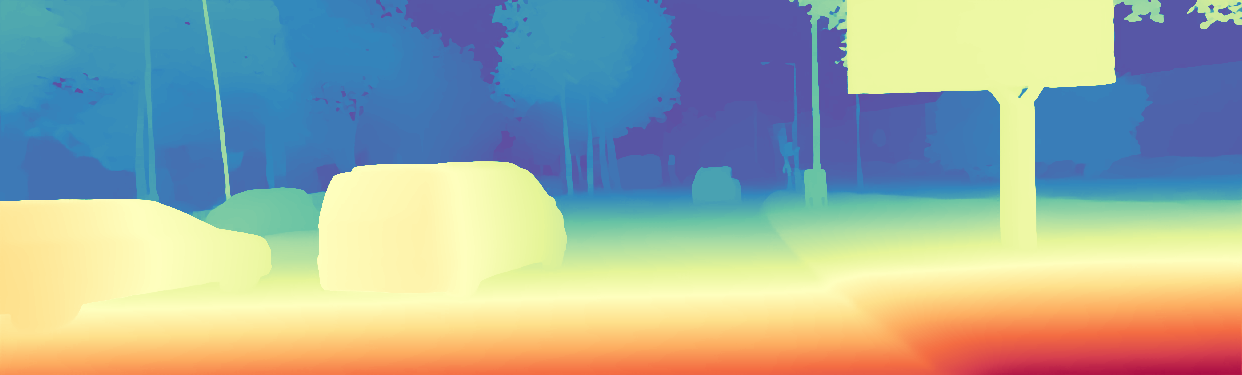} \\ \\

        \small RGB &
        \small RAFT-Stereo \cite{lipson2021raft} \\
        \includegraphics[width=0.48\textwidth]{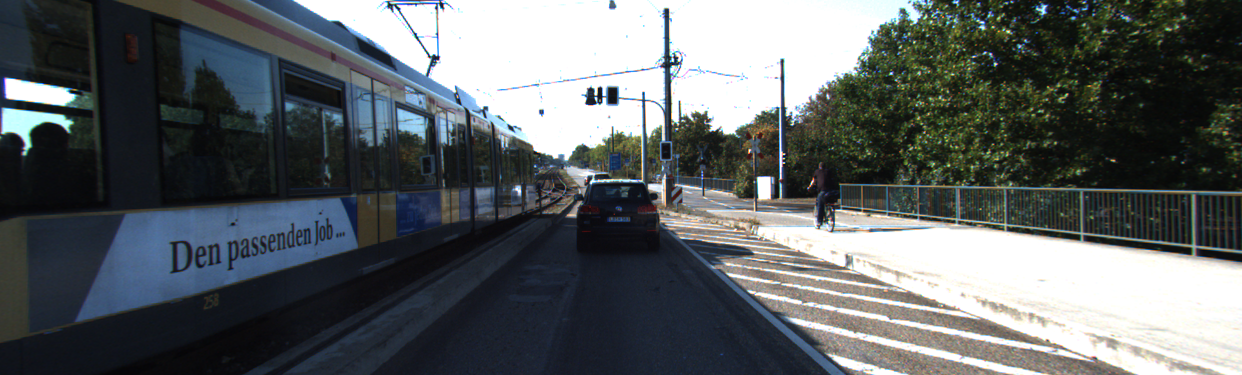} & 
        \includegraphics[width=0.48\textwidth]{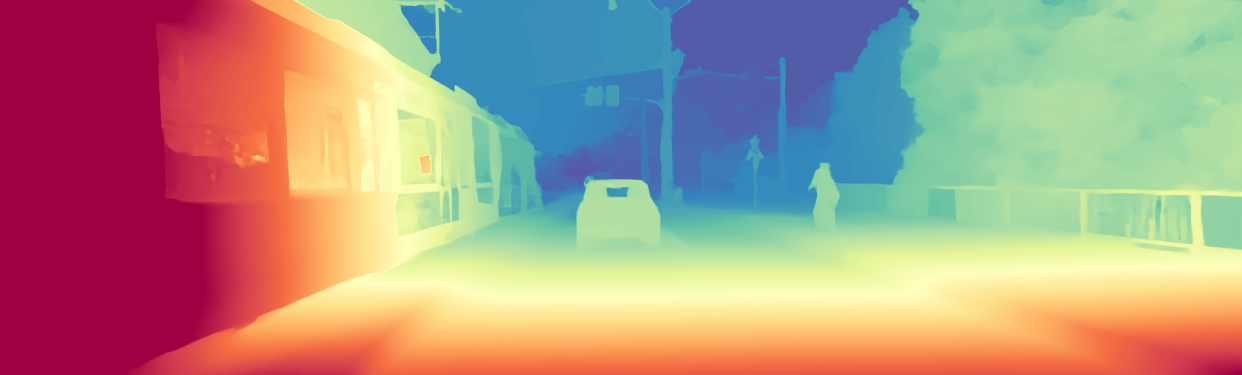} \\
        \small DLNR \cite{zhao2023high} &
        \small NMRF \cite{guan2024neural} \\
        \includegraphics[width=0.48\textwidth]{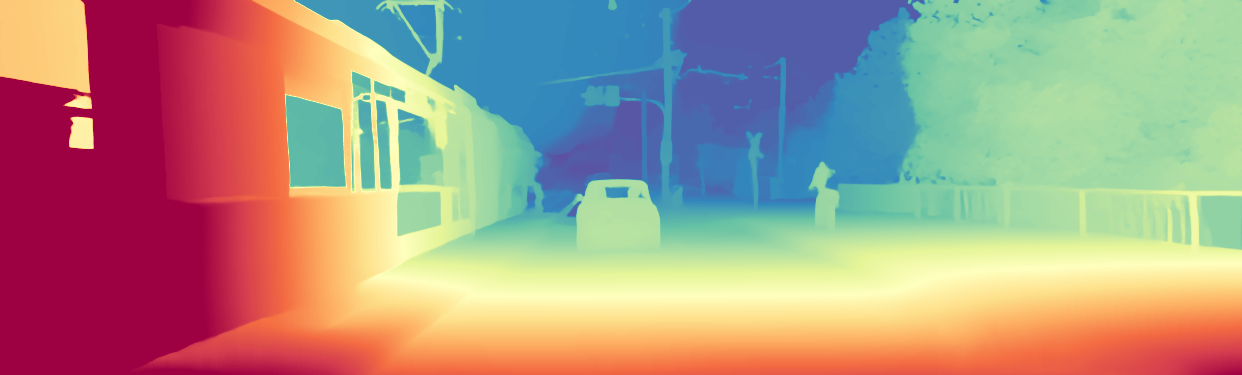} &
        \includegraphics[width=0.48\textwidth]{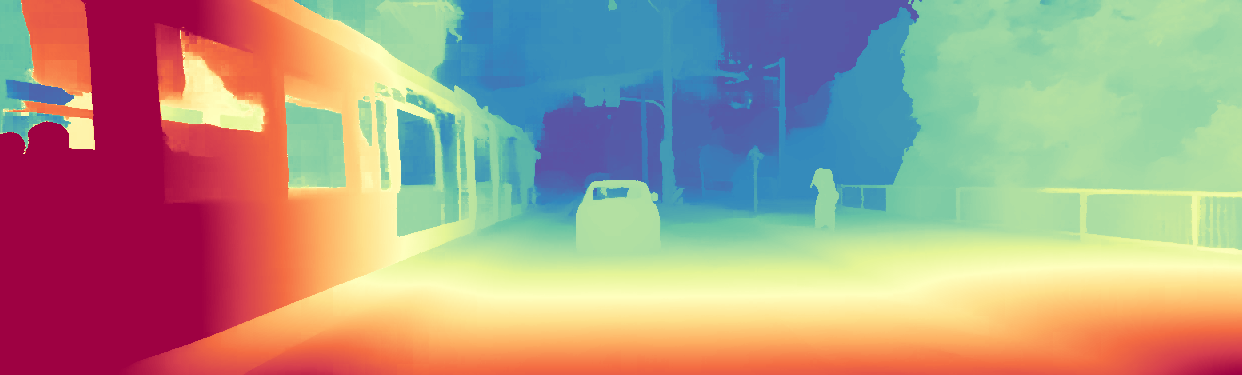} \\ 
        \small Selective-IGEV \cite{wang2024selective} &
        \textbf{\method (ours)} \\
        \includegraphics[width=0.48\textwidth]{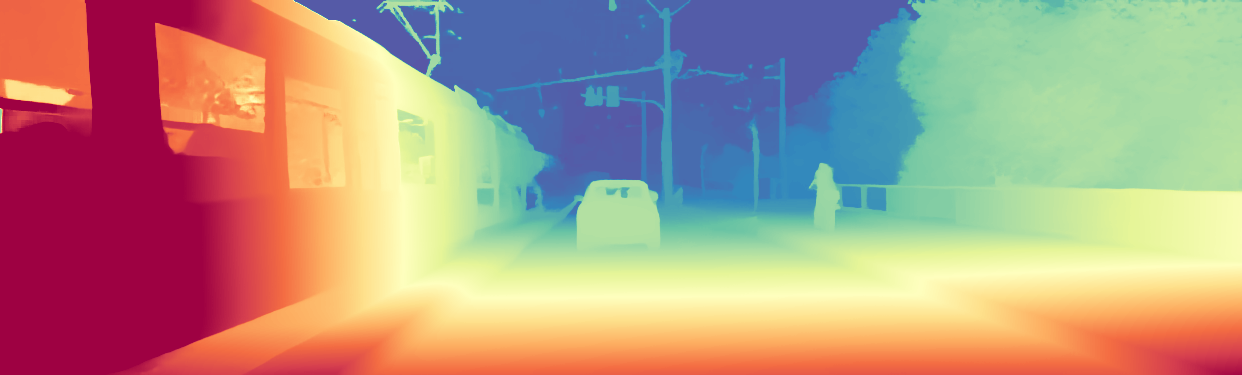} &
        \includegraphics[width=0.48\textwidth]{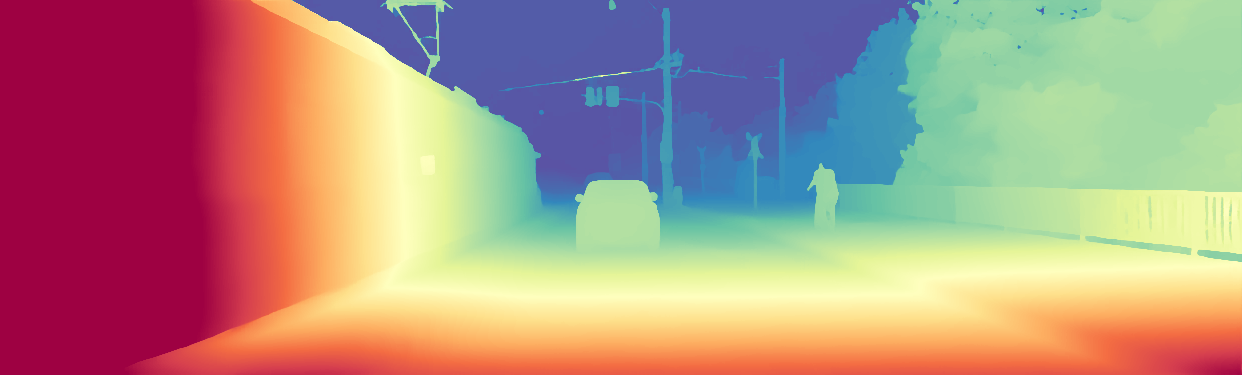} \\ 
    \end{tabular}\vspace{-0.3cm}
    \caption{\textbf{Qualitative Results -- KITTI 2015 (part 2).} Predictions by state-of-the-art models and \method.}
    \label{fig:qual_kitti15_2}\vspace{-0.3cm}
\end{figure*}

Figure \ref{fig:qual_kitti15_2} reports two additional samples from KITTI 2015 (respectively, \textit{000093} and \textit{000144}). These latter present both underexposed and transparent regions, respectively on the billboard and the tram in the two images. While existing stereo networks struggle at dealing with both, \method exposes unprecedented robustness. 

\clearpage

\begin{figure*}[h]
    \centering
    \renewcommand{\tabcolsep}{1pt}
    \begin{tabular}{ccc}
        
        \small RGB &
        \small RAFT-Stereo \cite{lipson2021raft} &
        \small DLNR \cite{zhao2023high} \\
        \includegraphics[width=0.32\textwidth]{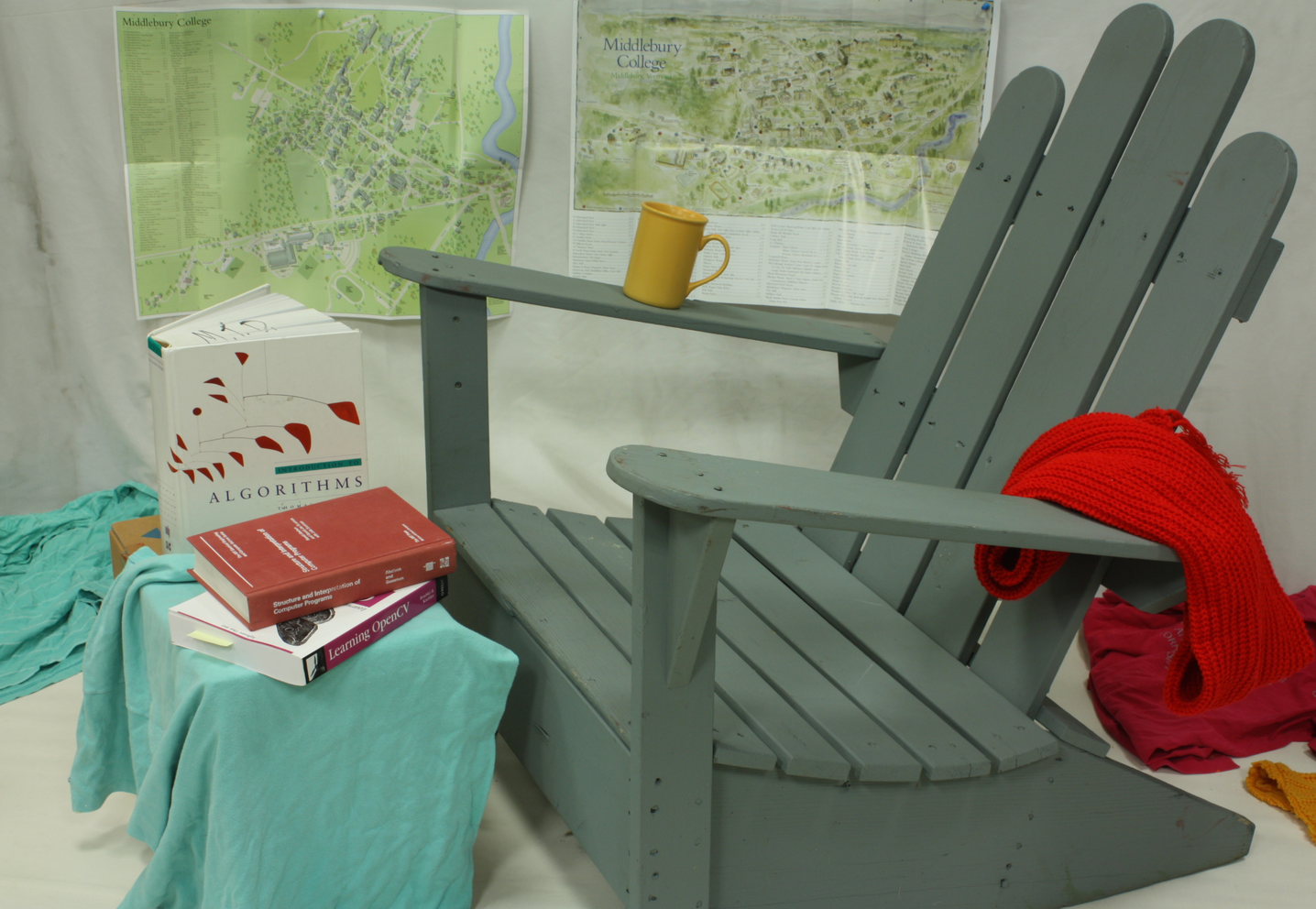} & 
        \includegraphics[width=0.32\textwidth]{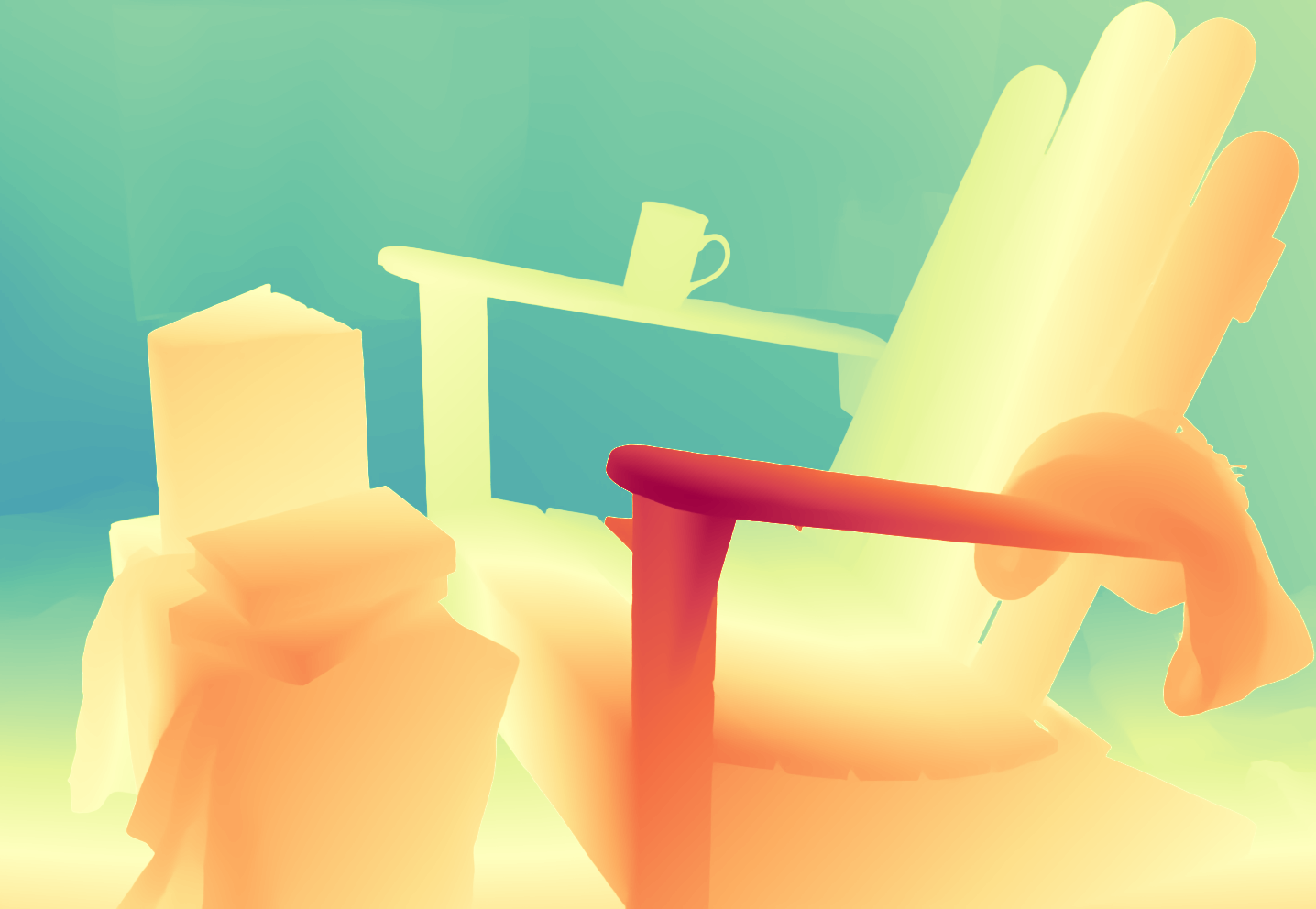} &
        \includegraphics[width=0.32\textwidth]{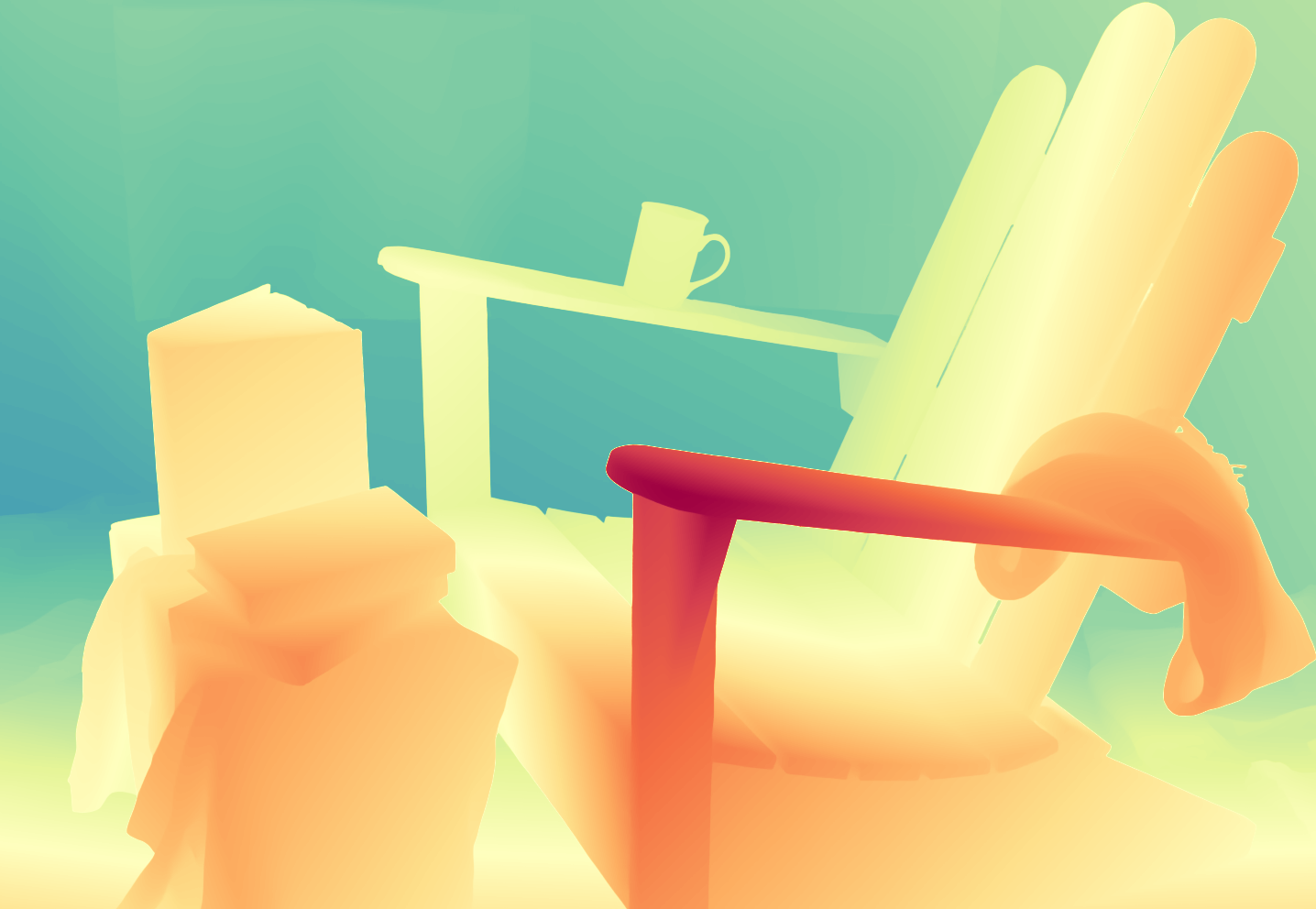} \\
        \small NMRF \cite{guan2024neural} &
        \small Selective-IGEV \cite{wang2024selective} &
        \textbf{\method (ours)} \\
        \includegraphics[width=0.32\textwidth]{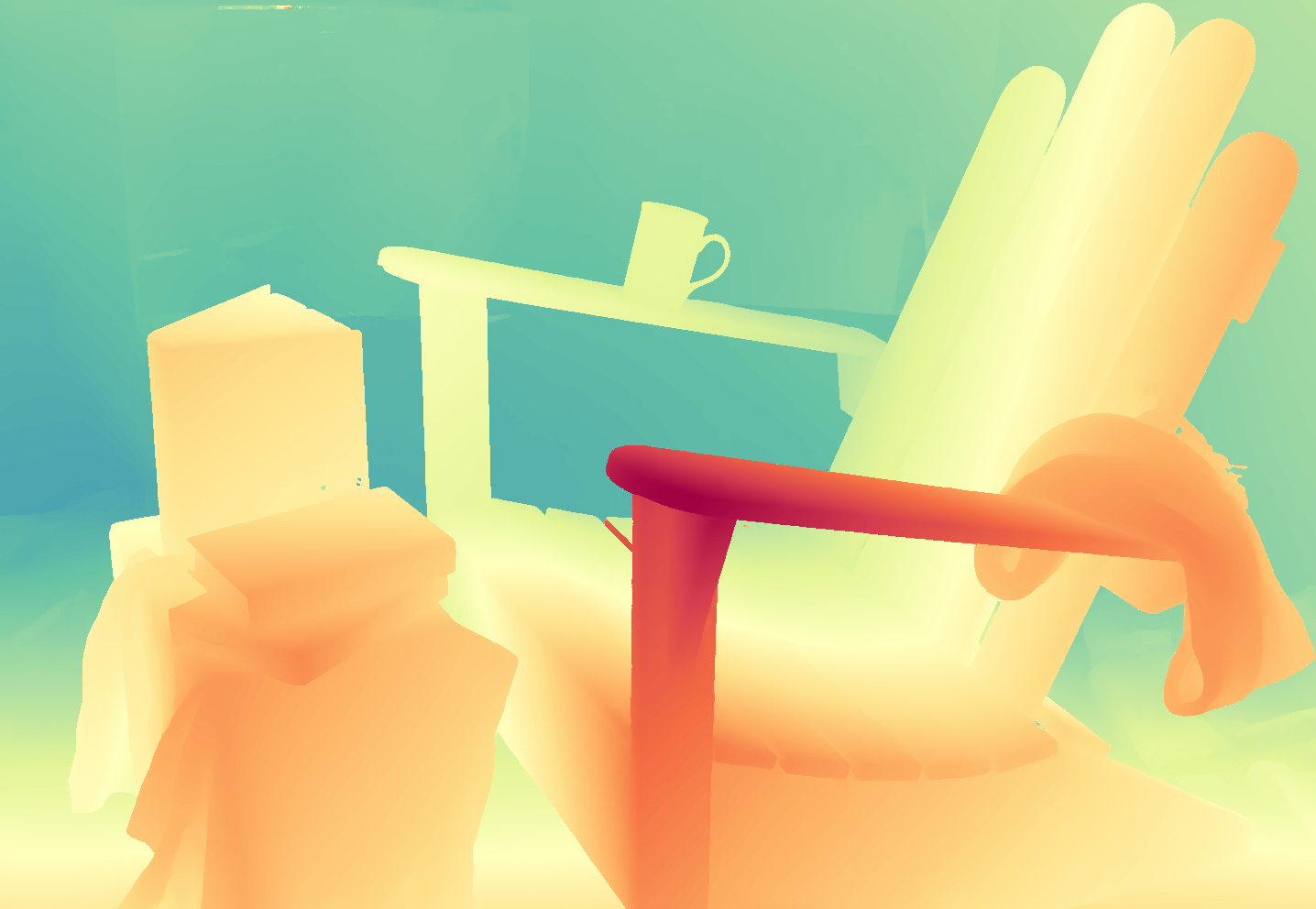} &
        \includegraphics[width=0.32\textwidth]{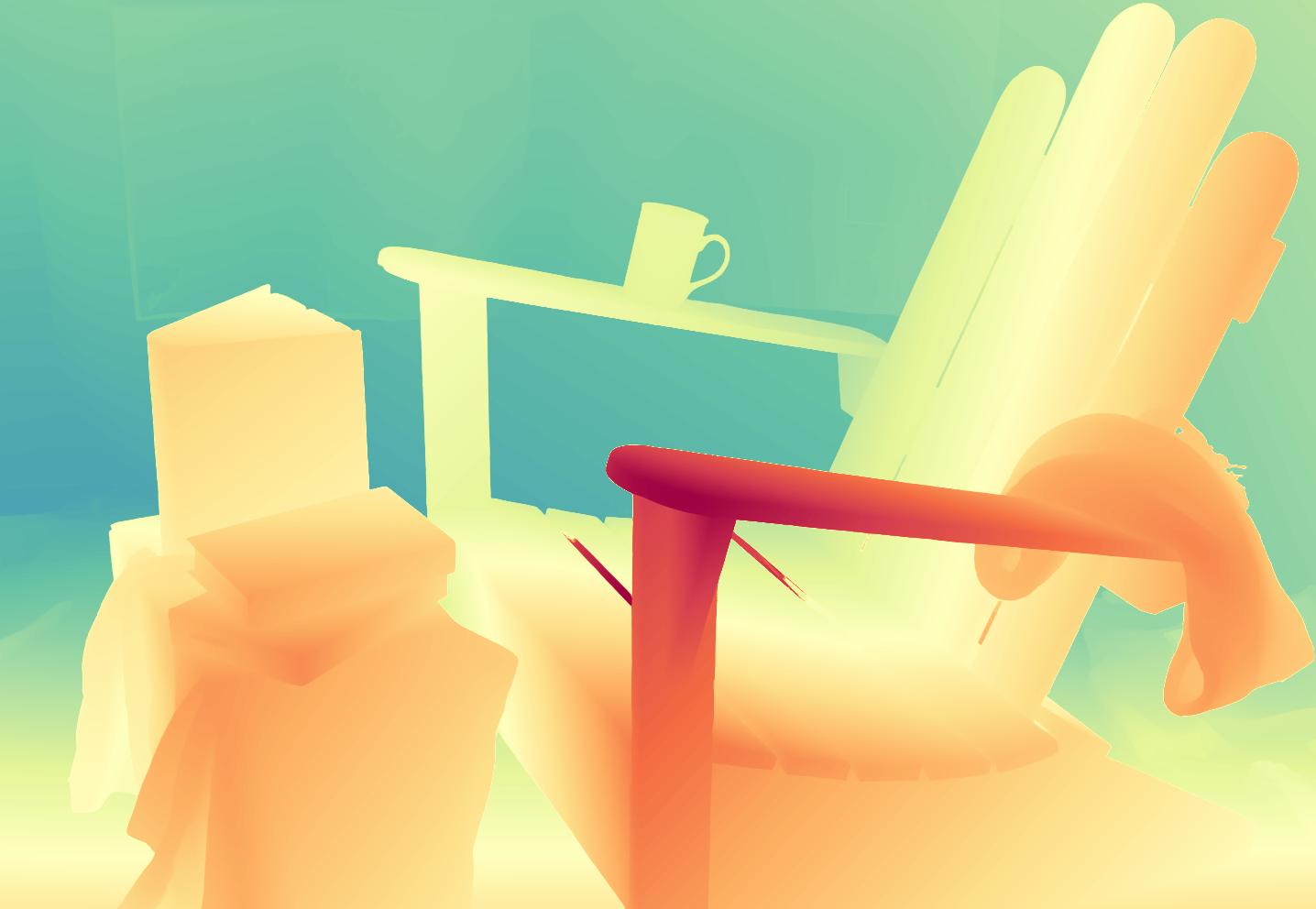} &
        \includegraphics[width=0.32\textwidth]{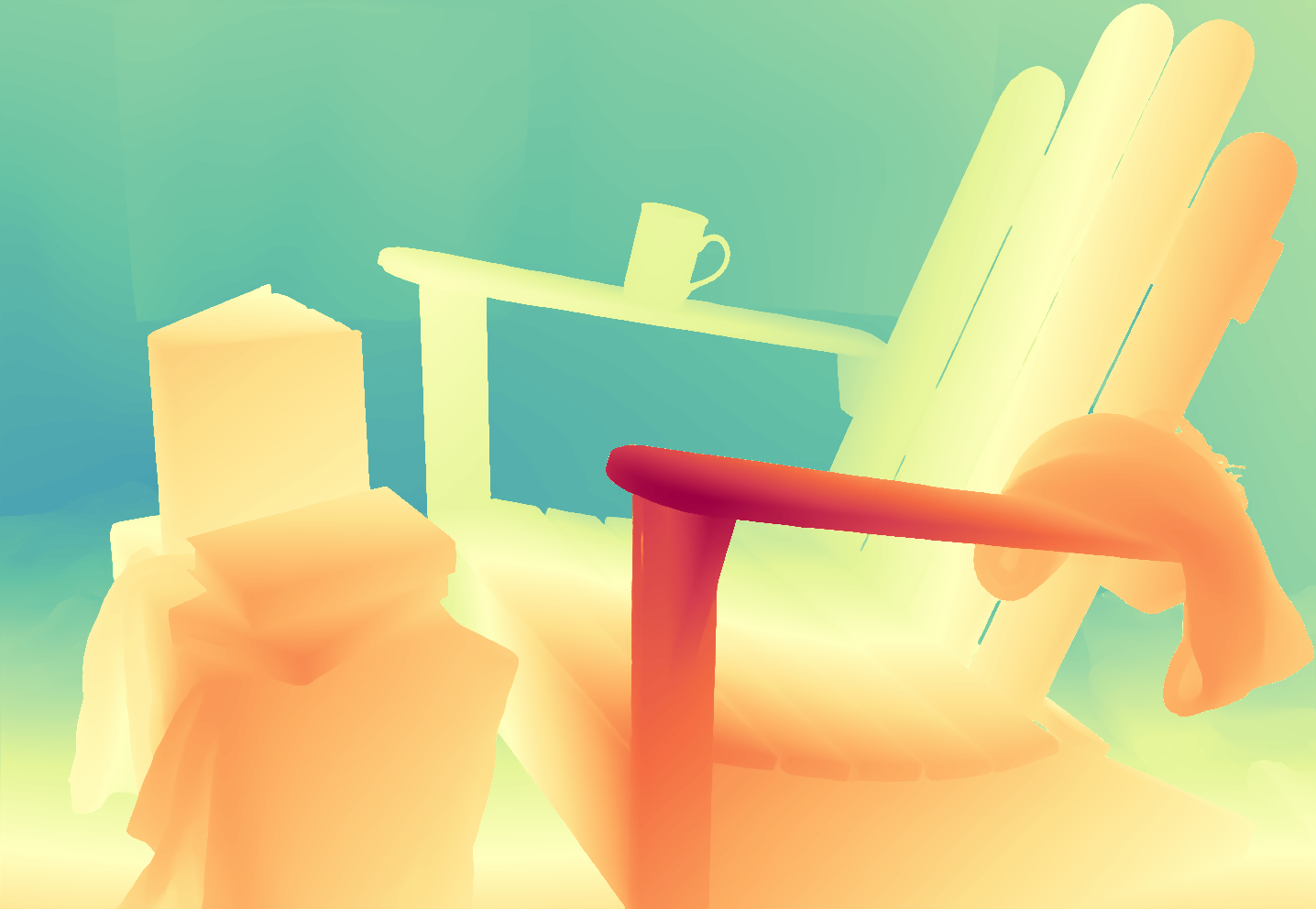} \\ \\
        
        \small RGB &
        \small RAFT-Stereo \cite{lipson2021raft} &
        \small DLNR \cite{zhao2023high} \\
        \includegraphics[width=0.32\textwidth]{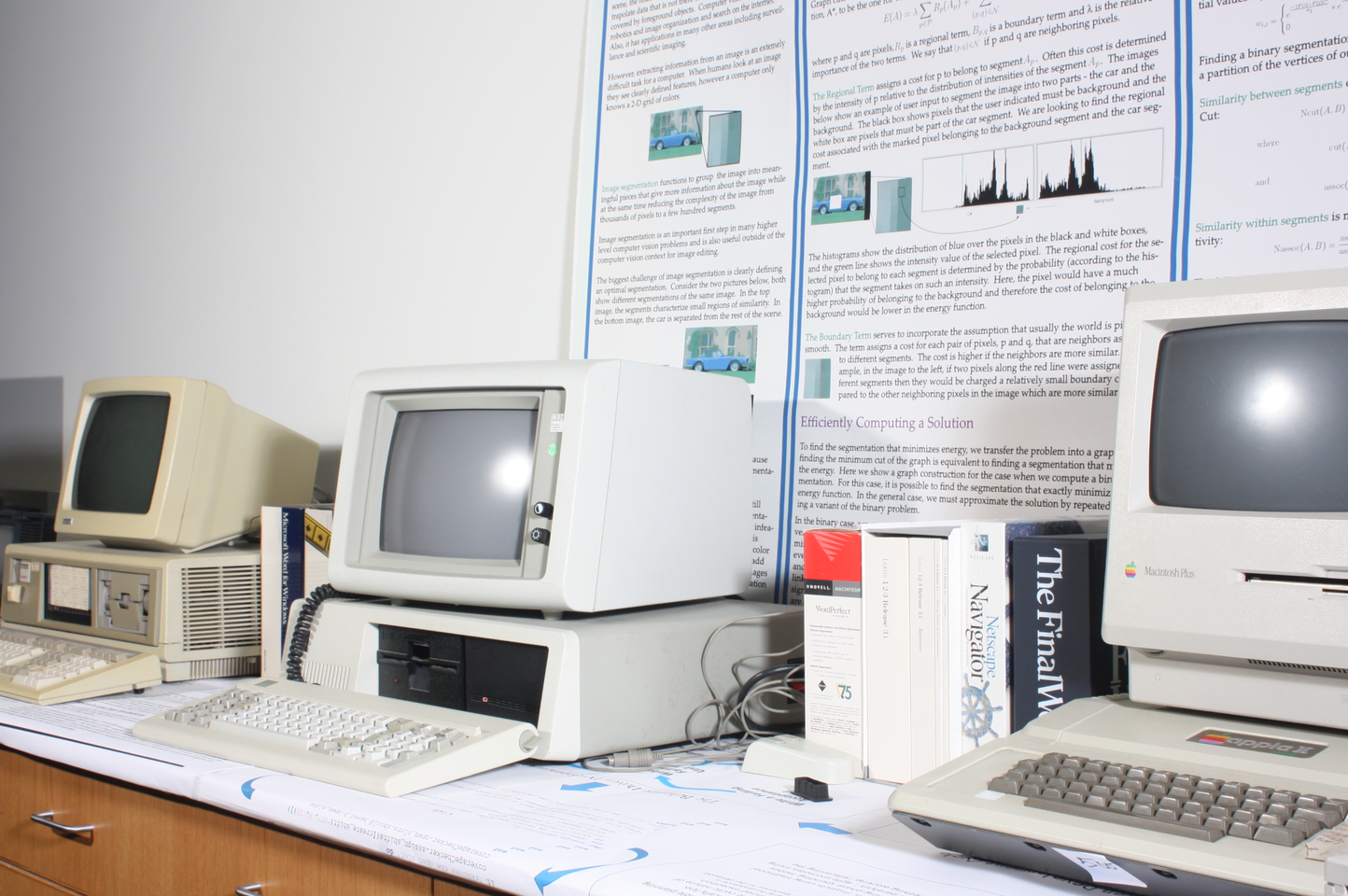} & 
        \includegraphics[width=0.32\textwidth]{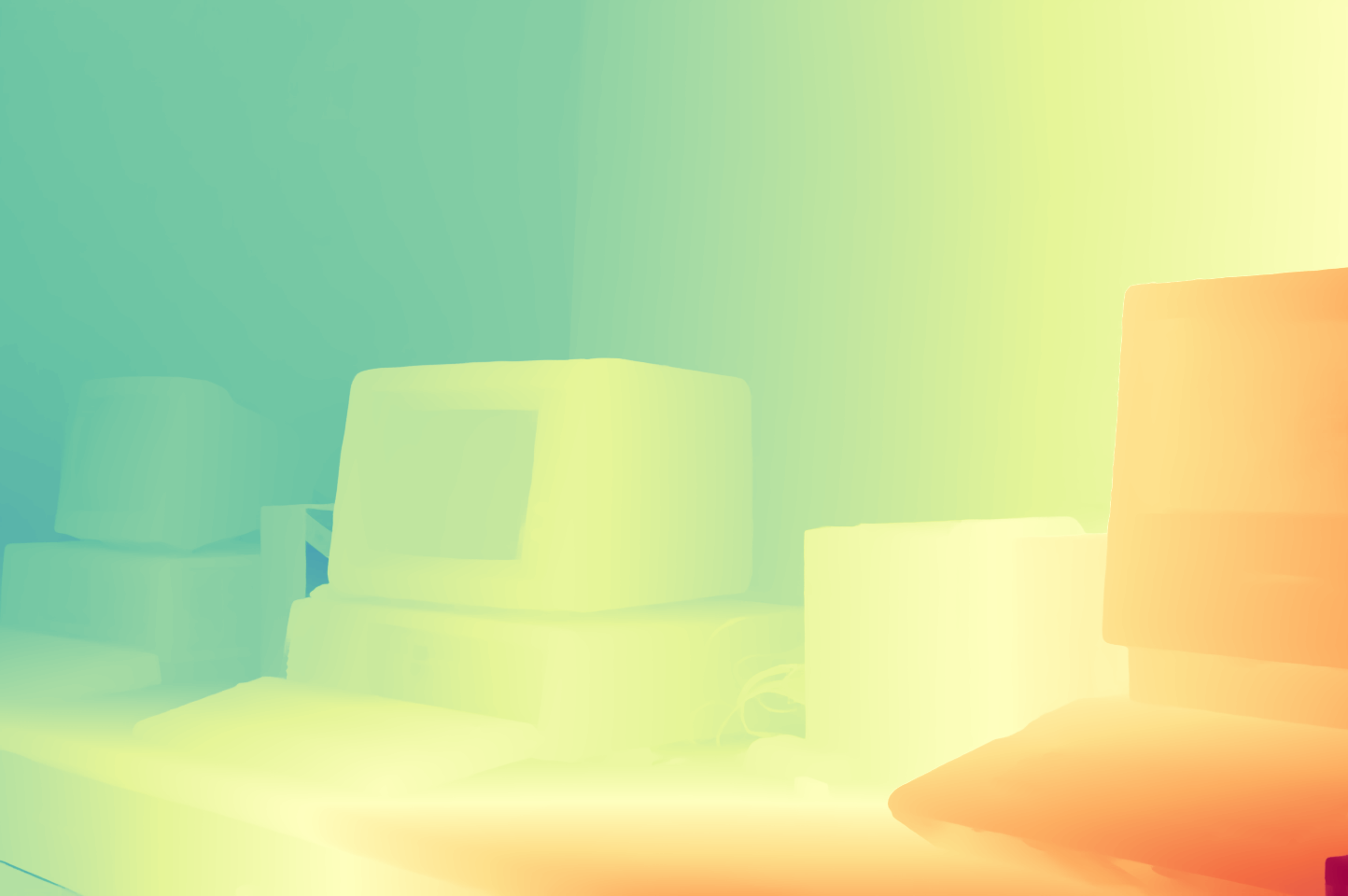} &
        \includegraphics[width=0.32\textwidth]{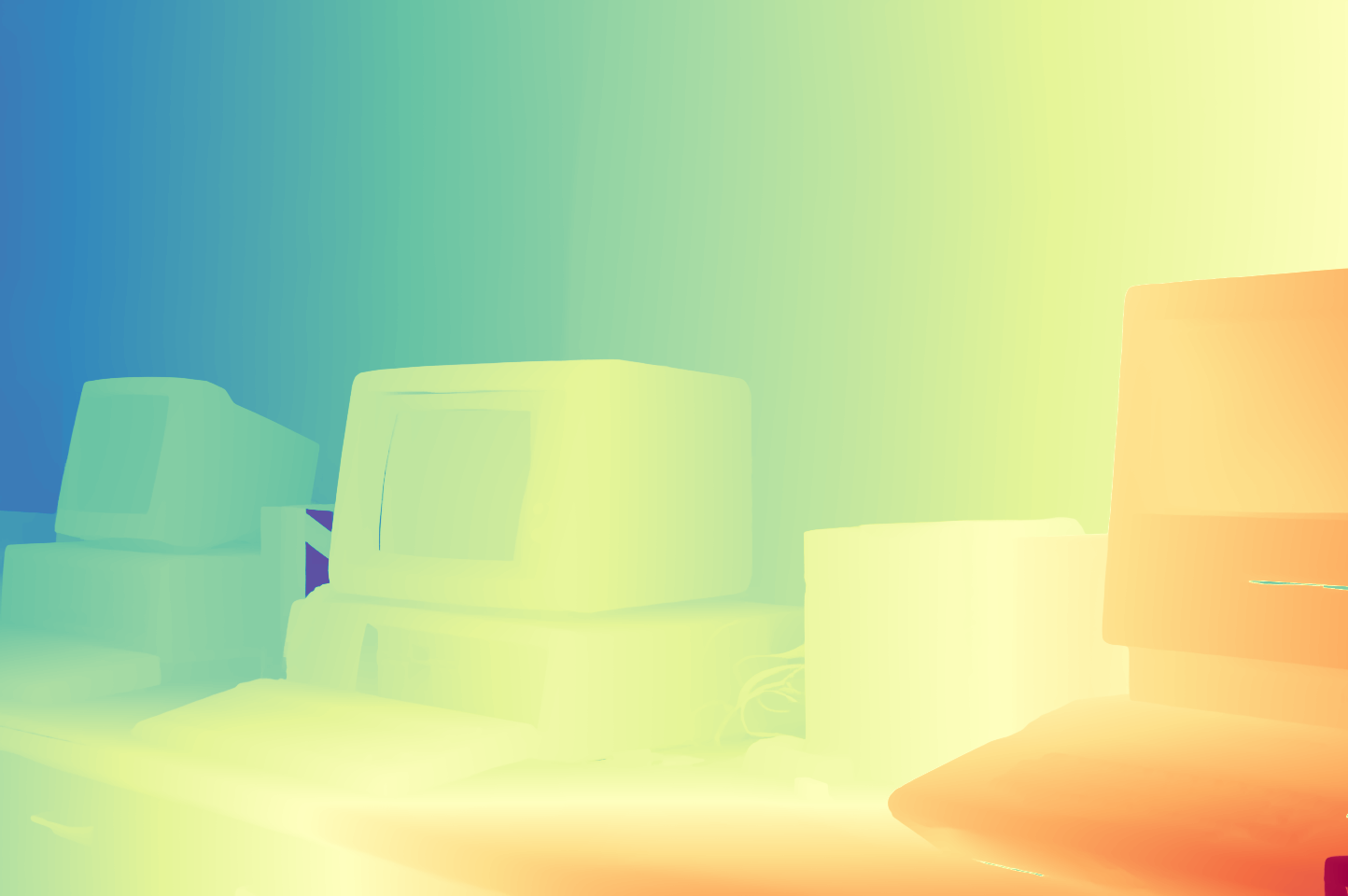} \\
        \small NMRF \cite{guan2024neural} &
        \small Selective-IGEV \cite{wang2024selective} &
        \textbf{\method (ours)} \\
        \includegraphics[width=0.32\textwidth]{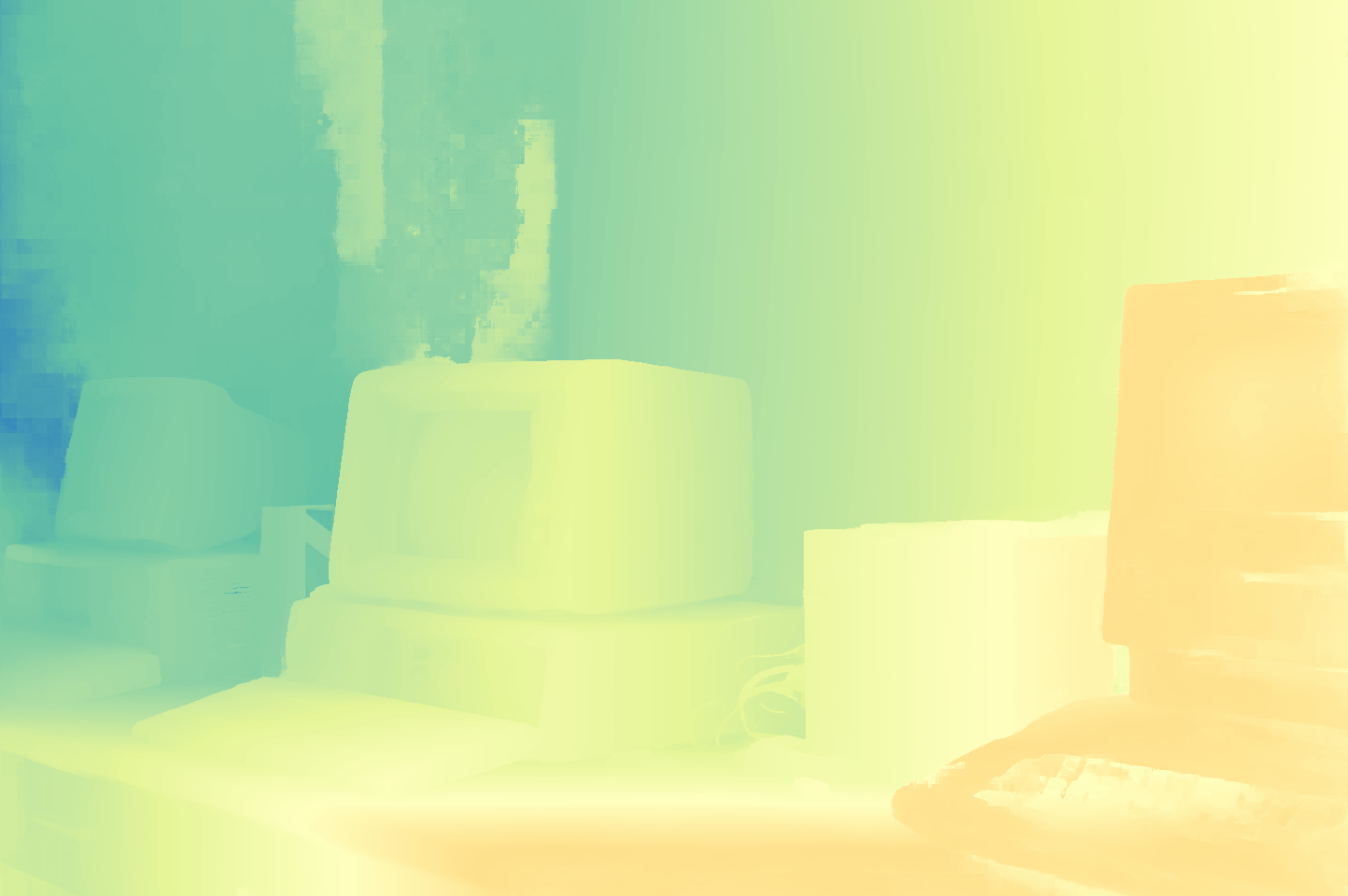} &
        \includegraphics[width=0.32\textwidth]{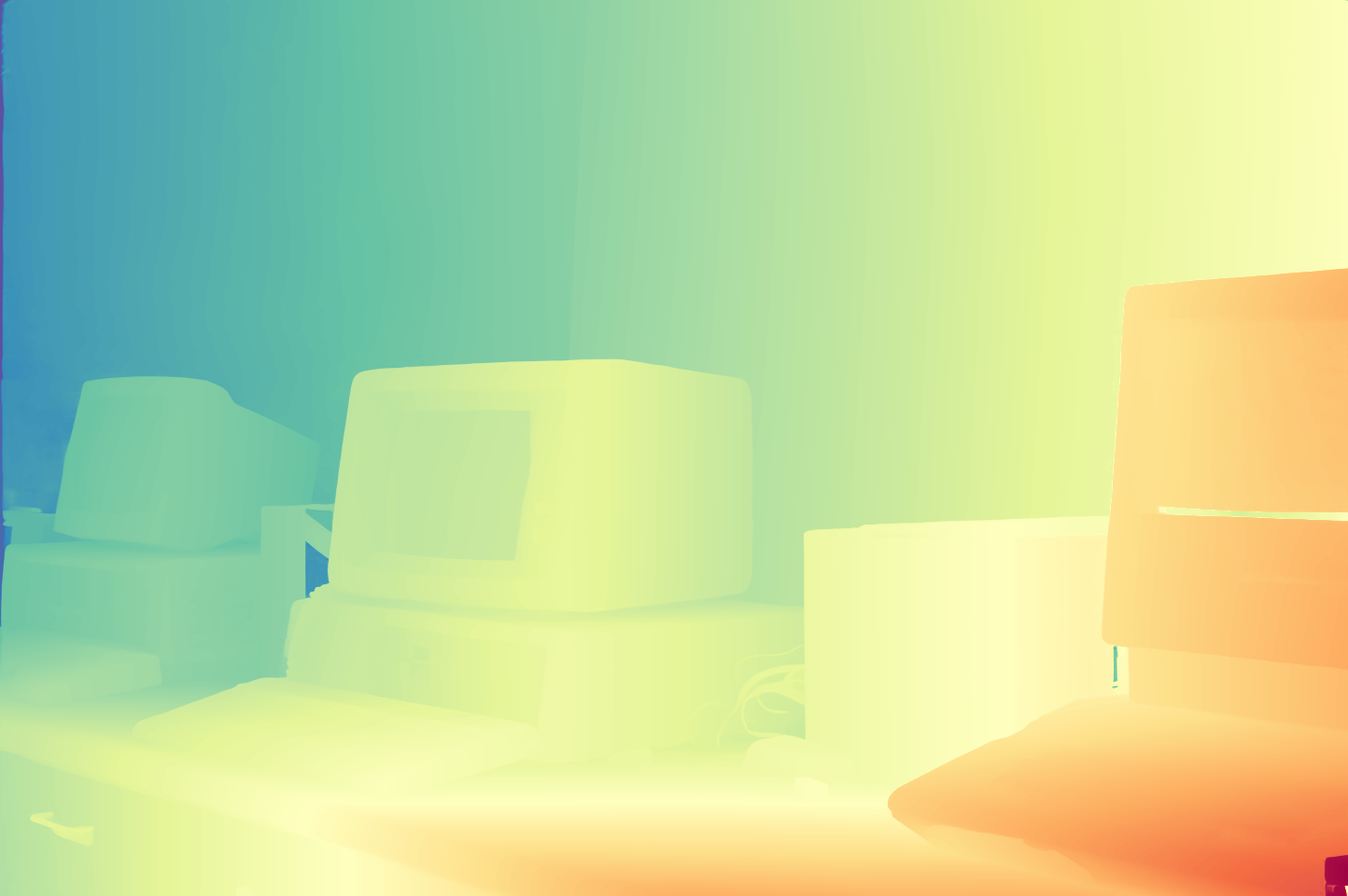} &
        \includegraphics[width=0.32\textwidth]{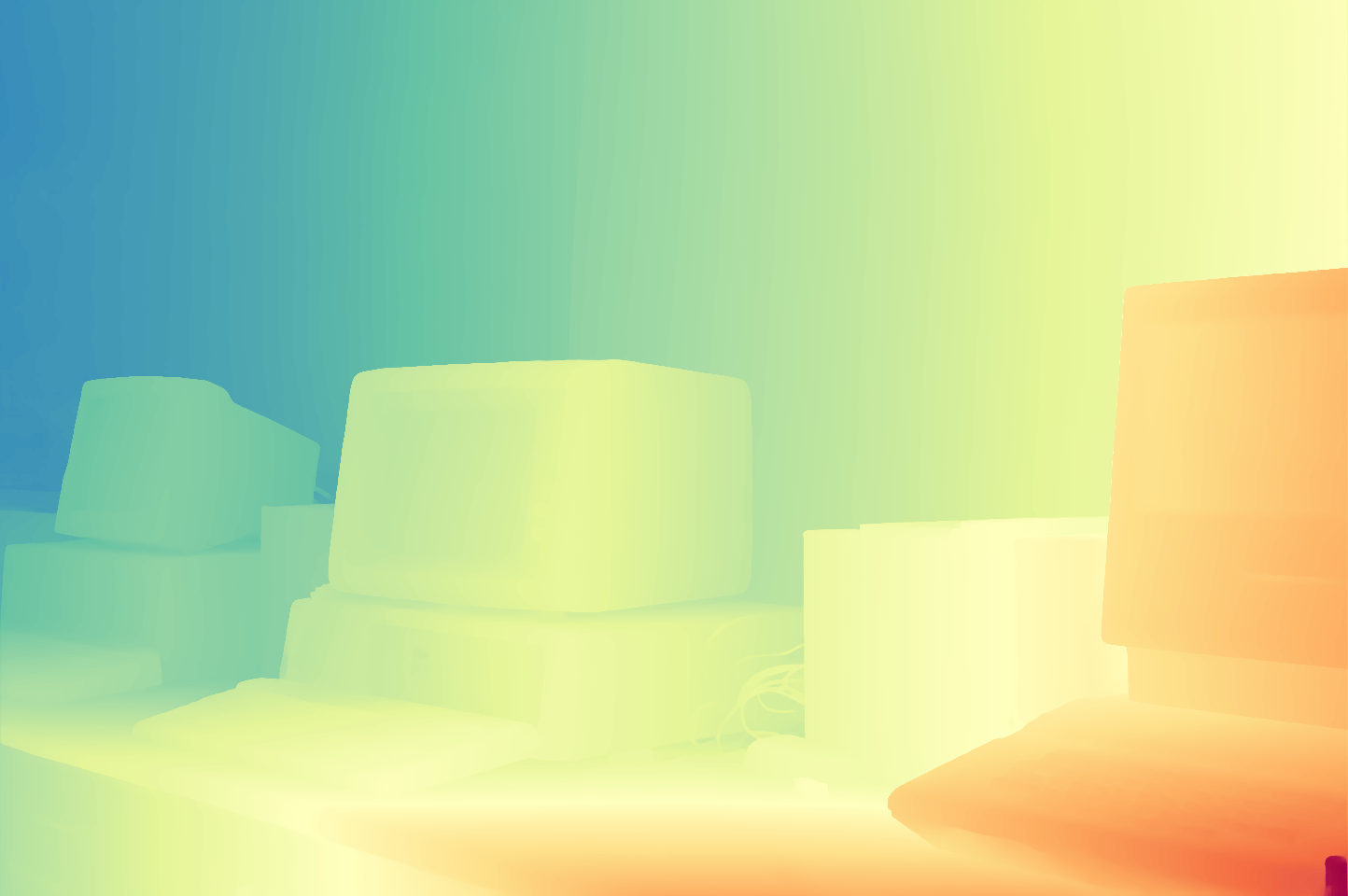} \\ 
    \end{tabular}\vspace{-0.3cm}
    \caption{\textbf{Qualitative Results -- Middlebury 2014.} Predictions by state-of-the-art models and \method.}
    \label{fig:qual_midd14}\vspace{-0.3cm}
\end{figure*}

Figure \ref{fig:qual_midd14} reports two image pairs from Middlebury 2014 (respectively, \textit{Adirondack} and \textit{Vintage}). On the former, \method preserves the very thin holes on the back of the chair, while on the latter it can properly estimate the disparity for the displays, where existing methods are fooled and predict holes.

\clearpage

\begin{figure*}[h]
    \centering
    \renewcommand{\tabcolsep}{1pt}
    \begin{tabular}{ccc}
        \small RGB &
        \small RAFT-Stereo \cite{lipson2021raft} &
        \small DLNR \cite{zhao2023high} \\
        \includegraphics[width=0.3\textwidth]{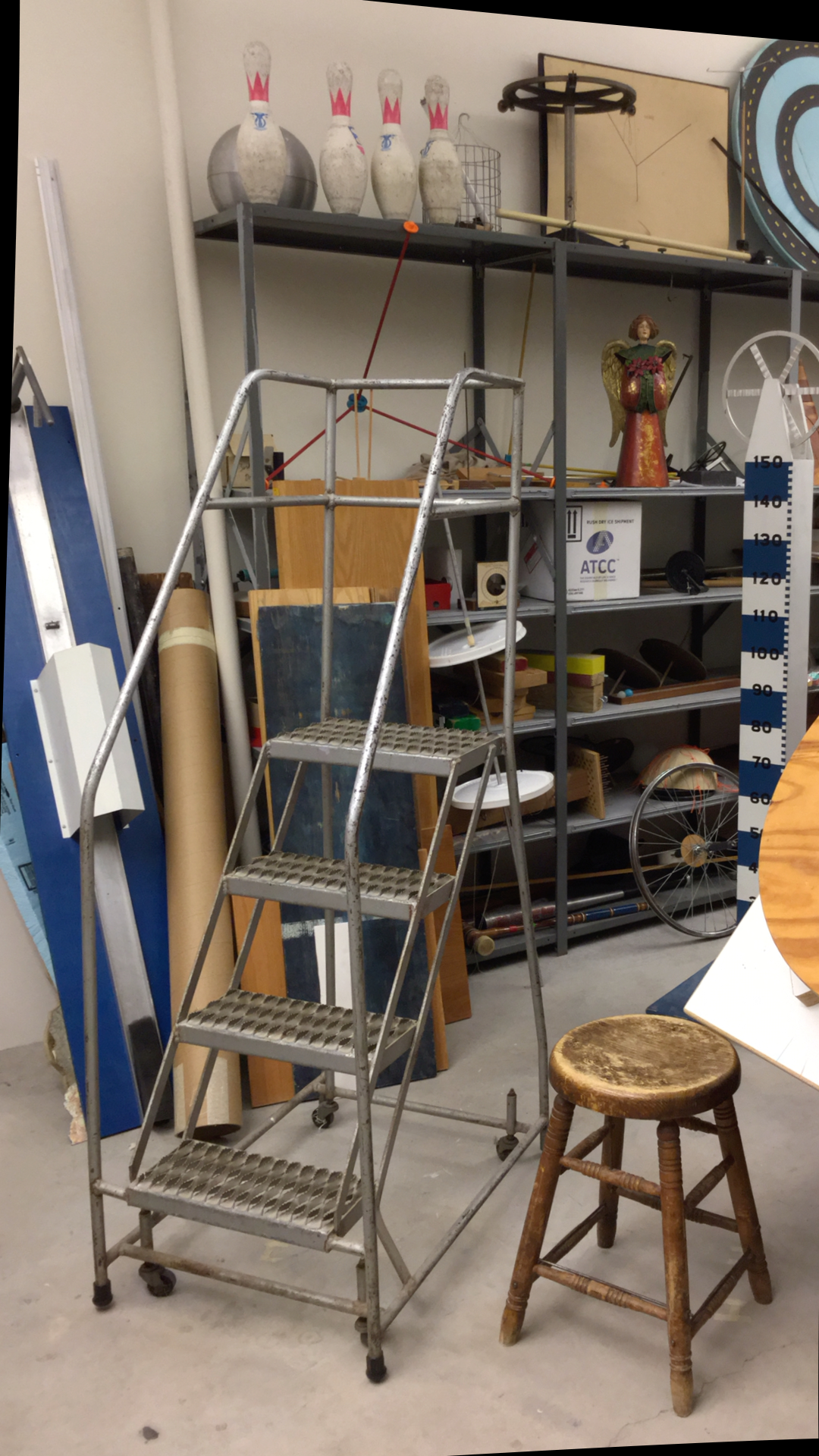} & 
        \includegraphics[width=0.3\textwidth]{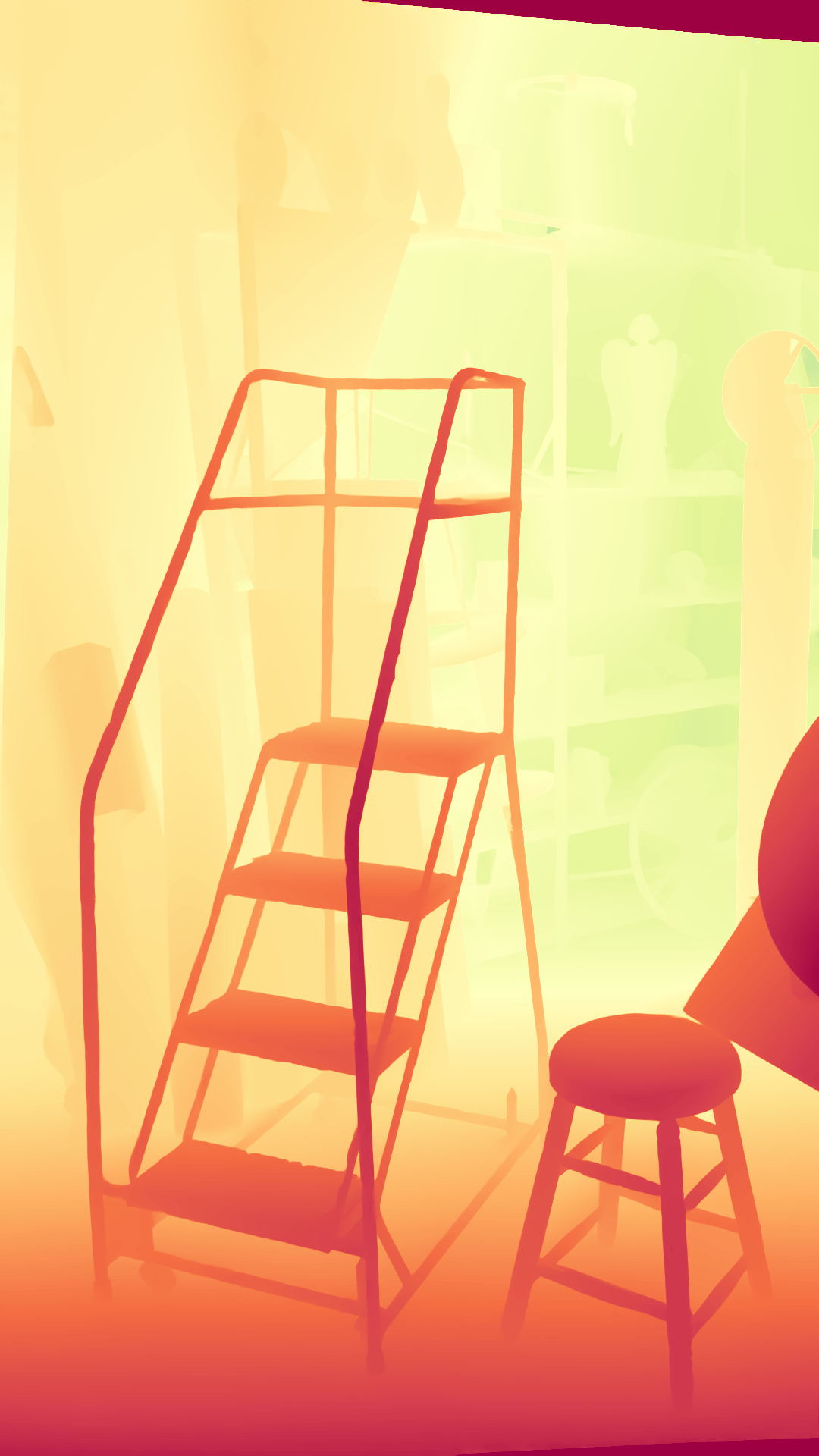} &
        \includegraphics[width=0.3\textwidth]{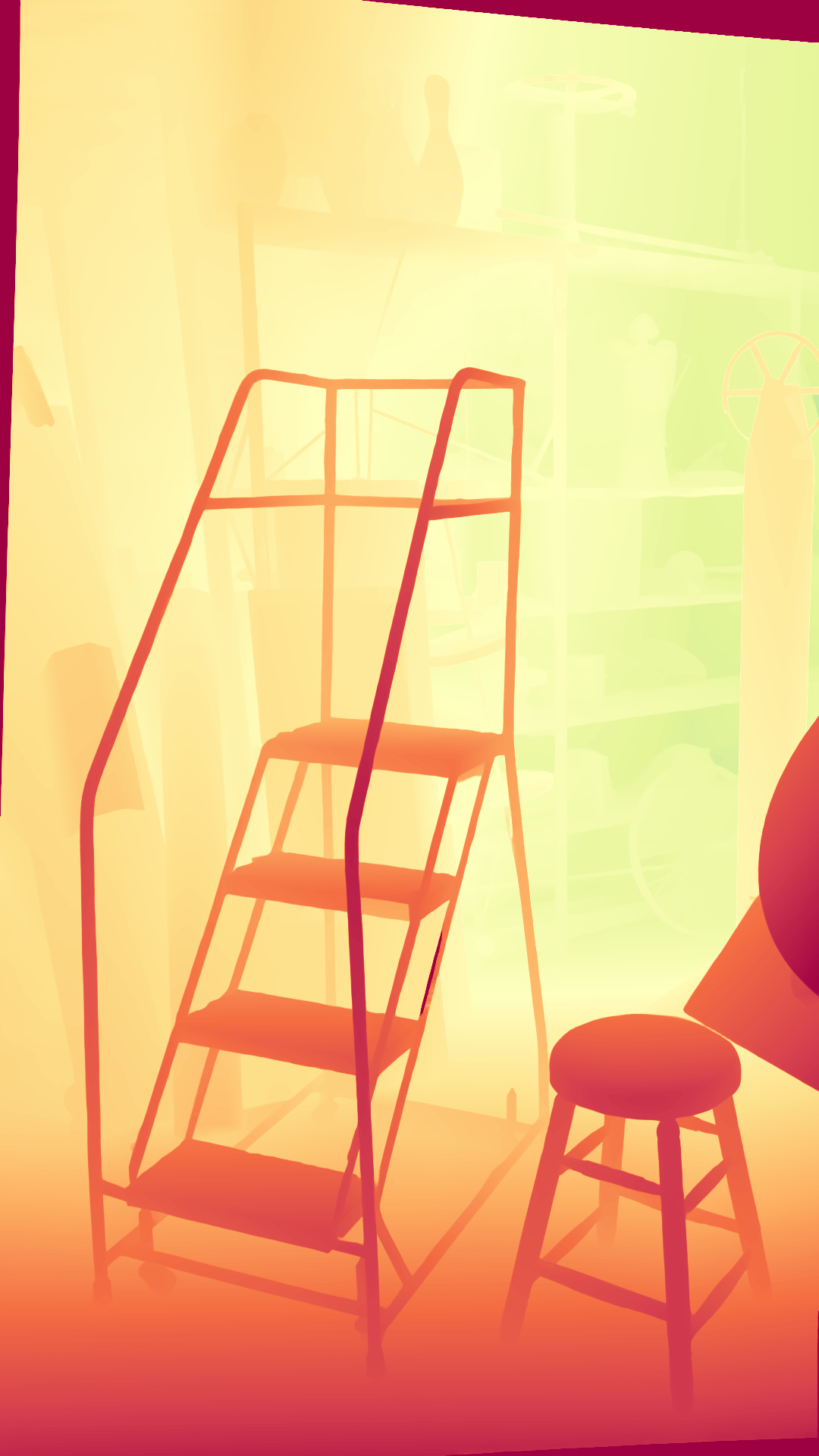} \\
        \small NMRF \cite{guan2024neural} &
        \small Selective-IGEV \cite{wang2024selective} &
        \textbf{\method (ours)} \\
        \includegraphics[width=0.3\textwidth]{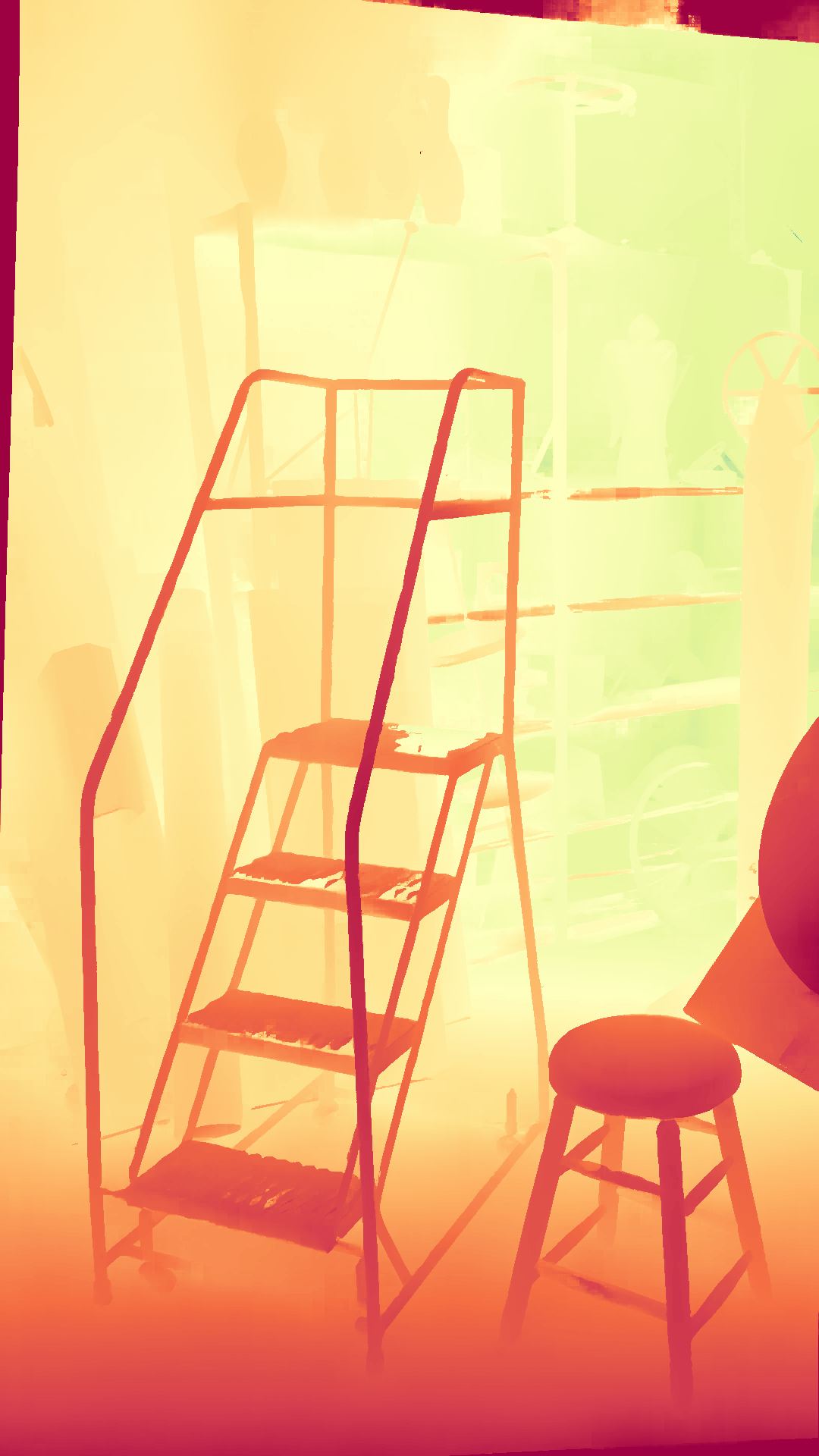} &
        \includegraphics[width=0.3\textwidth]{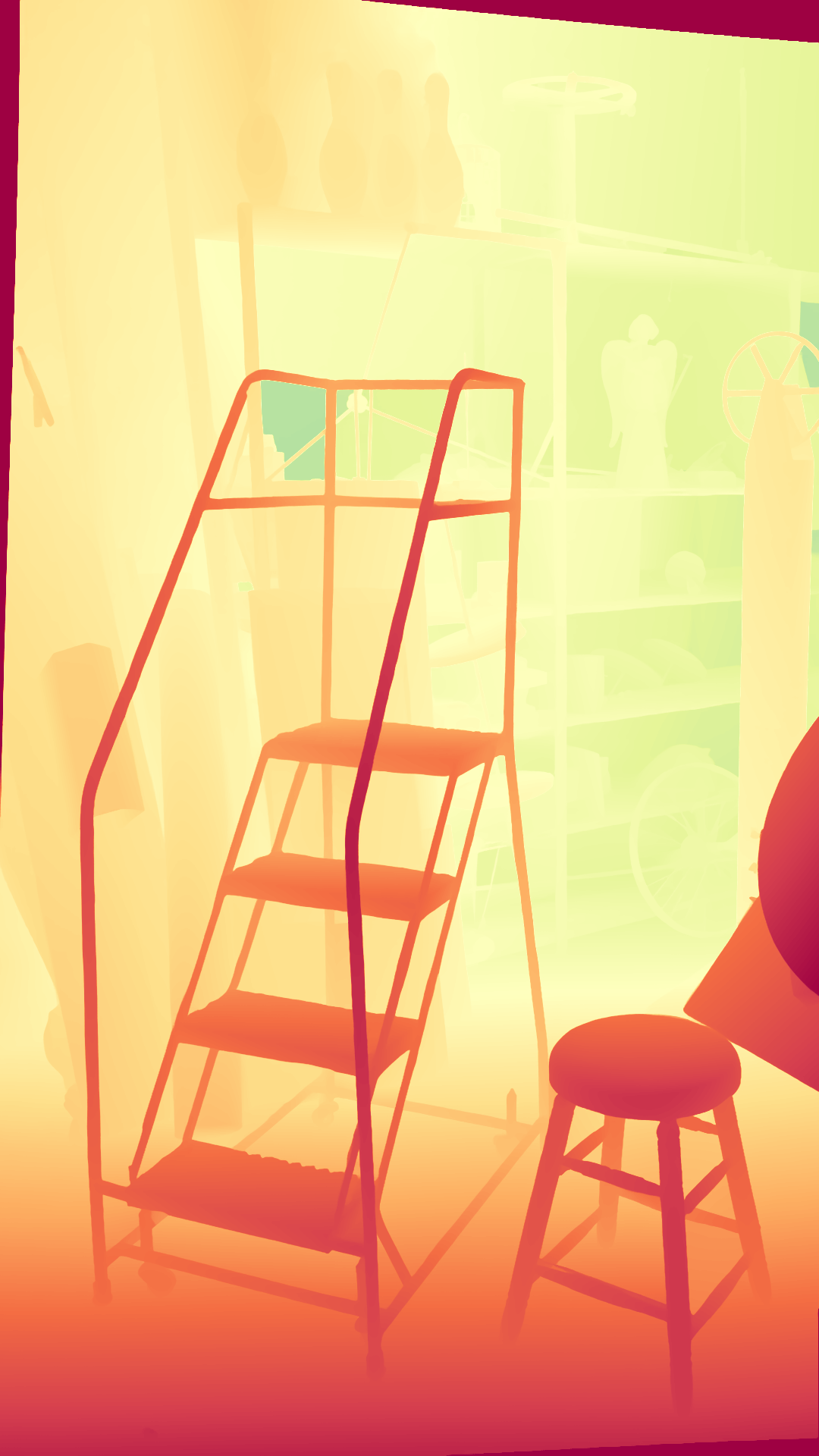} &
        \includegraphics[width=0.3\textwidth]{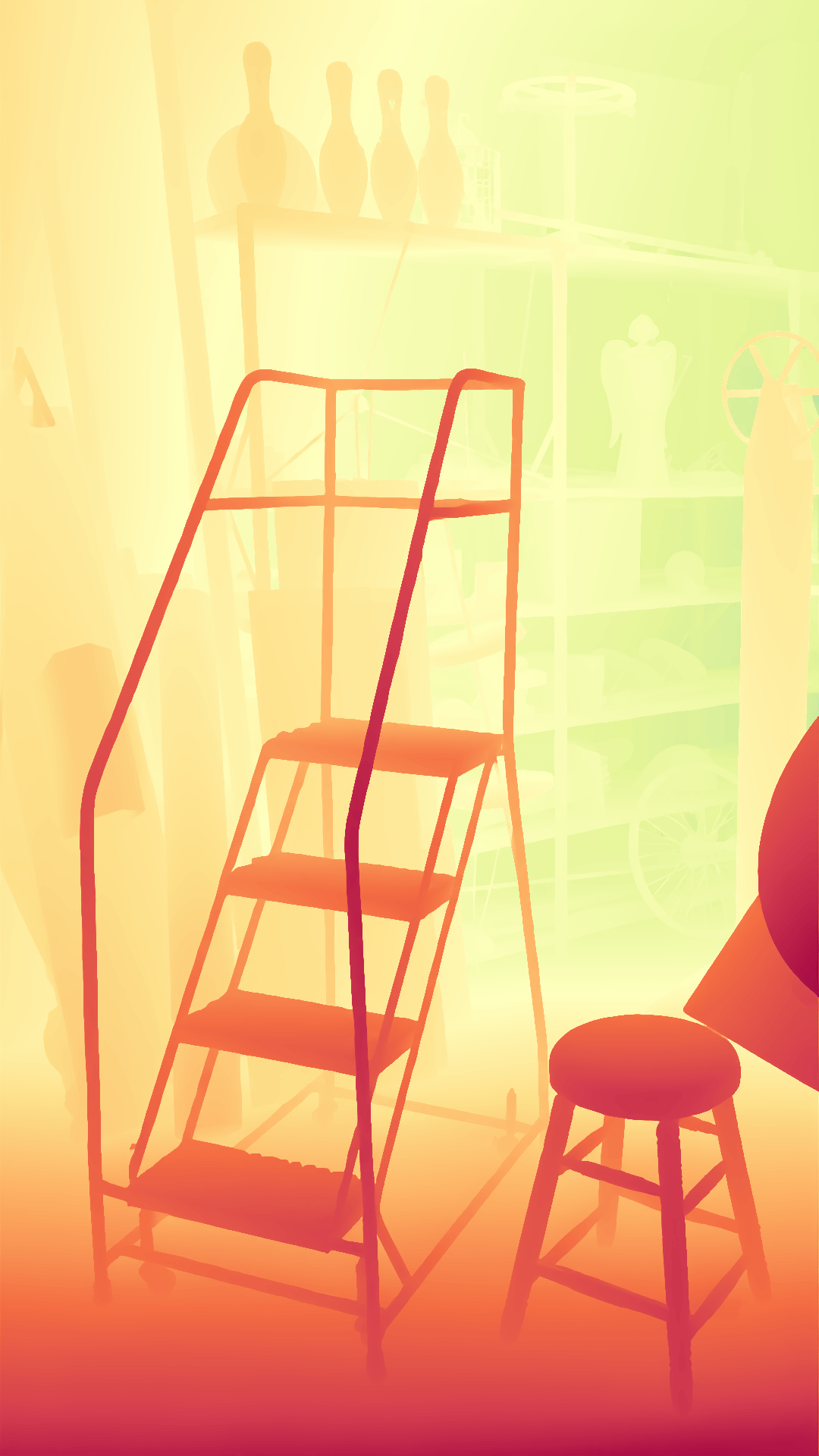} \\ 
    \end{tabular}\vspace{-0.3cm}
    \caption{\textbf{Qualitative Results -- Middlebury 2021 (part 1).} Predictions by state-of-the-art models and \method.}
    \label{fig:qual_midd21_1}\vspace{-0.3cm}
\end{figure*}
\begin{figure*}[h]
    \centering
    \renewcommand{\tabcolsep}{1pt}
    \begin{tabular}{ccc}
        \small RGB &
        \small RAFT-Stereo \cite{lipson2021raft} &
        \small DLNR \cite{zhao2023high} \\
        \includegraphics[width=0.3\textwidth]{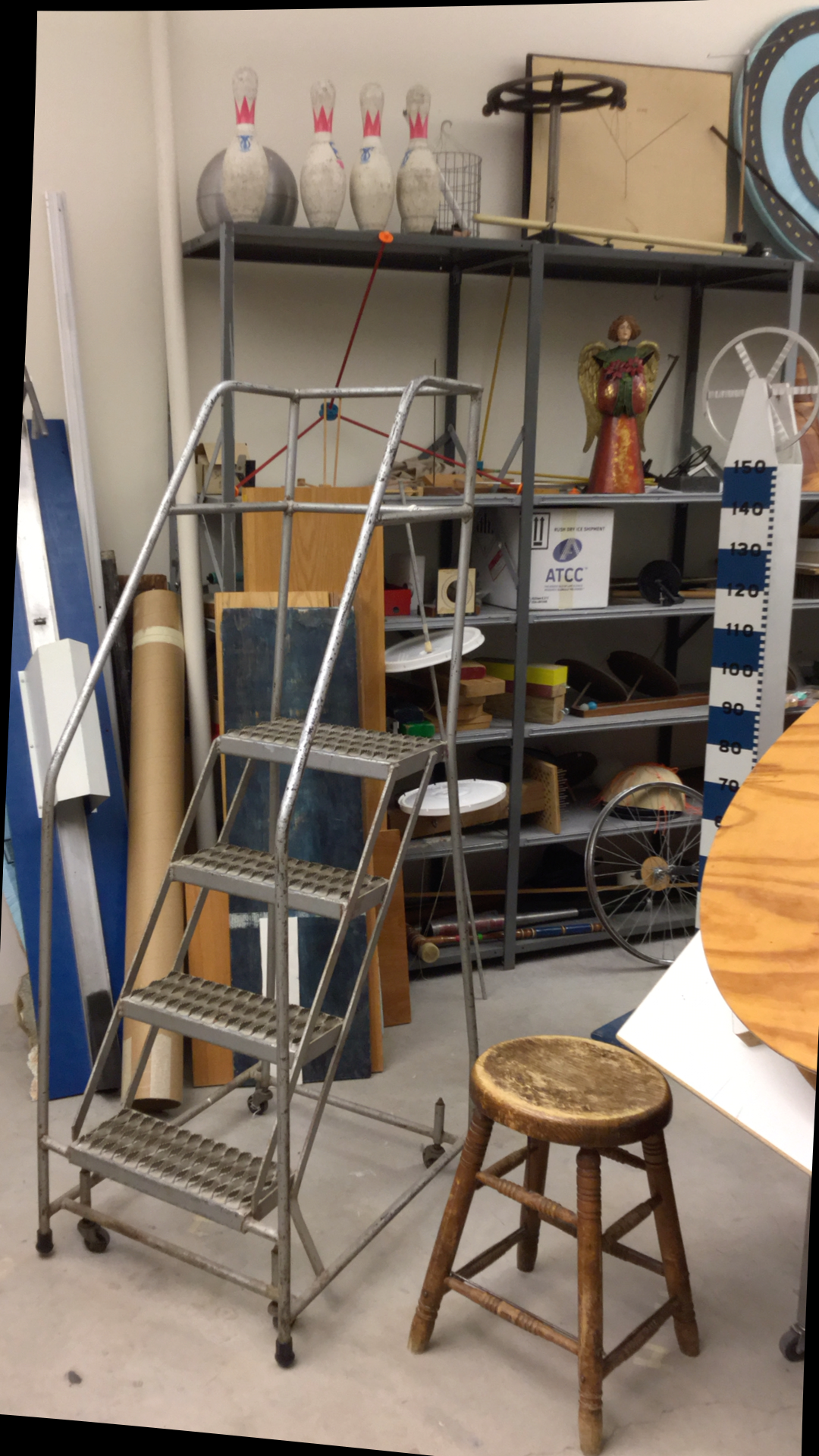} & 
        \includegraphics[width=0.3\textwidth]{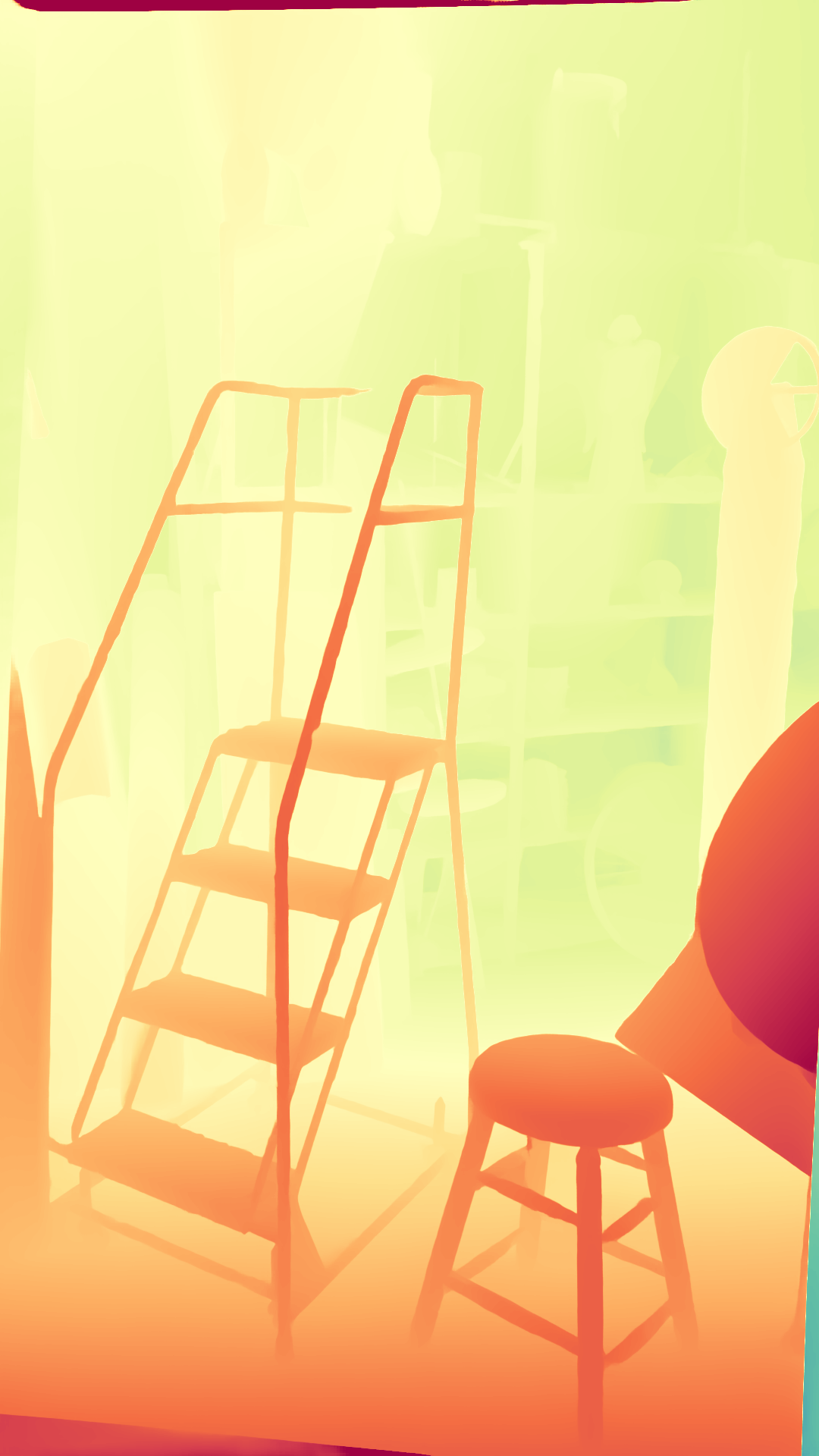} &
        \includegraphics[width=0.3\textwidth]{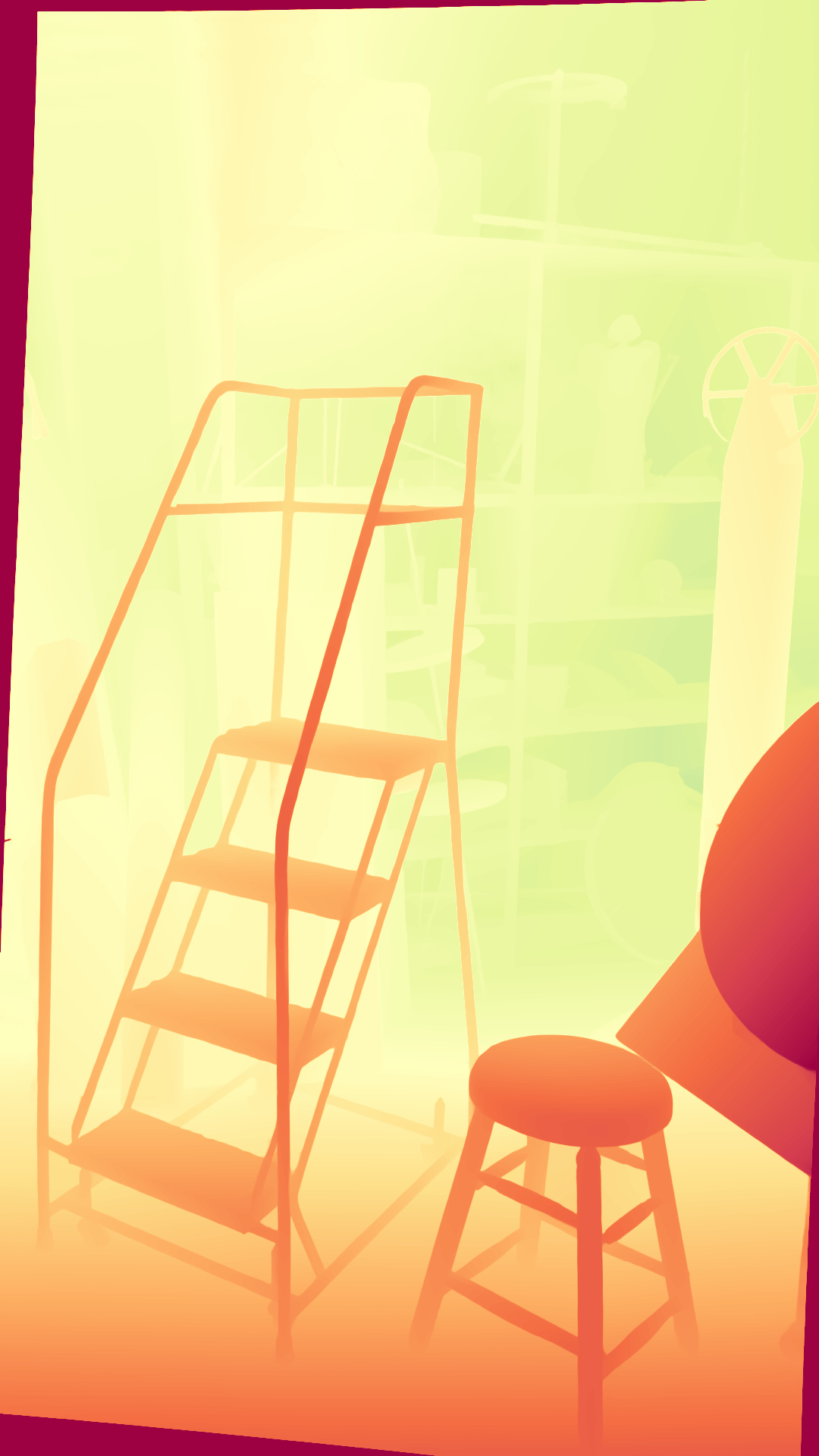} \\
        \small NMRF \cite{guan2024neural} &
        \small Selective-IGEV \cite{wang2024selective} &
        \textbf{\method (ours)} \\
        \includegraphics[width=0.3\textwidth]{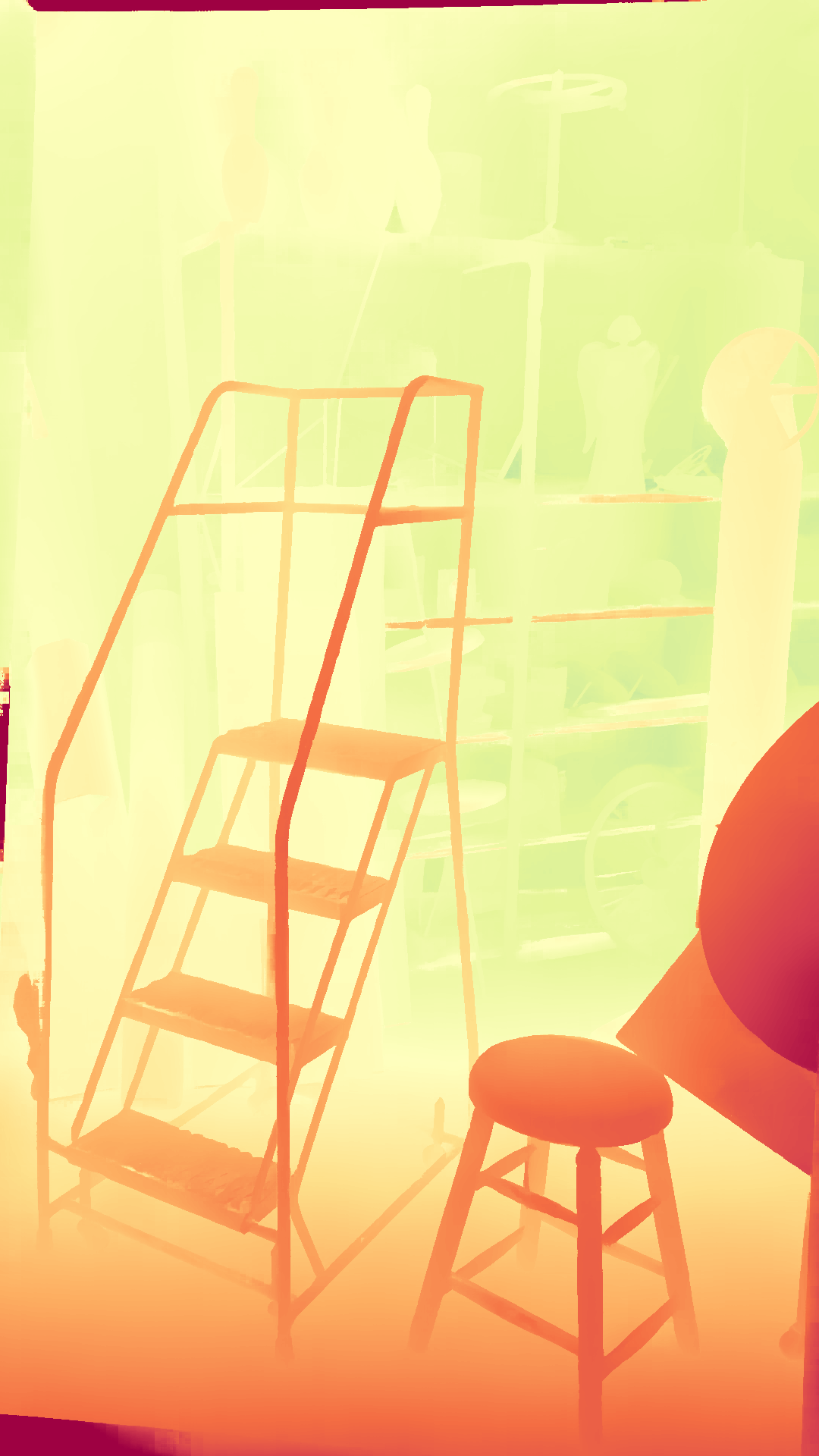} &
        \includegraphics[width=0.3\textwidth]{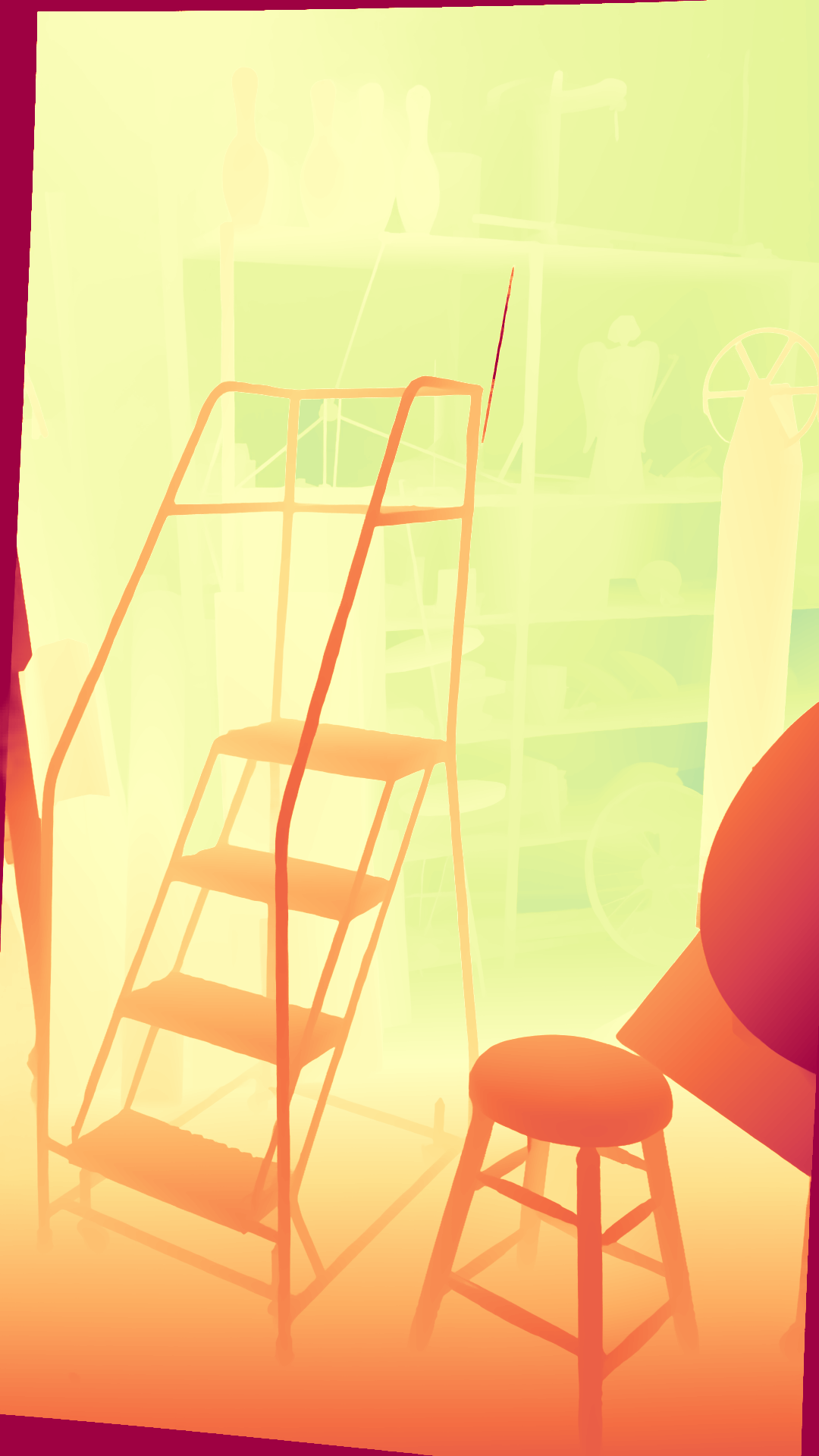} &
        \includegraphics[width=0.3\textwidth]{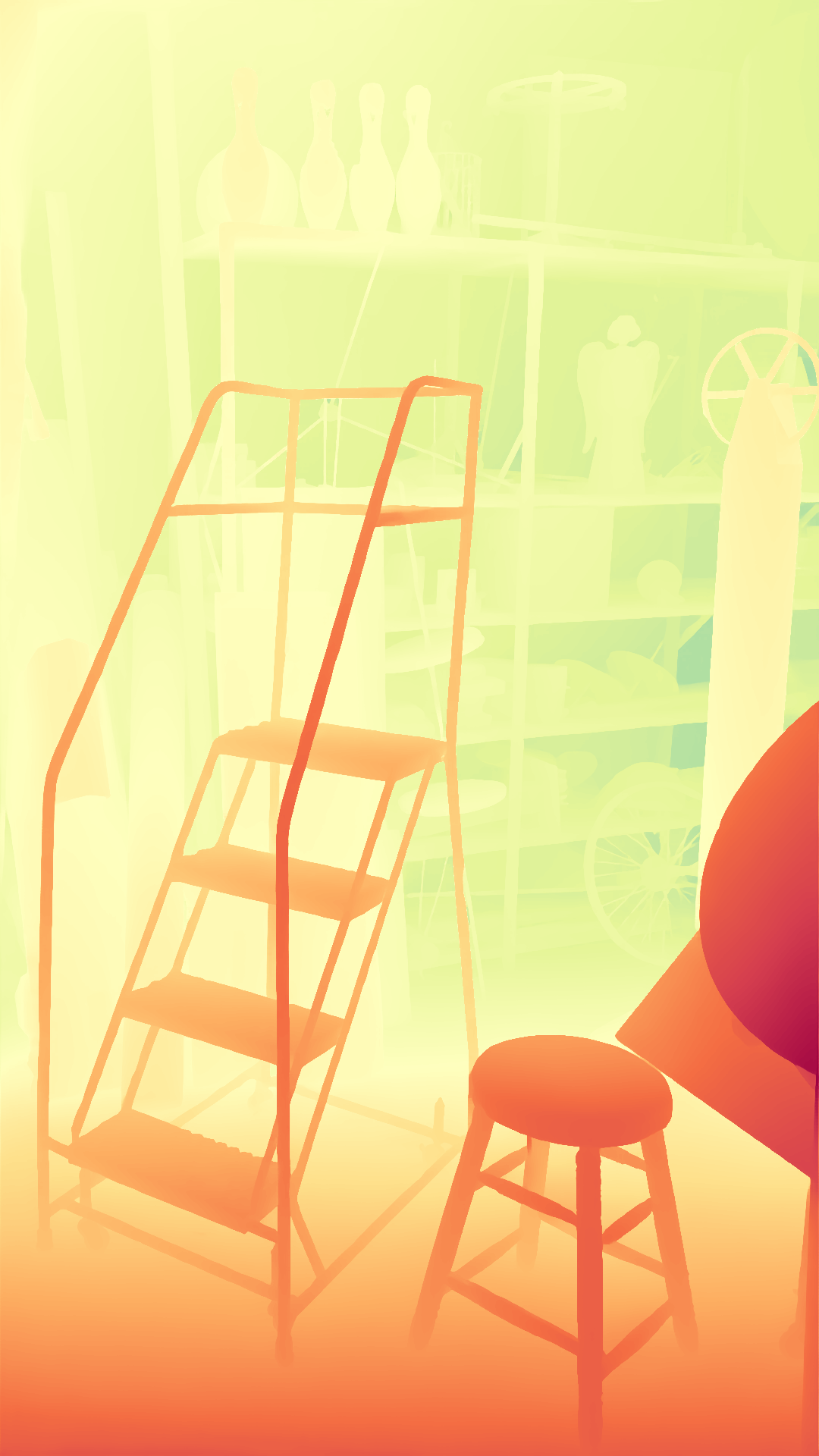} \\ 
    \end{tabular}\vspace{-0.3cm}
    \caption{\textbf{Qualitative Results -- Middlebury 2021 (part 2).} Predictions by state-of-the-art models and \method.}
    \label{fig:qual_midd21_2}\vspace{-0.3cm}
\end{figure*}

Figure \ref{fig:qual_midd21_1} and \ref{fig:qual_midd21_2} shows the results on two samples from Middlebury 2021, peculiar for their aspect ratio (respectively, \textit{ladder1} and \textit{ladder2}). Although existing models perform quite well on both, they fail to preserve the skittles on the top of the scene, whereas \method properly predicts their structure.

\clearpage

\clearpage

\begin{figure*}[h]
    \centering
    \renewcommand{\tabcolsep}{1pt}
    \begin{tabular}{ccc}
        \small RGB &
        \small RAFT-Stereo \cite{lipson2021raft} &
        \small DLNR \cite{zhao2023high} \\
        \includegraphics[width=0.31\textwidth]{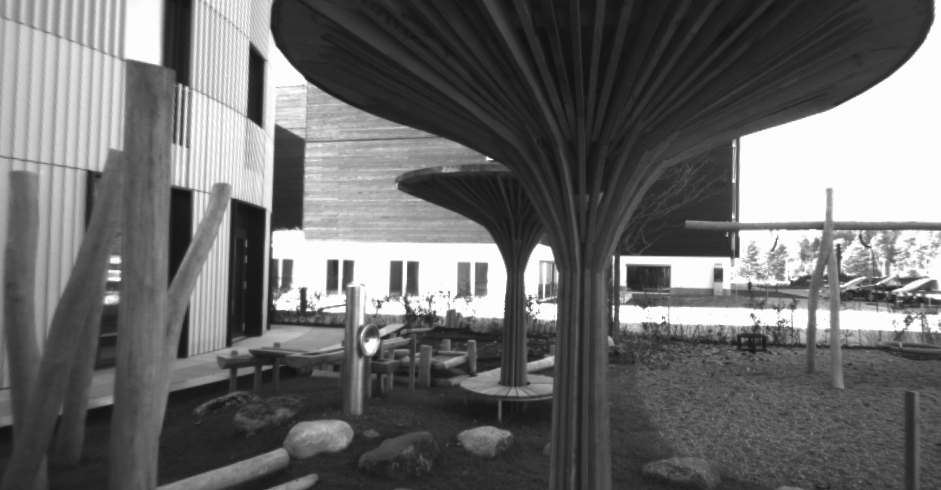} & 
        \includegraphics[width=0.31\textwidth]{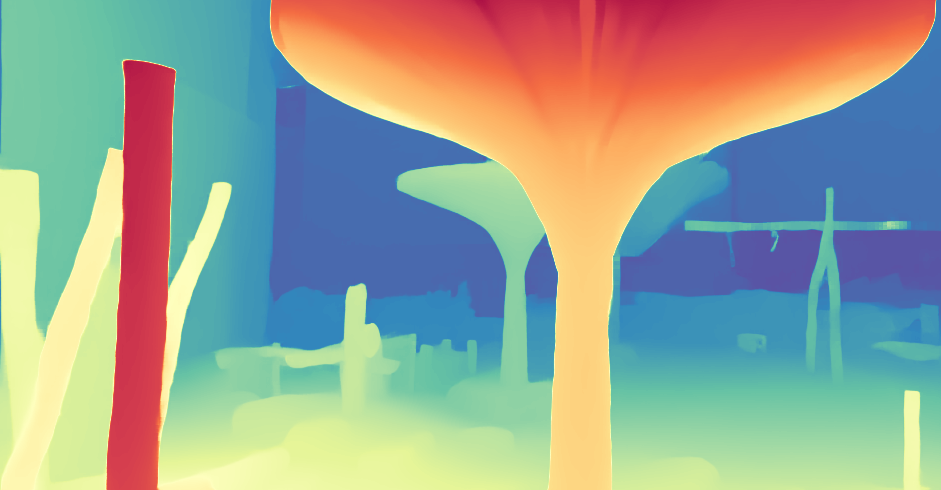} &
        \includegraphics[width=0.31\textwidth]{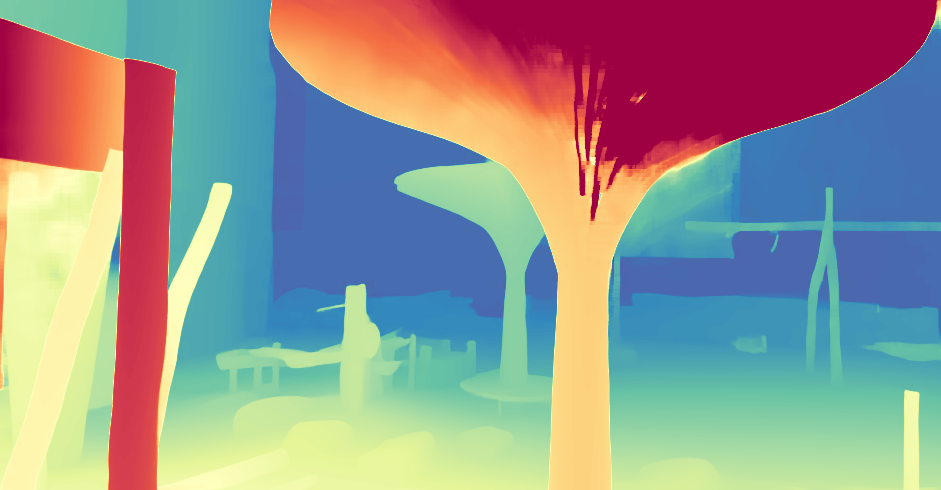} \\
        \small NMRF \cite{guan2024neural} &
        \small Selective-IGEV \cite{wang2024selective} &
        \textbf{\method (ours)} \\
        \includegraphics[width=0.31\textwidth]{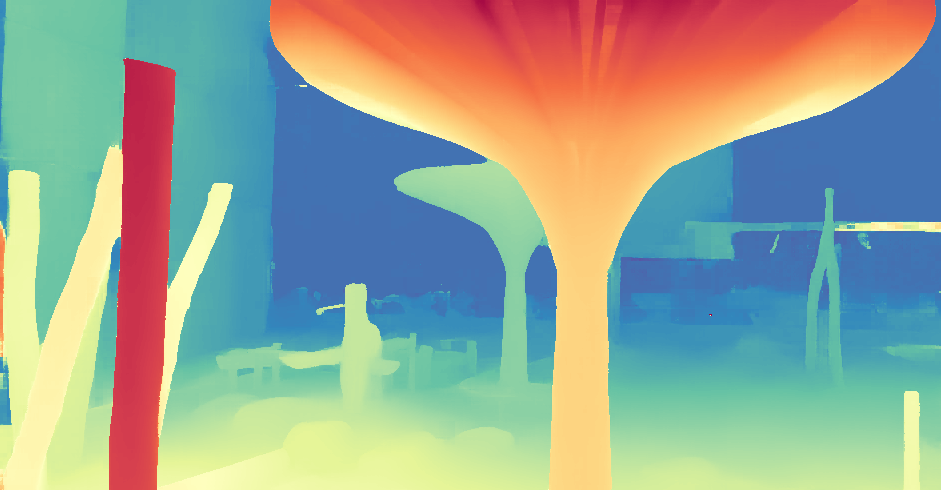} &
        \includegraphics[width=0.31\textwidth]{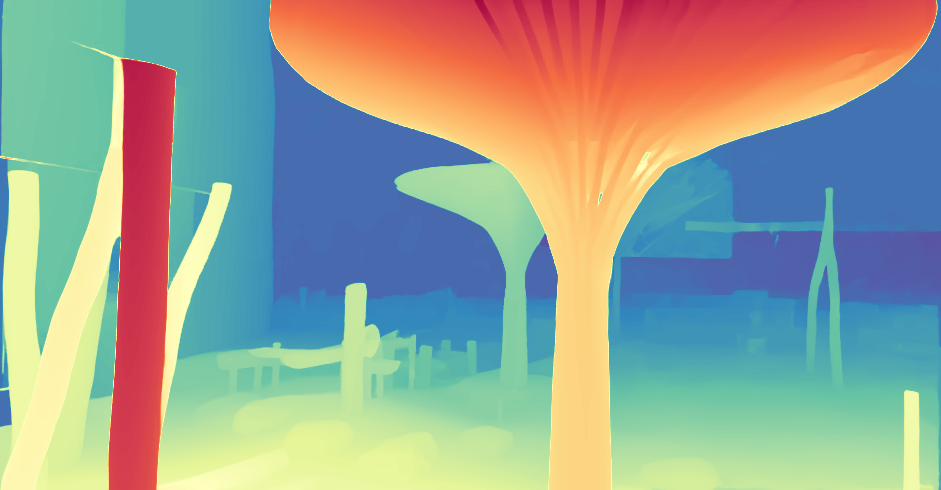} &
        \includegraphics[width=0.31\textwidth]{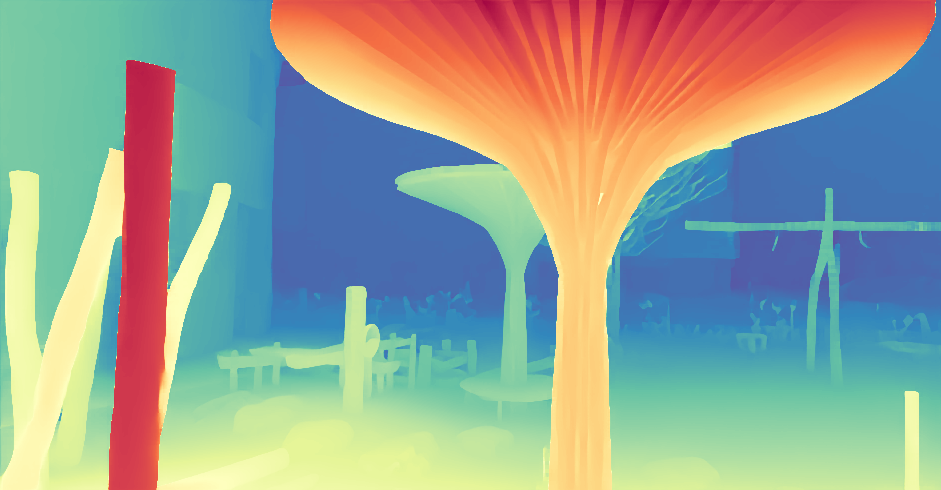} \\ \\
        
        \small RGB &
        \small RAFT-Stereo \cite{lipson2021raft} &
        \small DLNR \cite{zhao2023high} \\
        \includegraphics[width=0.31\textwidth]{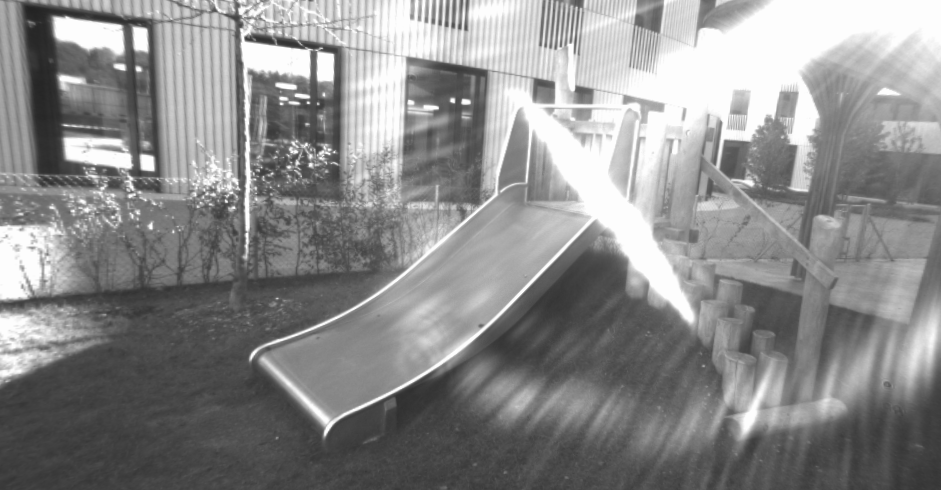} & 
        \includegraphics[width=0.31\textwidth]{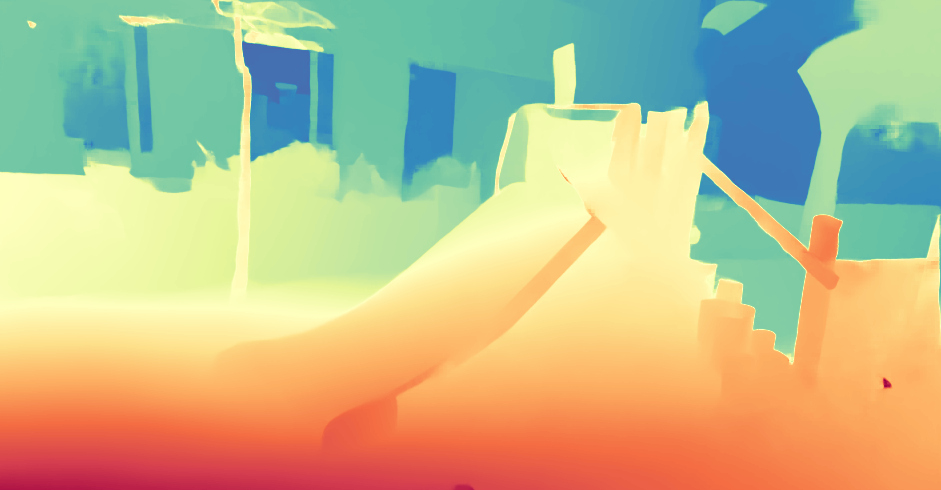} &
        \includegraphics[width=0.31\textwidth]{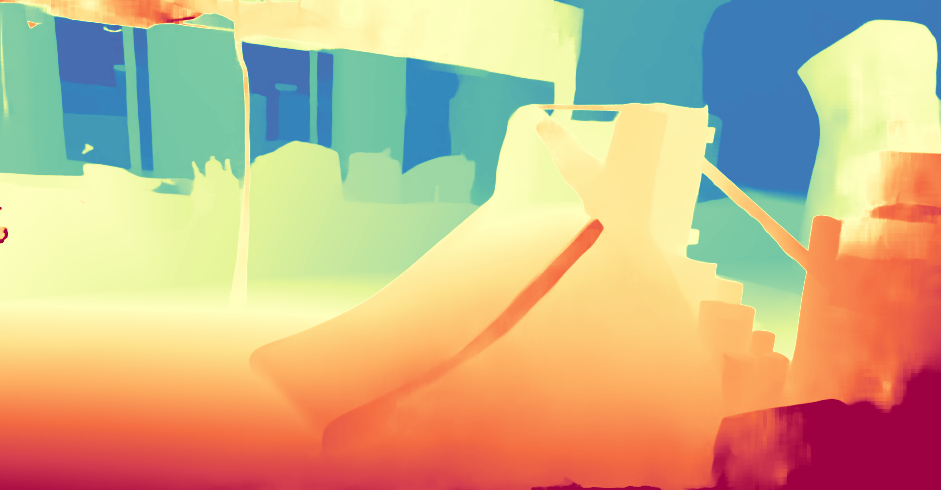} \\
        \small NMRF \cite{guan2024neural} &
        \small Selective-IGEV \cite{wang2024selective} &
        \textbf{\method (ours)} \\
        \includegraphics[width=0.31\textwidth]{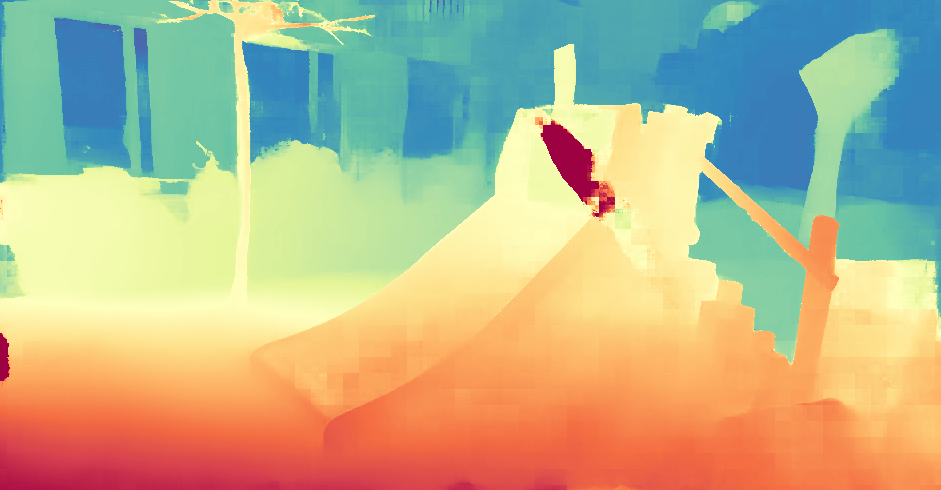} &
        \includegraphics[width=0.31\textwidth]{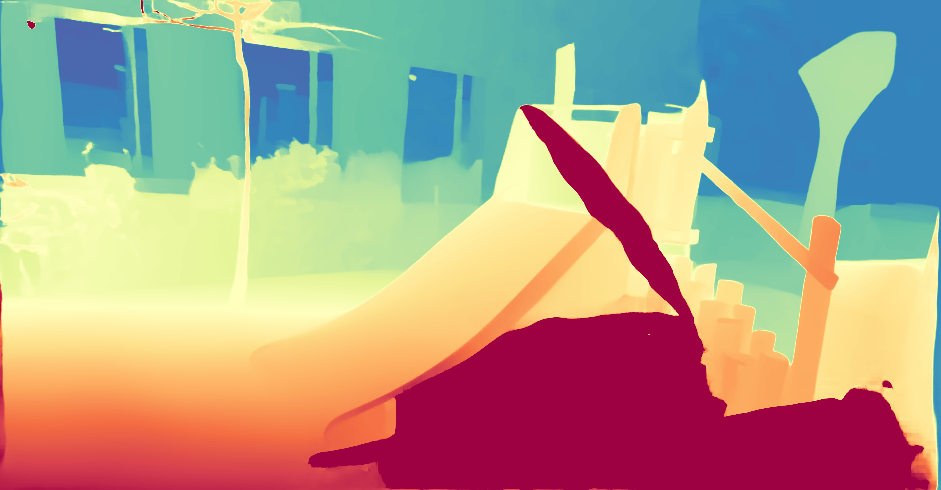} &
        \includegraphics[width=0.31\textwidth]{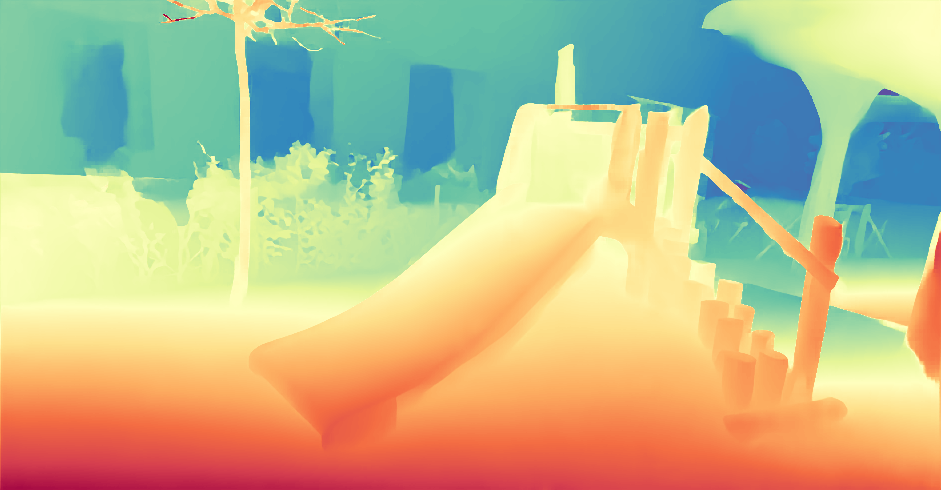} \\ \\

        \small RGB &
        \small RAFT-Stereo \cite{lipson2021raft} &
        \small DLNR \cite{zhao2023high} \\
        \includegraphics[width=0.31\textwidth]{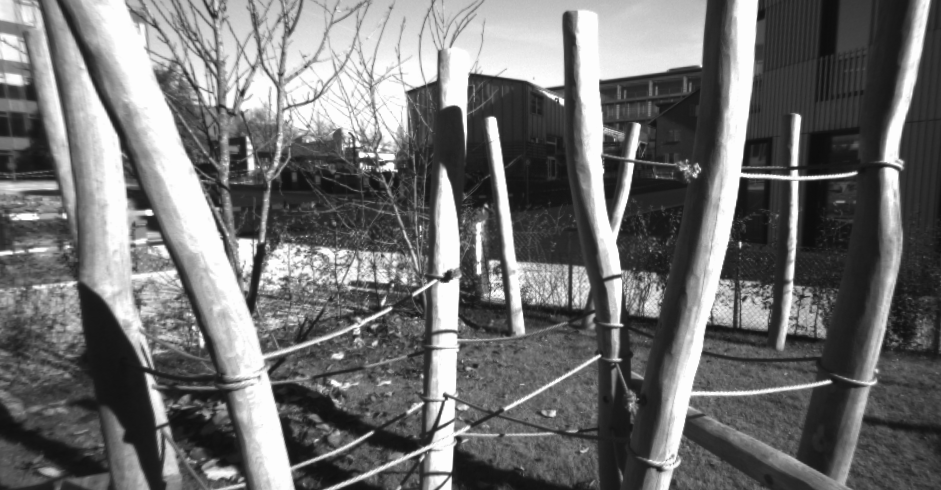} & 
        \includegraphics[width=0.31\textwidth]{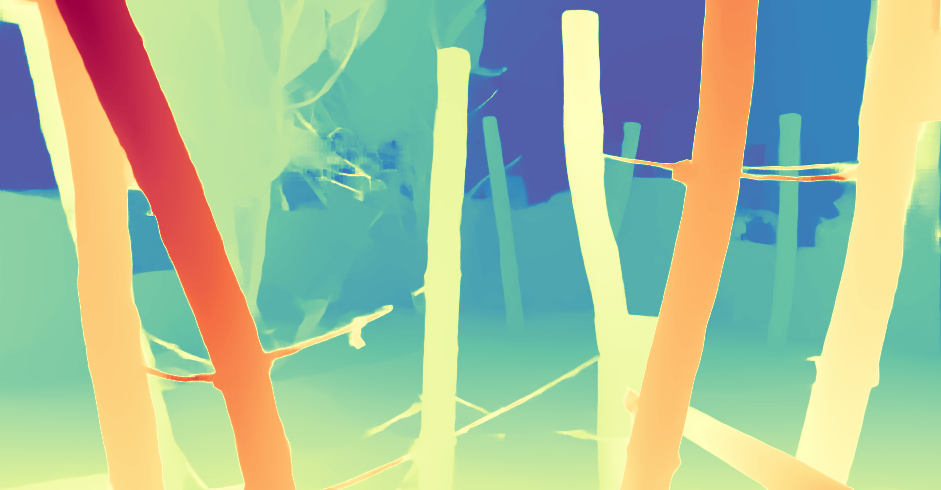} &
        \includegraphics[width=0.31\textwidth]{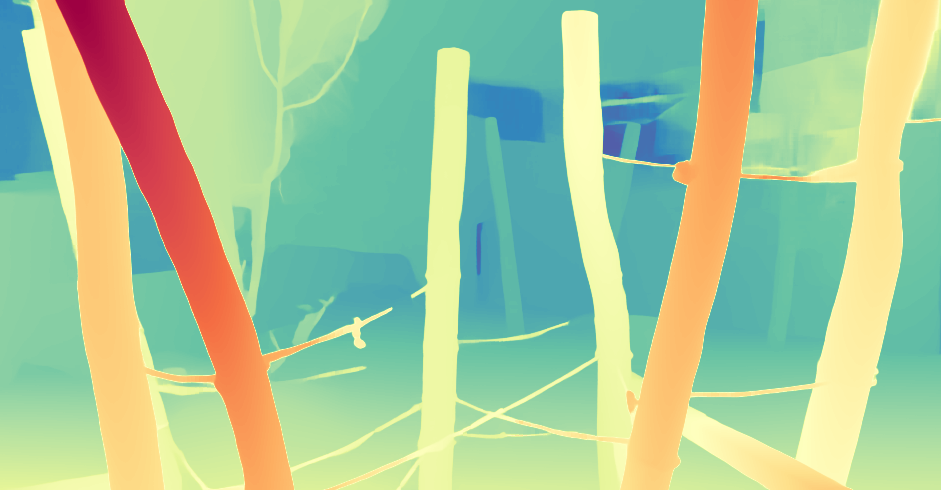} \\
        \includegraphics[width=0.31\textwidth]{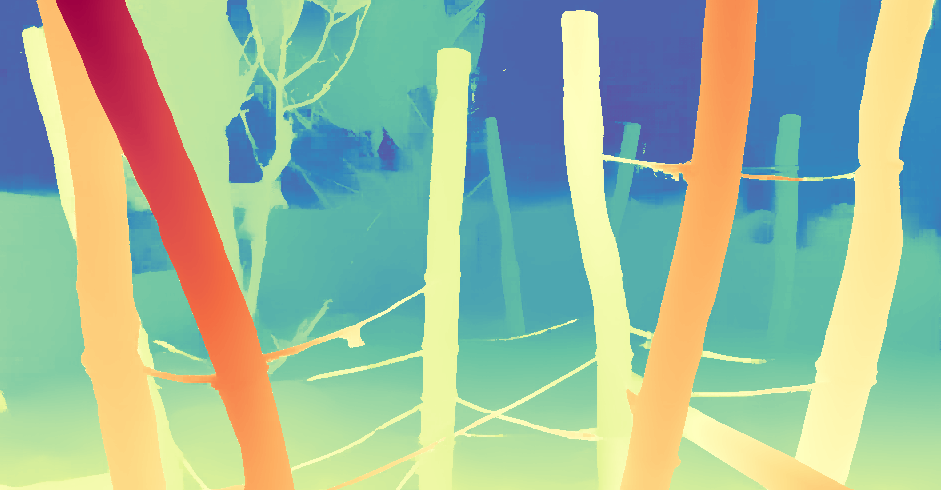} &
        \includegraphics[width=0.31\textwidth]{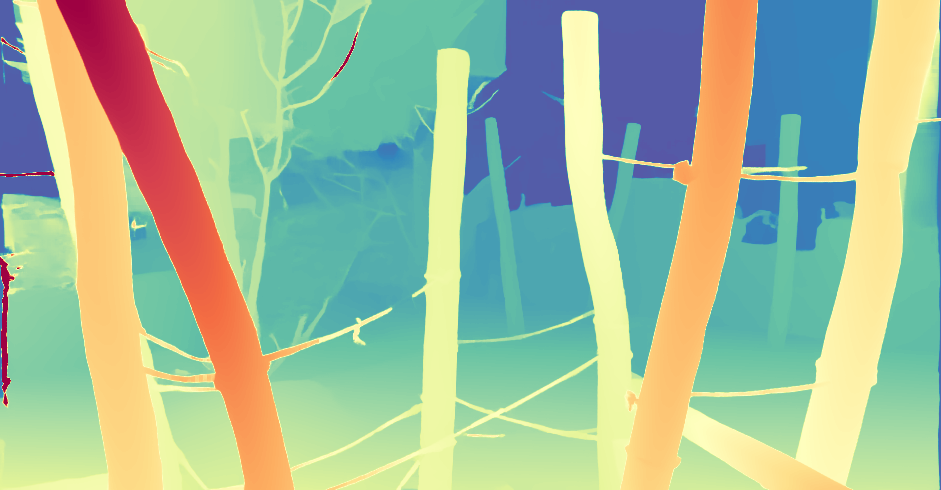} &
        \includegraphics[width=0.31\textwidth]{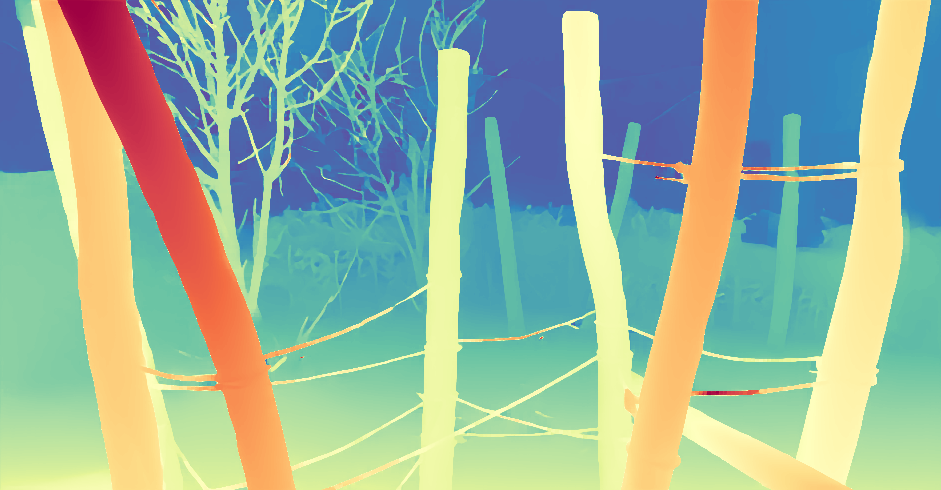} \\ 
    \end{tabular}\vspace{-0.3cm}
    \caption{\textbf{Qualitative Results -- ETH3D.} Predictions by state-of-the-art models and \method.}
    \label{fig:qual_eth3d}\vspace{-0.3cm}
\end{figure*}

Figure \ref{fig:qual_eth3d} collects three outdoor images from ETH3D (respectively, \textit{Playground1}, \textit{Playground2} and \textit{Playground3}). Once again, \method proves its supremacy at predicting fine details such as branches and poles, while resulting more robust to challenging illumination conditions such as the sun flare in \textit{Playground2}.

\clearpage

\begin{figure*}[h]
    \centering
    \renewcommand{\tabcolsep}{1pt}
    \begin{tabular}{ccc}
        
        \small RGB &
        \small RAFT-Stereo \cite{lipson2021raft} &
        \small DLNR \cite{zhao2023high} \\
        \includegraphics[width=0.32\textwidth]{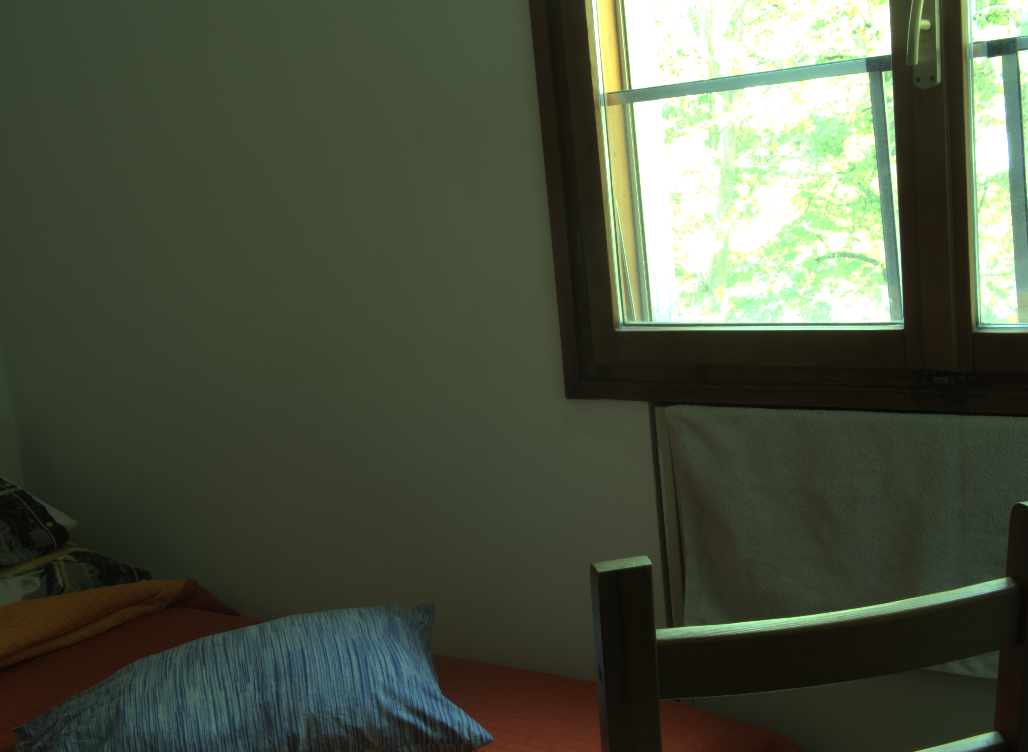} & 
        \includegraphics[width=0.32\textwidth]{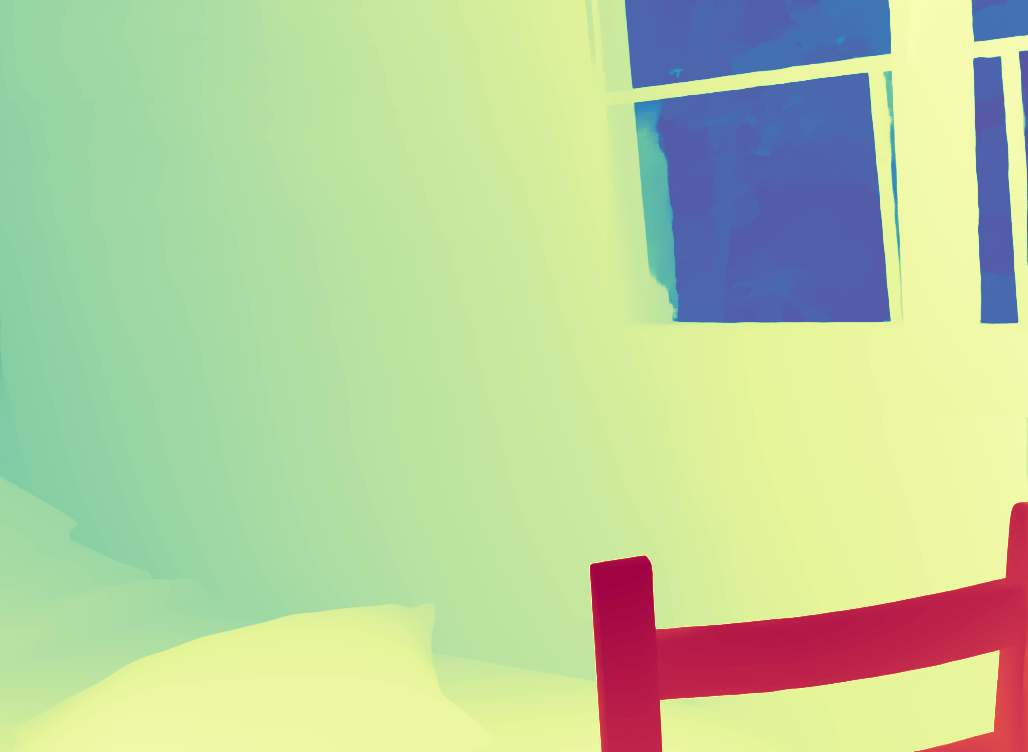} &
        \includegraphics[width=0.32\textwidth]{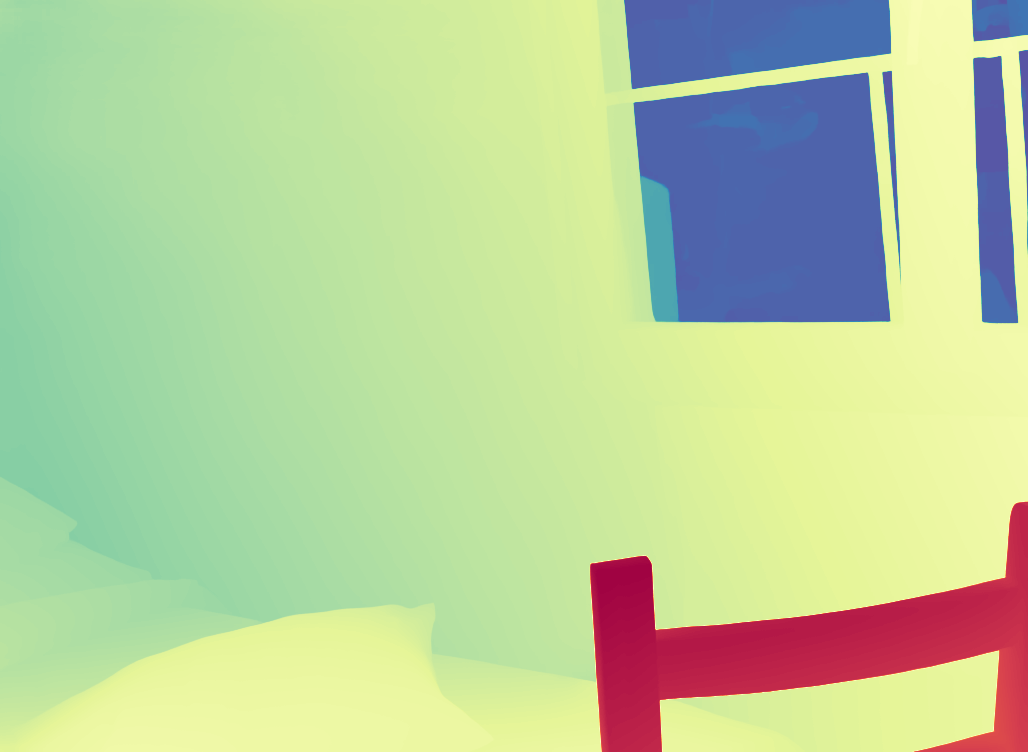} \\
        \small NMRF \cite{guan2024neural} &
        \small Selective-IGEV \cite{wang2024selective} &
        \textbf{\method (ours)} \\
        \includegraphics[width=0.32\textwidth]{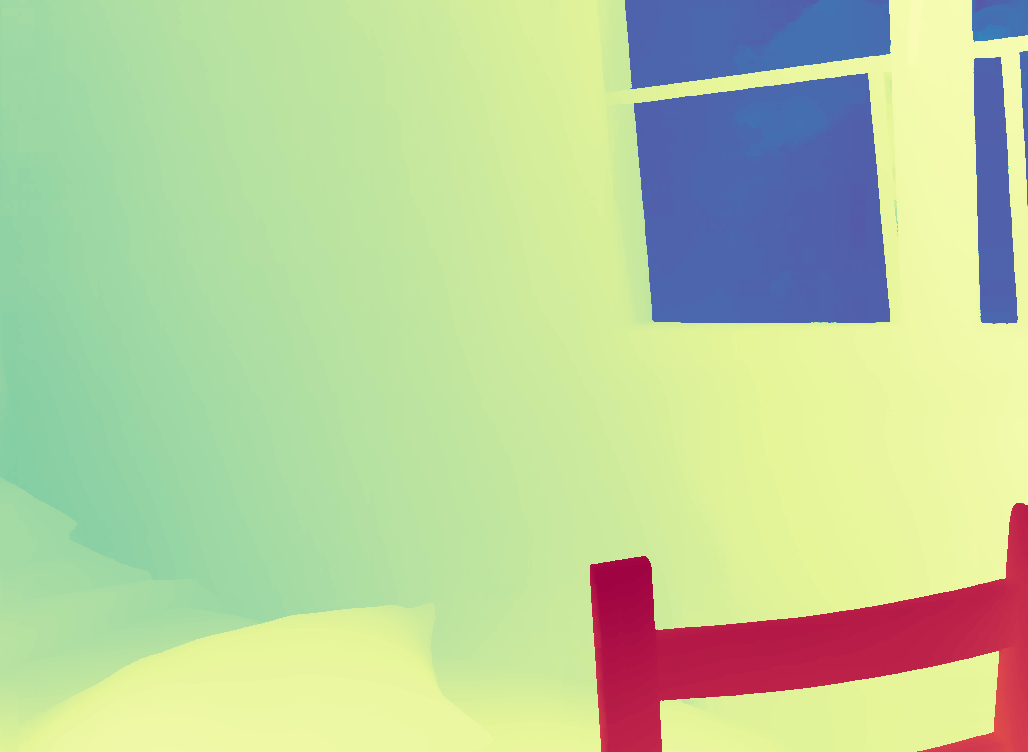} &
        \includegraphics[width=0.32\textwidth]{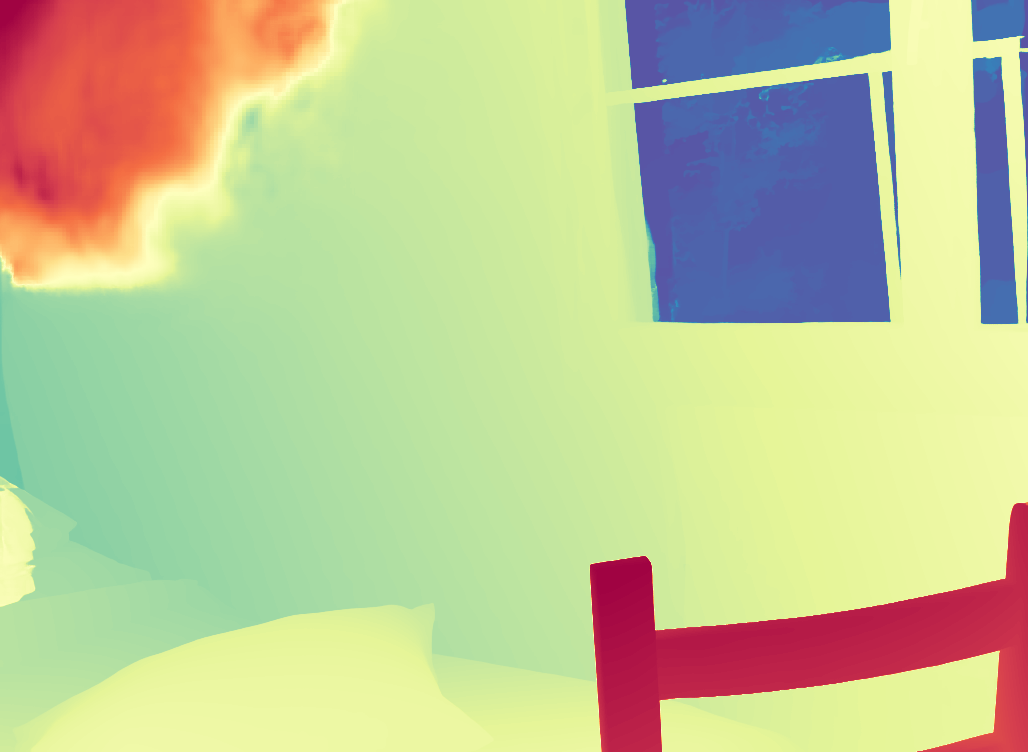} &
        \includegraphics[width=0.32\textwidth]{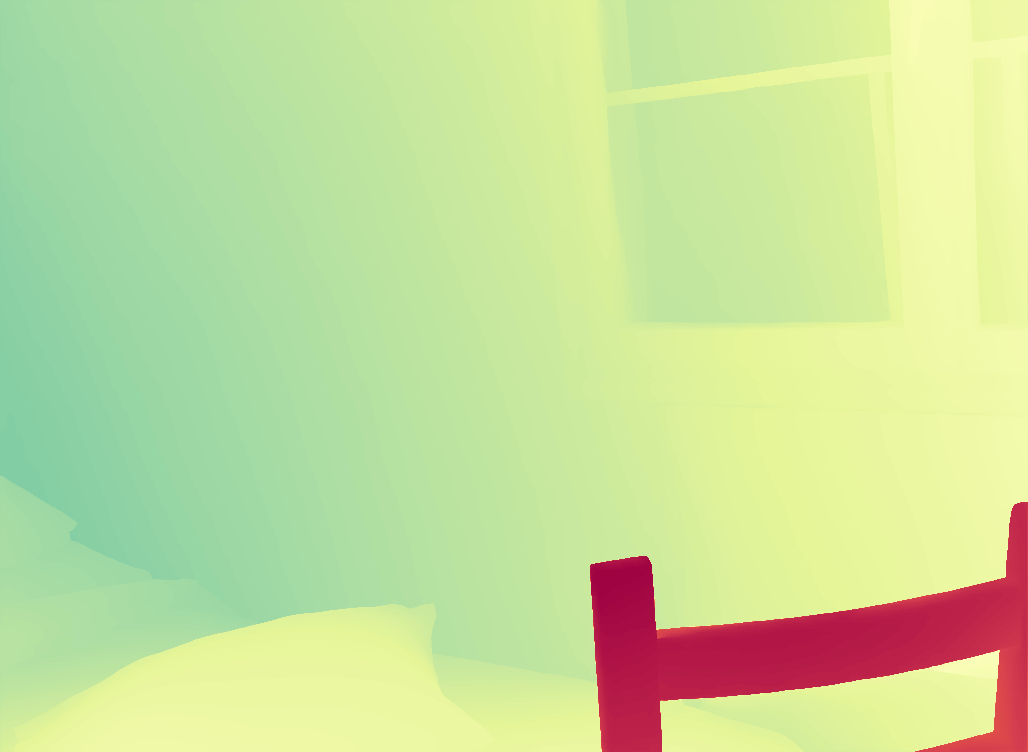} \\ \\
        
        \small RGB &
        \small RAFT-Stereo \cite{lipson2021raft} &
        \small DLNR \cite{zhao2023high} \\
        \includegraphics[width=0.32\textwidth]{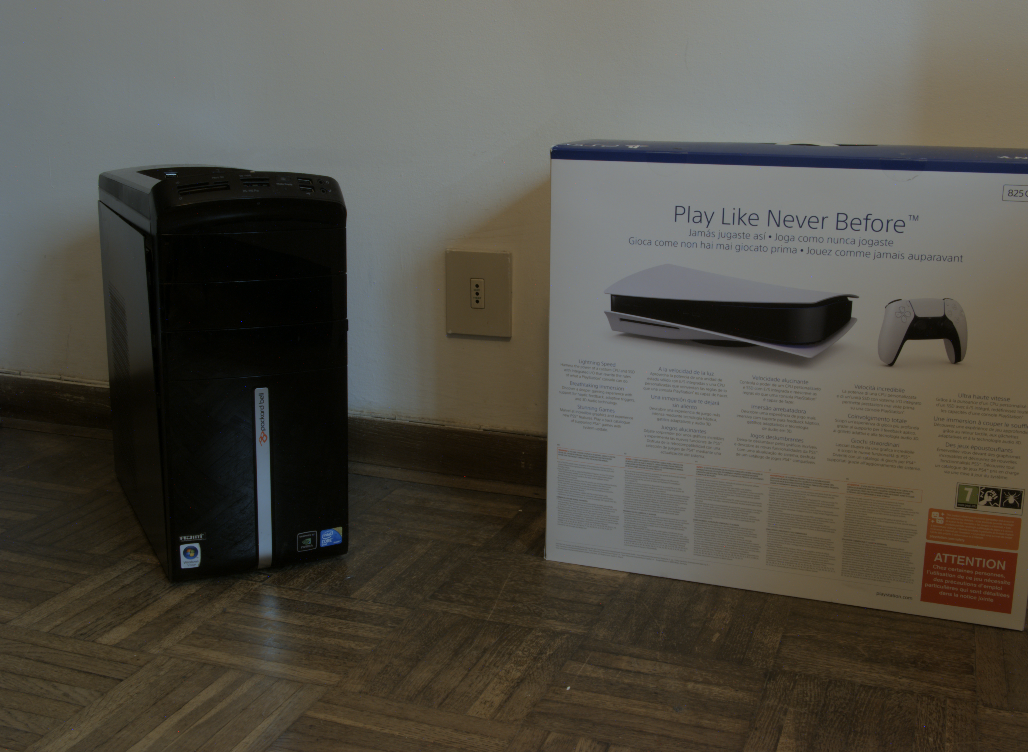} & 
        \includegraphics[width=0.32\textwidth]{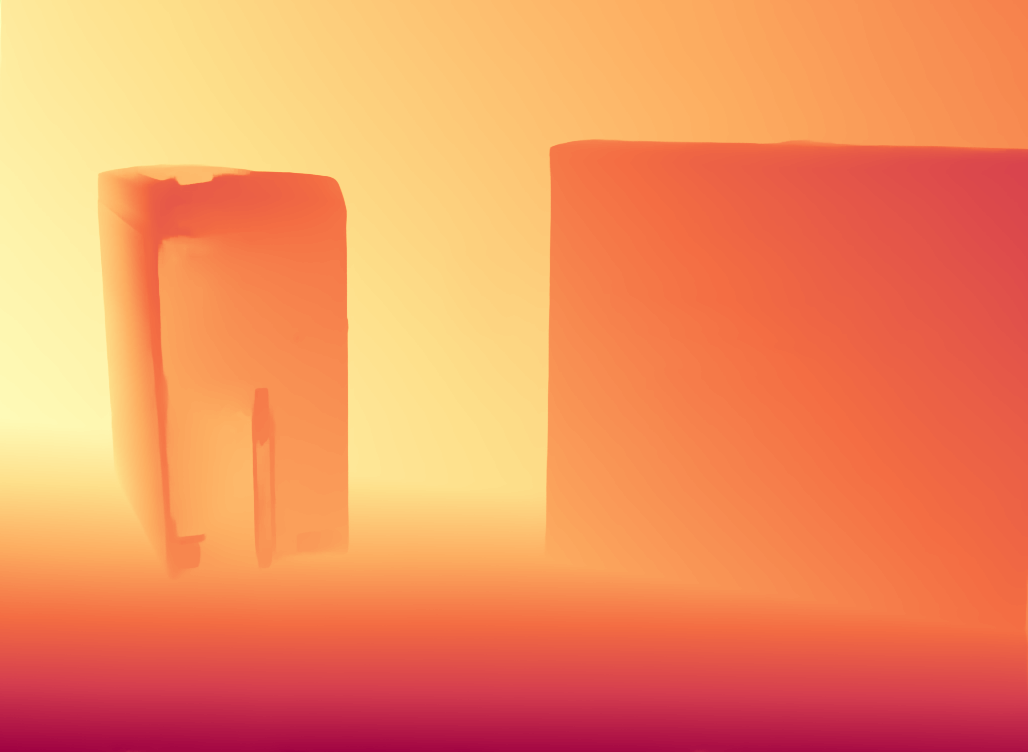} &
        \includegraphics[width=0.32\textwidth]{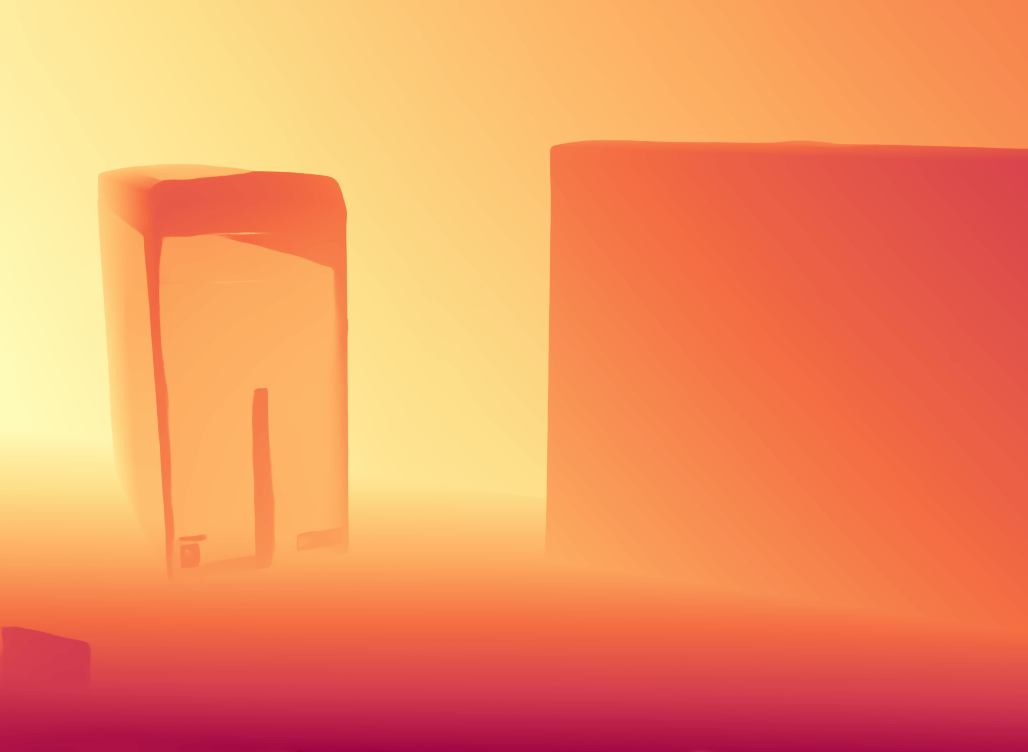} \\
        \small NMRF \cite{guan2024neural} &
        \small Selective-IGEV \cite{wang2024selective} &
        \textbf{\method (ours)} \\
        \includegraphics[width=0.32\textwidth]{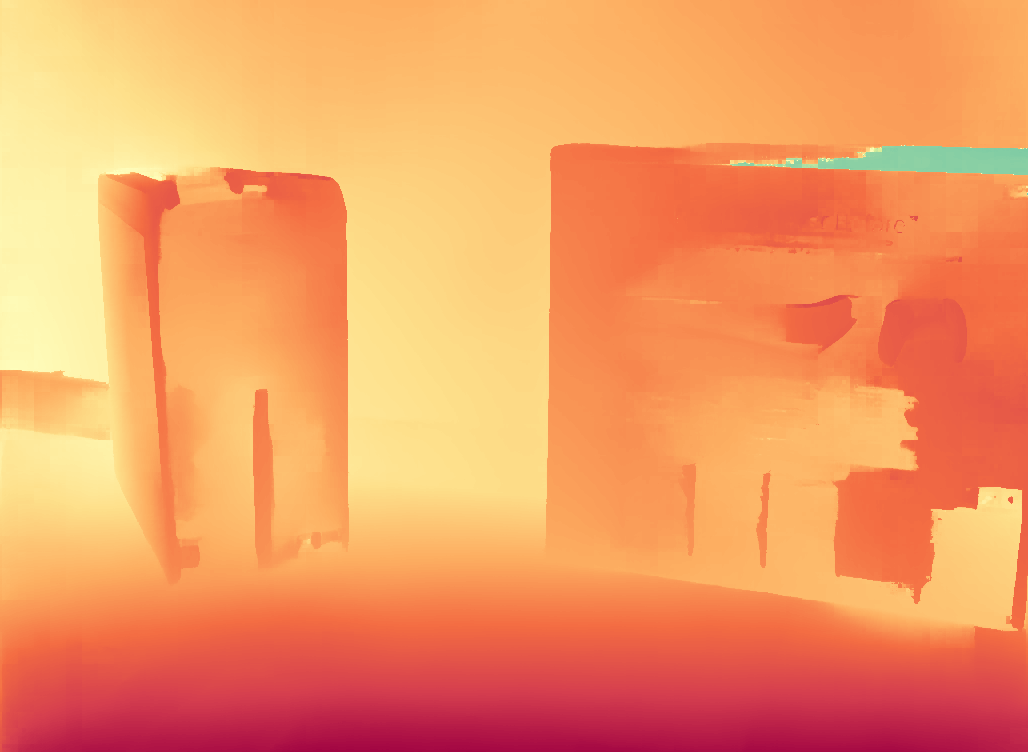} &
        \includegraphics[width=0.32\textwidth]{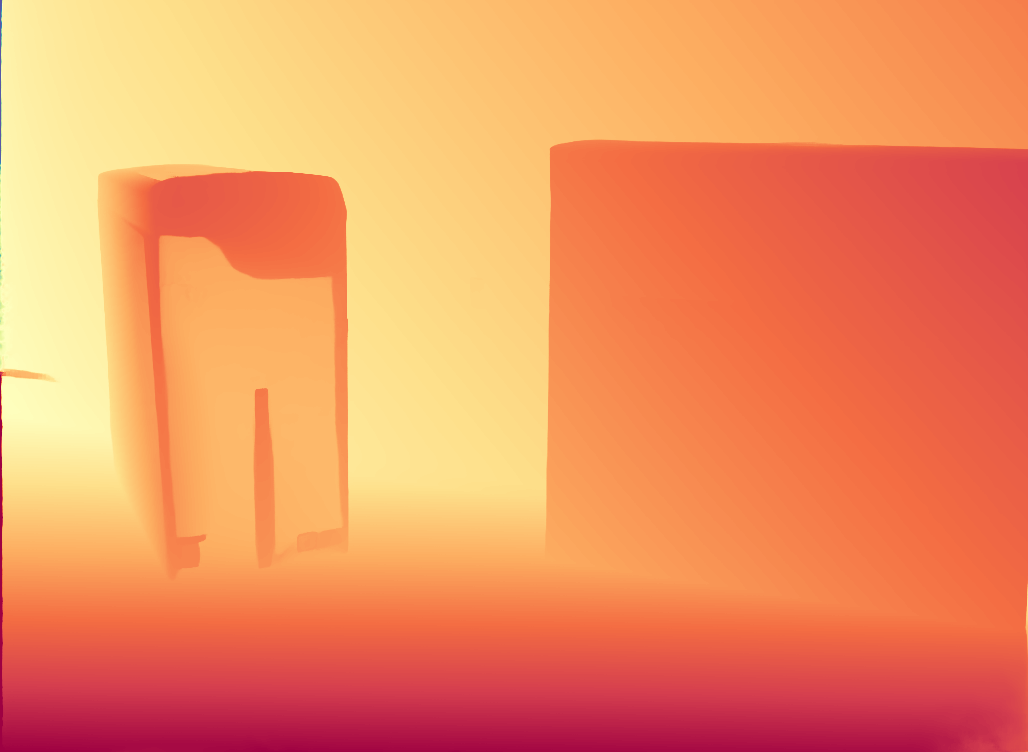} &
        \includegraphics[width=0.32\textwidth]{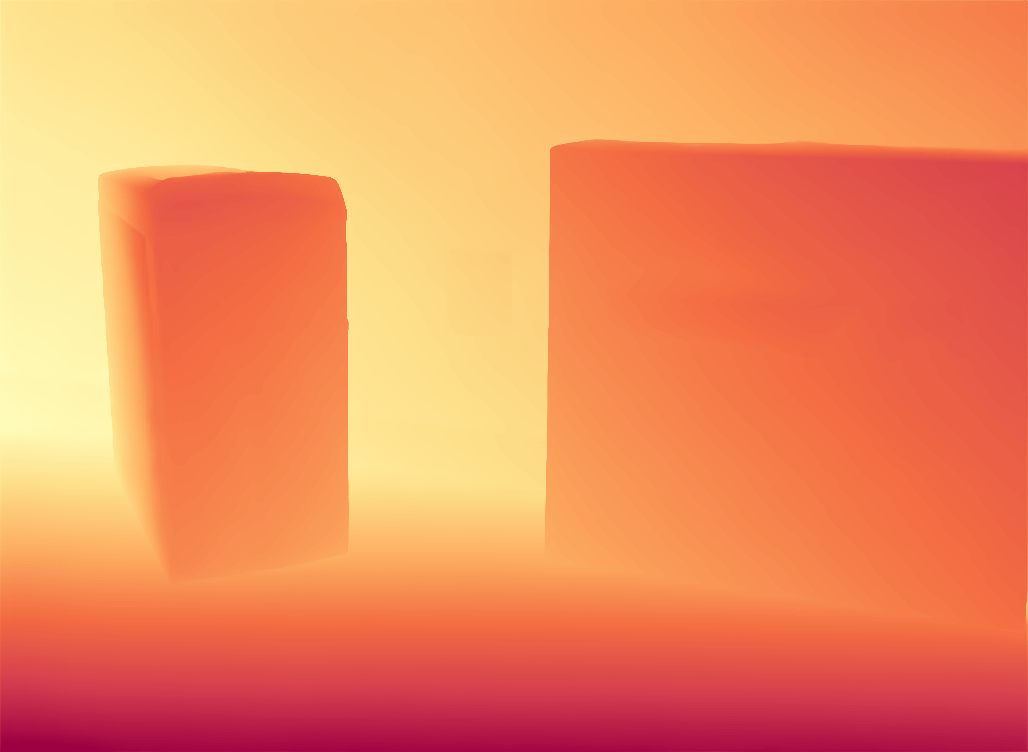} \\ 

    \end{tabular}\vspace{-0.3cm}
    \caption{\textbf{Qualitative Results -- Booster (part 1).} Predictions by state-of-the-art models and \method.}
    \label{fig:qual_booster_1}\vspace{-0.3cm}
\end{figure*}
\begin{figure*}[t]
    \centering
    \renewcommand{\tabcolsep}{1pt}
    \begin{tabular}{ccc}
        
        \small RGB &
        \small RAFT-Stereo \cite{lipson2021raft} &
        \small DLNR \cite{zhao2023high} \\
        \includegraphics[width=0.32\textwidth]{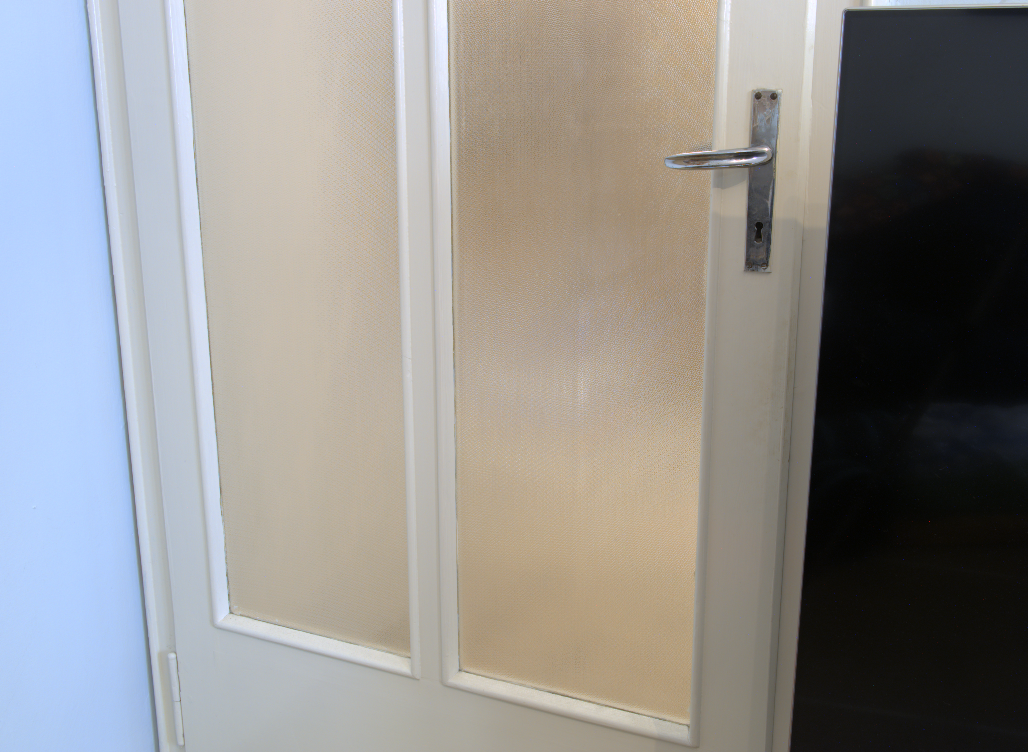} & 
        \includegraphics[width=0.32\textwidth]{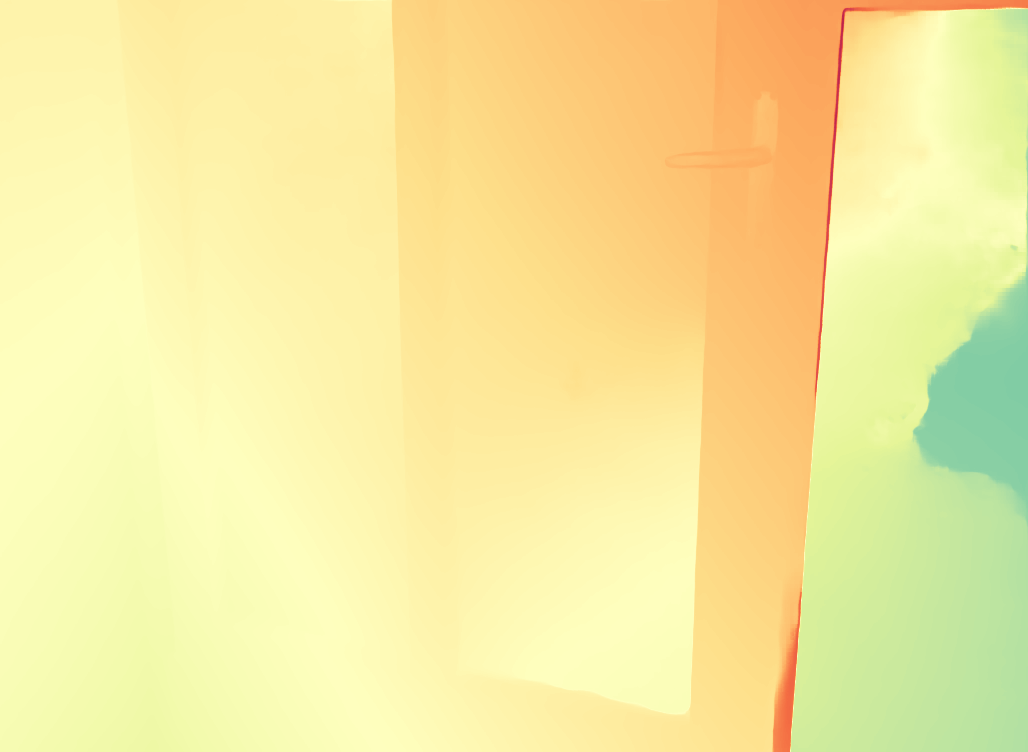} &
        \includegraphics[width=0.32\textwidth]{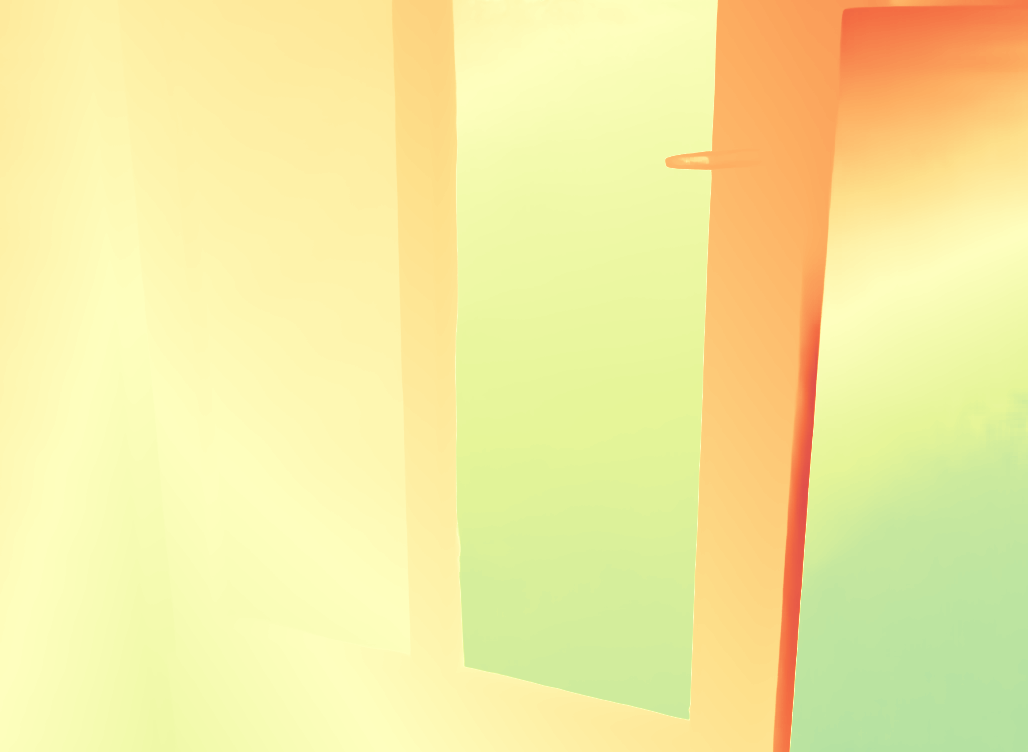} \\
        \small NMRF \cite{guan2024neural} &
        \small Selective-IGEV \cite{wang2024selective} &
        \textbf{\method (ours)} \\
        \includegraphics[width=0.32\textwidth]{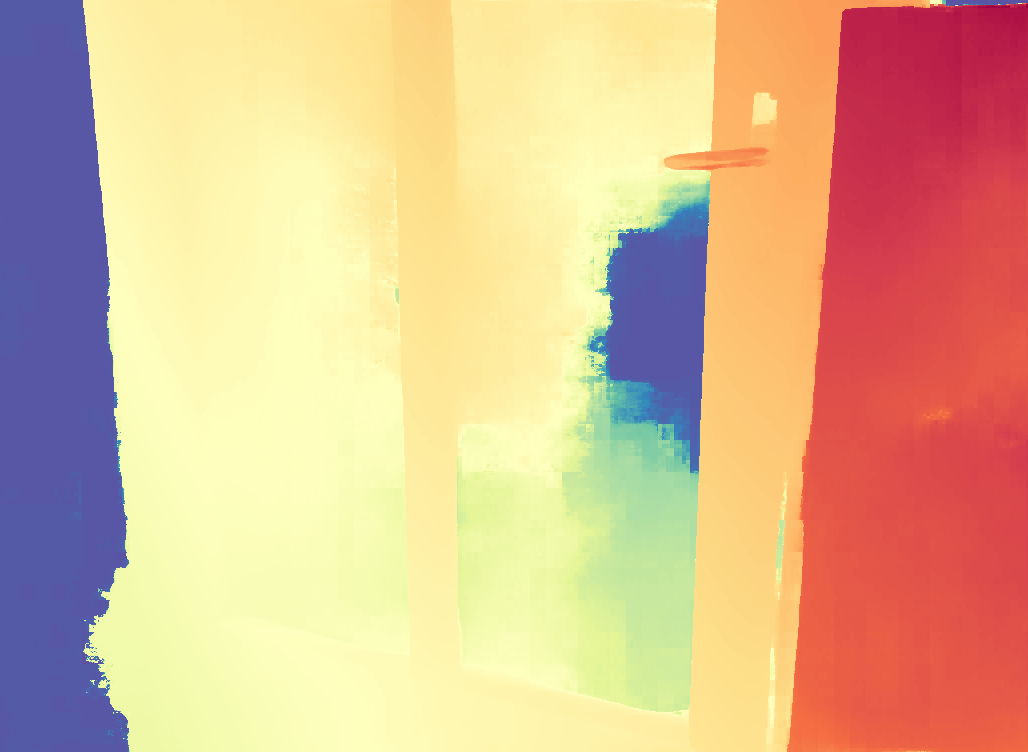} &
        \includegraphics[width=0.32\textwidth]{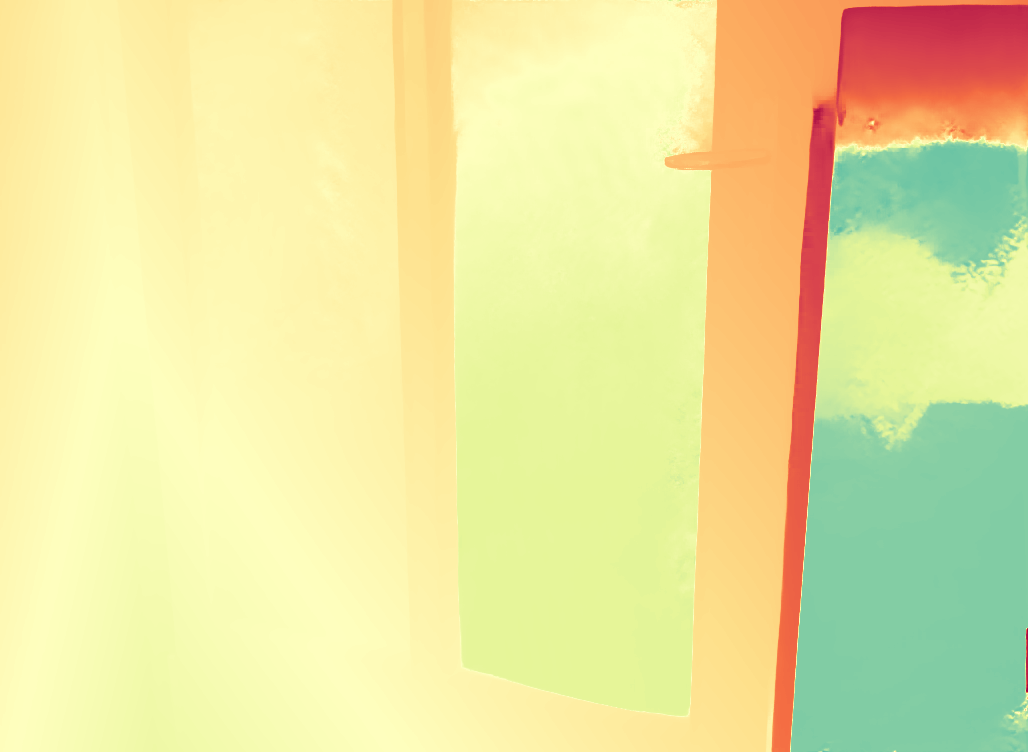} &
        \includegraphics[width=0.32\textwidth]{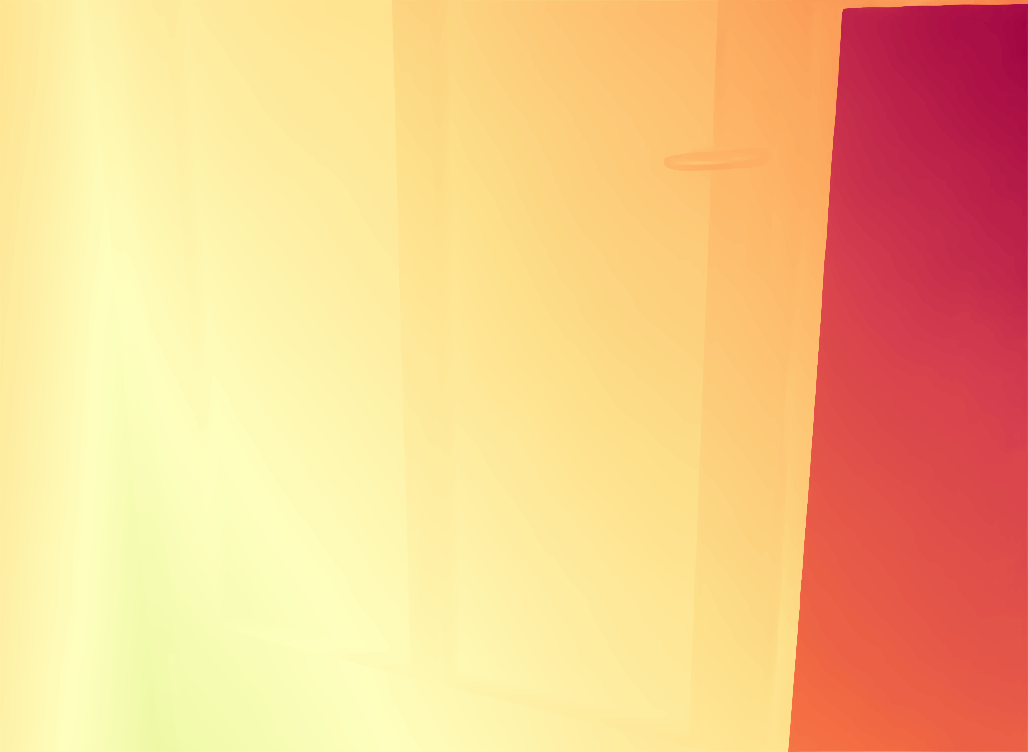} \\ \\
        
        \small RGB &
        \small RAFT-Stereo \cite{lipson2021raft} &
        \small DLNR \cite{zhao2023high} \\
        \includegraphics[width=0.32\textwidth]{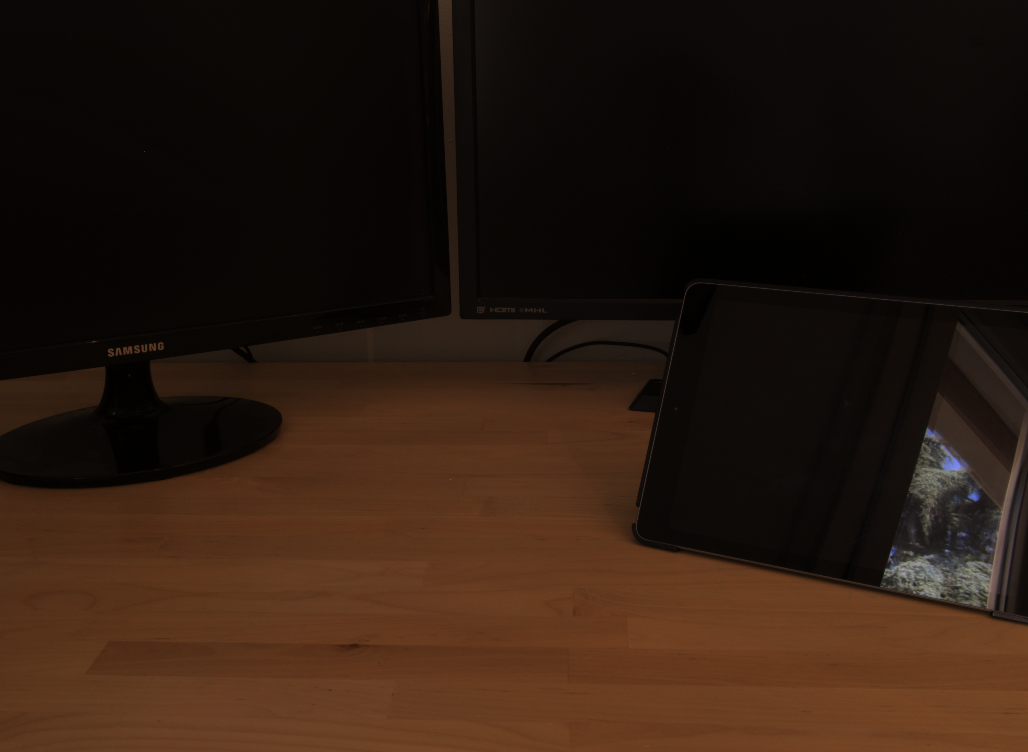} & 
        \includegraphics[width=0.32\textwidth]{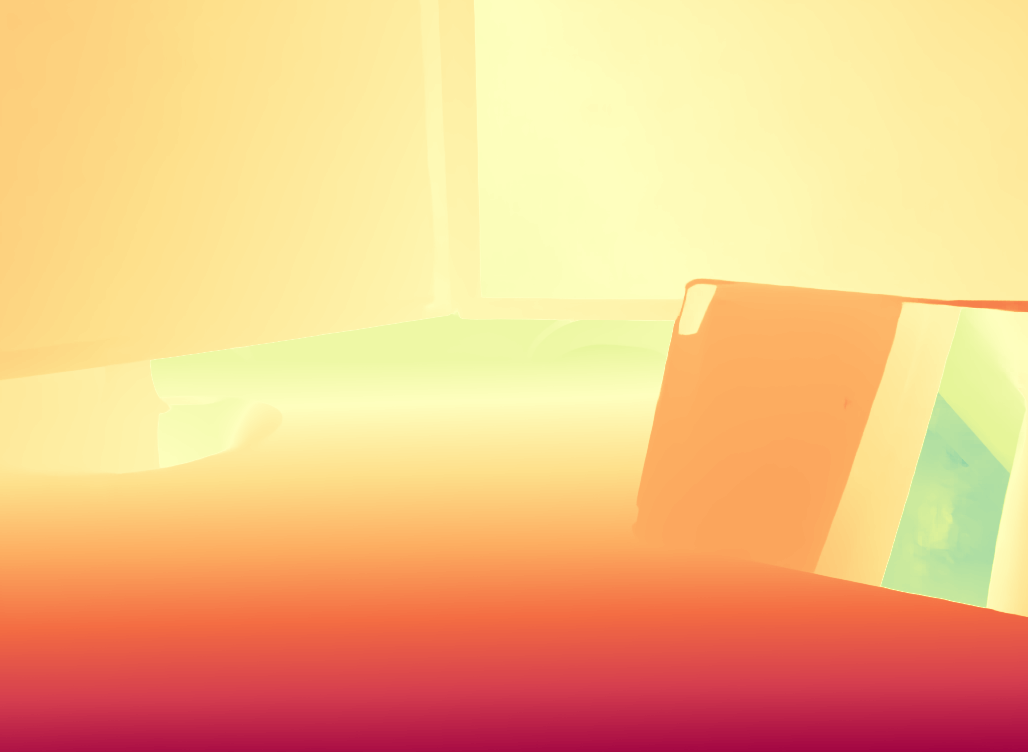} &
        \includegraphics[width=0.32\textwidth]{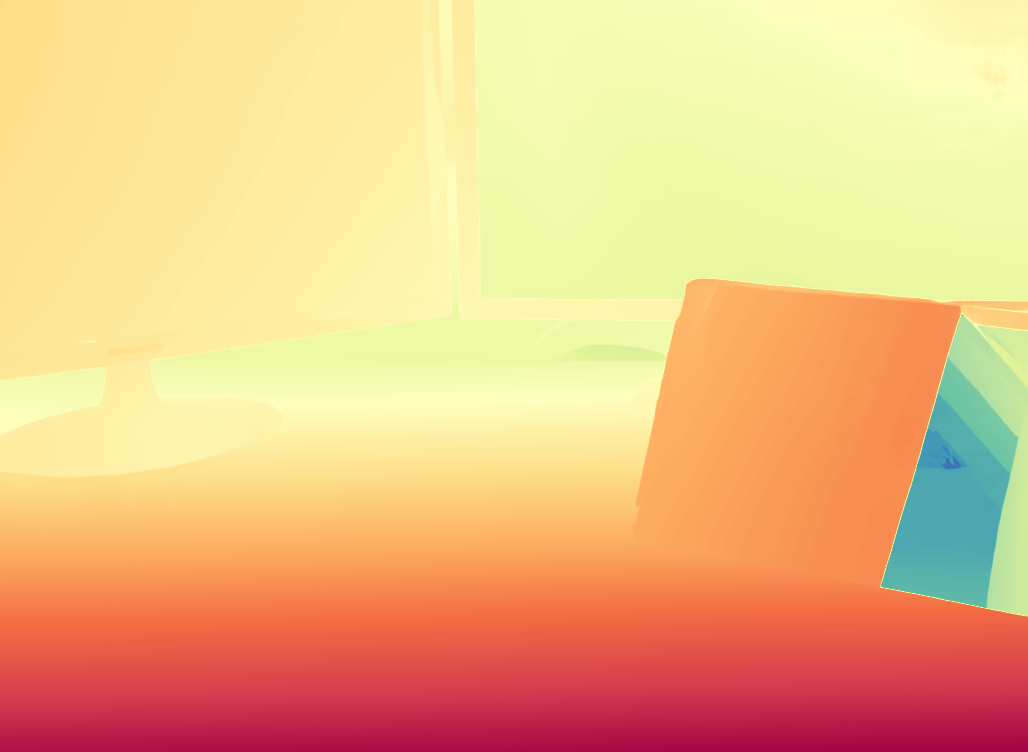} \\
        \small NMRF \cite{guan2024neural} &
        \small Selective-IGEV \cite{wang2024selective} &
        \textbf{\method (ours)} \\
        \includegraphics[width=0.32\textwidth]{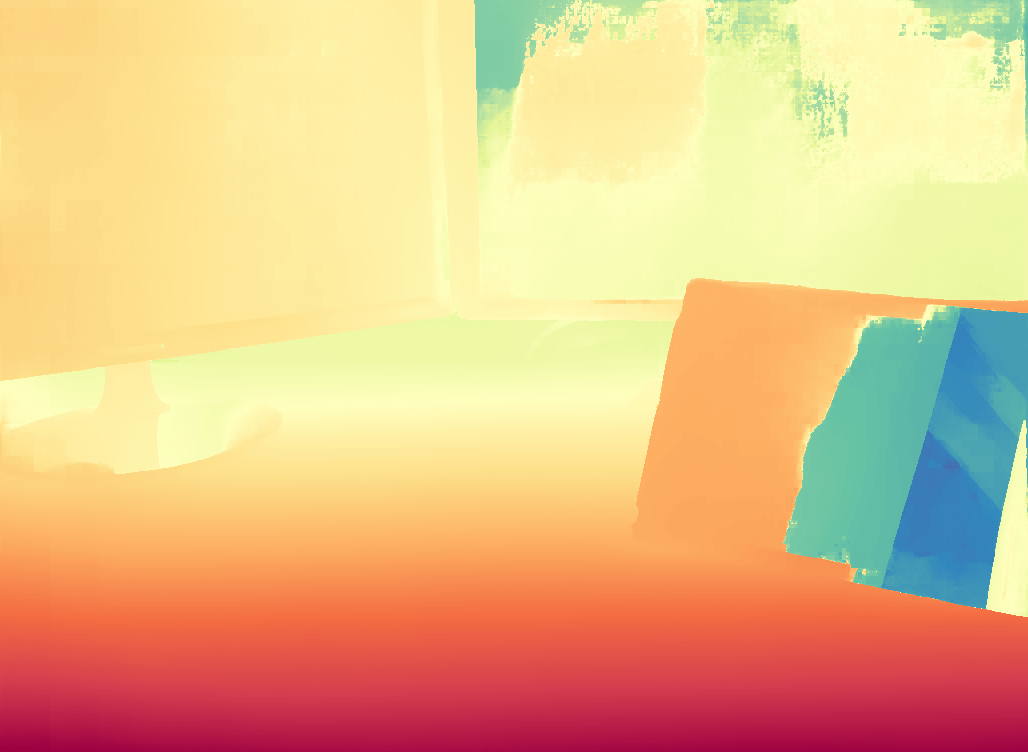} &
        \includegraphics[width=0.32\textwidth]{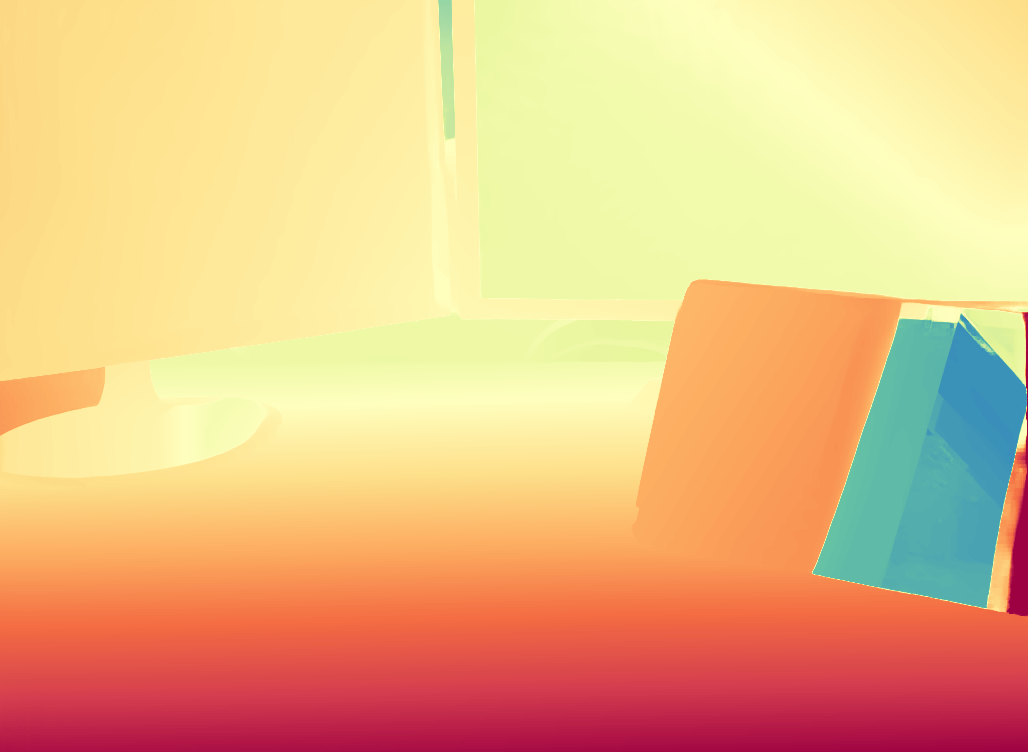} &
        \includegraphics[width=0.32\textwidth]{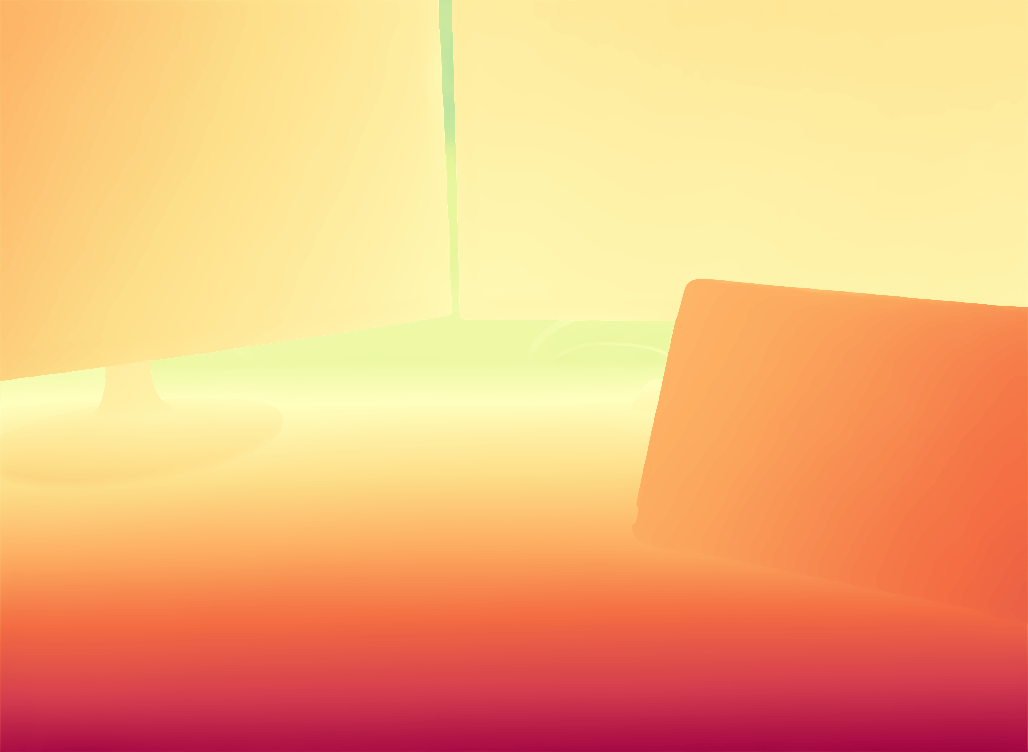} \\ 

    \end{tabular}\vspace{-0.3cm}
    \caption{\textbf{Qualitative Results -- Booster (part 2).} Predictions by state-of-the-art models and \method.}
    \label{fig:qual_booster_2}\vspace{-0.3cm}
\end{figure*}

Figure \ref{fig:qual_booster_1} and \ref{fig:qual_booster_2} report four examples from the Booster dataset, confirming how \method can exploit the strong priors provided by the VFM to properly perceive the glass surface on the window in the former image, as well as challenging, untextured black surfaces of the computer, the TV and the displays appearing in the remaining samples.

\clearpage

\begin{figure*}[h]
    \centering
    \renewcommand{\tabcolsep}{1pt}
    \begin{tabular}{ccc}
        
        \small RGB &
        \small RAFT-Stereo \cite{lipson2021raft} &
        \small DLNR \cite{zhao2023high} \\
        \includegraphics[width=0.27\textwidth]{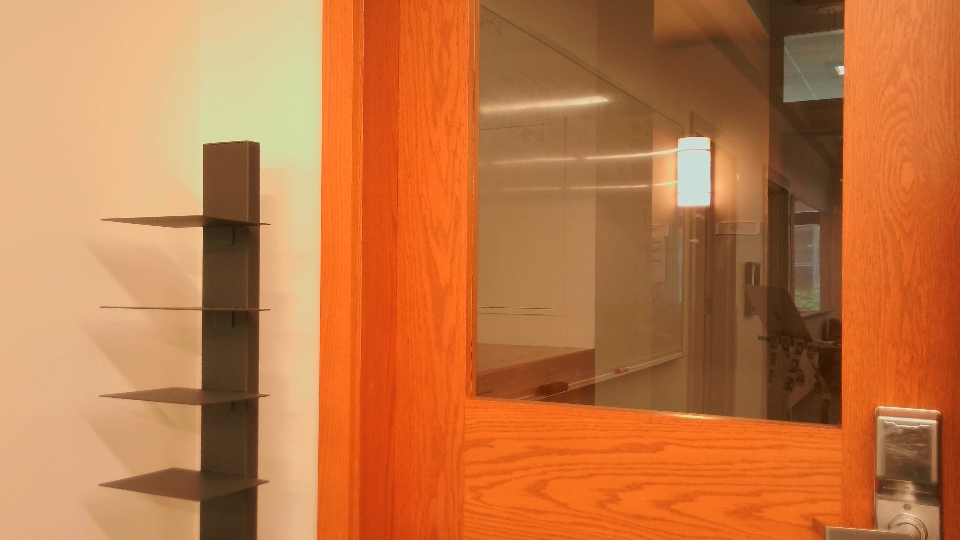} & 
        \includegraphics[width=0.27\textwidth]{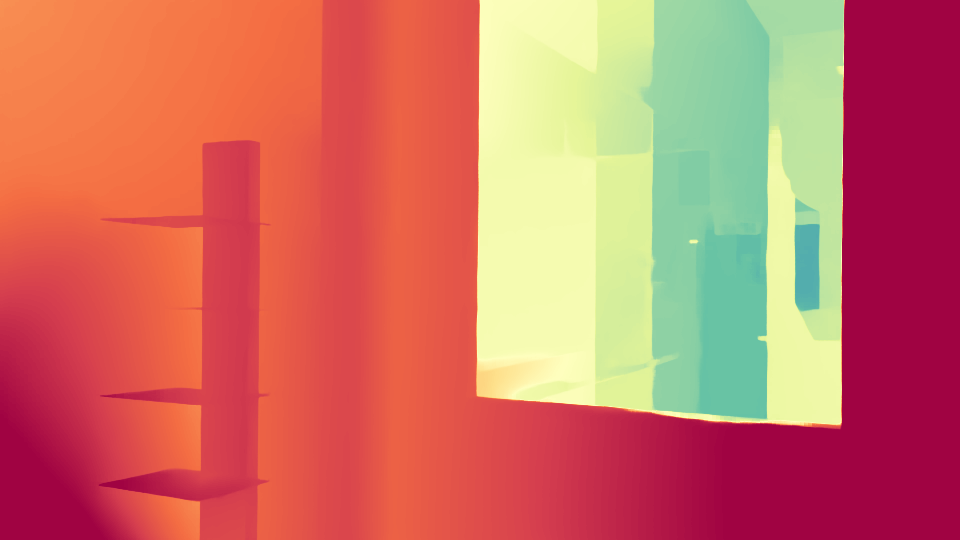} &
        \includegraphics[width=0.27\textwidth]{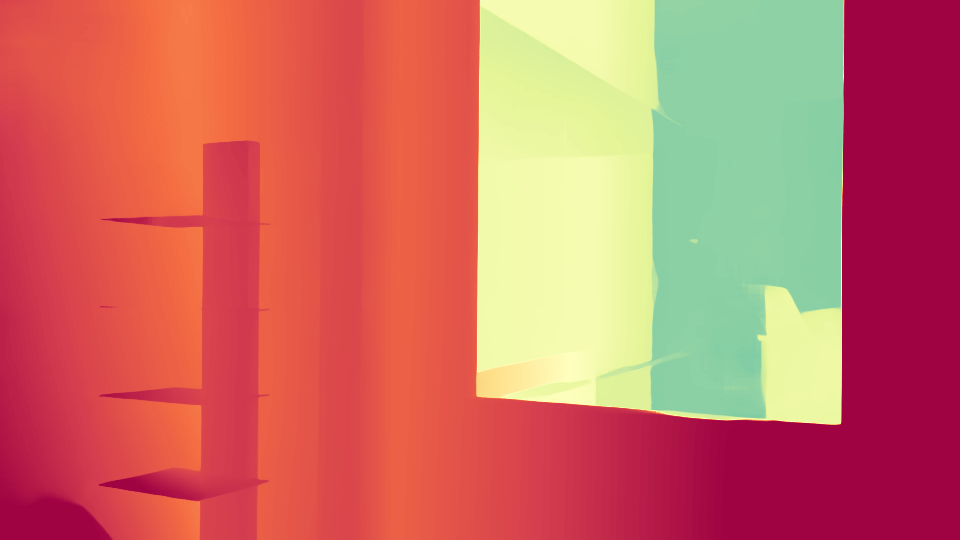} \\
        \small NMRF \cite{guan2024neural} &
        \small Selective-IGEV \cite{wang2024selective} &
        \textbf{\method (ours)} \\
        \includegraphics[width=0.27\textwidth]{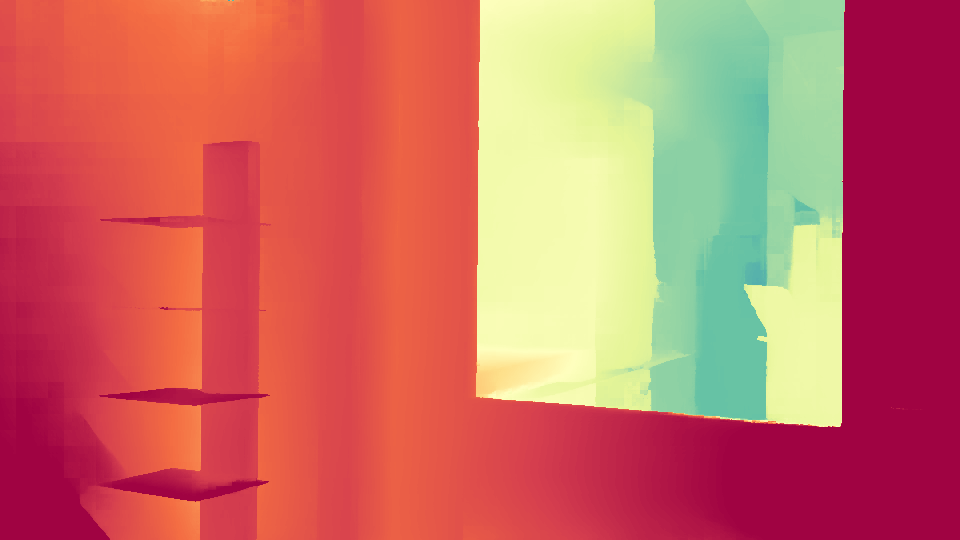} &
        \includegraphics[width=0.27\textwidth]{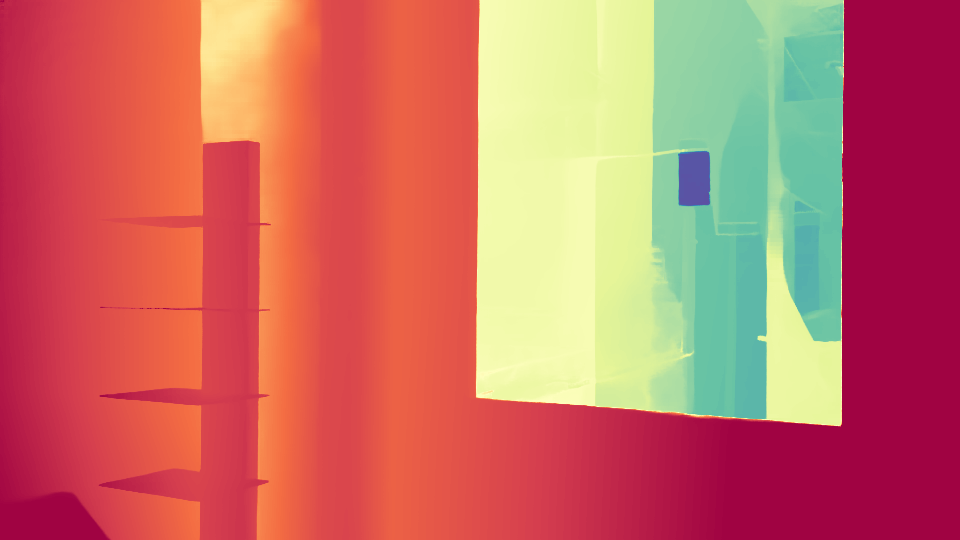} &
        \includegraphics[width=0.27\textwidth]{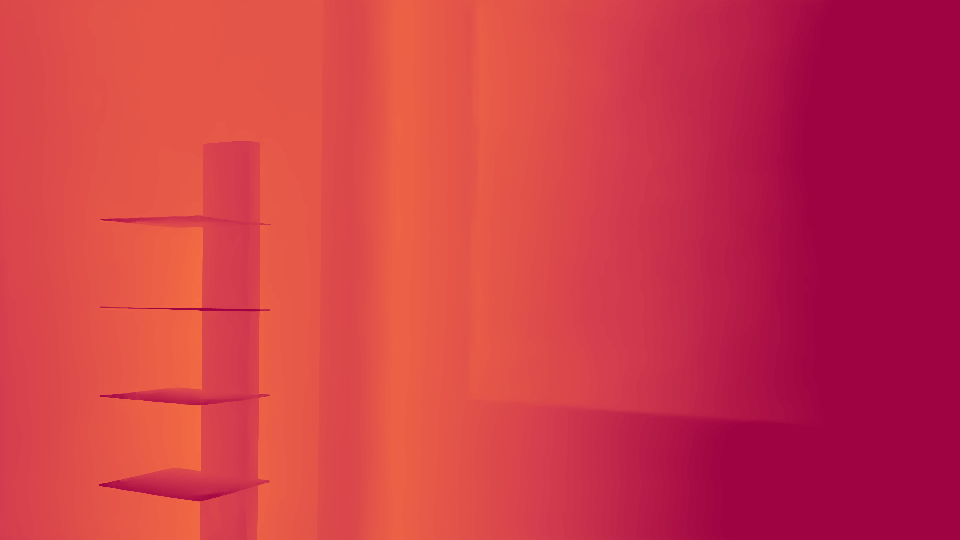} \\ \\
        
        \small RGB &
        \small RAFT-Stereo \cite{lipson2021raft} &
        \small DLNR \cite{zhao2023high} \\
        \includegraphics[width=0.27\textwidth]{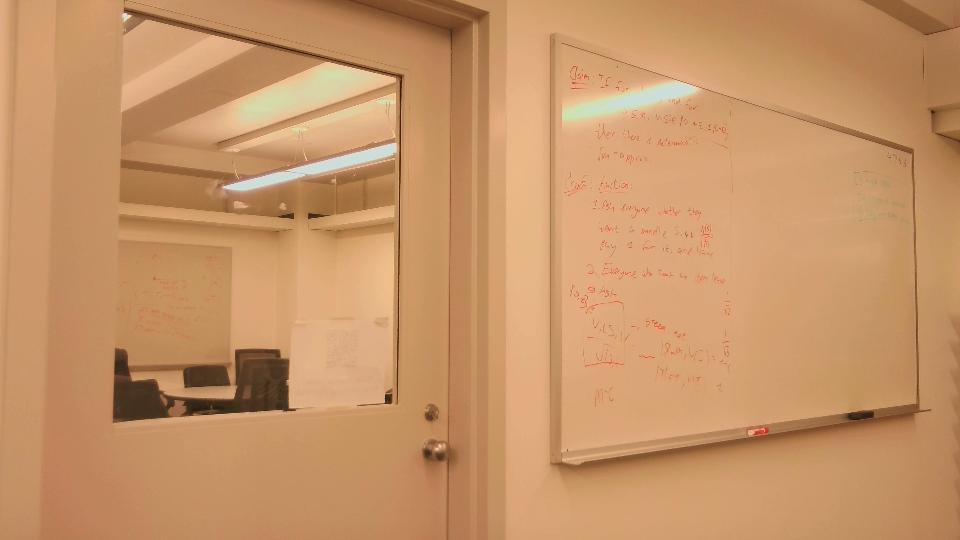} & 
        \includegraphics[width=0.27\textwidth]{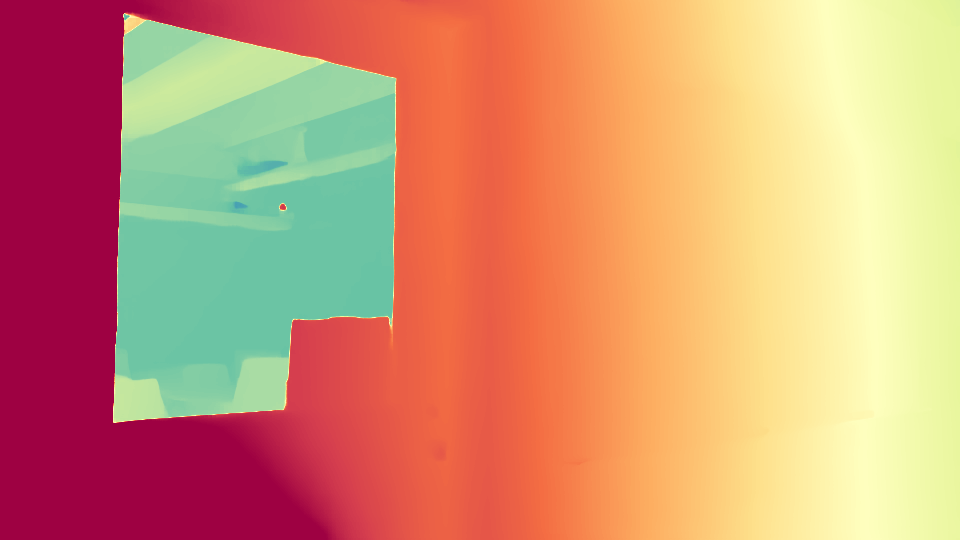} &
        \includegraphics[width=0.27\textwidth]{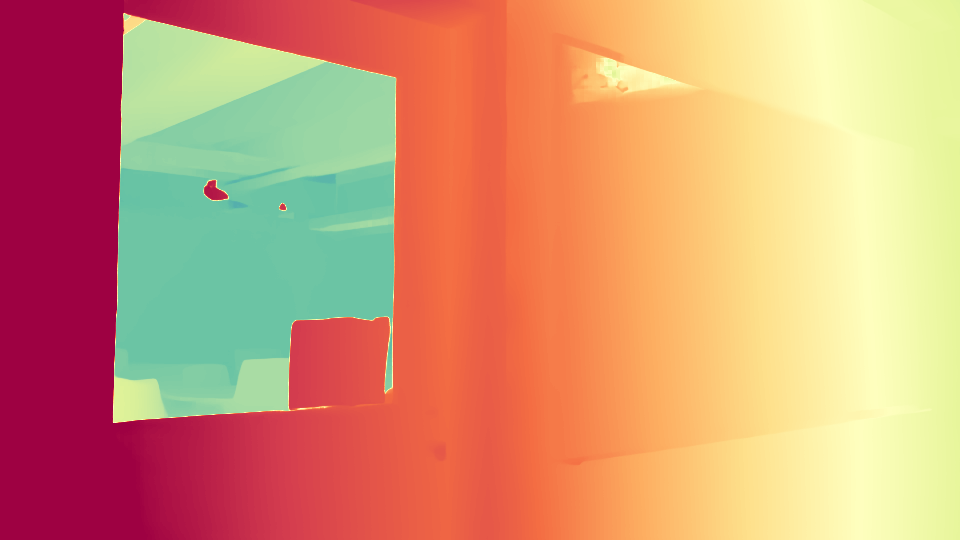} \\
        \small NMRF \cite{guan2024neural} &
        \small Selective-IGEV \cite{wang2024selective} &
        \textbf{\method (ours)} \\
        \includegraphics[width=0.27\textwidth]{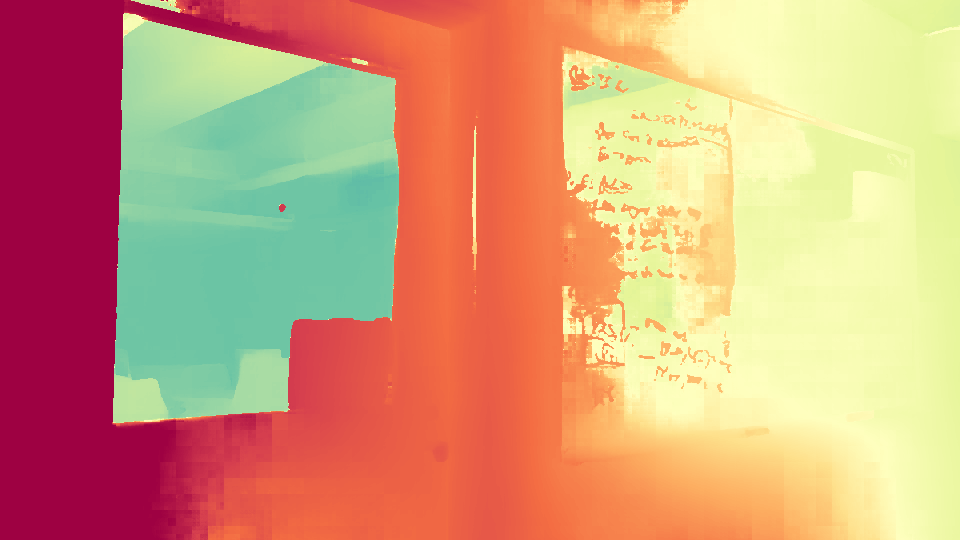} &
        \includegraphics[width=0.27\textwidth]{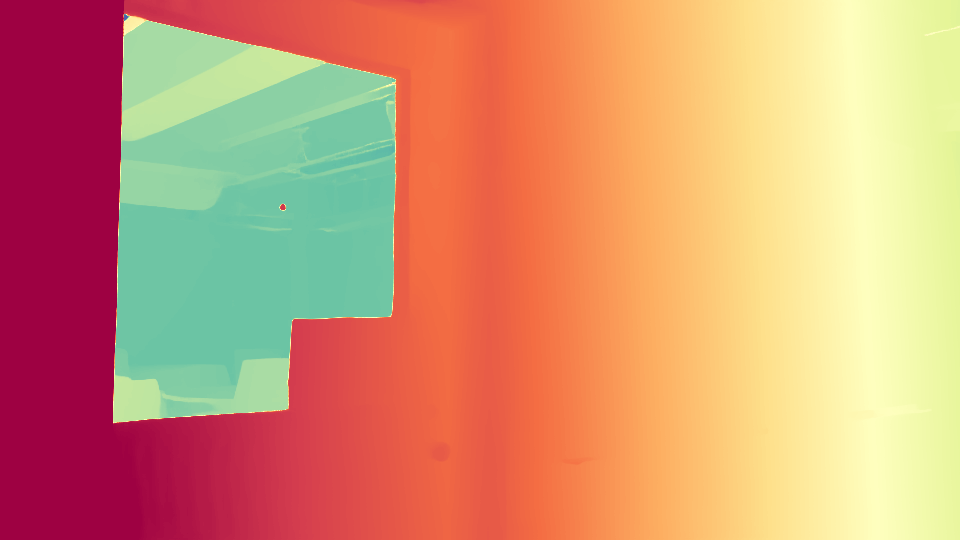} &
        \includegraphics[width=0.27\textwidth]{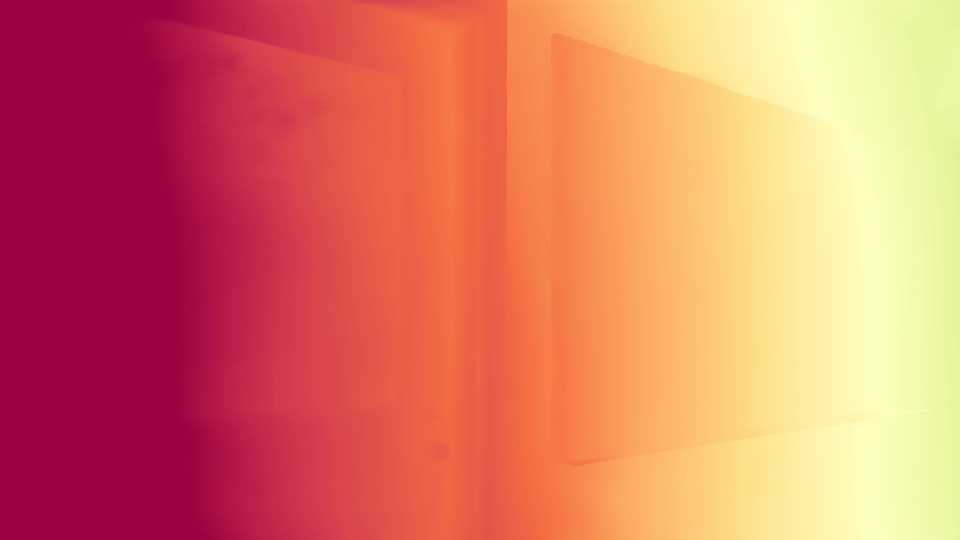} \\ \\

        \small RGB &
        \small RAFT-Stereo \cite{lipson2021raft} &
        \small DLNR \cite{zhao2023high} \\
        \includegraphics[width=0.27\textwidth]{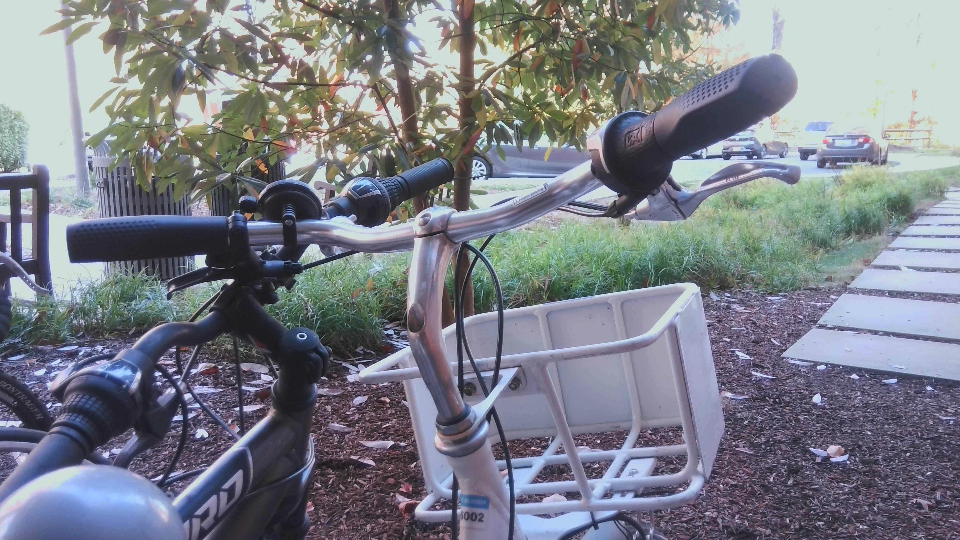} & 
        \includegraphics[width=0.27\textwidth]{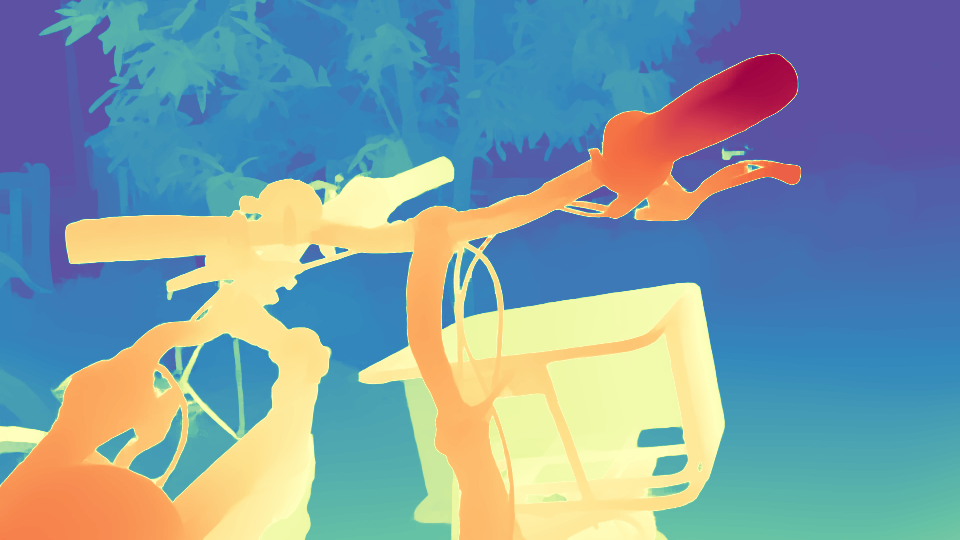} &
        \includegraphics[width=0.27\textwidth]{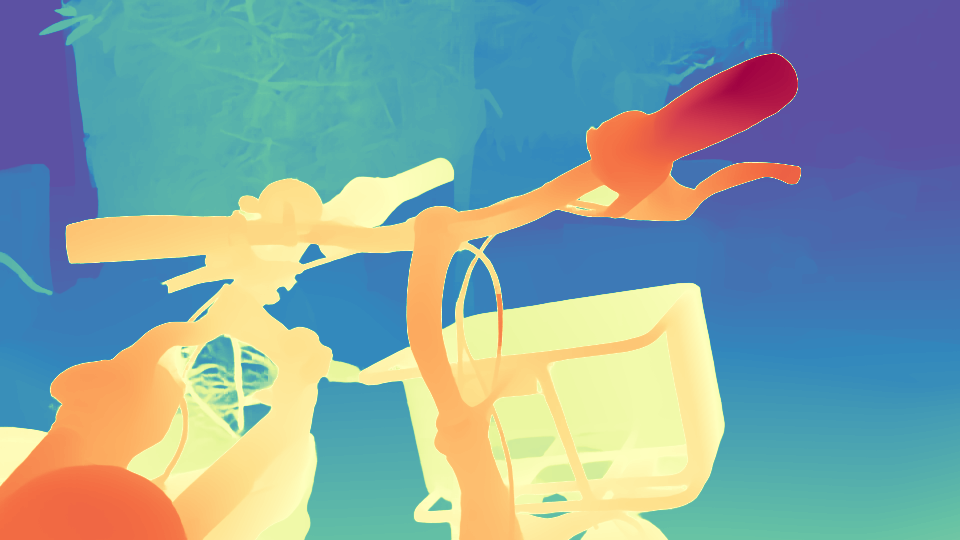} \\
        \small NMRF \cite{guan2024neural} &
        \small Selective-IGEV \cite{wang2024selective} &
        \textbf{\method (ours)} \\
        \includegraphics[width=0.27\textwidth]{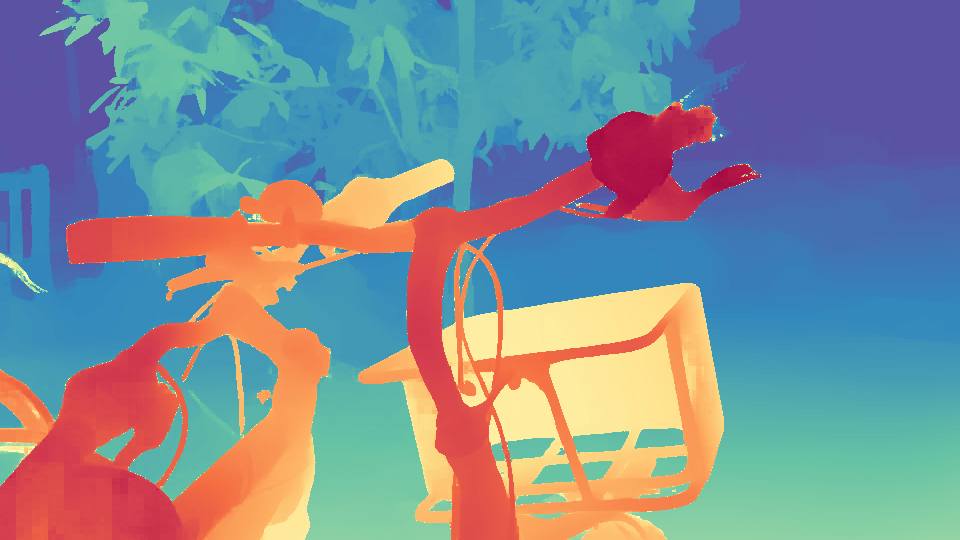} &
        \includegraphics[width=0.27\textwidth]{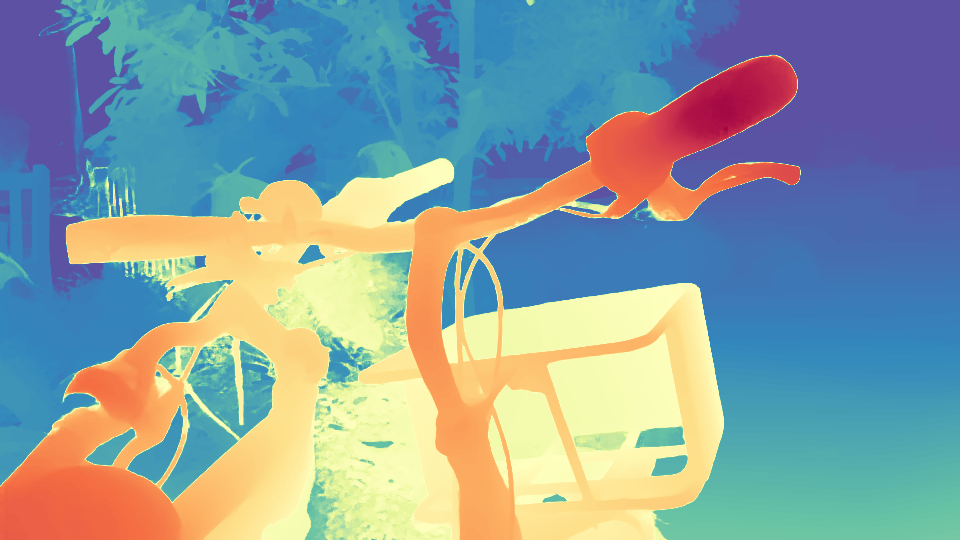} &
        \includegraphics[width=0.27\textwidth]{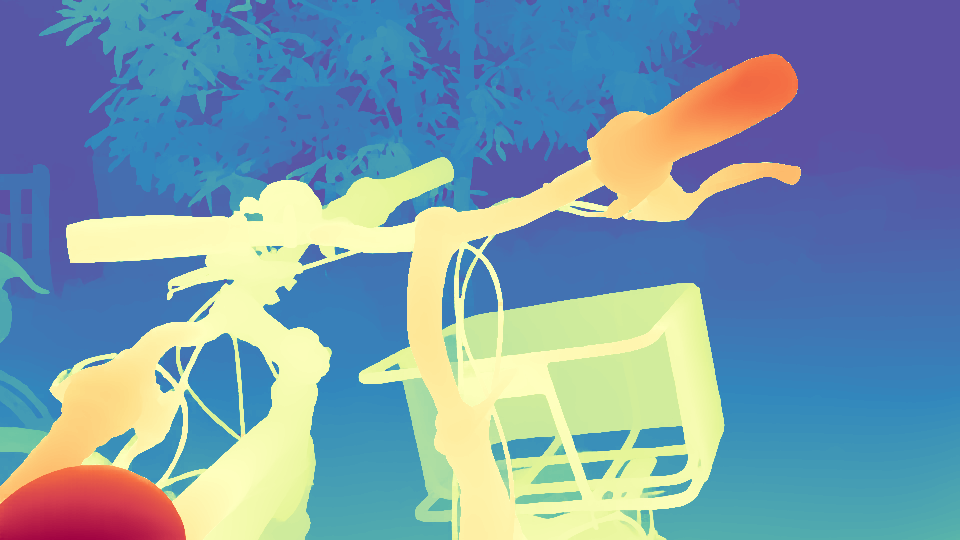} \\

    \end{tabular}\vspace{-0.3cm}
    \caption{\textbf{Qualitative Results -- LayeredFlow.} Predictions by state-of-the-art models and \method.}
    \label{fig:qual_layered}\vspace{-0.3cm}
\end{figure*}

Figure \ref{fig:qual_layered} showcases three images from the LayeredFlow dataset, highlighting once again the inability of the state-of-the-art networks to model even small, transparent surfaces as those in the doors from the first and second samples, conversely to \method which can properly identify their presence. Finally, the third sample further highlights the high level of detail in \method predictions  once again.

\clearpage

\begin{figure*}[h]
    \centering
    \renewcommand{\tabcolsep}{1pt}
    \begin{tabular}{ccc}
        
        \small RGB &
        \small Depth Anything v2 \cite{depth_anything_v2} &
        \small DepthPro \cite{depthpro} \\
        \includegraphics[width=0.23\textwidth]{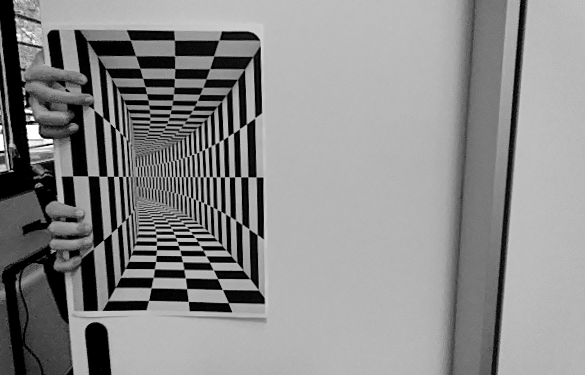} & 
        \includegraphics[width=0.23\textwidth]{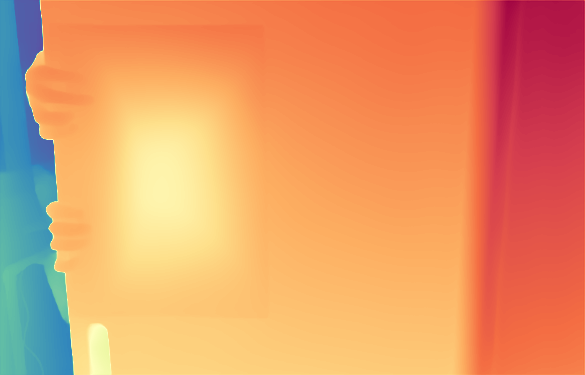}  &
        \includegraphics[width=0.23\textwidth]{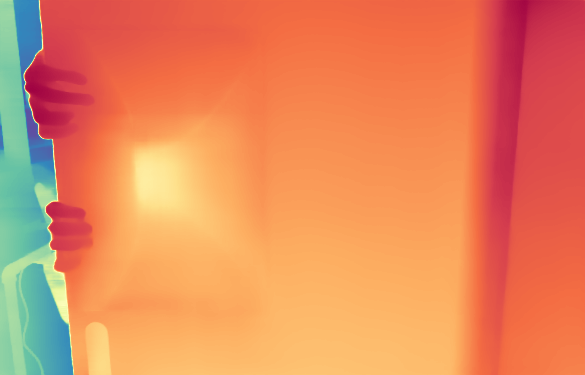} \\
        \small RAFT-Stereo \cite{lipson2021raft} &
        \small Selective-IGEV \cite{wang2024selective} &      
        \textbf{\method (ours)} \\
        \includegraphics[width=0.23\textwidth]{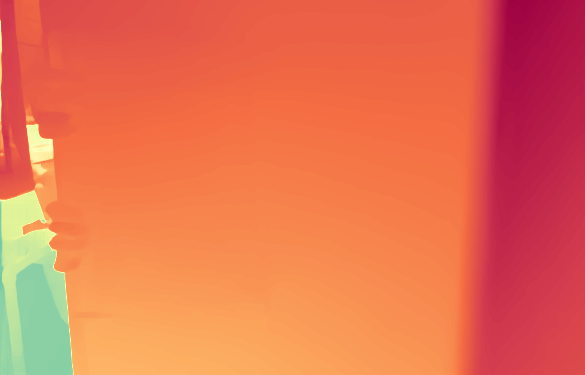} &
        \includegraphics[width=0.23\textwidth]{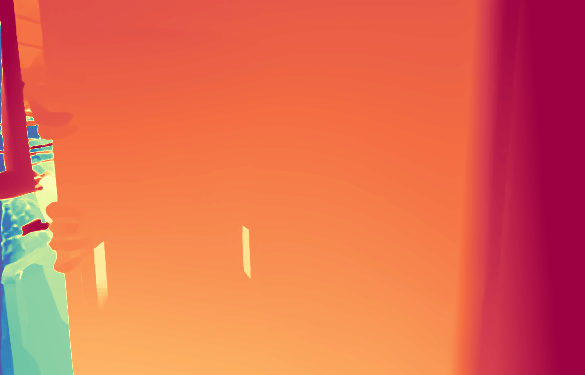} &
        \includegraphics[width=0.23\textwidth]{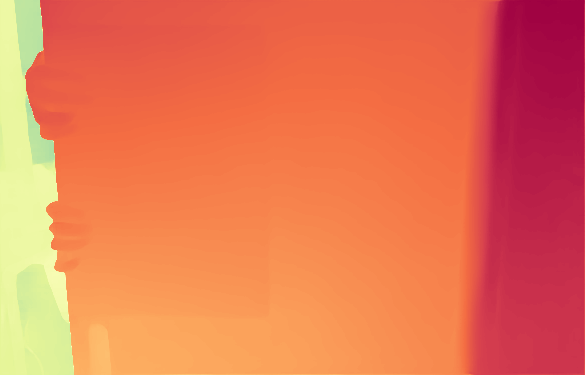} \\ \\

        \small RGB &
        \small Depth Anything v2 \cite{depth_anything_v2} &
        \small DepthPro \cite{depthpro} \\
        \includegraphics[width=0.23\textwidth]{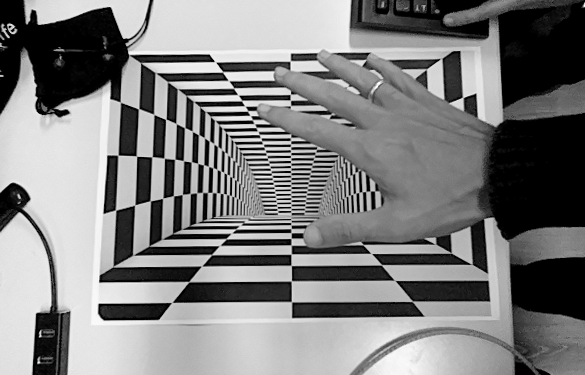} & 
        \includegraphics[width=0.23\textwidth]{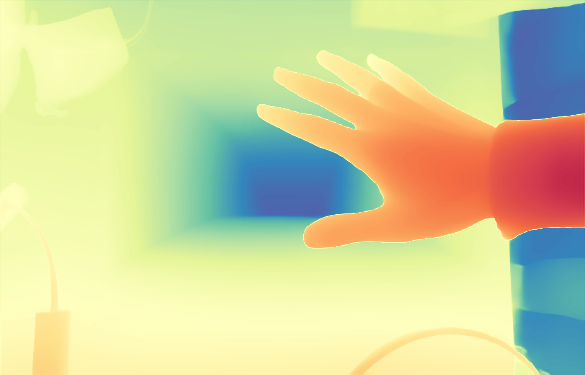}  &
        \includegraphics[width=0.23\textwidth]{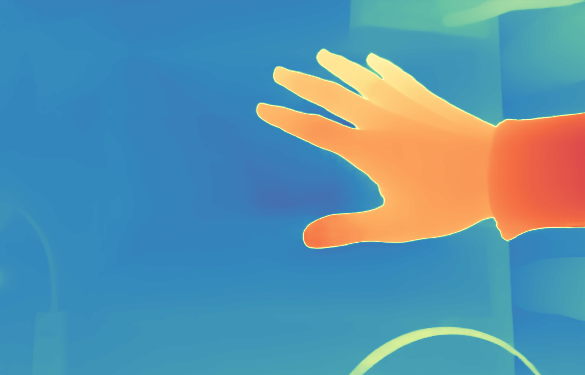} \\
        \small RAFT-Stereo \cite{lipson2021raft} &
        \small Selective-IGEV \cite{wang2024selective} &      
        \textbf{\method (ours)} \\
        \includegraphics[width=0.23\textwidth]{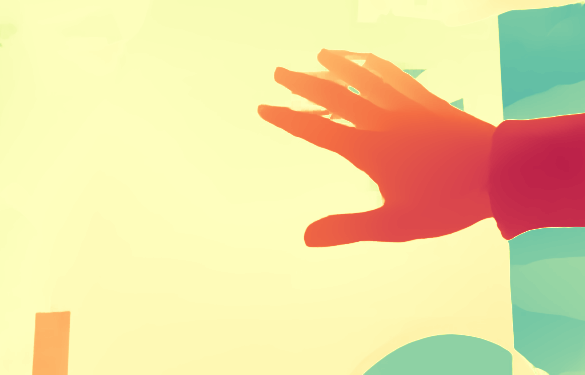} &
        \includegraphics[width=0.23\textwidth]{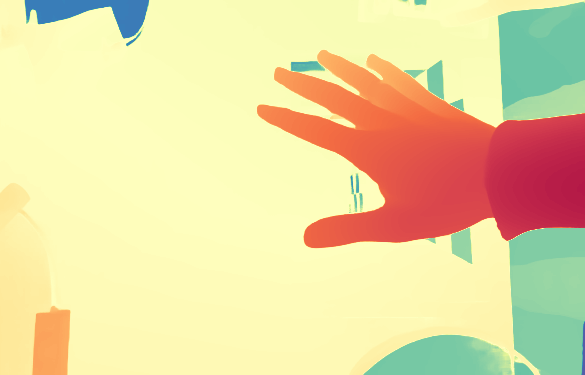} &
        \includegraphics[width=0.23\textwidth]{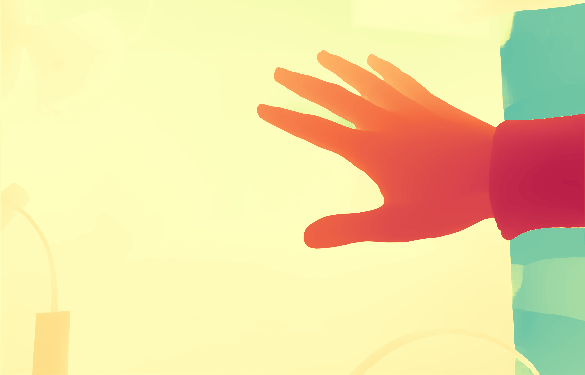} \\ \\

        \small RGB &
        \small Depth Anything v2 \cite{depth_anything_v2} &
        \small DepthPro \cite{depthpro} \\
        \includegraphics[width=0.23\textwidth]{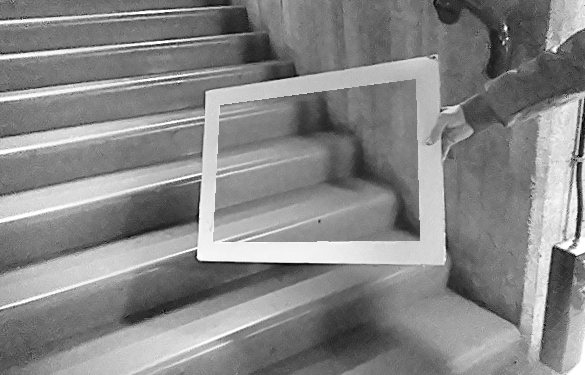} & 
        \includegraphics[width=0.23\textwidth]{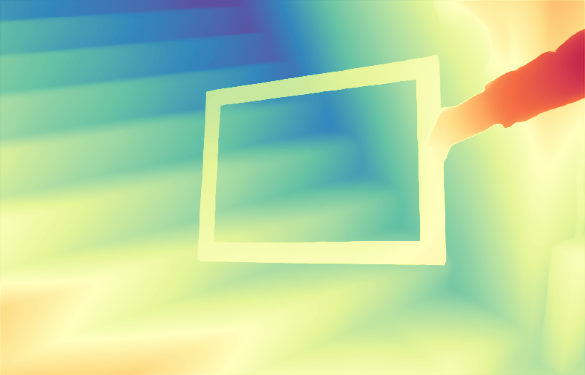}  &
        \includegraphics[width=0.23\textwidth]{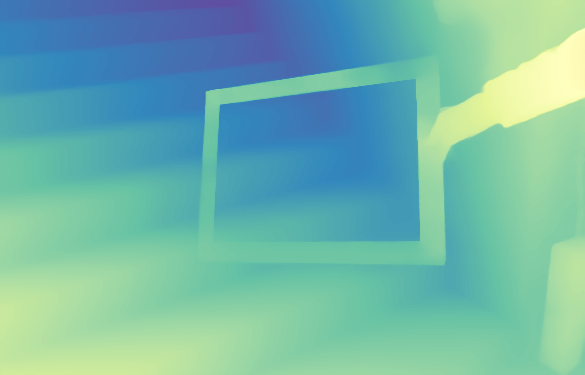} \\
        \small RAFT-Stereo \cite{lipson2021raft} &
        \small Selective-IGEV \cite{wang2024selective} &      
        \textbf{\method (ours)} \\
        \includegraphics[width=0.23\textwidth]{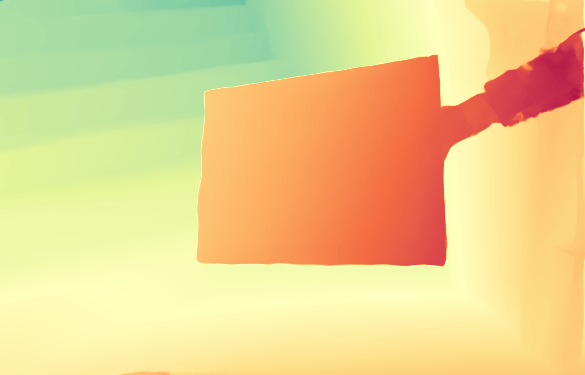} &
        \includegraphics[width=0.23\textwidth]{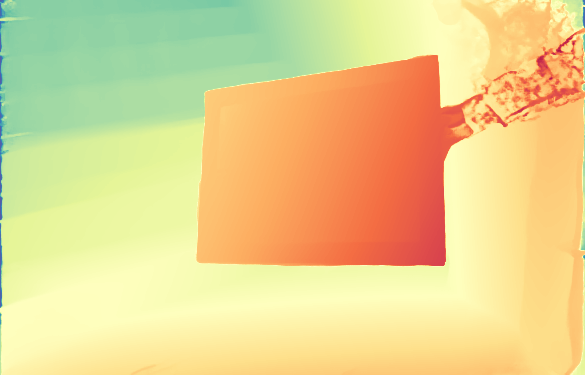} &
        \includegraphics[width=0.23\textwidth]{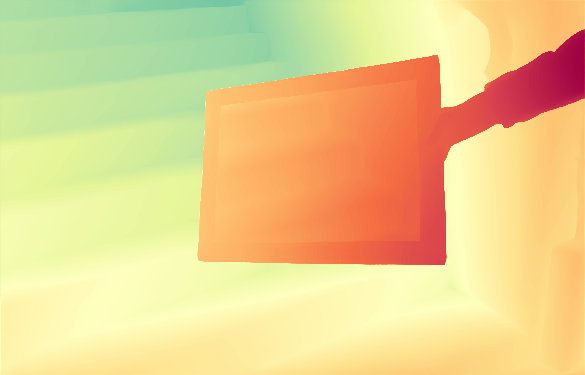} \\ 
    \end{tabular}\vspace{-0.3cm}
    \caption{\textbf{Qualitative Results -- MonoTrap.} Predictions by state-of-the-art models and \method.}
    \label{fig:qual_monotrap2}\vspace{-0.3cm}
\end{figure*}

To conclude, Figure \ref{fig:qual_monotrap2} collects three scenes from our novel \dataset dataset. In this case, we report predictions by both state-of-the-art monocular and stereo models, as well as by \method. 
The perspective illusions fooling monocular methods, unsurprisingly, do not affect stereo networks, which however are inaccurate near the left border of the image (first sample) or in the absence of texture (second sample). 
\method effectively combines the strength of both worlds, while being not affected by any of their weaknesses. 

\end{document}